\documentclass[11pt,oneside]{mnthesis}
\usepackage{epsfig} % Allows the inclusion of eps files
\usepackage{epic} % Enhanced picture mode
\usepackage{eepic} % Extensions for epic
\usepackage{units} % SI unit typesetting
\usepackage{url} % URL handling
\usepackage{longtable} % Tables that continue onto multiple pages
\usepackage{mathrsfs} % Support for \mathscr script
\usepackage{multirow} % Span rows in tables
\usepackage{bigstrut} % Space struts in tables up and down
\usepackage{amssymb} % AMS math symbols and helpers
\usepackage{graphicx} % Enhanced graphics support
\usepackage{setspace} % Adjust spacing in captions, single by default
\usepackage{xspace} % Automatically adjusting space after macros
\usepackage{amsmath} % \text, and other math formatting options
\usepackage{siunitx} % \num{} formatting and SI unit formatting
\usepackage{booktabs} % Enhanced tables with \toprule, etc.
\usepackage{hyperref} % Add clickable links to other parts of the document
\usepackage{subfigure}
\usepackage{algorithm,algorithmic}
\usepackage{setspace}
\usepackage{mdwlist}
\usepackage[title]{appendix}

\usepackage{comment}
\usepackage{makecell}
\usepackage{array}
\usepackage{amsthm}

\allowdisplaybreaks
\usepackage{bm,amsmath,amssymb,amsfonts,graphicx,amsthm,color}

\usepackage{bbold}
\usepackage{caption}
\usepackage{url}            % simple URL typesetting
\usepackage{booktabs}       % professional-quality tables
\usepackage{amsfonts}       % blackboard math symbols
\usepackage{nicefrac}       % compact symbols for 1/2, etc.
\usepackage{microtype}      % microtypography

\usepackage{times}
\usepackage{graphicx} % more modern
\usepackage{subfigure} 
\newcommand{\RN}[1]{%
	\textup{\uppercase\expandafter{\romannumeral#1}}%
}

\usepackage{wrapfig}
\usepackage{accents}
\makeatletter
\def\wid{\check{{\cc@style\underline{\mskip9.5mu}}}}
\def\Wideubar{\underaccent{{\cc@style\underline{\mskip6mu}}}}
\makeatother

\makeatletter
\def\wideubar{\underaccent{{\cc@style\underline{\mskip9.5mu}}}}
\def\Wideubar{\underaccent{{\cc@style\underline{\mskip6mu}}}}
\makeatother

\makeatletter
\def\widebar{\accentset{{\cc@style\underline{\mskip9.5mu}}}}
\def\Widebar{\accentset{{\cc@style\underline{\mskip6mu}}}}
\makeatother

\newcommand{\minimize}{{\rm minimize}}

\newtheorem{proposition}{Proposition}
\newtheorem{lemma}{Lemma}

\newtheorem{theorem}{Theorem}
\newtheorem{definition}{Definition}
\newtheorem{assumption}{Assumption}
\theoremstyle{remark}\newtheorem{remark}{Remark}

\begin{document}
	\bibliographystyle{IEEEtranS}
%\bibliographystyle{hunsrt} % style of bibliography
%%%%%%%%%%%%%%%%%%%%%%%%%%%%%%%%%%%%%%%%%%%%%%%%%%%%%%%%%%%%%%%%%%%%%%%%%%%%%%%%

%%%%%%%%%%%%%%%%%%%%%%%%%%%%%%%%%%%%%%%%%%%%%%%%%%%%%%%%%%%%%%%%%%%%%%%%%%%%%%%%
% Title and other sections that come before the body of the document
%%%%%%%%%%%%%%%%%%%%%%%%%%%%%%%%%%%%%%%%%%%%%%%%%%%%%%%%%%%%%%%%%%%%%%%%%%%%%%%%
%%%%%%%%%%%%%%%%%%%%%%%%%%%%%%%%%%%%%%%%%%%%%%%%%%%%%%%%%%%%%%%%%%%%%%%%%%%%%%%%
% title.tex - Set up the beginning of thesis.
%%%%%%%%%%%%%%%%%%%%%%%%%%%%%%%%%%%%%%%%%%%%%%%%%%%%%%%%%%%%%%%%%%%%%%%%%%%%%%%%

% Uncomment to turn on draft mode, which changes the title page to have a draft
% label and date of compilation
%\draft

% Set the type of thesis
\phd % use if for a Ph.D. dissertation
%\ms % use if for a Master of Science thesis

% Set the title and your name. Remember that the guidelines state:
%
% "The title of the thesis must not contain chemical or mathematical formulas,
% symbols, superscripts, subscripts, Greek letters, or other non-standard
% characters; words must be substituted."
%\title{\bf Achieving Spectrum Efficient Networking Design under Cross-Technology Interference}
\title{Robust, Deep, and Reinforcement Learning for Management of Communication and Power Networks}
%	Algorithms and Theory for Non-Convex Phase Retrieval}
\author{Alireza Sadeghi}
% Advisor name, put co-advisors here as well separated by commas
\director{Prof. Georgios B. Giannakis, Advisor}

% Specify the month and year; if commented out then these default to the
% current month and year
\submissionmonth{August}
\submissionyear{2021}

% Pages after the title page
\abstract{%%%%%%%%%%%%%%%%%%%%%%%%%%%%%%%%%%%%%%%%%%%%%%%%%%%%%%%%%%%%%%%%%%%%%%%%%%%%%%%%
% abstract.tex: Abstract
%%%%%%%%%%%%%%%%%%%%%%%%%%%%%%%%%%%%%%%%%%%%%%%%%%%%%%%%%%%%%%%%%%%%%%%%%%%%%%%%

%%%%%%%%%%%%%%%%%%%%%%%%%%%%%%%%%%%%%%%%%%%%%%%%%%%%%%%%%%%%%%%%%%%%%%%%%%%%%%%%
 
%\linespread{1.2}

{ \setstretch{1.269}

Data-driven machine learning algorithms have effectively handled standard learning tasks. This has become possible due mainly to the abundance of data in today's world, ranging from century-long climate records, high-resolution brain maps, all-sky survey observations, to billions of texts, images, videos, and search queries that we generate daily. By learning from big data, the state-of-the-art machine learning algorithms have successfully dealt with abundant, high-dimensional data in various domains. Although being capable of handling classical learning tasks, they fall short in managing and optimizing the next-generation highly complex cyber-physical systems, operational self-deriving cars, and self-surgical systems, which desperately need ground-breaking
control, monitoring, and decision making schemes that can guarantee \textit{robustness}, \textit{scalability}, and \textit{situational} awareness. For example, the  
emergence of multimedia services and Internet-friendly portable devices, the advent of diverse networks of
intelligent nodes such as those deployed to monitor the smart power grid, and transportation
networks will transform the cyder-physical infrastructures to an even more complex and heterogeneous one.  

To cope with the arising challenges in cyber-physical system control and management using machine learning toolkits, the present thesis first develops principled methods to make generic machine learning models robust against distributional uncertainties and adversarial data. Particular focus will be on parametric models where some training data are being used to learn a parametric model. The developed framework is of high interest especially when training and testing data are drawn from ``slightly'' \textit{different} distribution. We then introduce distributionally robust learning frameworks to minimize the worst-case expected loss over a prescribed ambiguity set of training distributions quantified via Wasserstein distance. Later, we build on this robust framework to design robust semi-supervised learning over graph methods. 

The second part of this thesis aspires to mange and control cyber-physical systems using machine learning toolkits. Especially, to fully unleash the potential of next-generation wired and wireless networks, we design ``smart'' network entities using (deep) reinforcement learning approaches. These network entities are capable of learning, tracking, and adapting to unknown dynamics or environments such as spatio-temporal dynamics of popular contents, network topologies, and diverse network resource allocation policies deployed across network entities. These smart units are such that they can pro-actively store reusable information during off-the peak periods and then reuse it during peak time instances. 

Finally, this thesis enhances the power system operation and control. Our contribution is on sustainable distribution grids with high penetration of renewable sources and demand response programs.~To account for unanticipated and rapidly changing renewable generation and load consumption scenarios, we specifically delegate reactive power compensation to both utility-owned control devices (e.g., capacitor banks), as well as smart inverters of distributed generation units with cyber-capabilities.~We further advocate an architectural paradigm shift from (stochastic) optimization-based power management to data-driven ``learning-to-control'' the system with reinforcements, by coupling physics-based optimization schemes with deep reinforcement learning (DRL) advances.~Our dynamic control algorithms are scalable and adaptive to real-time changes of renewable generation and load consumption. To further enhance the situational awareness in power network, we also develop robust power system state estimation solvers (PSSE). Our algorithms will leverage power network topology information, as well as data-driven priors to learn more fine tuned voltages across all the buses in the network from only limited available information. Numerical tests of pertinent tasks are provided for each chapter of this thesis. These tests well demonstrated the merits of the proposed algorithms. 
}}

% Copyright: Uncomment one of the following:
\copyrightpage       % Full copyright
%\copyrightpageccby   % Full copyright with Creative Commons CC-BY 4.0 license
%\copyrightpageccbysa % Full copyright with Creative Commons CC-BY-SA 4.0 license

% Acknowledgments and dedication
\acknowledgments{%%%%%%%%%%%%%%%%%%%%%%%%%%%%%%%%%%%%%%%%%%%%%%%%%%%%%%%%%%%%%%%%%%%%%%%%%%%%%%%%
% acknowledge.tex: Acknowledgements
%%%%%%%%%%%%%%%%%%%%%%%%%%%%%%%%%%%%%%%%%%%%%%%%%%%%%%%%%%%%%%%%%%%%%%%%%%%%%%%%

First and foremost, my deepest gratitude goes to my advisor Prof. Georgios B. Giannakis for providing me with the opportunity to be a part of SPiNCOM research group, as well as ECE/CS graduate program of University of Minnesota. He has helped me in developing clear and scientific thought and expression, and without his support, the completion of this PhD Thesis would not have been possible. 
%His advice and feedback on the formalization, investigation, and presentation of original research has been extraordinary to me by all means. His invaluable guidance as well as constant encouragement through dedicating extensive amounts oftime has not only made me a better researcher, but also a better person. 
His vision and enthusiasm about innovative research and beyond, his broad and deep knowledge, and his unbounded energy have constantly been a true inspiration for me.
Because of him, I have been very fortunate to be always surrounded by other wonderful students and colleagues. 

%I would also like to extend my sincerest appreciation of him for truly understanding of my concerns, and many other reasons which cannot be expressed in the space provided.

%%%%%%%%%%%%%%%%%%%%%%%%%%%%%%%%%%%%%%%%%%%%%%%%%%%%%%%%%%%%%%%%%%%%%%%%%%%%%%%%

Due thanks go to Profs. Mostafa Kaveh, Zhi-Li Zhang, and Mehmet Ak\c{c}akaya for agreeing to serve on my committee as well as all their valuable comments and feedback on my research and thesis.  
Thanks also go to other professors in the Departments of Electrical Engineering and Computer Science whose
graduate level courses helped me build the necessary background to embark on this journey.

During my PhD studies, I had the opportunity to collaborate with several excellent
individuals, and I have greatly benefited from their critical thinking, brilliant ideas, and vision. Particularly, I would like to
express my greatest gratitude to my friend and collaborator Dr. Fatemeh Sheikholeslami who was patient enough to train me in the first couple of difficult years at UMN, and Prof. Gang Wang, with whom I was fortunate to collaborate with and learn from. I greatly benefited from his vision, ideas, and insights. I would also like to extend my due credit and warmest thanks to Prof. Antonio G. Marques (King Juan Carlos University) for his valuable contribution and insights to our fruitful collaborations. The material in this thesis has also benefited from collaborating with Qiuling Yang.  

I would like to extend my gratitude to current and former members of the SPiNCOM group at UMN: Dr. Siavash Ghavami, Dr. Brian Baingana, Dr. Yanning Shen, Dr. Dimitris Berberidis, Dr. Jia Chen, Dr. Meng Ma, Prof. Tianyi Chen, Dr. Vassilis Ioannidis, Dr. Georgios V. Karanikolas, Dr. Donghoon Lee,  Dr. Athanasios Nikolakopoulos, Dr. Panagiotis Traganitis, Prof. Daniel Romero, Dr. Liang Zhang, and Seth Barrash. I am truly grateful to these people for their continuous help. I would also wish to acknowledge the grants that support financially our research. 

I am not forgetting my friends, some of which I have already mentioned above, both the ones here in Minneapolis, and my old-term friends that are far away, in particular: Danial Panahandeh-Shahraki, Amirhossein Hosseini, Mojtaba Kadkhodaei Elyaderani, Mohsen Mahmoodi, Fazel Zare Bidoky, Abolfazl Zamanpour Kiasari, Hamed Mosavat, Mehrdad Damsaz, Movahed Jamshidi, Javad Ansari, Meysam Mohajer, and Ali Ghaffarpour.

Finally, gift of a family is incomparable. They are the source of my strength, motivation, and sustenance. A special feeling of gratitude to my loving mother Fattaneh, whose heart I know is sick of my long distance. Special thanks also goes to my lovely sister Maedeh, who never left my side. I am eternally grateful to my mother, who encouraged me to pursue academic endeavor. 
%for their endless love, support, and inspiration throughout all my endeavors. 
Without you, I would not be standing here today.

\vspace{2em}
\begin{flushright}
\emph{Alireza Sadeghi, Minneapolis, April, 2021.}
\end{flushright}}
\dedication{This dissertation is dedicated to my mother and sister for their unconditional love and support. 
}

% Use a special preface
%\extra{\input{preface}}

% The \beforepreface command actually causes insertion of the title,
% abstract, signature, and copyright pages into the new document.
\beforepreface

% Define the text to go before the table of contents
\figurespage
\tablespage

% The \afterpreface command actually causes insertion of the
% contents, list of figures, etc. into the new document.
\afterpreface
%%%%%%%%%%%%%%%%%%%%%%%%%%%%%%%%%%%%%%%%%%%%%%%%%%%%%%%%%%%%%%%%%%%%%%%%%%%%%%%%

%%%%%%%%%%%%%%%%%%%%%%%%%%%%%%%%%%%%%%%%%%%%%%%%%%%%%%%%%%%%%%%%%%%%%%%%%%%%%%%%

%%%%%%%%%%%%%%%%%%%%%%%%%%%%%%%%%%%%%%%%%%%%%%%%%%%%%%%%%%%%%%%%%%%%%%%%%%%%%%%%
% Now lets include the body of the document...
%%%%%%%%%%%%%%%%%%%%%%%%%%%%%%%%%%%%%%%%%%%%%%%%%%%%%%%%%%%%%%%%%%%%%%%%%%%%%%%%
%%%%%%%%%%%%%%%%%%%%%%%%%%%%%%%%%%%%%%%%%%%%%%%%%%%%%%%%%%%%%%%%%%%%%%%%%%%%%%%
% intro.tex: Introduction to the thesis
%%%%%%%%%%%%%%%%%%%%%%%%%%%%%%%%%%%%%%%%%%%%%%%%%%%%%%%%%%%%%%%%%%%%%%%%%%%%%%%%
\chapter{Introduction}\label{chap:problem}
%%%%%%%%%%%%%%%%%%%%%%%%%%%%%%%%%%%%%%%%%%%%%%%%%%%%%%%%%%%%%%%%%%%%%%%%%%%%%%%%

\section{Motivation and Context}
Nowadays, overcoming emerging engineering challenges in cyber-physical systems requires successfully performing various learning tasks. At the same time, although machine learning algorithms have been successful in dealing with standard learning tasks with the sheer volume and high dimensionality of data, they are defenseless against adversarially manipulated input data, and sensitive to dynamically changing environments. Recent advancements in non-linear function approximation, optimal transport theory, and minimax optimization techniques provide a timely opportunity to transform machine learning algorithms to a scalable, reliable, secure, and safe technology to control and manage complex cyber-physical systems. In this context, the present thesis aspires to develop principled methods  incorporating \textit{scalability} along with \textit{robustness} in machine learning paradigms, having as ultimate goal to enhance their prediction, control, and tracking performance in unknown, dynamic, and possibly adversarial settings.

By putting forth an analytical and algorithmic framework for learning and inferring from data, with applications in management of cyber-physical systems, this thesis will develop a suit of machine learning based tools to optimally control these systems. our vision is to effect technical advances in function approximation, machine learning, optimal transport theory, and optimization to develop state-of-the-art algorithms for managing cyber-physical systems. The central goal is to theoretically, algorithmically, and numerically developed online, robust, and scalable algorithms. Specifically, the following research thrusts will be pursued:

\begin{itemize}
	\item[\textbf{(T1)}] \noindent Robust supervised learning under distributional uncertainties due to adversaries;
	{
		\item[\textbf{(T2)}] Distributionally robust semi-supervised learning and inference over graphs; and,
	}
	\item[\textbf{(T3)}] Deep- and reinforced-learning for network resource management;
	
	\item[\textbf{(T4)}] Data-driven, reinforced, and robust learning approaches for smart power grid.

\end{itemize}

\section{Learning Robust against distributional uncertainties and adversarial data}
It has been recently recognized that learning function models is vulnerable to adversarially manipulated input data, which discourages their use in safety-critical applications. In addition, learning algorithms often rely on the premise that training and testing data are drawn from the \textit{same} distribution, which may not hold in practice. Major efforts have been devoted to robustifying learning models to uncertainties arising from e.g., distributional mismatch using data pre-processing techniques such as compression, sparsification, and variance minimization~\cite{guo2017countering,krishnan2018combat,miyato2018regul}. While these can handle structured outliers, they are challenged by adversarial attacks, which can be mitigated by augmenting the training set with adversarially manipulated data \cite{fellow2014adv,ifgm2016adversarial,madry2017towards}. Despite their effectiveness, the latter fall short in performance guarantees, which motivates distributionally robust alternatives that minimize the worst-case expected loss over a prescribed ambiguity set of training distributions~\cite{quantifying19blanchet}. Means of quantifying uncertainty include momentum, likelihood, Kullback-Leibler (KL), and the Wasserstein distance~\cite{delage2010moment,hu2013kl,bandi2014robust}. Unfortunately, all robust approaches so far result in suboptimal solvers. In this context, the distributionally robust optimization framework is developed in this thesis for
training a parametric model, both in centralized and federated
learning settings. The objective is to endow the trained model
with robustness against adversarially manipulated input data, or,
distributional uncertainties, such as mismatches between training
and testing data distributions, or among datasets stored at
different workers. To this aim, the data distribution is assumed
unknown, and lies within a Wasserstein ball centered around
the empirical data distribution. This robust learning task entails
an infinite-dimensional optimization problem, which is challenging.
Leveraging a strong duality result, a surrogate is obtained,
for which three stochastic primal-dual algorithms are developed:
i) stochastic proximal gradient descent with an $\epsilon$-accurate oracle,
which invokes an oracle to solve the convex sub-problems; ii)
stochastic proximal gradient descent-ascent, which approximates
the solution of the convex sub-problems via a single gradient
ascent step; and, iii) a distributionally robust federated learning
algorithm, which solves the sub-problems locally at different
workers where data are stored. Compared to the empirical
risk minimization and federated learning methods, the proposed
algorithms offer robustness with little computation overhead.
Numerical tests using image datasets showcase the merits of the
proposed algorithms under several existing adversarial attacks
and distributional uncertainties.

\section{Robust semi-supervised inference over graphs.}
Inference tasks over social, brain, communication, biological, transportation, and sensor networks, have well-documented success by capitalizing on inter-dependencies captured by graphs~\cite{shuman2013mag, kolaczyk2014statistical}. In practice however, data are only available at a subset of nodes, due to e.g. sampling costs, and computational or privacy constraints. As inference is desired across all network nodes, such SSL tasks over networks can benefit from the underlying graph topology~\cite{sslbook, adadif2019berb, qin20120icassp}. Recent advances in graph neural networks (GNNs), offer parametric models that leverage the topology-guided structure of network data to form nested architectures that conveniently express processes over graphs~\cite{gnn2018survey, gnn2020comprehensivesurvey, antonio2019tsp}. 
%By leveraging the underlying irregular structure of networked data, the GNNs enjoy lower computational complexity, less parameters for training, and improved generalization capabilities relative to traditional DNNs, making them appealing for learning over graphs 

By succinctly encoding local graph structures and features of nodes, state-of-the-art GNNs can scale linearly with the size of graph. Despite their success in practice, most of existing methods are unable to handle graphs with uncertain nodal attributes. Specifically whenever mismatches between training and testing data distribution exists, these models fail in practice. Challenges also arise due to distributional uncertainties associated with data acquired by noisy measurements.  For instance, small perturbations to input data could significantly deteriorate the regression performance or result in classification error \cite{gnn2020adversarial, gnn2020adversarialsurvey}, just to name a couple of undesirable consequences. Hence, it is critical to endow learning and inference of processes over graphs with robustness against distributional uncertainties and adversarial data, especially in safety-critical applications, such as robotics \cite{multirobot2020learning} and transportation~\cite{gnn2020transport}. In this context, a distributionally robust learning framework is developed, where the objective is to train models that exhibit quantifiable robustness against perturbations. The data distribution is considered unknown, but lies within a Wasserstein ball centered around empirical data distribution. A robust model is obtained by minimizing the worst expected loss over this ball. However, solving the emerging functional optimization problem is challenging, if not impossible. Advocating a strong duality condition, we develop a principled method that renders the problem tractable and efficiently solvable. Experiments assess the performance of the proposed method.

\section{Deep- and reinforced-Learning for resource management} 
Consider the Internet, where millions of users rely on to access millions of terabyte of content such as Netflix or Amazon movies, music, social media on a daily bases. Serving end users with high quality of service in such huge-scale is no an easy task. In reality, to meet the ever-increasing data demand, novel technologies are required. Recognized as a key component is the so-called \textit{caching}, which refers to storing reusable popular contents across geographically distributed storage-enabled network entities. The rationale here is to alleviate unfavorable surges of data traffic by pro-actively storing anticipated highly popular contents at local storage devices during off-peak periods. Such resource pre-allocation is envisioned to provide significant savings in terms of network resources such as energy, bandwidth, and cost, in addition to increased user satisfaction.  To fully unleash its potential, a content-agnostic caching entity needs to rely on available observations to learn what and when to cache. A part of my research is to empower next generation networks with ``smart'' caching units, capable of learning, tracking, and adapting to unknown dynamics or environments such as spatio-temporal dynamics of content popularities, network topologies, and diverse caching policies deployed across network entities. By leveraging contemporary (deep) reinforcement learning tools, novel algorithms will be developed which are capable of progressively improving network performance in online and decentralized settings. 

Specifically we start with considering the caching problem in wireless networks, where small basestations (SBs) equipped with caching
units have potential to handle the unprecedented demand growth
in heterogeneous networks. Through low-rate, backhaul connections
with the backbone, SBs can prefetch popular files during
off-peak traffic hours, and service them to the edge at peak
periods. To intelligently prefetch, each SB must learn what and
when to cache, while taking into account SB memory limitations,
the massive number of available contents, the unknown popularity
profiles, as well as the space-time popularity dynamics
of user file requests. In this work, local and global Markov
processes model user requests, and a RL
(RL) framework is put forth for finding the optimal caching
policy when the transition probabilities involved are unknown.
Joint consideration of global and local popularity demands along
with cache-refreshing costs allow for a simple, yet practical
asynchronous caching approach. The novel RL-based caching
relies on a Q-learning algorithm to implement the optimal policy
in an online fashion, thus enabling the cache control unit at the
SB to learn, track, and possibly adapt to the underlying dynamics.
To endow the algorithm with scalability, a linear function
approximation of the proposed Q-learning scheme is introduced,
offering faster convergence as well as reduced complexity and
memory requirements. Numerical tests corroborate the merits of
the proposed approach in various realistic settings.

Then we build on this framework and consider a network of caches. In this context,
distributing the limited storage capacity across network entities
calls for decentralized caching schemes. Many practical caching
systems involve a parent caching node connected to multiple leaf
nodes to serve user file requests. To model the two-way interactive
influence between caching decisions at the parent and leaf nodes,
a RL framework is put forth. To handle
the large continuous state space, a scalable deep RL approach is
pursued. The novel approach relies on a hyper-deep Q-network
to learn the Q-function, and thus the optimal caching policy, in an
online fashion. Reinforcing the parent node with ability to learnand-
adapt to unknown policies of leaf nodes as well as spatiotemporal
dynamic evolution of file requests, results in remarkable
caching performance, as corroborated through numerical tests.

Finally, we design adaptive caching mechanism wedding tools from optimization and RL. We introduce simple but flexible generic time-varying fetching and caching costs, which are then used to formulate a constrained minimization of the  aggregate cost across files and time. Since caching decisions per time slot influence the content availability in future slots, the novel formulation for optimal fetch-cache decisions falls into the class of dynamic programming. Under this generic formulation, first by considering stationary distributions for the costs as well as file popularities, an efficient RL-based solver known as value iteration algorithm can be used to solve the emerging optimization problem. Then, it is shown that practical limitations on cache capacity can be handled using a particular instance of this generic dynamic pricing formulation. Under this setting, to provide a light-weight online solver for the corresponding optimization, the well-known RL algorithm, $Q$-learning, is employed to find optimal fetch-cache decisions. Numerical tests corroborating the merits of the proposed approach.

\section{Data-driven, reinforced, and robust learning approaches for smart power grid}
Given solar generation and load consumption predictions, voltage and reactive power control aims at optimizing reactive power injections to minimize a certain loss (e.g., power, voltage deviations), while respecting physical and operating constraints.~Proper redistribution of reactive power sources
can result in local correction of the power factor, increase system capacity, and improve power quality.~Traditionally, reactive compensation is provided by utility-owned equipment such as tap-changing under load transformers, voltage regulators, and manually-controlled capacitor banks \cite{salem1997voltage}, whose slow responses and limited lifespan render them ineffective in dealing with the variability introduced by distributed energy resources.~Advances in smart power inverters offer new opportunities, which can provide fast and continuously-valued reactive power injection or consumption.~Methods for compensating reactive power using the inverters of PV and storage systems have been advocated in \cite{farivar2011inverter,ZDGT13,kekatos2015stochastic,zhu2016fast,wang2016ergodic,VZ,lin2018real,magnusson2017voltage}.~Unfortunately, \textit{joint} control of both traditional utility-owned devices as well as contemporary smart inverters is challenging and has not been explored thus far, primarily because they operate in different timescales (e.g., hourly versus every few seconds), and involve discrete and continuous actions.~In addition, several fundamental challenges remain. How should one split the reactive power compensation duty equitably between the smart inverters and traditionally utility devices?~Should the control law be centralized (potentially vulnerable), distributed (more robust), or hybrid?~Whether centralized or decentralized, what variables
should be used as inputs to the control algorithms?~e.g., what should be the states, actions, and rewards of a RL algorithm? The main challenge arise due to the fact that the discrete on-off commitment of capacitor units is often configured on an hourly or daily basis, yet smart inverters can be controlled within milliseconds, thus challenging joint control of these two types of assets. In this context, a novel two-timescale voltage regulation scheme is developed for distribution grids by judiciously coupling data-driven with physics-based optimization. On a faster timescale, say every second, the optimal setpoints of smart inverters are obtained by minimizing instantaneous bus voltage deviations from their nominal values, based on either the exact alternating current power flow model or a linear approximant of it; whereas, on the slower timescale (e.g., every hour), shunt capacitors are configured to minimize the long-term discounted voltage deviations using a deep RL algorithm. Extensive numerical tests on a real-world $47$-bus distribution network as well as the IEEE $123$-bus test feeder using real data corroborate the effectiveness of the novel scheme. Finally, we finish this thesis by considering fas and robust state estimation (SE) to maintain a comprehensive view of the system in real
time. Conventional PSSE solvers typically entail minimizing a
nonlinear and nonconvex least-squares cost using e.g., the Gauss-
Newton method. Those iterative solvers however, are sensitive to
initialization, and may converge to local minima. To overcome
these hurdles, this thesis adapts and leverages recent
advances on image denoising to introduce a PSSE approach
with a regularizer capturing a deep neural network (DNN)
prior. For the resultant regularized PSSE objective, a ``Gauss-
Newton-type'' alternating minimization solver is developed.
To accommodate real-time monitoring, a novel end-to-end DNN is
constructed subsequently by unrolling the proposed alternating
minimization solver. The deep PSSE architecture can further
account for the power network topology through a graph neural
network (GNN) based prior. To further endow the physics-based
DNN with robustness against bad data, an adversarial DNN
training method is put forth. Numerical tests using real load data
on the IEEE 118-bus benchmark system showcase the improved
estimation and robustness performance of the proposed scheme
compared with several state-of-the-art alternatives.

\section{Thesis outline}
The reminder of this thesis is organized as follows.
Chapter \ref{chap:af} puts forth distributionally robust supervised and federated learning methods. Chapter \ref{chap:sslovergraphs} builds on the developed distributionally robust supervised learning framework to arrive at a distributionally robust semi-supervised learning over graphs. Chapter \ref{chap:saf} deals with deep and RL approaches to manage limited network resources. Finally, the objective of Chapter \ref{chap:sparse} is to design efficient learning approaches for smart grid management and control. Finally Chapter \ref{chap:concl} presents a concluding discussion of the proposed approaches, along with future research directions.

\section{Notational Conventions}

Unless otherwise noted, the following notation will be used throughout the subsequent chapters. Lower- (upper-) case boldface letters denote vectors (matrices). Calligraphic letters are reserved for sets, e.g., $\mathcal{S}$. For vectors, $\|\!\cdot\!\|_2$ or $\|\!\cdot\!\|$ represents the Euclidean norm, while $\|\!\cdot\!\|_0$ denotes the $\ell_0$ pseudo-norm counting the number of nonzero entries. The $n\times n$ identity matrix is denoted by $\mathbf{I}_n$, and all-one vector by $\mathbf{1}$, and all-zero vector $\mathbf{0}$. The size of the matricies (vectors) is omitted if it is obvious from the context; otherwise it is indicated by a subscript. Operator $(\cdot)^\top$ stands for matrix transpose, $|\cdot|$ the cardinality of a set, or the absolute value of a number.

%Base Chapters
%%%%%%%%%%%%%%%%%%%%%%%%%%%%%%%%%%%%%%%%%%%%%%%%%%%%%%%%%%%%%%%%%%%%%%%%%%%%%%%
% intro.tex: Introduction to the chapter
%%%%%%%%%%%%%%%%%%%%%%%%%%%%%%%%%%%%%%%%%%%%%%%%%%%%%%%%%%%%%%%%%%%%%%%%%%%%%%%%
\chapter{Learning Robust against Distributional Uncertainties and Adversarial Data}  \label{chap:af}
%%%%%%%%%%%%%%%%%%%%%%%%%%%%%%%%%%%%%%%%%%%%%%%%%%%%%%%%%%%%%%%%%%%%%%%%%%%%%%%%

\section{Introduction}
%\label{sec:intro}	
	
Machine learning models and tasks hinge on the premise that the training data are trustworthy, reliable, and representative of the testing data.~In practice however, data are usually generated and stored at geographically distributed devices (a.k.a., workers) each equipped with limited computing capability, and adhering to privacy, confidentiality, and possibly cost constraints \cite{federated_mag}.~Furthermore, the data quality is not guaranteed due to adversarially generated examples and distribution drifts across workers or from the training to testing phases~\cite{fedlearn_challenges}. Visually imperceptible perturbations to a dermatoscopic image of a benign mole can render the first-ever artificial intelligence (AI) diagnostic system approved by the U.S. Food and Drug Administration in 2018, to classify it as cancerous with 100\% confidence~\cite{mole}.~A stranger wearing pixelated sunglasses can fool even the most advanced facial recognition software in a home security system to mistake it for the homeowner~\cite{stranger}.~Hackers indeed manipulated readings of field devices and control centers of the Ukrainian supervisory control and data acquisition system to cause the first ever cyberattack-caused power outage in 2015 \cite{outage,tac2020wu}.~Examples of such failures in widely used AI-enabled safety- and security-critical systems today could put national infrastructure and even lives at risk. \footnote{Results of this Chapter are published in \cite{sadeghi2020learning}} 

Recent research efforts have focused on devising defense strategies against adversarial attacks.~These strategies fall under two groups: attack detection, and attack recovery.~The former identifies whether a given input is adversarially perturbed \cite{gu2014arxiv, lu2017iccv}, while the latter trains a model to gain robustness against such adversarial inputs \cite{guo2017countering, schmidt2018nips}, which is also the theme of the present contribution.~To robustify learning models against adversarial data, a multitude of data pre-processing schemes have been devised~\cite{miyato2018regul,sheikholeslami2019minimum}, to identify anomalies not adhering to postulated or nominal data.~Adversarial training on the other hand, adds imperceptible well-crafted noise to clean input data to gain robustness~\cite{fellow2014adv}; see also e.g., \cite{madry2017towards,moosavi2017iccvpr, papernotattack}, and \cite{advsurvey1} for a recent survey.~In these contributions, optimization tasks are formulated to craft adversarial perturbations.~Despite their empirical success, solving the resultant optimization problems is challenging.~Furthermore, analytical properties of these approaches have not been well understood, which hinders explainability of the obtained models.~In addition, one needs to judiciously tune hyper parameters of the attack model, which tends to be cumbersome in practice. 

On the other hand, data are typically generated and/or stored at geographically distributed sites, each having subsets of data with different distributions.~While keeping data localized to e.g., respect privacy, as well as reduce communication- and computation-overhead, the federated learning (FL) paradigm targets a global model, whereby multiple devices are coordinated by a central parameter server \cite{federated_mag}.~Existing FL approaches have mainly focused on the communicating versus computing tradeoff by aggregating model updates from the learners; see e.g., \cite{fedavg,fedsanjabi,fedpoor,shlezinger2020uveqfed} and references therein.~From the few works dealing with robust FL, \cite{fedrobustensemble} learns from dependent data through e.g., sparsification, while \cite{feduntrustedsources} entails an ensemble of untrusted sources.~These methods are rather heuristic, and rely on aggregation to gain robustness.~This context, motivates well a principled approach that accounts for the uncertainties associated with the underlying data distributions.

\section{Our Contribution}
Tapping on a distributionally robust optimization perspective, this Chapter develops robust learning procedures that respect privacy and ensure robustness to distributional uncertainties and adversarial attacks. Independent, identically distributed (i.i.d.) samples can be drawn from the known data distribution.~Building on~\cite{sinha2017certify}, the adversarial input perturbations are constrained to lie in a Wasserstein ball, and the sought robust model minimizes the worst-case expected loss over this ball.~As the resulting formulation leads to a challenging infinite-dimensional optimization problem, we leverage strong duality to arrive at a tractable and equivalent unconstrained minimization problem, requiring solely the empirical data distribution.~To solve the latter, a stochastic proximal gradient descent (SPGD) algorithm is developed based on an  $\epsilon$-accurate oracle, along with its lightweight stochastic proximal gradient descent-ascent (SPGDA) iteration.~The first algorithm relies on the oracle to solve the emerging convex sub-problems to $\epsilon$-accuracy, while the second simply approximates its solution via a single gradient ascent step.~To accommodate communication constraints and private or possibly untrusted datasets distributed across multiple workers, we further develop a distributionally robust federated learning (DRFL) algorithm. In a nutshell, the main contributions of this Chapter are as follows. 
\begin{itemize}
	\item A regularized distributionally robust learning framework to endow machine learning models with robustness against adversarial input perturbations;
	
	\item Two efficient proximal-type distributionally robust optimization algorithms with finite-sample convergence guarantees; and,
	
	\item A distributionally robust federated learning implementation to account for untrusted and possibly anonymized data from distributed sources. 
	
\end{itemize}

\section{Outline and notation}
Bold lowercase letters denote column vectors, while calligraphic uppercase fonts are reserved for sets; $\mathbb{E}[\cdot]$ represents expectation; $\nabla$ denotes the gradient operator; $(\cdot)^\top$ denotes transposition, and $\|\bm x\|$ is the $2$-norm of the vector $\bm x$.

The rest of this Chapter is structured as follows. Problem formulation and its robust surrogate are the subjects of Section \ref{sec:probstate}. The proposed SPGD with $\epsilon$-accurate oracle and SPGDA algorithms with their convergence analyses are presented in Sections \ref{sec:spgdorac} and \ref{sec:spgda}, respectively. The DRFL implementation is discussed in Section \ref{sec:distrib}.~Numerical tests are given in Section~\ref{sec:experiments} with conclusions drawn in Section \ref{sec:concls}. Technical proofs are deferred to the Appendix.

\section{Problem Statement}
\label{sec:probstate}
Consider the standard regularized statistical learning task
\begin{align}
& \underset{\bm{\theta}\in\Theta
}{{\rm min}}\;\;\mathbb{E}_{\bm{z}\sim P_0}\!\big[\ell(\bm{\theta};\bm{z})\big] + r(\bm{\theta})
\label{eq:reglearn}
\end{align}
where $\ell(\bm{\theta}; \bm{z})$ denotes the loss of a model parameterized by the unknown parameter vector $\bm{\theta}$ on a datum $\bm{z} =\! (\bm{x},y) \sim P_0$, with feature $\bm x$ and label $y$, drawn from some nominal distribution $P_0$. Here, $\Theta$ denotes the feasible set for model parameters. To prevent over fitting or incorporate prior information, regularization term $r(\bm{\theta})$ is oftentimes added to the expected loss. Popular regularizers include $r(\bm{\theta}):=\beta \|\bm{\theta}\|_1^2$ or $\beta \|\bm{\theta}\|_2^2$, where $\beta \ge 0$ is a hyper-parameter controlling the importance of the regularization term relative to the expected loss.  

In practice, the nominal distribution $P_0$ is typically unknown. Instead, we are given some data samples $\{\bm{z}_n\}_{n=1}^N \! \sim \! \widehat P_0^{(N)}$ (a.k.a. training data), which are drawn i.i.d. from $P_0$. Upon replacing $P_0$ with the so-called empirical distribution $\widehat P_0^{(N)}$ in \eqref{eq:reglearn}, we arrive at the empirical loss minimization 
\begin{align}
\underset{\bm{\theta}\in\Theta
}{{\rm min}}\;\; \bar{\mathbb{E}}_{\bm{z}\sim \widehat{P}^{(N)}_0}\!\big[\ell(\bm{\theta};\bm{z})\big] + r(\bm{\theta}) 
\label{eq:emlmin}
\end{align}
where $ \bar{\mathbb{E}}_{\bm{z}\sim \widehat{P}^{(N)}_0}\![\ell(\bm{\theta};\bm{z})] \!\!=\!\! N^{-1}  \sum_{n=1}^N \!\ell(\bm{\theta}; \bm{z}_n)$. 
Indeed, a variety of machine learning tasks can be cast as \eqref{eq:emlmin}, including e.g., ridge and Lasso regression, logistic regression, and reinforcement learning. The resultant models obtained by solving \eqref{eq:emlmin} however, have been shown vulnerable to adversarially corrupted data in $\widehat{P}_0^{(N)}$. Furthermore, the testing data distribution often deviates from the available $\widehat{P}_0^{(N)}$. For this reason, targeting an adversarially robust model against a set of distributions corresponding to perturbations of the underlying data distribution, has led to the  formulation~\cite{sinha2017certify}  
\begin{equation}
\underset{\bm{\theta}\in\Theta
}{{\rm min}}\;\,\sup_{P\in\mathcal{P}}\,\mathbb{E}_{\bm{z}\sim P}\!\left[\ell(\bm{\theta};\bm{z})
\right] + r(\bm{\theta})\label{eq:minsup0}
\end{equation}
where $\mathcal P$ represents a set of distributions centered around the data generating distribution $\widehat{P}^{(N)}_0$. Compared with \eqref{eq:reglearn}, the worst-case formulation \eqref{eq:minsup0}, yields models ensuring reasonable performance across a continuum of distributions characterized by $\mathcal{P}$. In practice, different types of ambiguity sets $\mathcal{P}$ can be considered, and they lead to different robustness guarantees and computational requirements. Popular choices of $\mathcal P$ include momentum \cite{delage2010moment, wiesemann2014moment}, KL divergence \cite{hu2013kl}, staatistical test \cite{bandi2014test}, and Wasserstein distance-based ambiguity sets \cite{bandi2014test,sinha2017certify}; see e.g., \cite{quantifying19blanchet} for a recent overview. Among all choices, it has been shown that the Wasserstein ambiguity set $\mathcal{P}$ results in a tractable realization of \eqref{eq:minsup0}, thanks to the strong duality result of  \cite{bandi2014test} and \cite{sinha2017certify}, which also motivates this work. 

To formalize this, consider two probability measures $P$ and $Q$ supported on set $\mathcal{Z}$, and let $\Pi(P,Q)$ be the set of all joint measures supported on $\mathcal{Z}^2$, with marginals $P$ and $Q$. Let $c: \mathcal{Z} \times \mathcal{Z}  \rightarrow [0, \infty)$ measure the cost of transporting a unit of mass from $\bm{z}$ in $P$ to another element $\bm{z}'$ in $Q$. The celebrated optimal transport
problem is given by \cite[page 111]{vinali08opttrans}
\begin{align}
W_c(P,Q) := \; \underset{\pi \in \Pi}{\inf} \, \mathbb{E}_\pi \big[ c(\bm{z},\bm{z}')\big]. 
\label{eq:wassdist}
\end{align}

\begin{remark}
	If $c(\cdot,\cdot)$ satisfies the axioms of distance, then $W_c$ defines a distance on the space of probability measures. For instance, if $P$ and $Q$ are defined over a Polish space equipped with metric $d$, then choosing $c(\bm z, \bm z') = d^p(\bm z, \bm z')$ for some $p\in [1, \infty)$ asserts that $W_c^{1/p}(P,Q)$ is the well-known Wasserstein distance of order $p$ between probability measures $P$ and $Q$ \cite[Definition 6.1]{vinali08opttrans}.  
\end{remark}
For a given empirical distribution $\widehat{P}^{(N)}_0$, define the uncertainty set $\mathcal{P}:= \{P| W_c(P,\widehat{P}^{(N)}_0) \le \rho\}$ to include all probability distributions having at most $\rho$-distance from $P_0^{(N)}$. Incorporating this ambiguity set into \eqref{eq:minsup0}, yields the following reformulation  
\begin{subequations}\label{eq:robform}
	\begin{equation}
	\underset{\bm{\theta}\in\Theta
	}{{\rm min}} \;\sup_{P}\,\mathbb{E}_{\bm{z}\sim P}\!\left[\ell(\bm{\theta};\bm{z})
	\right] + r(\bm{\theta}) \label{eq:robforma}
	\end{equation}
	\begin{equation}
	\qquad {\rm s.t.} \quad W_c(P,\widehat{P}^{(N)}_0) \le \rho. 
	\label{eq:robformb}
	\end{equation}
\end{subequations}

Observe that the inner supremum in \eqref{eq:robforma} runs over all joint probability measures $\pi$ on $\mathcal{Z}^2$ implicitly characterized by \eqref{eq:robformb}. Intuitively, directly solving this optimization over the infinite-dimensional space of distribution functions is challenging, if not impossible. Fortunately, for a broad range of losses as well as transport costs, it has been shown that the inner maximization satisfies a strong duality condition \cite{quantifying19blanchet}; that is, the optimal objective of this inner maximization and its Lagrangian dual optimal objective, are equal. In addition, the dual problem involves optimization over a one-dimensional dual variable. These two observations make it possible to solve \eqref{eq:minsup0} in the dual domain. To formally obtain a tractable surrogate to \eqref{eq:robform}, we make the following assumptions. 
\begin{assumption}
	\label{as:cfunc}
	The transportation cost function $c:\mathcal{Z}\times\mathcal{Z} \to [0, \infty)$, is a lower semi-continuous function satisfying $c(\bm{z}, \bm{z})=0$ for $\bm{z} \in \mathcal{Z}$\footnote{A simple example satisfying these constraints is the Euclidean distance $c(\bm z, \bm z') = \|\bm z - \bm z' \|$.}.
\end{assumption}

\begin{assumption}
	\label{as:lfunc}
	The loss function $\ell: \Theta \times \mathcal{Z} \to [0, \infty)$, is upper semi-continuous, and integrable.
\end{assumption}

\noindent The following proposition provides a tractable surrogate for \eqref{eq:robform}, whose proof can be found in \cite[Theorem 1]{quantifying19blanchet}.  
\begin{proposition}
	\label{prop:strongdual}
	Let $\ell: \Theta \times \mathcal{Z} \rightarrow [0,\infty)$, and $c:\mathcal{Z} \times \mathcal{Z} \rightarrow [0, \infty)$ satisfy Assumptions $1$ and $2$, respectively. Then, for any given $\widehat{P}^{(N)}_0$, and $\rho > 0$, it holds that
	\begin{equation}
	\sup_{P\in\mathcal{P}} \,\mathbb{E}_{\bm{z}\sim P}\!\left[\ell(\bm{\theta};\bm{z})
	\right] 
	 \quad = \inf_{\gamma \ge 0} \big\{  \bar{\mathbb{E}}_{\bm z \sim \widehat{P}^{(N)}_0} \big[\sup_{\bm \zeta \in {\mathcal Z}} \!\left\{ \ell(\bm \theta; \bm \zeta) -\gamma (c(\bm z, \bm \zeta)-\rho)\right\} \big] \big\}
	\label{eq:strongduality}
	\end{equation}
	where $\mathcal{P}:= \big\{P| W_c(P,\widehat{P}^{(N)}_0) \le \rho \big\}$.
\end{proposition}

\begin{remark}
	Thanks to strong duality, the right-hand side in \eqref{eq:strongduality} simply is a univariate dual reformulation of the primal problem represented in the left-hand side. In sharp contrast with the primal formulation, the expectation in the dual domain is taken only over the empirical $\widehat{P}^{(N)}_0$ rather than any $P \in \mathcal{P}$. In addition, since this reformulation circumvents the need for finding the optimal $\pi \in \Pi$ to form $\mathcal{P}$, and  characterizing the primal objective $\forall P \in \mathcal{P}$, it is practically more convenient.
\end{remark}

Upon relying on Proposition \ref{prop:strongdual}, the following distributionally robust surrogate is obtained 
\begin{align}
\min_{\bm{\theta}\in \Theta} \inf_{\gamma \ge 0} \! \big\{  \bar{\mathbb{E}}_{\bm z \sim \widehat{P}^{(N)}_0} \!\big[\sup_{\bm \zeta \in {\mathcal Z}} \!\left\{ \ell(\bm \theta; \bm \zeta) \!+\! \gamma (\rho - c(\bm z, \bm \zeta))\right\} + r(\bm \theta) \big] \!\big\}.
\label{eq:robustdual}
\end{align}

\begin{remark}
	\label{rm:minmax}
	The robust surrogate in \eqref{eq:robustdual} boils down to minimax (saddle-point) optimization which has been widely studied in e.g., \cite{jordan2019minmax}. However, \eqref{eq:robustdual} requires the supremum to be solved separately for each sample $\bm{z}$, and the problem cannot be handled through existing methods.  
	
	A relaxed (hence suboptimal) version of \eqref{eq:robustdual} with a fixed $\gamma$ value 
	has recently been studied in \cite{sinha2017certify}. Unfortunately, one has to select an appropriate $\gamma$ value using cross validation over a grid search that is also application dependent. Heuristically choosing a $\gamma$ does not guarantee optimality in solving the distributionally robust surrogate \eqref{eq:robustdual}. Clearly, the effect of heuristically selecting $\gamma$ is more pronounced when training deep neural networks. Instead, we advocate algorithms that optimize $\gamma$ and $\bm{\theta}$ simultaneously. 
	
	Our approach to addressing this, relies on the structure of \eqref{eq:robustdual} to \textit{iteratively} update parameters $\bar{\bm \theta}:=[\bm \theta^\top~ \gamma]^\top$ and $\bm \zeta$. To end up with a differentiable function of $\bar{\bm \theta}$ after maximizing over $\bm \zeta$, Danskin's theorem requires the sup-problem to have a unique solution~\cite{bertsekas1997nonlinear}. For this reason, we design the inner maximization to involve a strongly concave objective function through the selection of a strongly convex transportation cost, such as $c(\bm{z}, \bm{z'}) :=\| \bm{z} - \bm{z'}\|^2_p$ for $p \ge 1$. For the maximization over $\bm \zeta$ to rely on a strongly concave objective, we let $\gamma \in \Gamma :=\{\gamma| \gamma >\gamma_0\}$, where $\gamma_0$ is large enough. Since $\gamma$ is the dual variable corresponding to the constraint in \eqref{eq:robform}, having $\gamma \in \Gamma$ is tantamount to tuning $\rho$, which in turn \emph{controls} the level of \emph{robustness}. Replacing $\gamma \ge 0$ in \eqref{eq:robustdual} with $\gamma \in \Gamma$, our \emph{robust learning model} is obtained as the solution of 
\begin{align}
\min_{\bm{\theta}\in \Theta}~\inf_{\gamma \in \Gamma}~\bar{\mathbb E}_{\bm{z} \sim \widehat{P}^{(T)}_0}\big[ \sup_{\bm{\zeta}\in\mathcal{Z}} \psi(\bar{\bm{\theta}}, \bm{\zeta}; \bm {z}) \big] + r(\bar{\bm{\theta}})
\label{eq:objective}
\end{align}
	where $ \psi(\bar{\bm{\theta}}, \bm{\zeta}; \bm {z}):=\ell(\bm \theta;\bm \zeta) + \gamma (\rho -  c(\bm{z}, \bm \zeta))$.
	Intuitively, input $\bm{z}$ in \eqref{eq:objective} is pre-processed by maximizing $\psi$ accounting for the adversarial perturbation. To iteratively solve our objective in \eqref{eq:objective}, the ensuing sections provide efficient solvers under some mild conditions. Those include cases, every inner maximization (supremum) can be solved to $\epsilon$-optimality by an oracle.   
\end{remark}

Before developing our algorithms, we make several standard assumptions; see also \cite{sinha2017certify}, \cite{jordan2019minmax}.

\begin{assumption}
	\label{as:cstrongcvx}
	Function $c(\bm z, \cdot)$ is $L_c$-Lipschitz and $\mu$-strongly convex for any given $\bm{z} \in \mathcal{Z}$, with respect to the norm $\|\cdot\|$. 
\end{assumption}

\begin{assumption}
	\label{as:losslipschitz}
	The loss function $\ell(\bm \theta; \bm z)$ obeys the following Lipschitz smoothness conditions 
	\begin{subequations}
		\label{eq:as2}
		\begin{align}
		& 
		\|\nabla_{\bm \theta} \ell(\bm{\theta}; \bm{z})- \nabla_{\bm{\theta}} \ell(\bm{\theta}';\bm{z})\|_{\ast} \le L_{\bm{\theta} \bm{\theta}} \|\bm{\theta}-\bm{\theta}'\|
		\label{eq:as2th}
		\\ 
		&
		\| \nabla_{\bm{\theta}} \ell(\bm{\theta; \bm{z}}) - \nabla_{\bm{\theta}} \ell(\bm{\theta}; \bm{z}')\|_\ast \le L_{\bm{\theta}\bm{z}} \|\bm{z}-\bm{z}'\| 
		\\ 
		&
		\label{eq:as2z}
		\| \nabla_{\bm{z}} \ell(\bm{\theta; \bm{z}}) - \nabla_{\bm{z}} \ell(\bm{\theta}; \bm{z}')\|_\ast \le L_{\bm{z}\bm{z}} \|\bm{z}-\bm{z}'\|
		\\
		&
		\label{eq:as2ztheta}
		\|\nabla_{\bm z} \ell(\bm{\theta}; \bm{z})- \nabla_{\bm{z}} \ell(\bm{\theta}';\bm{z})\|_{\ast} \le L_{\bm{z} \bm{\theta}} \|\bm{\theta}-\bm{\theta}'\|
		\end{align}
		and it is continuously differentiable with respect to $\bm{\theta}$.
	\end{subequations}
\end{assumption}

Assumption  \eqref{as:losslipschitz} guarantees that the supremum in \eqref{eq:robustdual} results in a smooth function of $\bar{\bm{\theta}}$; thus, one can execute gradient descent to update $\bm \theta$ upon solving the supremum. This will further help to provide convergence analysis of our proposed algorithms. To elaborate more on this, the following lemma characterizes the smoothness and gradient Lipschitz properties obtained upon solving the maximization problem in \eqref{eq:objective}. 

\begin{lemma}
	\label{lem:smooth}
	For each $\bm{z}\!\in\mathcal{Z}$, 
	%let us
	define 
	%$\psi(\bar{\bm  \theta}, \bm  \zeta; \bm{z}) :=  \ell(\bm \theta; \bm \zeta) - \gamma (c(\bm z, \bm \zeta) - \rho)$ and define 
	$\bar{\psi} (\bar{\bm  \theta}; \bm{z})\!=\! \sup_{\bm{\zeta}} \psi (\bar{\bm  \theta}, \bm  \zeta; \bm{z})$ with $\bm{\zeta}_\ast (\bar{\bm{\theta}}; \bm{z}) = \arg \max_{\bm{\zeta} \in \mathcal{Z}} \psi(\bar{\bm  \theta}, \bm  \zeta; \bm{z})$. Then $\bar{\psi}(\cdot)$ is differentiable, and its gradient is $\nabla_{\bar{\bm{\theta}}} \bar{\psi}(\bar{\bm{\theta}};\bm{z}) = \nabla_{\bar{\bm{\theta}}} \psi(\bar{\bm{\theta}}, \bm{\zeta}_\ast(\bar{\bm{\theta}}; \bm{z});\bm{z})$. Moreover, the following conditions hold  
	\begin{subequations}
		\begin{align}
		& \big \|\bm \zeta_\ast(\bar{\bm{\theta}}_1; \bm{z}) -  \bm \zeta_\ast(\bar{\bm{\theta}}_2; \bm{z}) \big \| 
		\le \frac{L_{\bm{z}\bm{\theta}}}{\lambda}  \|\bm{\theta}_2 -\bm{\bm{\theta}}_1 \| + \frac{L_c}{\lambda}\, \| \gamma_2 -\gamma_1 \| \label{eq:lemmax}\\
		%\end{equation}
		%and
		%\begin{align}
		& \big\| \nabla_{\bar{\bm{\theta}}} \bar{\psi}(\bar{\bm{\theta}}_1; \bm{z}) -   \nabla_{\bar{\bm{\theta}}} \bar{\psi}(\bar{\bm{\theta}}_2; \bm{z})  \big\|  \le \frac{ L_{\bm \theta z }L_c + L^2_c}{\lambda}\,  \| \gamma_2 -\gamma_1 \|   
		\\
		& +  \Big(L_{\bm{\theta \theta}} + \frac{L_{\bm \theta z }L_{\bm{z}\bm{\theta}} + L_c L_{\bm{z}\bm{\theta}}}{\lambda}\Big)  \|\bm{\theta}_2 - \, \bm{\bm{\theta}}_1 \| \label{eq:lemmapsi}
		\end{align}
	\end{subequations}
	where $\gamma^{1,2}\in \Gamma$, and  $\psi(\bar{\bm{\theta}}, \cdot; \bm{z})$ is $\lambda$-strongly concave.
\end{lemma}
Proof: See Appendix \ref{app:lemm1} for the proof.

Lemma \ref{lem:smooth} paves the way for iteratively solving the surrogate optimization \eqref{eq:objective}, intuitively because it guarantees a differentiable and smooth objective upon solving the inner supremum to its optimum.
\begin{remark}
	Equation \eqref{eq:lemmax} is appealing in practice.~Indeed, if $\bar{\bm{\theta}}^t\!=\![\bm{\theta}^t, \gamma^t]$ is updated with a small enough step size, the corresponding $\bm{\zeta}_\ast(\bm{\theta}^{t+1};\bm{z})$ is close enough to $\bm{\zeta}_\ast(\bm{\theta}^{t};\bm{z})$.~Building on this observation, instead of using an oracle to find the optimum $\bm{\zeta}_\ast(\bm{\theta}^{t+1};\bm{z})$, an $\epsilon$-accurate solution $\bm{\zeta}_\epsilon(\bm{\theta}^{t+1};\bm{z})$ suffices to obtain comparable performance.~This also circumvents the need to find the optimum for the inner maximization per iteration, which could be computationally demanding. 
\end{remark}

\section{Stochastic Proximal Gradient Descent with $\epsilon$-accurate Oracle}
\label{sec:spgdorac}

\begin{algorithm}[t]
	{{\bfseries Input}: Initial guess $\bar{\bm{\theta}}^0$, step size sequence $\{\alpha_t >0 \}_{t=0}^{T}$, $\epsilon$-accurate oracle}
	
%	{{\bfseries Output}: Initial guess $\bar{\bm{\theta}}^0$, step size sequence $\{\alpha_t >0 \}_{t=0}^{T}$, $\epsilon$-accurate oracle}
	
%	\SetKwInOut{Output}{Output}
%	\SetKw{Set}{Set}	
%	\Input{Initial guess $\bar{\bm{\theta}}^0$, step size sequence $\{\alpha_t >0 \}_{t=0}^{T}$, $\epsilon$-accurate oracle}
	
	{$t = 1, \ldots, T$}
	{ {Draw i.i.d samples $\{\bm{z}_n\}_{n=1}^N$} \\ 
		{Find $\epsilon$-optimizer $\bm{\zeta}_\epsilon(\bar{\bm{\theta}^t}; \bm{z}_n)$ via the oracle} \\
		{Update:  \newline $\bar{\bm{\theta}}^{t+1} = {\rm prox}_{\alpha_t r} \Big[ \bar{\bm{\theta}}^t-\frac{\alpha_t}{N}\sum_{n=1}^N \nabla_{\bar{\bm{\theta}}} \psi (\bar{\bm{\theta}}, {\bm \zeta}_\epsilon(\bar{\bm{\theta}}^t; \bm{z}_n); \bm{z}_n)\big|_{\bar{\bm{\theta}}=\bar{\bm{\theta}}^t} \Big]$}
	}	  
	\caption{SPGD with $\epsilon$-accurate oracle}
	\label{alg:spgd}
\end{algorithm}

A standard solver of regularized optimization problems is the proximal gradient algorithm.~In this section, we develop a variant of it to tackle the robust surrogate \eqref{eq:objective}. For convenience, let us define
\begin{equation}
f(\bm{\theta}, \gamma) := 
%\inf_{\gamma \ge 0} \! \big\{  
{\mathbb{E}} \;
%_{\bm z \sim \widehat{P}^{(N)}_0}
\!\big[\sup_{\bm{\zeta} \in {\mathcal Z}} \!\left\{ \ell(\bm \theta; \bm{\zeta}) \!+\! \gamma (\rho - c(\bm z, \bm{\zeta}) ) \right\} \big] \label{eq:fdef}
%\big\}. 
\end{equation}

and rewrite our objective as 
\begin{align}
\min_{\bm{\theta}\in \Theta} \inf_{\gamma \in \Gamma} \; F(\bm{\theta}, \gamma) := f(\bm{\theta}, \gamma) + r(\bm{\theta}) \label{eq:sorrogate}
\end{align}
where $f(\bm{\theta}, \gamma)$ is the smooth function in \eqref{eq:fdef}, and $r(\bm{\cdot})$ is a non-smooth and convex regularizer, such as the $\ell_1$-norm. With a slight abuse of notation, upon introducing $\bar{\bm \theta}:=[\bm \theta~\gamma]$,  we define  $f(\bar{\bm \theta}):= f(\bm \theta, \gamma)$ and $F(\bar{\bm \theta}):= F(\bm \theta, \gamma)$. 

The proximal gradient algorithm the updates $\bar{\bm{\theta}}^t$ as  
\begin{equation*}
\bar{\bm \theta}^{t+1} = \arg \min_{\bm \theta} \, \alpha_t r(\bm \theta) + \alpha_t \big\langle \bm \theta - \bar{\bm \theta}^t, \bm g (\bar{\bm \theta}^t) \big \rangle + \frac{1}{2} \big\|\bm \theta - \bar{\bm \theta}^{t}\big\|^2 
%\label{eq:proxupdate}
\end{equation*}
where $\bm g (\bar{\bm \theta}^t) := \nabla f(\bar{\bm{\theta}})|_{\bar{\bm{\theta}} = \bar{\bm{\theta}^t}}$, and $ \alpha_t > 0$ is some step size. The last update is expressed in the compact form 
\begin{equation}
\bar{\bm \theta}^{t+1} = \textrm{prox}_{\alpha_t r} \big[\bar{\bm \theta}^{t} - \alpha_t \bm g(\bar{\bm  \theta}^{t})\big]\label{eq:thetaupdt}
\end{equation}
where the proximal gradient operator is given by
\begin{equation}
\textrm{prox}_{\alpha r} [\bm  v] := \arg  \min_{\bm  \theta} \; \alpha r(\bm  \theta) + \frac{1}{2}\|\bm  \theta - \bm  v\|^2. \label{eq:proximal}
\end{equation}
The working assumption is that this optimization problem can be solved efficiently using off-the-shelf solvers.

Starting from the guess $\bar{\bm \theta}^0$, the proposed SPGD with $\epsilon$-accurate oracle executes two steps per iteration $t=1,$ $2,\ldots$. First, it relies on an $\epsilon$-accurate maximum oracle to solve the inner problem $\sup_{\bm \zeta \in {\mathcal Z}} \! \{ \ell(\bm \theta^t; \bm \zeta) \!-\! \gamma^t c(\bm z, \bm \zeta)\}$ for randomly drawn samples $\{\bm{z}_n\}_{n=1}^{N}$ to yield $\epsilon$-optimal  $\bm{\zeta}_\epsilon(\bar{\bm{\theta}}^t,\bm{z}_n)$ with the corresponding objective values $\psi(\bar{\bm{\theta}}^t,\bm  \zeta_{\epsilon}(\bar{\bm \theta}^t, \bm  z_n); \bm  z_n)$. Next, $\bar{\bm{\theta}}^t$ is updated using a stochastic proximal gradient step as 
\begin{equation}
\bar{\bm{\theta}}^{t+1} = {\text prox}_{\alpha_t r} \Big[ \bar{\bm{\theta}}^t-\frac{\alpha_t}{N}\sum_{n=1}^N \nabla_{\bar{\bm{\theta}}} \psi (\bar{\bm{\theta}}, \bm \zeta_\epsilon(\bar{\bm{\theta}}^t; \bm{z}_n); \bm{z}_n) \Big]. \nonumber
\end{equation}

For implementation, the proposed SPGD algorithm with $\epsilon$-accurate oracle is summarized in Alg. \ref{alg:spgd}.~Convergence performance of this algorithm is analyzed in the ensuing subsection.

\subsection{Convergence of SPGD with $\epsilon$-accurate oracle} 
\label{suc:convspgd}
In general, the postulated model is nonlinear, and the robust surrogate $F(\bar{\bm{\theta}})$ is nonconvex.~In this section, we characterize the convergence performance of Alg. \ref{alg:spgd} to a stationary point.~However, lack of convexity and smoothness implies that stationary points must be understood in the sense of the Fr\`echet subgradient.~Specifically, the Fr\`echet subgradient
${\partial} F(\check{\bm{\theta}})$ for the composite optimization in \eqref{eq:sorrogate}, is the set \cite{rockafellar2009variational}
\begin{equation}
{\partial} F(\check{\bm{\theta}})\! :=\! \Big\{\bm{v} \, \big| \lim_{\bar{\bm \theta} \to \check{\bm{\theta}}} \inf \frac{F(\bar{\bm{\theta}})-F(\check{\bm{\theta}})-\bm{v}^\top (\bar{\bm{\theta}}-\check{\bm{\theta}}))}{\|\bar{\bm{\theta}} - \check{\bm{\theta}}\|} \ge 0 \Big\}. \nonumber
\end{equation}
Consequently, the distance between vector $\bm 0$ and the set $\partial F(\check{\bm{\theta}})$ is a measure characterizing whether a point is stationary or not. To this end, define the distance between a vector $\bm v$ and a set $\mathcal{S}$ as ${\rm dist}(\bm{v}, \mathcal{S}):= \min_{\bm{s} \in \mathcal{S}}\|\bm{v}-\bm{s}\|$, and the notion of $\delta$-stationary points as defined next.
\begin{definition}
	Given a small $\delta > 0$, we call vector $\check{\bm{\theta}}$ a $\delta$-stationary point if and only if
	${\rm dist}(\bm 0, \partial F(\check{\bm{\theta}})) \le \delta$.
	\label{de:stationary}
\end{definition}

Since $f(\cdot)$ in \eqref{eq:fdef} is smooth, we have that ${\partial} F(\bar{\bm{\theta}}) = \nabla f(\bar{\bm{\theta}}) + {\partial} r(\bar{\bm{\theta}})$ \cite{rockafellar2009variational}.~Hence, it suffices to prove that the algorithm converges to a $\delta$-stationary point $\check{\bm{\theta}}$ satisfying
\begin{align}
{\rm dist}\big(\bm{0},\nabla f(\check{\bm{\theta}}) + {
	\partial} r(\check{\bm{\theta}})\big) \le \delta.
\end{align}  
We further adopt the following assumption that is standard in stochastic optimization. 
\begin{assumption}
	\label{as:grdestimate}
	Function $f$ satisfies the next two conditions. 
	\begin{enumerate}	
		\item Gradient estimates are unbiased and have a bounded variance, i.e., $\mathbb{E} [  \bm g^\ast(\bar{\bm \theta}^t) - \nabla f (\bar{\bm \theta}^t) ] = \bm{0} $, and there is a constant $\sigma^2 <\infty$, so that $\mathbb{E} [\|\nabla f (\bar{\bm \theta}^t) \! - \! \bm g^\ast(\bar{\bm \theta}^t)\|_2^2] \le \sigma^2$. 
		\item Function $f (\bar{\bm \theta})$ is smooth with $L_f$-Lipschitz continuous gradient, i.e., \newline 
		$\| \nabla f (\bar{\bm \theta}_1) - \nabla f (\bar{\bm \theta}_2)\| \le L_f \|\bar{\bm \theta}_1 - \bar{\bm \theta}_2 \|$.
	\end{enumerate}
\end{assumption}
We are now ready to claim the convergence guarantees for Alg. \ref{alg:spgd}; see Appendix \ref{app:thm1} for the proof.
\begin{theorem} Let Alg. \ref{alg:spgda} run for $T$ iterations with constant step sizes $\alpha, \eta >0$. Under Assumptions \ref{as:cfunc}--\ref{as:grdestimate}, Alg. \ref{alg:spgda} generates a sequence of $\{\bar{\bm{\theta}}^t\}$ that satisfies
	\begin{align}
	\nonumber
	\mathbb{E}\,[{\textrm dist}(\bm{0}, \partial F(\bar{\bm \theta}^{t'}))^2] 
	& \le  
	\Big(\frac{2}{\alpha} +\beta \Big) \frac{\Delta_F}{T} + \Big( \frac{\beta}{\eta}+2\Big) \sigma^2   
	\\
	& \quad + \frac{(\beta+2)L_{\bar{\bm \theta z}}^2 \epsilon}{\lambda_0}
	\end{align} 
	where $t'$ is uniformly sampled from $\{ 1,\ldots, T\}$; here, $\Delta_F := F(\bar{\bm{\theta}}^0) - F(\bar{\bm{\theta}}^{T+1})$; $ L_{\bar{\bm \theta} \bm  z}^2 := L_{\bm \theta \bm  z}^2 + \lambda_0 L_c$, and $\beta, \,\lambda_0>0$ are some constants.
	\label{thm:convspgd}
\end{theorem}

Theorem \ref{thm:convspgd} asserts that $\{\bar{\bm{\theta}}^t\}_{t=1}^{T}$ generated by Alg. \ref{alg:spgd} converges to a stationary point on average.~The upper bound here is characterized by the initial error $\Delta_F$, which decays at the rate of $\mathcal{O}(1/T)$; and, the constant bias terms induced by the gradient estimate variance $\sigma^2$ as well as the oracle accuracy~$\epsilon$.

\begin{remark}[Oracle implementation]
	The $\epsilon$-accurate oracle can be implemented in practice by several optimization algorithms, with gradient ascent being a desirable one due to its simplicity.~Assuming $\gamma_{0} \ge {L_{\bm{z z}}}/{\mu}$, gradient ascent with constant step size $\eta$ obtains an $\epsilon$-accurate solution within at most $\mathcal{O}(\log({d_0^2}/{\epsilon \eta}))$ iterations, where $d_0$ is the diameter of set $\mathcal{Z}$.
	
	The computational complexity of Alg. \ref{alg:spgd} can grow prohibitively when dealing with large-size datasets and complex models.~This motivates lightweight, scalable, yet efficient methods.~To this end, we introduce next a stochastic proximal gradient descent-ascent (SPGDA) algorithm.
\end{remark}

\section{Stochastic Proximal Gradient Descent-Ascent}
\label{sec:spgda}
Leveraging the strong concavity of the inner maximization problem and Lemma \ref{lem:smooth}, a lightweight variant of the SPGD with $\epsilon$-accurate oracle is developed here. Instead of optimizing the inner maximization problem to $\epsilon$-accuracy by an oracle, we approximate its solution after only a \textit{single} gradient ascent step. Specifically, for a batch of data $\{\bm{z}^t_m\}_{m=1}^{M}$ per iteration $t$, our SPGDA algorithm first perturbs each datum 
via a gradient ascent step
\begin{align}
\bm{\zeta}_m^t =  \bm{z}_m^t  + \eta_t \nabla_{\bm{\zeta}} \psi(\bar{\bm \theta}^t, \bm  \zeta; \bm{z}_m^t)\big|_{\bm{\zeta}={\bm{z}_m^t}},~\forall m=1,\ldots,M \label{eq:sgaupdt}
\end{align}
and then forms
\begin{align}
\bm g^{t}(\bar{\bm \theta}^t)  := \frac{1}{M} \sum_{m=1}^{M} \! \nabla_{\bar{\bm{\theta}}} \psi (\bar{\bm \theta}, \bm \zeta_m^{t}; \bm z_m^t)\big|_{\bar{\bm \theta}=\bar{\bm \theta}^{t}}. \label{eq:gsgda}
\end{align}
Using \eqref{eq:gsgda}, an extra proximal gradient step is taken to obtain
\begin{align}
\bar{\bm{\theta}}^{t+1} = \textrm{prox}_{\alpha_t r}
\big[ \bar{\bm{\theta}}^{t} -  \alpha_t \bm g^{t}(\bar{\bm \theta}^t) \big]. 
\label{eq:spgdaupdt}
\end{align}
The SPGDA steps are summarized in Alg. \ref{alg:spgda}. Besides its simplicity and scalability, SPGDA enjoys convergence to a stationary point as elaborated next.

\subsection{Convergence of SPGDA}
\label{sec:convspgda}

\begin{algorithm}[t]
	{{\bfseries Input}: Initial guess $\bar{\bm{\theta}}^0$, step size sequence $\{\alpha_t, \eta_t >0 \}_{t=0}^{T}$, batch size $M$}
	
%	\SetKwInOut{Input}{Input}
%	\SetKwInOut{Output}{Output}
%	\SetKw{Set}{Set}	
%	\Input{Ini/\eta_t >0 \}_{t=0}^{T}$, batch size $M$}
	
	{For $t = 1, \ldots, T$}
		{Draw a batch of i.i.d samples $\{\bm{z}_m\}_{m=1}^M$} 
		\\
		{Find $\{\bm{\zeta}_m^t\}_{m=1}^{M}$ via gradient ascent:}
		{$\bm{\zeta}_m^t =  \bm{z}_m^t  + \eta_t \nabla_{\bm{\zeta}} \psi(\bar{\bm \theta}^t, \bm  \zeta;\bm{z}_m^t)\big|_{\bm{\zeta}=\bm{z}_m^t}, \quad m = 1, \ldots, M$}
		\newline 
		{Update:   $\bar{\bm{\theta}}^{t+1} = {\rm prox}_{\alpha_t r} \Big[ {\bar{\bm{\theta}}}^t-\frac{\alpha_t}{M}\sum_{m=1}^M \nabla_{\bar{\bm{\theta}}} \psi (\bar{\bm{\theta}}^t, {\bm \zeta}_m^t; \bm{z}_m^t)\big|_{\bar{\bm{\theta}}=\bar{\bm{\theta}}^t} \Big]$}	  
	\caption{SPGDA}
	\label{alg:spgda}
\end{algorithm}

To prove convergence of Alg. \ref{alg:spgda}, let us start by defining
\begin{align}
\bm g^\ast(\bar{\bm \theta}^t) := \frac{1}{M} \sum_{m=1}^{M} \nabla_{\bar{\bm{\theta}}} \bm \psi^\ast (\bar{\bm \theta}^{t}, \bm \zeta_m^\ast; \bm z_m^t). \label{eq:gastsgda}
\end{align}
Different from \eqref{eq:gsgda}, the gradient here is obtained
at the optimum $\bm{\zeta}_m^\ast$.~To establish convergence, one more assumption is needed.

\begin{assumption}
	Function $f$ satisfies the following conditions. 
	\begin{enumerate}
		\item[1)] Gradient estimates $\nabla_{\bar{\bm{\theta}}} \bm \psi^\ast (\bar{\bm \theta}^{t}, \bm \zeta_m^\ast; \bm z_m)$ at $\bm{\zeta}_m^\ast$~are unbiased and have bounded variance. That is, for $m = 1 \cdots M$, we have $\mathbb E \, [ \nabla_{\bar{\bm{\theta}}} \psi^\ast (\bm \theta, \bm \zeta_m^\ast; \bm z_m) - \nabla_{\bar{\bm{\theta}}} f(\bm \theta)] = \bm{0}$ and  $\mathbb E \, [\|\nabla_{\bar{\bm{\theta}}} \psi^\ast (\bm \theta, \bm \zeta_m^\ast; \bm z_m)-\nabla f(\bm \theta)\|^2] \le \sigma^2$. 
		\item[2)] The expected norm of $\bm{g}^t(\bar{\bm{\theta}})$ is bounded, that is, $\mathbb{E} \| \bm{g}^t(\bar{\bm{\theta}}) \|^2 \le B^2$. 
	\end{enumerate}
	\label{as:spgda}
\end{assumption}  
We now present a theorem on the convergence of Alg. \ref{alg:spgda}; see Appendix \ref{thm:spgda} for the proof.  
\begin{theorem}[Convergence of Alg. \ref{alg:spgda}]
	Let $\Delta_F := F(\bar{\bm{\theta}}^0) - \inf_{\bar{\bm{\theta}}}F(\bar{\bm{\theta}})$, and $D$ denote the diameter of the feasible set $\Theta$. 
	%$ L_{\bar{\bm \theta} \bm  z}^2   = L_{\bm \theta \bm  z}^2 + (\gamma_{\rm min}-L_{\bm{z z}}) L_c$. 
	% Assume $\mathbb{E} [\|\nabla f (\bar{\bm \theta}^t) - \bm g^\ast(\bar{\bm \theta}^t)\|_2^2] \le \sigma^2$. 
	Under As. \ref{as:cfunc}--\ref{as:losslipschitz} and \ref{as:spgda}, for a constant step size $\alpha > 0$, and a fixed batch size $M>0$,  after $T$ iterations, Alg. \ref{alg:spgda} satisfies 	
	\begin{align}
	\label{eq:convspgda}
	\mathbb E  \big[   {\rm dist}  (\bm 0, {\partial} F(\bar{\bm{\theta}}^T))^2 \big] 
	& \le \frac{\upsilon}{T+1} \Delta_F + \frac{4 \sigma^2}{M} +  \frac{2  L^2_{\bm {\theta z}} \nu}{M} \! \left[ \left(1-\alpha \mu \right) D^2 + \alpha^2 B^2 \right] 
	\end{align}
	where $\upsilon$, $\nu$, and $\mu = \gamma_{0} -L_{\bm{z z}}$ are some positive constants.
	\label{thm:convspgda}
\end{theorem}

Theorem \ref{thm:convspgda} implies that the sequence $\{\bar{\bm \theta}^t\}_{t=1}^{T}$ generated by Alg. \ref{alg:spgda}\, converges to a stationary point. The upper bound in \eqref{eq:convspgda} is characterized by a vanishing term induced by initial error $\Delta_F$, and constant bias terms.    

\section{Distributionally Robust Federated Learning}
\label{sec:distrib}

In practice, massive datasets are distributed geographically across multiple sites, where scalability, data privacy and integrity, as well as bandwidth scarcity typically discourage uploading them to a central server.~This has propelled the so-called federated learning framework, where multiple workers exchange information with a server to learn a centralized model using data locally generated and/or stored across workers~\cite{fedavg,fedlearn_challenges,federated_mag,chen2019joint}.~Workers in this learning framework communicate \textit{iteratively} with the server.~Albeit appealing for its scalability, one needs to carefully address the bandwidth bottleneck associated with server-worker links.~Furthermore, the workers' data may have (slightly) different underlying distributions, which further challenges the learning task.~To seek a model robust to distribution drifts across workers, we will adapt our novel SPGDA approach to design a privacy-respecting and robust algorithm. 

To that end, consider $K$ workers with each worker $k \in \mathcal{K}$ collecting samples $\{\bm{z}_n(k)\}_{n=1}^{N}$.~A globally shared model parameterized by $\bm{\theta}$ is to be updated at the server by aggregating gradients computed locally per worker.~For simplicity, we consider workers having the same number of samples $N$.~The goal is to learn a single global model from stored data at all workers by minimizing the following objective function
\begin{align}
\underset{\bm{\theta}\in\Theta
}{{\rm min}}\; \bar{\mathbb{E}}_{\bm{z}\sim \widehat{P}}\!\left[\ell(\bm{\theta};\bm{z}) 
\right] + r(\bm{\theta}) 
\end{align} 
where $\bar{\mathbb{E}}_{\bm{z}\sim \widehat{P}}\!\left[\ell(\bm{\theta};\bm{z}) 
\right] := \frac{1}{NK} \sum_{n=1}^N\sum_{k=1}^{K} \ell(\bm{\theta}, \bm{z}_n(k))$.~To endow the learned model with robustness against distributional uncertainties, our novel formulation will solve the following problem in a distributed fashion    
\begin{align}
& \underset{\bm{\theta}\in\Theta
}{{\rm min}}\;\sup_{P\in\mathcal{P}}\;\mathbb{E}_{\bm{z}\sim P}\!\left[\ell(\bm{\theta};\bm{z}) 
\right] + r(\bm{\theta})  \nonumber \\ 
& {\textrm{s. to.}}~ \mathcal{P}:= \Big\{P \Big| \sum_{k=1}^{K} W_c(P,\widehat{P}^{(N)}(k)) \le \rho \Big\}
\label{eq:minsupdistr}
\end{align} 
where $W_c(P,\widehat{P}^{(N)}(k))$ denotes the Wasserstein distance between distribution $P$ and the local $\widehat{P}^{(N)}(k)$, per worker $k$. 

Clearly, the constraint $P \in \mathcal{P}$, couples the optimization in \eqref{eq:minsupdistr} across all workers.~To offer distributed implementations, we resort to Proposition \ref{prop:strongdual}, to arrive at the equivalent reformulation 
\begin{align}
\label{eq:robustdualdistr}
\min_{\bm{\theta}\in \Theta} \, \inf_{\gamma \in \Gamma}   \sum_{k=1}^{K}   \bar{\mathbb{E}}_{\bm z(k) \sim \widehat{P}^{(N)}(k)}   \big[\sup_{\bm \zeta \in {\mathcal Z}} \!\left\{ \ell(\bm \theta; \bm \zeta)  \qquad \qquad +    \gamma (\rho - c(\bm z(k), \bm \zeta))\right\} \big] + r(\bm{\theta}). 
\end{align}

Next, we present our communication- and computation-efficient DRFL that builds on the SPGDA scheme in Sec. \ref{sec:spgda}.

\begin{algorithm}[t]
	{{\bfseries Input}: Initial guess $\bar{\bm{\theta}}^1$, a set of workers $\mathcal{K}$ with data samples $\{\bm{z}_n(k)\}_{n=1}^{N}$ per worker $k \in \mathcal{K}$, step size sequence $\{\alpha_t, \eta_t > 0 \}_{t=1}^{T}$}
	
{{\bfseries Output}: $\bar{\bm{\theta}}^{T+1}$} 

%	\SetKwInOut{Input}{Input}
%	\SetKwInOut{Output}{Output}
%	\SetKw{Set}{Set}	
%	\Input{Initial guess $\bm{\bar{\theta}}^1$, a set of workers $\mathcal{K}$ with data samples $\{\bm{z}_n(k)\}_{n=1}^{N}$ per worker $k \in \mathcal{K}$, step size sequence $\{\alpha_t, \eta_t > 0 \}_{t=1}^{T}$ 
%		
%		\Output{$\bm{\bar{\theta}}^{T+1}$}	
%	}
	
	For{$t = 1, \ldots, T$}
	{
		\newline
		{{\bf Each worker}:
			
			{Samples a minibatch $\mathcal{B}^t(k)$ of samples} 
			
			{Given $\bar{\bm{\theta}}^t$ and $\bm{z} \in \mathcal{B}^t(k)$, forms local perturbed loss \[\psi_k({\bar{\bm{\theta}}}^t, \bm \zeta; \bm{z}) :=  \ell(\bar{\bm{\theta}}^t; \bm \zeta) + \gamma^t (\rho - c(\bm z, \bm \zeta))\]}
			{Lazily maximizes $\psi_k({\bar{\bm{\theta}}}^t, \bm \zeta; \bm{z})$ over $\bm{\zeta}$ to find \[{\bm{\zeta}}(\bar{\bm{\theta}}^t; \bm{z}) = \bm{z} + \eta_t \nabla_{\bm{\zeta}} \psi_k (\bar{\bm{\theta}}^t, \bm{\zeta}; \bm{z})|_{\bm{\zeta}=\bm{z}}\]}
			{Computes stochastic gradient  \[\frac{1}{|\mathcal{B}^t(k)|} \sum_{\bm{z}  \in \mathcal{B}^t(k)} \!\!\! \nabla_{\bar{\bm \theta}}\psi_k (\bar{\bm{\theta}}^t, \bm{\zeta}(\bar{\bm{\theta}}^t; \bm{z}); \bm{z})\big|_{\bar{\bm{\theta}}=\bar{\bm{\theta}}^t}\] 
				and uploads to server
			}
		}

		{\bf Server}:
		
		{Updates $\bar{\bm{\theta}}^t$ according to \eqref{eq:tildthetadistr}}
		
		{Broadcasts $\bar{\bm{\theta}}^{t+1}$ to workers}
	}
	
	\caption{DRFL}
	\label{alg:drfl}
\end{algorithm}

Specifically, our DRFL hinges on the fact that with fixed server parameters $\bar{\bm{\theta}}^t:=[{\bm{\theta}^t}^\top, \gamma^t]^\top$ per iteration $t$, the optimization problem becomes \textit{separable} across all workers. Hence, upon receiving $\bar{\bm{\theta}}^t$ from the server, each worker $k \in \mathcal{K}$: i) samples a minibatch $\mathcal{B}^t(k)$ of data from $\widehat{P}^{(N)}(k)$; ii) forms the \textit{perturbed} loss $\psi_k({\bar{\bm{\theta}}}^t, \bm \zeta; \bm{z}) :=  \ell(\bm{\theta}^t; \bm \zeta) + \gamma^t (\rho - c(\bm z, \bm \zeta))$ for each $\bm{z} \in \mathcal{B}^t(k)$; iii) lazily maximizes $\psi_k({\bar{\bm{\theta}}}^t, \bm \zeta; \bm{z})$ over $\bm{\zeta}$ using a single gradient ascent step to yield ${\bm{\zeta}}(\bar{\bm{\theta}}^t; \bm{z}) = \bm{z} + \eta_t \nabla_{\bm{\zeta}} \psi_k (\bar{\bm{\theta}}^t, \bm{\zeta}; \bm{z})|_{\bm{\zeta}=\bm{z}}$;
%, that is   $\bm{\zeta}(\bar{\bm{\theta}}^t; \bm{z}^k_n)$ per datum $\bm{z}^k_n$ (using $\epsilon$-accurate oracle or a gradient ascent step), 
and, iv) sends the stochastic gradient $|\mathcal{B}^t(k)|^{-1}\sum_{\bm{z} \in \mathcal{B}^t(k)} \nabla_{\bar{\bm \theta}}\psi_k (\bar{\bm{\theta}}^t, \bm{\zeta}(\bar{\bm{\theta}}^t; \bm{z}); \bm{z})\big|_{\bar{\bm{\theta}}=\bar{\bm{\theta}}^t}$ back to the server. Upon receiving all local gradients, the server updates $\bar{\bm{\theta}}^t$ using a proximal gradient descent step to find $\bar{\bm{\theta}}^{t+1}$, that is
\begin{align}
\bar{\bm{\theta}}^{t+1}    =      {\rm prox}_{\alpha_t r}    \left[ \bar{\bm{\theta}}^t   -  \frac{\alpha_t}{K}   \sum_{k=1}^{K} \frac{1}{|\mathcal{B}^t(k)|} 
\times   \sum_{\bm{z} \in \mathcal{B}^t(k)}   \nabla_{\bar{\bm{\theta}}} \psi_k (\bar{\bm{\theta}}^t, \bm{\zeta}(\bar{\bm{\theta}}^t; \bm{z}); \bm{z})\big|_{\bar{\bm{\theta}}=\bar{\bm{\theta}}^t} \right]
\label{eq:tildthetadistr}
\end{align}
which is then broadcast to all workers to begin a new round of local updates.~Our DRFL approach is tabulated in Alg. \ref{alg:drfl}.

\begin{figure}[t]
	\centering
	\begin{subfigure} 
		\centering
		\includegraphics[width= .31 \textwidth]{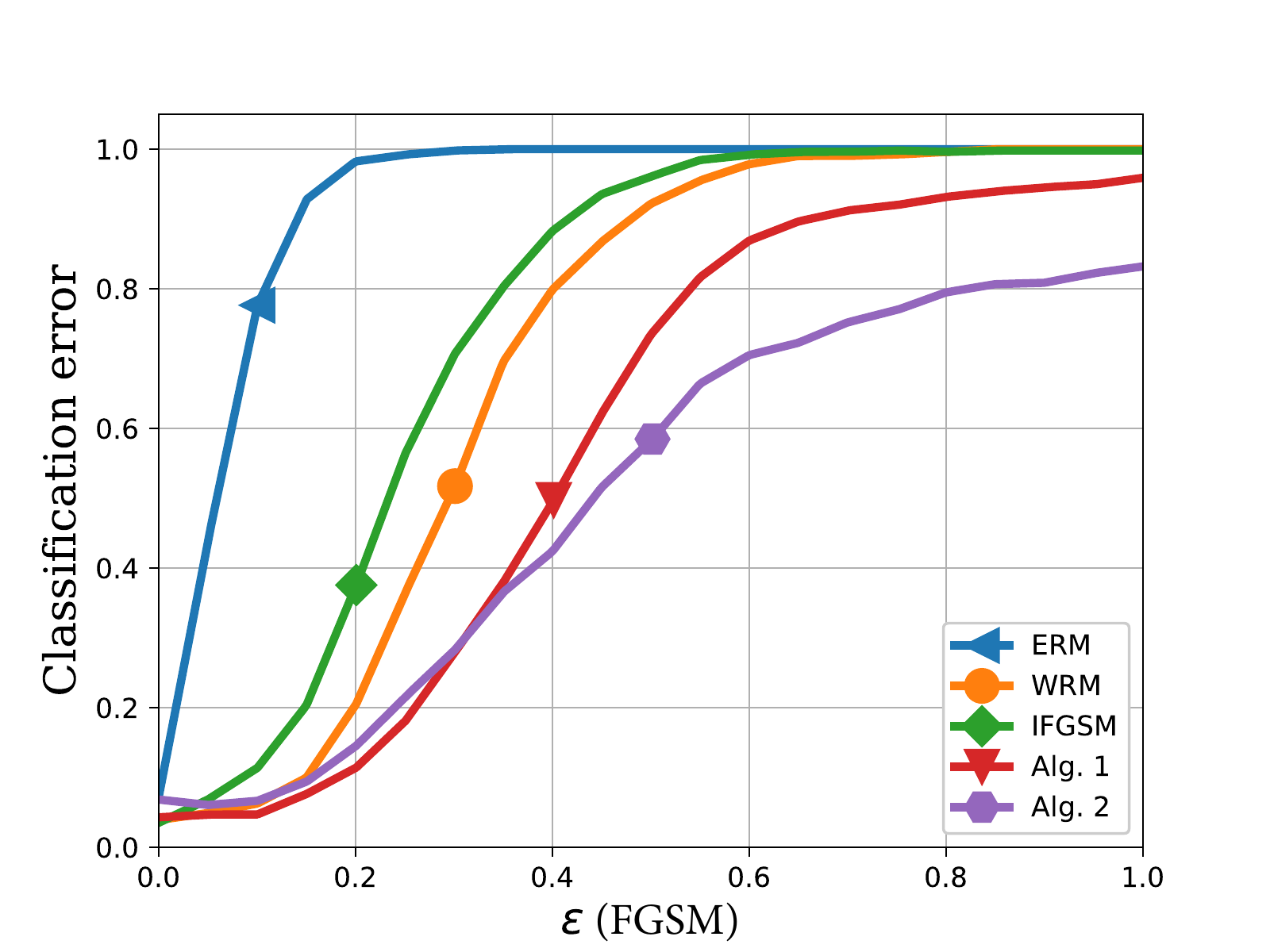}
	\end{subfigure}%
	~ 
	\begin{subfigure} 
		\centering
		\includegraphics[width= .31 \textwidth]{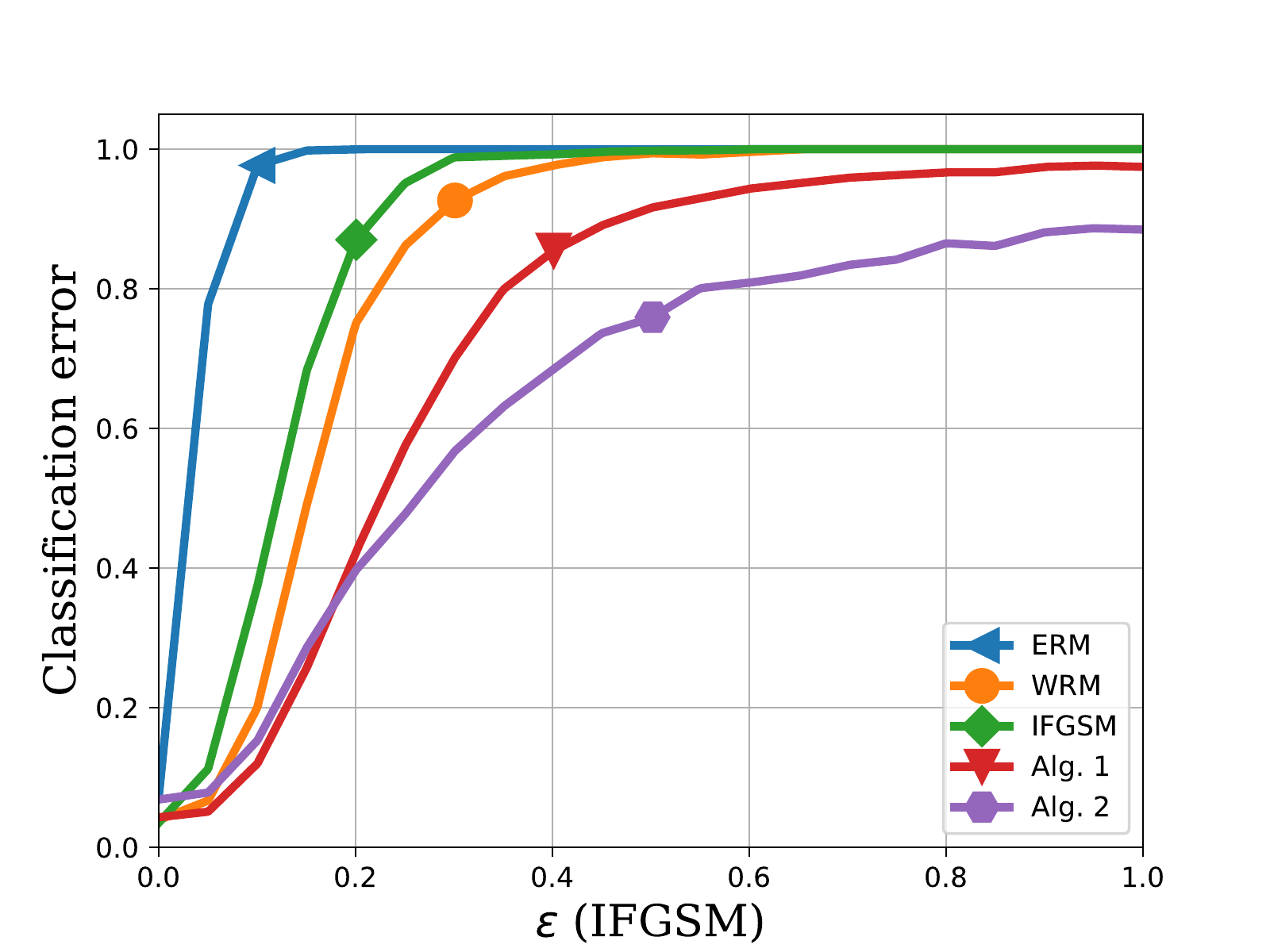}
	\end{subfigure}%
	~
	\begin{subfigure} 
		\centering
		\includegraphics[width= .31 \textwidth]{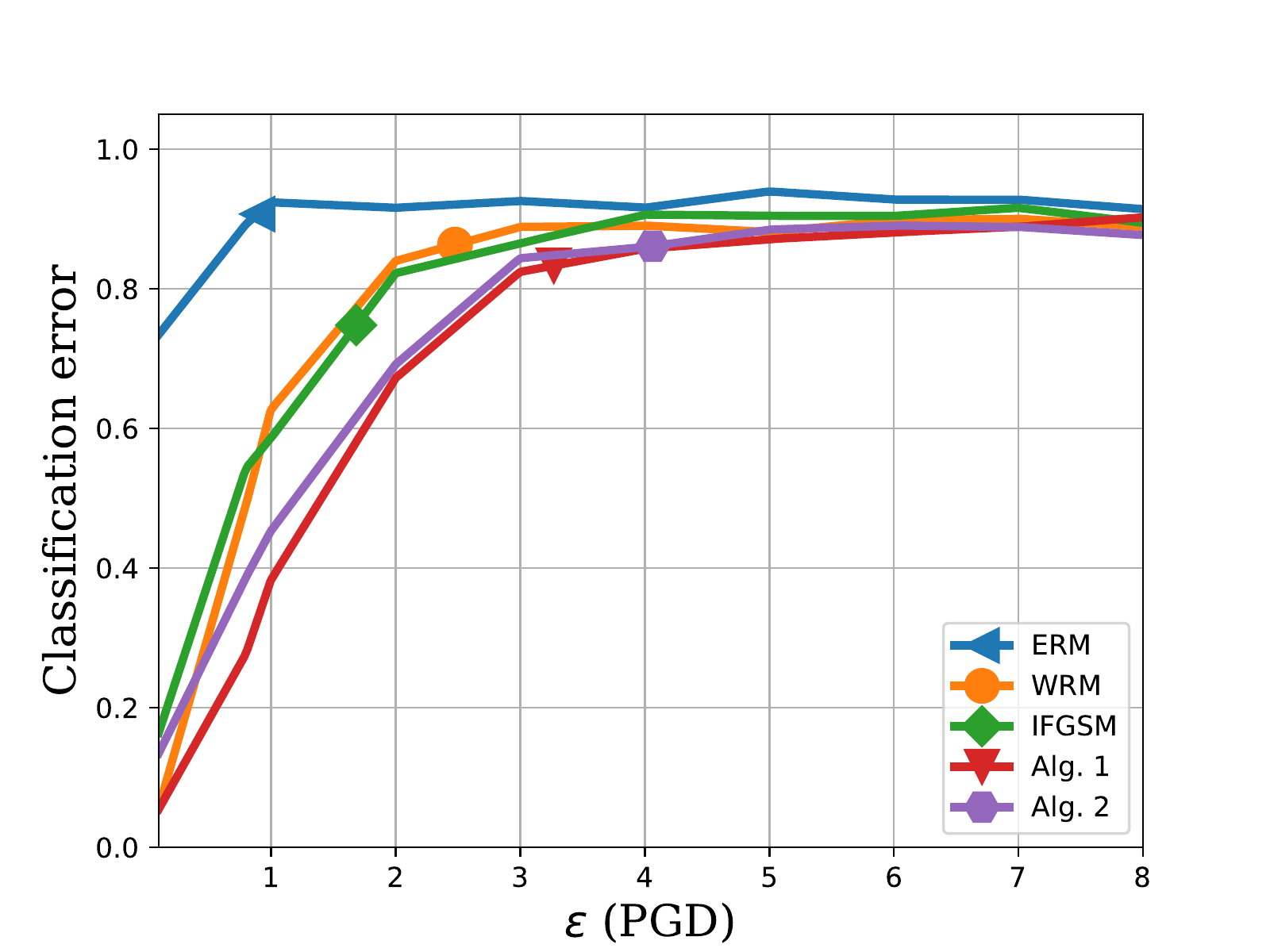}
	\end{subfigure}%
	\caption{Misclassification error rate for different training methods using MNIST dataset; Left: FGSM attack, Middle: IFGSM attack, Right: PGD attack} 
	\label{fig:mnist}
\end{figure}

\begin{figure}[t]
	\centering
	\begin{subfigure} 
		\centering
		\includegraphics[width= .32 \textwidth]{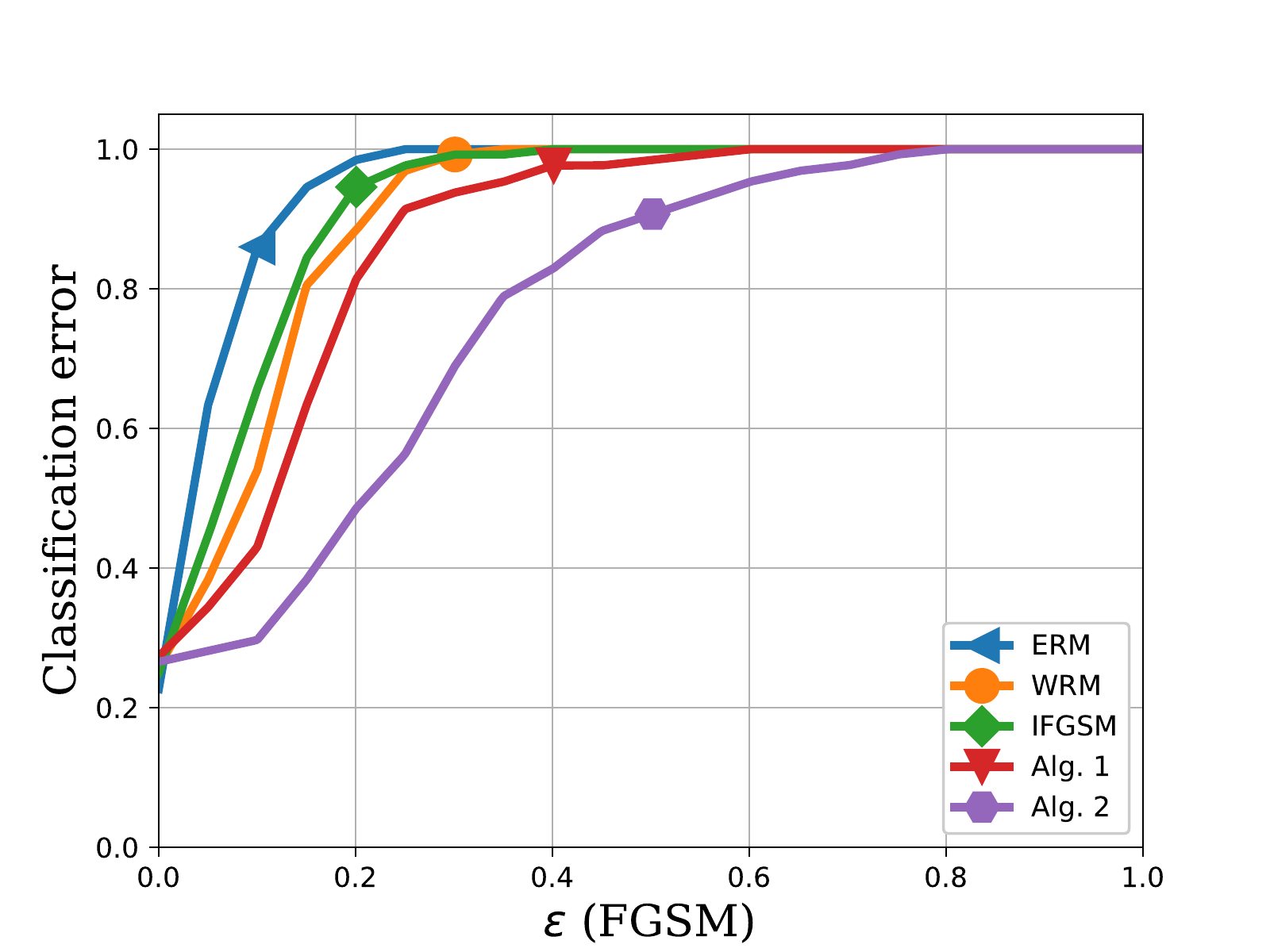}
	\end{subfigure}
	\begin{subfigure} 
		\centering
		\includegraphics[width= .32 \textwidth]{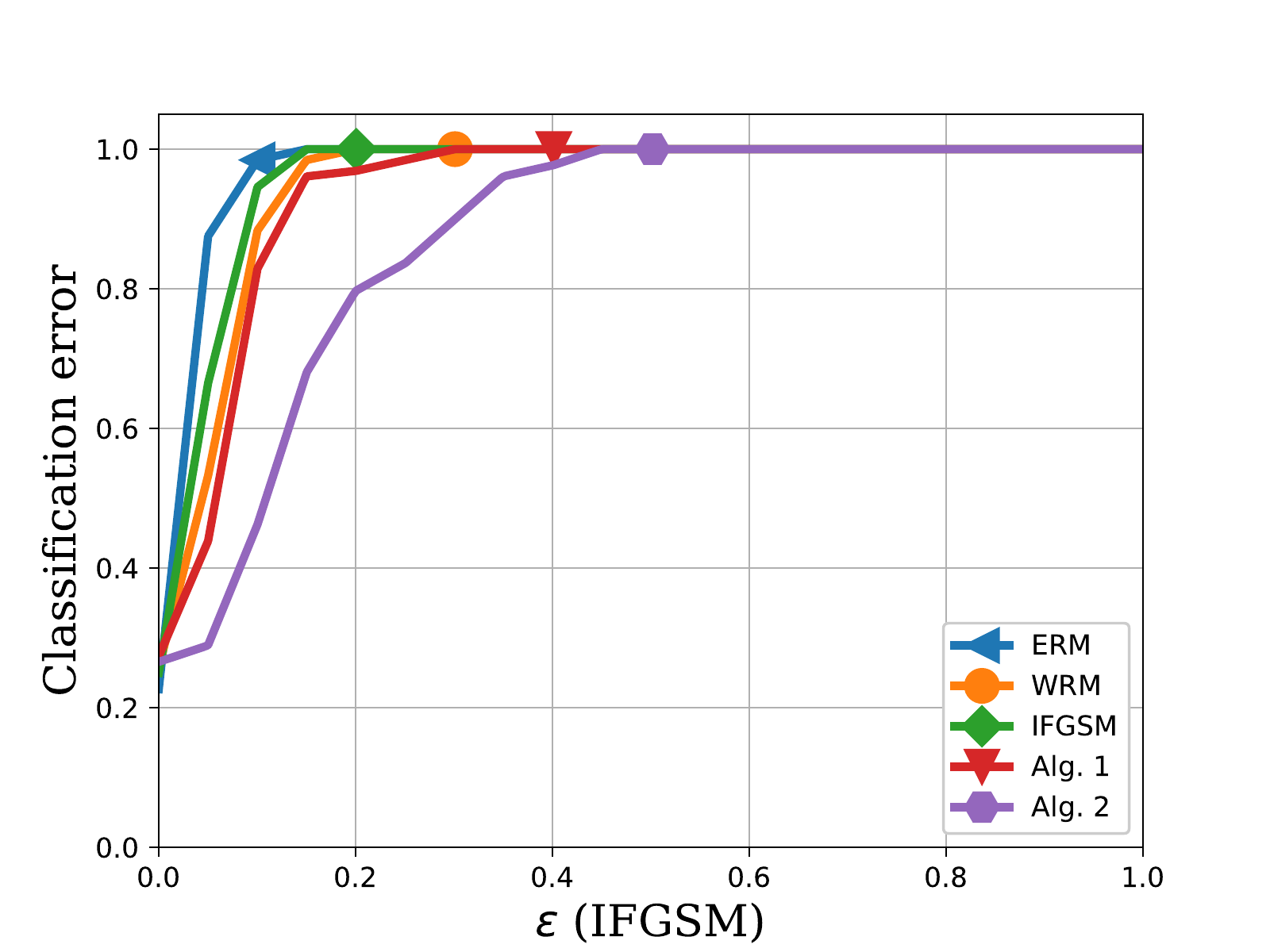}
	\end{subfigure}
	~
	\begin{subfigure} 
		\centering
		\includegraphics[width= .32 \textwidth]{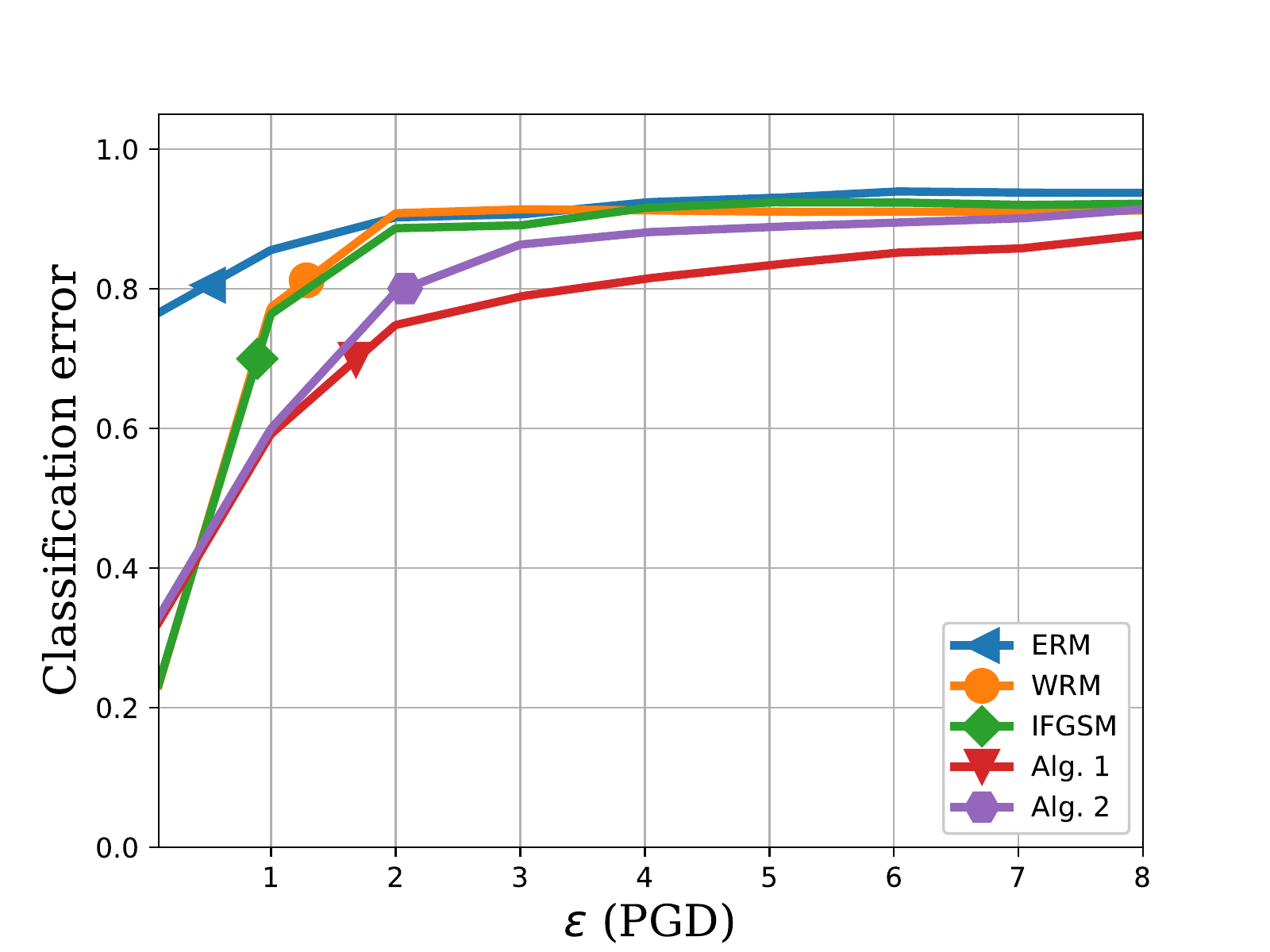}
	\end{subfigure}
	\caption{Misclassification error rate for  different training methods using F-MNIST dataset; Left: FGSM attack, Middle: IFGSM attack, Right: PGD attack}
	\label{fig:fmnist}
\end{figure}

\section{Numerical Tests}
\label{sec:experiments}
To assess the performance in the presence of distribution drifts and adversarial perturbations, we will rely on empirical classification of standard MNIST and Fashion- (F-)MNIST datasets.~Specifically, we compare performance using models trained with empirical risk minimization (ERM), the fast-gradient method (FGSM) \cite{fellow2014adv}, its iterated variant (IFGM) \cite{ifgm2016adversarial}, and the Wasserstein robust method (WRM) \cite{sinha2017certify}. We further evaluate the testing performance using the projected gradient descent (PGD) attack \cite{madry2017towards}. We first test the performance of SPGD with $\epsilon$-accurate oracle, and the SPGDA algorithm on standard classification tasks.

\subsection{SPGD with $\epsilon$-accurate oracle and SPGDA}  
The FGSM attack performs one step gradient update along the direction of the gradient's sign to find an adversarial sample; that is,
\begin{equation}
\label{eq:fgsm}
\bm{x}_{\rm adv} = {\rm Clip}_{[-1,1]}\{\bm{x} + \epsilon_{\rm adv} {\rm sign} (\nabla \ell_{\bm{x}} (\bm{\theta}; (\bm{x}, y)))\}
\end{equation}   
where $\epsilon_{\rm avd}$ controls the maximum $\ell_{\infty}$ perturbation of adversarial samples. The element-wise ${\rm Clip}_{[a, b]}\{\}$ operator forces its input to reside in the prescribed range $[-1, 1]$. By running $T_{\rm adv}$ iterations of \eqref{eq:fgsm} iterative (I) FGSM attack samples are generated \cite{fellow2014adv}.~Starting with an initialization $\bm{x}^{0}_{\rm adv}=\bm{x}$, and considering the $\ell_{\infty}$ norm, the PGD attack iterates \cite{madry2017towards}
\begin{equation}
\label{eq:ifgsm}
\bm{x}^{t+1}_{\rm adv} =  \Pi_{\mathcal{B}_{\epsilon} (\bm{x}_{\rm adv}^t)} \Big\{\bm{x}^{t}_{\rm adv} + \alpha {\rm sign} (\nabla \ell_{\bm{x}} (\bm{\theta}; (\bm{x}^{t}_{\rm adv}, y)))\Big\}
\end{equation} 
for $T_{\rm adv}$ steps, where $\Pi$ denotes projection onto the ball $\mathcal{B}_{\epsilon}(\bm{x}_{\rm adv}^t) := \{\bm{x}: \|\bm{x} - \bm{x}_{\rm adv}^t \|_\infty \le \epsilon_{\rm adv} \}$, and $\alpha>0$ is the stepsize set to $1$ in our experiments.~We use $T_{\rm adv} = 10$
iterations for all iterative methods both in training and attack samples. The PGD can also be interpreted as an iterative algorithm that solves the optimization problem $ \max_{\bm{x'}} \ell(\bm{\theta};(\bm{x}',y))$ subject to $\| \bm{x}' - \bm{x} \|_{\ell_{\infty}} \le \alpha$.~The Wasserstein attack on the other hand, generates adversarial samples by solving a perturbed training loss with an $\ell_2$-based transportation cost associated with the Wasserstein distance between the training and adversarial data distributions~\cite{sinha2017certify}.

For the MNIST and F-MNIST datasets, a convolutional neural network (CNN) classifier consisting of $8 \times 8$, $6 \times 6$, and $5 \times 5$ filter layers with rectified linear units (ReLU) and the same padding, is used.~Its first, second, and third layers have $64$, $128$, and $128$ channels, respectively, followed by a fully connected layer, and a softmax layer at the output. 

CNNs with the same architecture are trained, using different adversarial samples.~Specifically, to train a Wasserstein robust CNN model (WRM), $\gamma=1$ was used to generate Wasserstein adversarial samples, $\epsilon_{\rm adv}$ was set to $0.1$
%$0.1 \, \mathbb{E} \, \|\bm{x}\|_{\infty} = 0.1$ 
for the other two methods, and $\rho=25$ was used to define the uncertainty set for both Algs. \ref{alg:spgd} and \ref{alg:spgda}.~Unless otherwise noted, we set the batch size to $128$, the number of epochs to $30$, the learning rates to $\alpha=0.001$ and $\eta=0.02$, and used the Adam optimizer \cite{adam}.~Fig.~\ref{fig:mnist}(Left) shows the classification error on the MNIST dataset.~The error rates were obtained using testing samples generated according to the FGSM method with $\epsilon_{\rm adv}$.~Clearly all training methods outperform ERM, and our proposed Algs. \ref{alg:spgd} and \ref{alg:spgda}  offer improved performance over competing alternatives.~The testing accuracy of all methods using samples generated according to an IFGSM attack is presented in  Fig.~\ref{fig:mnist}(Middle).~Likewise, Algs.~\ref{alg:spgd} and \ref{alg:spgda} outperform other methods in this case.~Fig.~\ref{fig:mnist}(Right) depicts the testing accuracy of the considered methods under different levels of a PGD attack.~The plots in Fig.~\ref{fig:mnist} showcase the improved performance obtained by CNNs trained using Algs.~\ref{alg:spgd} and \ref{alg:spgda}.

%%%%%%%%%%%%%%%%%%%%%%%
%%%%%%%%FIGS%%%%%%%%%%%
\begin{figure}[t]
	\centering
	\begin{subfigure}
		\centering
		\includegraphics[width= .32\textwidth]{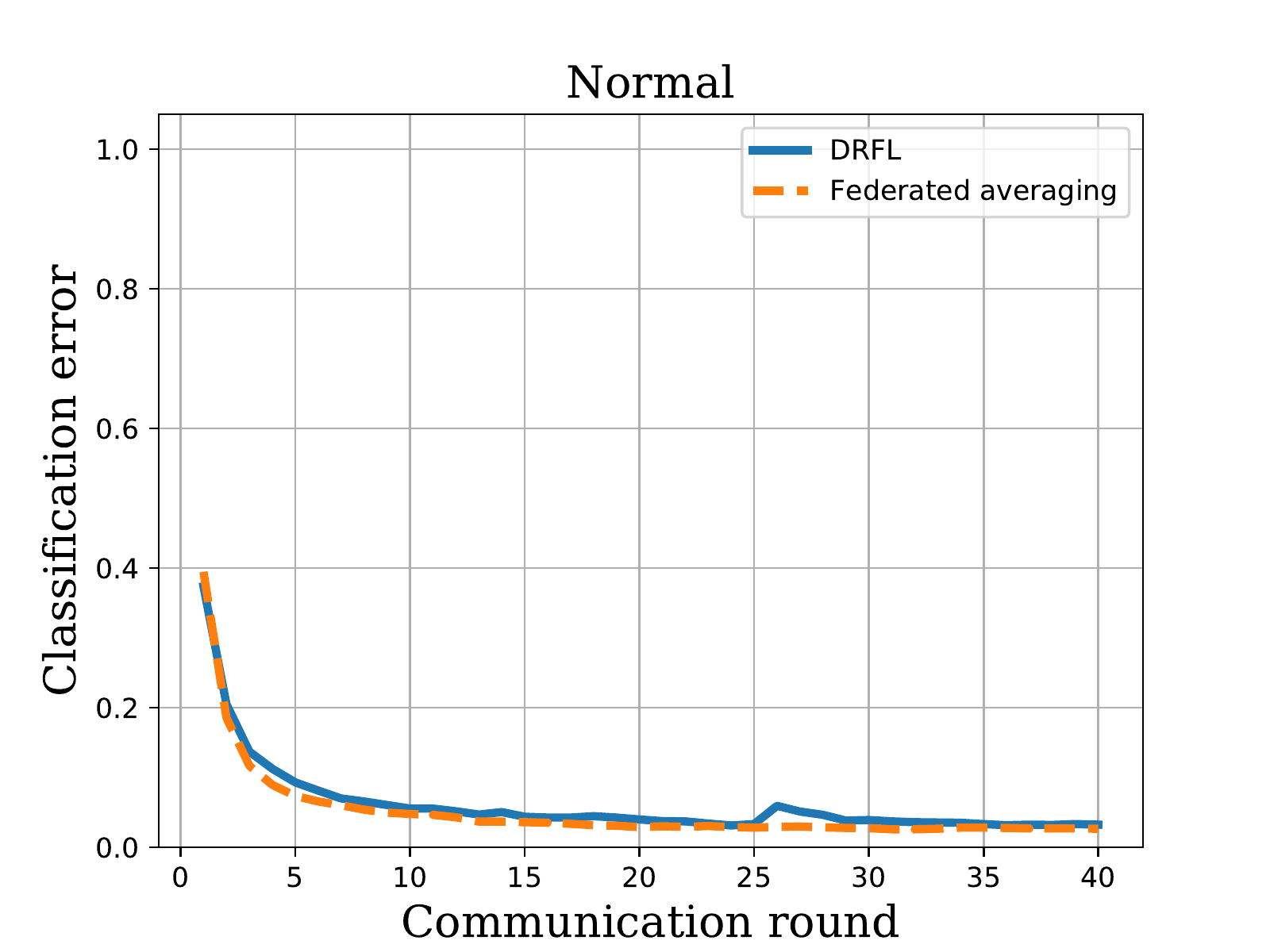}
	\end{subfigure}
	\hspace{-12pt}
	\begin{subfigure}
		\centering
		\includegraphics[width= .32 \textwidth]{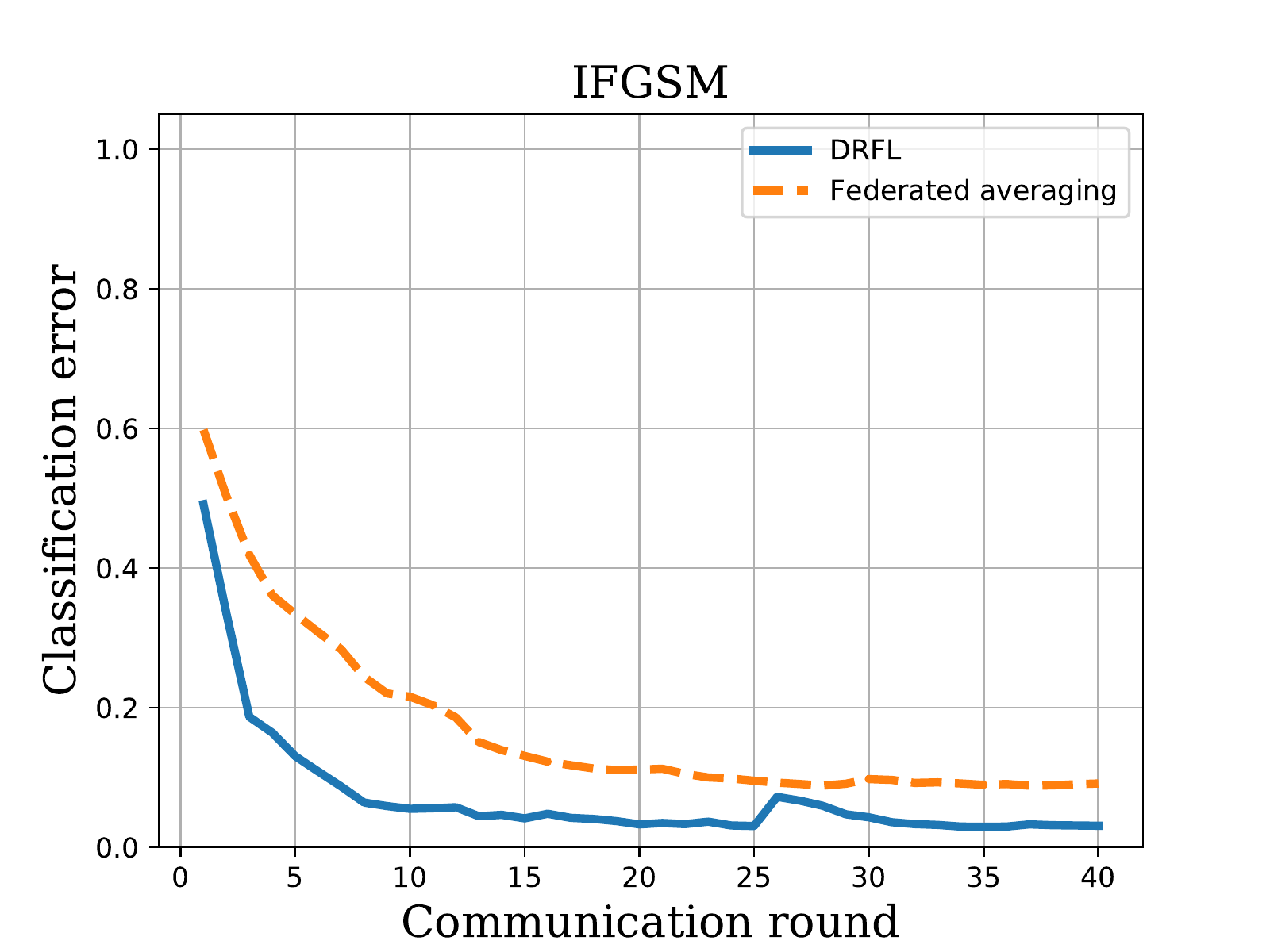}
	\end{subfigure}
	\hspace{-12pt}
	\begin{subfigure}
		\centering
		\includegraphics[width= .32 \textwidth]{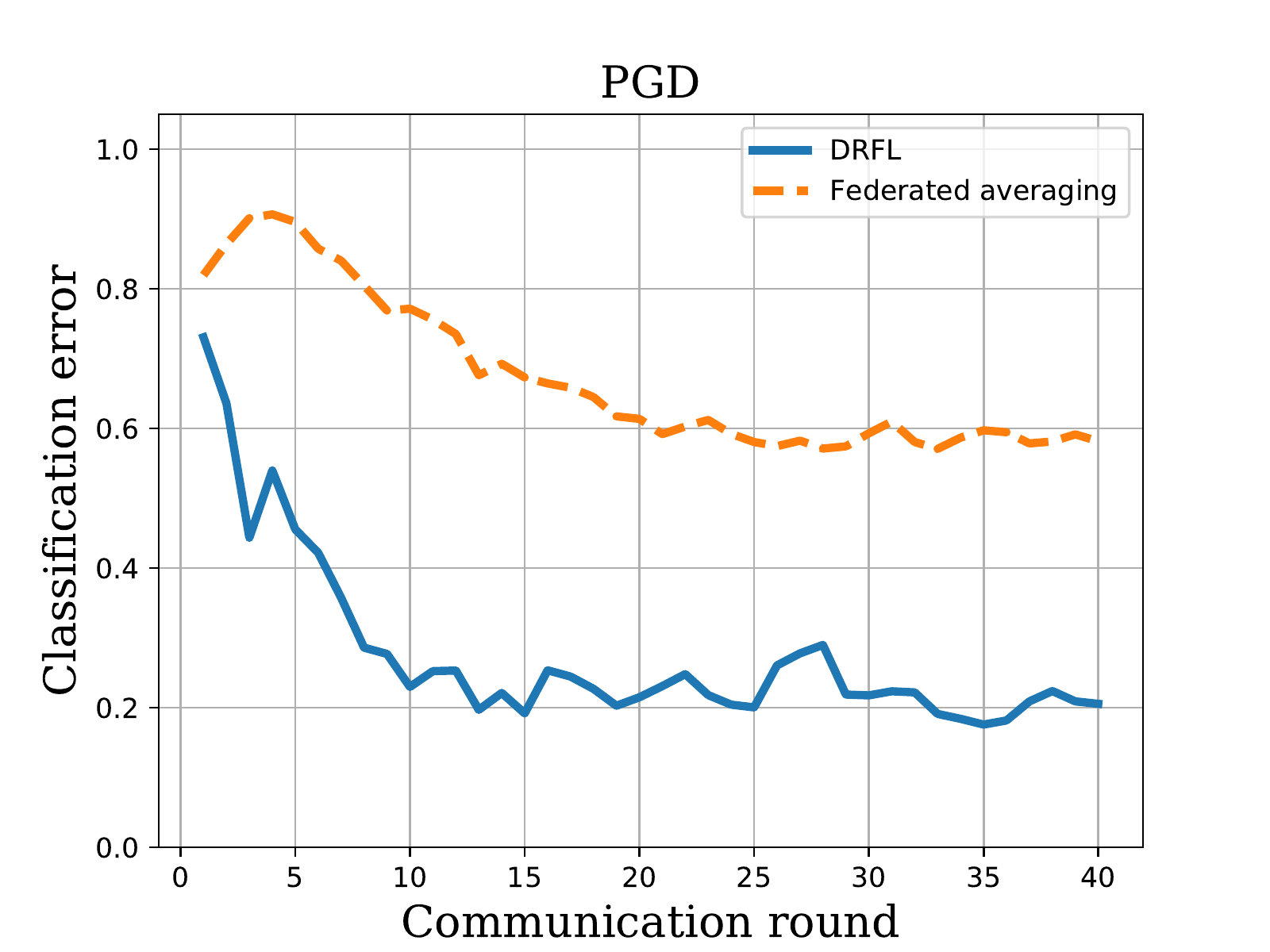}
	\end{subfigure}
	\caption{Distributionally robust federated learning for image classification using the non-i.i.d. F-MNIST dataset; Left: No attack, Middle: IFGSM attack, Right: PGD attack}
	\label{fig:fedfmnist_biased}
\end{figure}
\begin{figure}[t]
	\centering
	\begin{subfigure}
		\centering
		\includegraphics[width= .31 \textwidth]{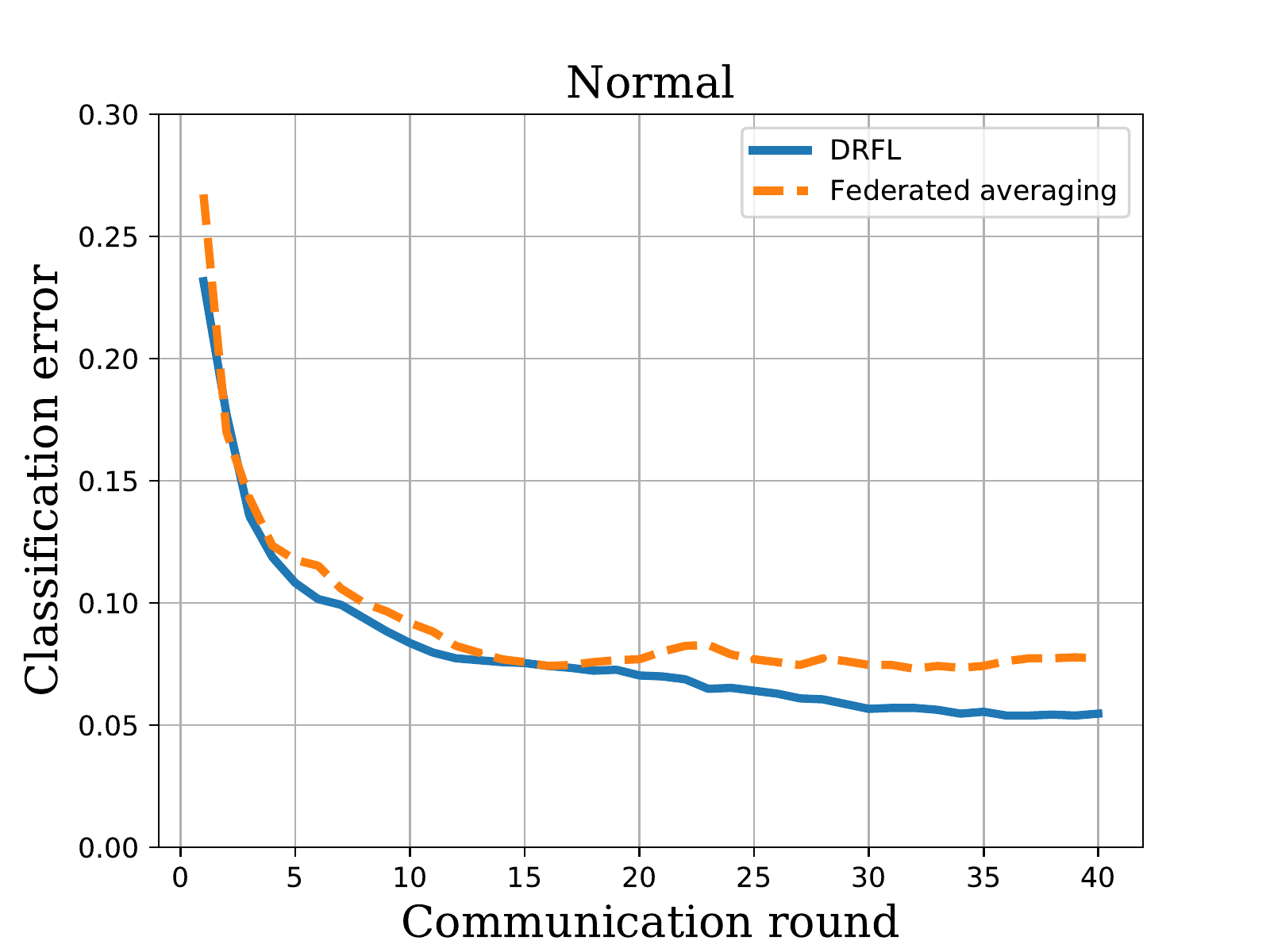}
	\end{subfigure}%
	\begin{subfigure} 
		\centering
		\includegraphics[width= .31 \textwidth]{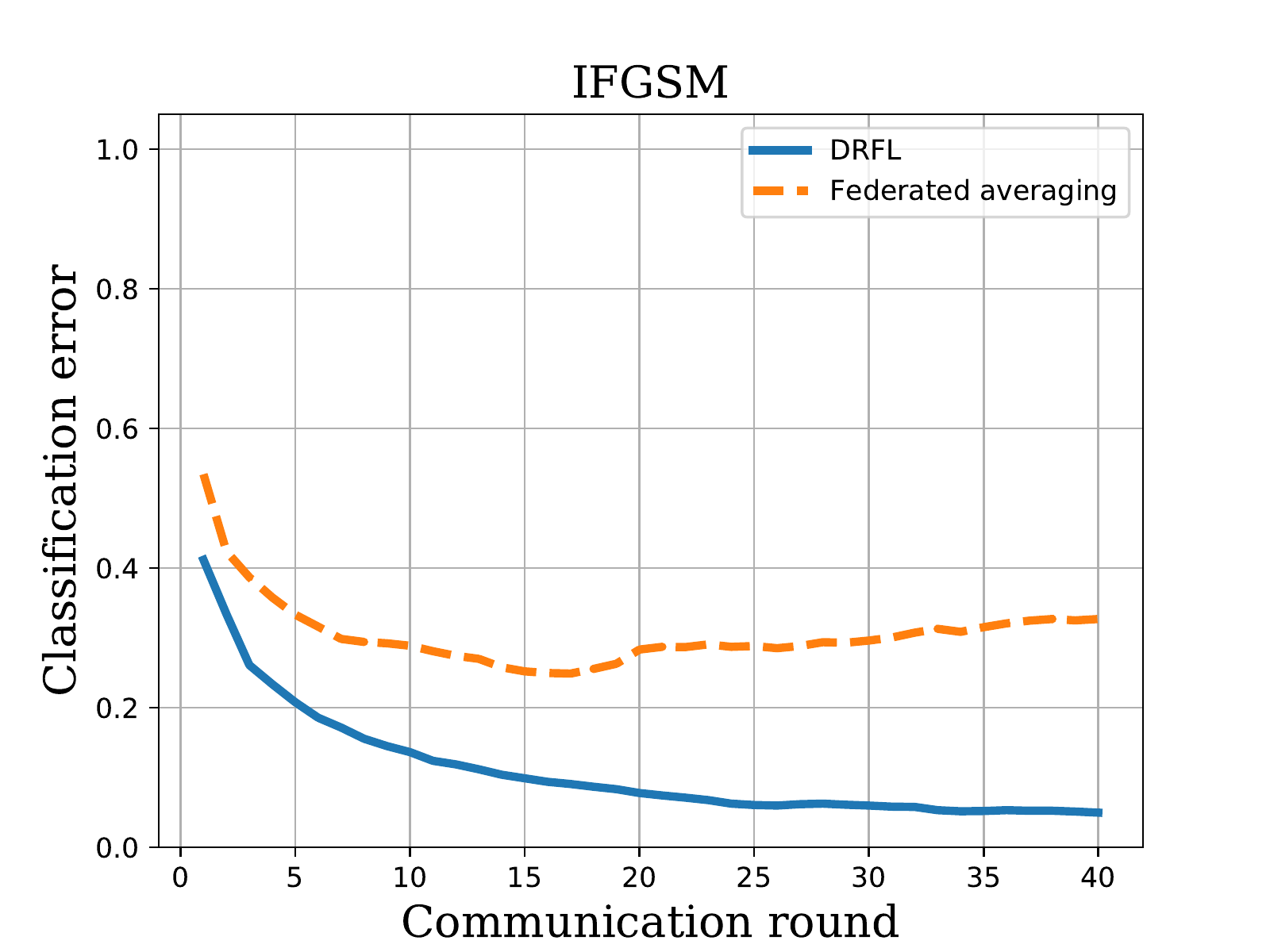}
	\end{subfigure}
	\begin{subfigure}[t] 
		\centering
		\includegraphics[width= .31 \textwidth]{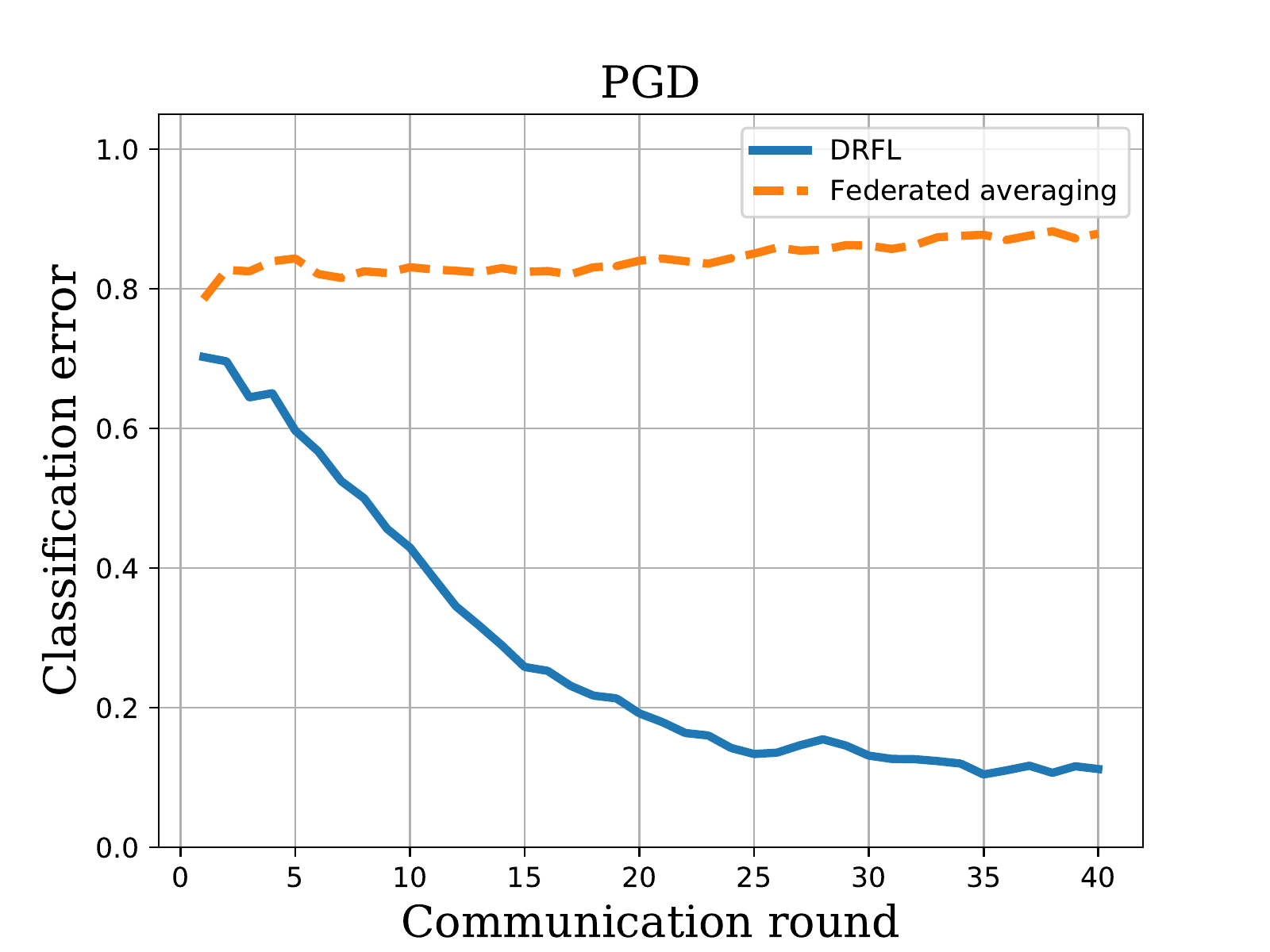}
	\end{subfigure}
	\caption{Federated learning for image classification using the MNIST dataset; Left: No attack, Middle: IFGSM attack, Right: PGD attack}
	\label{fig:fedmnist}
\end{figure}

\begin{figure}[t]
	\centering
	\begin{subfigure}
		\centering
		\includegraphics[width= 0.32 \textwidth]{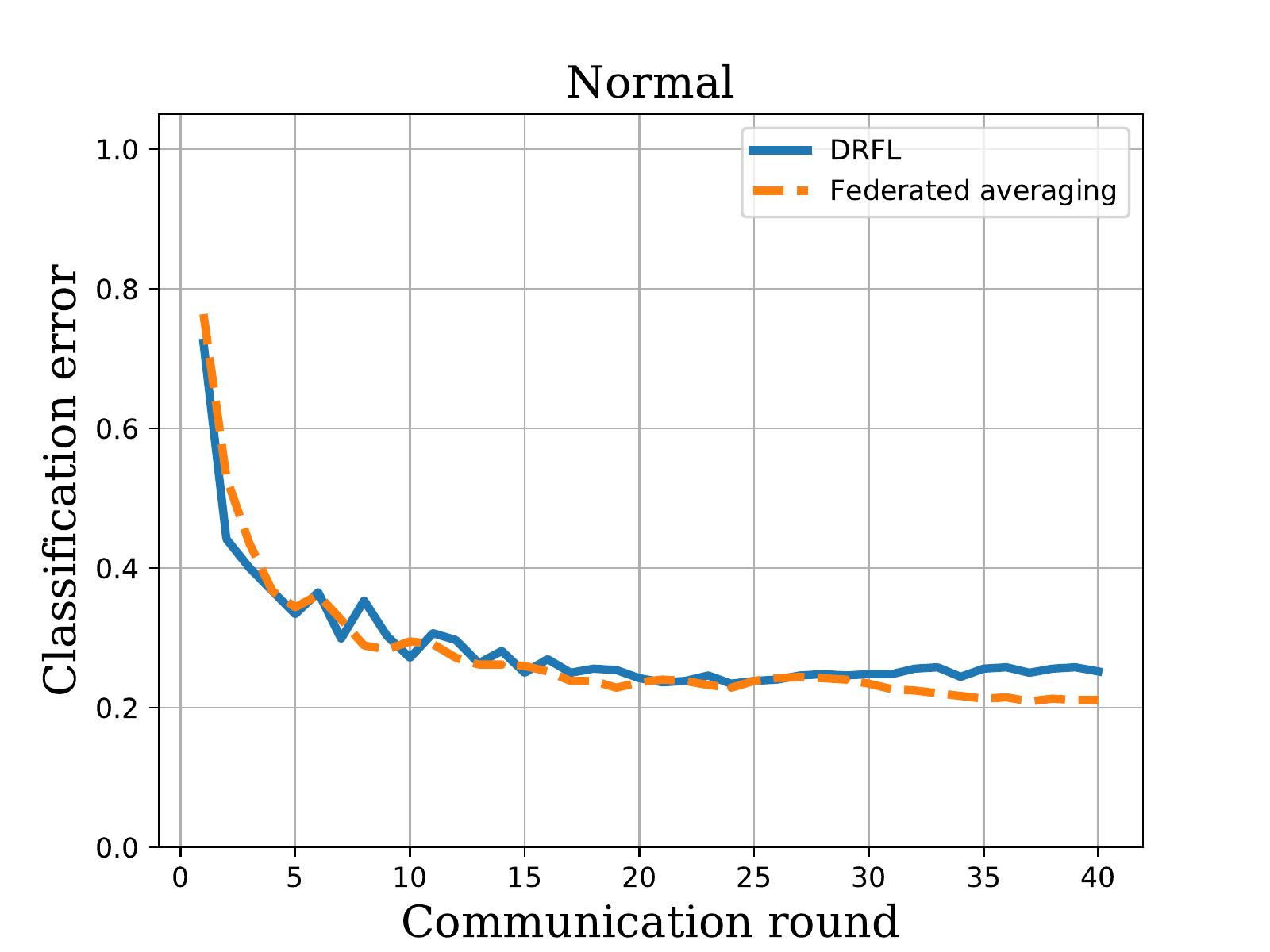}
	\end{subfigure}%
	\begin{subfigure} 
		\centering
		\includegraphics[width= 0.32 \textwidth]{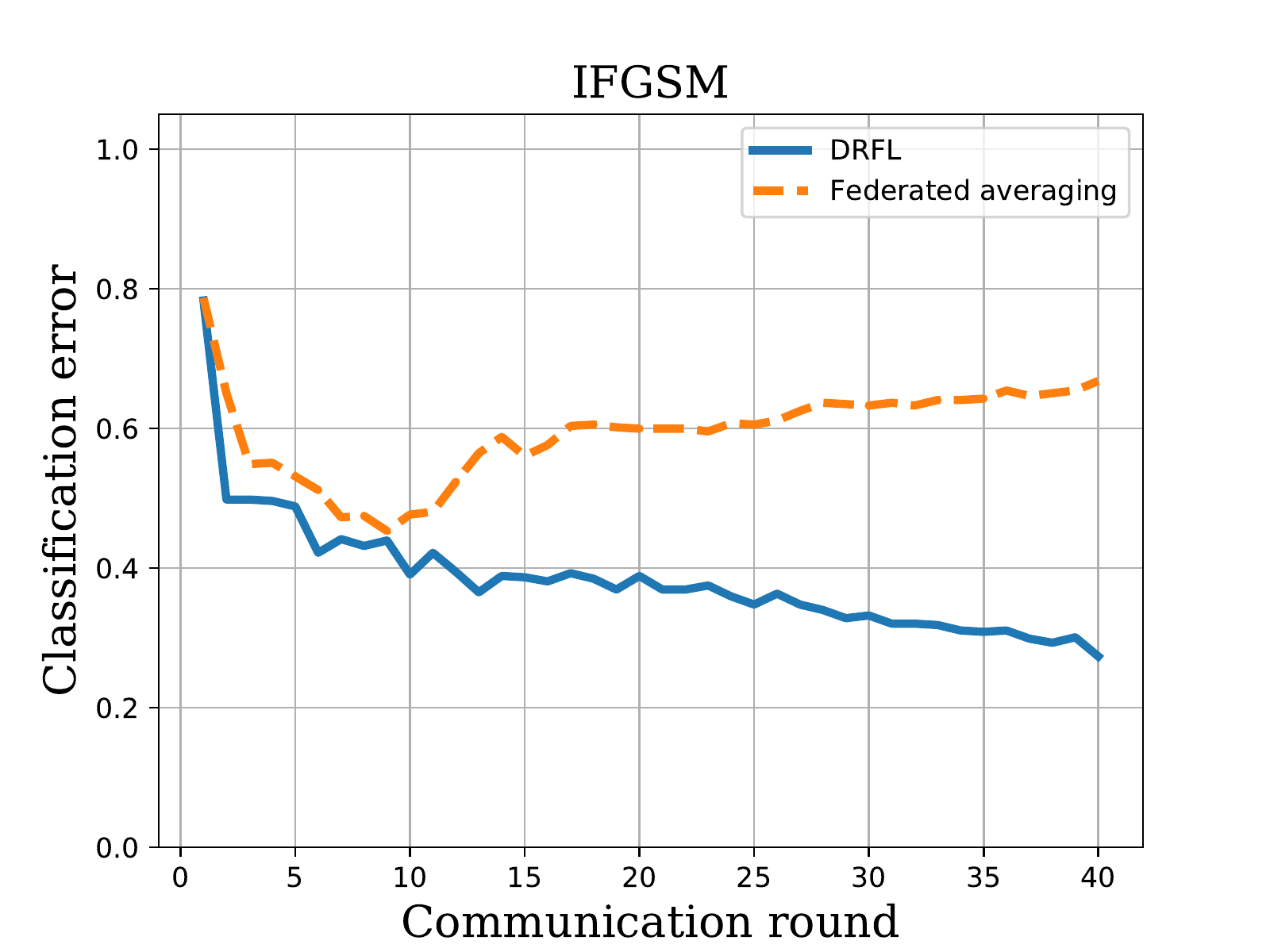}
	\end{subfigure}
	\begin{subfigure} 
		\centering
		\includegraphics[width= 0.32  \textwidth]{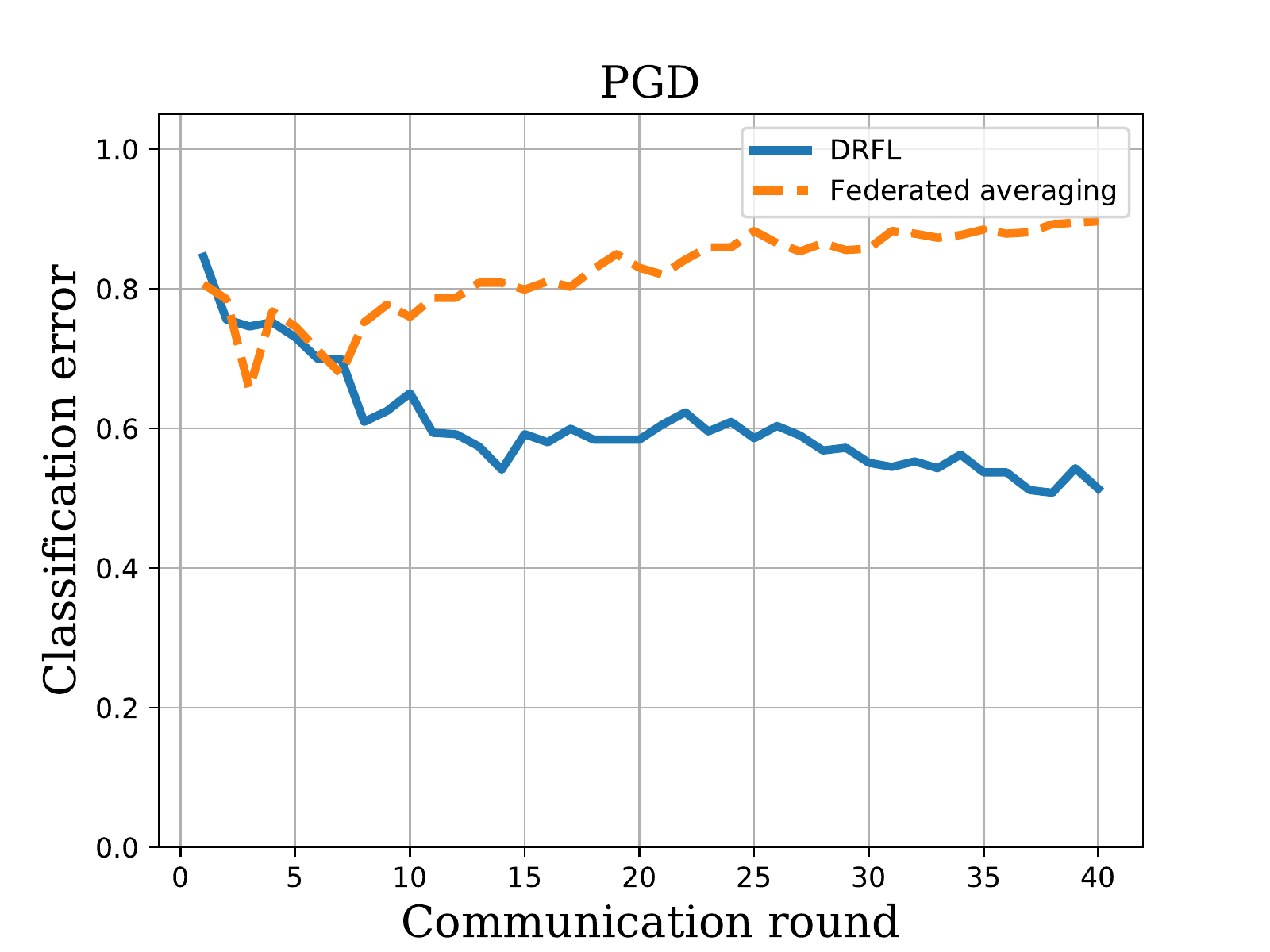}
	\end{subfigure}
	\caption{Distributionally robust federated learning for image classification using F-MNIST dataset; Left: No attack, Middle: IFGSM attack, Right: PGD attack}
	\label{fig:fedfmnist}
\end{figure}

The F-MNIST article image dataset is used in our second experiment. Similar to MNIST dataset, each example in F-MNIST is also a $28 \times 28$ gray-scale image, associated with a label from $10$ classes.~F-MNIST is a modern replacement for the original MNIST dataset for benchmarking machine learning algorithms.~Using CNNs with the same architectures as before, the classification error is depicted for different training methods in Fig.~\ref{fig:fmnist}.  Three different attacks, namely FGSM, IFGSM, and PGD are used during testing. The proposed SPGD and SPGDA algorithms outperform the other methods, verifying the superiority of Algs. \ref{alg:spgd} and \ref{alg:spgda} in terms of yielding robust models.

\subsection{Distributionally robust federated learning}
\label{sec:result_distr}
To validate the performance of our DRFL algorithm, we considered an FL environment consisting of a server and $10$ workers, with local batch size $64$, and assigned to every worker an equal-sized subset of training data containing i.i.d. samples from $10$ different classes.~All workers participated in each communication round.~To benchmark the DRFL, we simulated the federated averaging method \cite{fedavg}.~The testing accuracy on the MNIST dataset per communication round using clean (normal) images is depicted in Fig.~\ref{fig:fedmnist}.~Clearly, both DRFL and federated averaging algorithms exhibit reasonable performance when the data is not corrupted.~The performance is further tested against IFGSM and PGD attacks with a fixed $\epsilon_{\rm adv}=0.1$ during each communication round, and the corresponding misclassification error rates are shown in Figs.~\ref{fig:fedmnist}(Middle) and \ref{fig:fedmnist}(Right), respectively.~The classification performance using federated averaging does not improve in Fig. \ref{fig:fedmnist}(Middle), whereas the DRFL performance keeps improving across communication rounds.~This is a direct consequence of accounting for the data uncertainties during the learning process.~Moreover, Fig.~\ref{fig:fedmnist}(Right) showcases that the federated averaging becomes even worse as the model gets progressively trained under the PGD attack.~This indeed motivates our DRFL approach when data are from untrusted entities with possibly adversarial input perturbations.~Similarly, Fig.~\ref{fig:fedfmnist} depicts the misclassification rate of the proposed DRFL method compared with federated averaging, when using the F-MNIST dataset.

As the distribution of data across devices may influence performance, we further considered a biased local data setting.~In particular, each worker $k\!=\! 1,\ldots, 10$ has data from only one class, so the distributions at workers are highly perturbed, and data stored across workers are thus non-i.i.d.~The testing error rate for normal inputs is reported in Fig. \ref{fig:fedfmnist_biased}, while the test error against adversarial attacks is depicted in Figs.~\ref{fig:fedfmnist_biased}(Middle) and \ref{fig:fedfmnist_biased}(Right).~This additional set of tests shows that having distributional shifts across workers can indeed enhance testing performance when the samples are adversarially manipulated.

\section{Conclusions}
\label{sec:concls}
A framework to robustify parametric machine learning models against distributional uncertainties was put forth.~The learning task was cast as a distributionally robust optimization problem, for which two scalable stochastic optimization algorithms were developed.~The first algorithm relies on an $\epsilon$-accurate maximum-oracle to solve the inner convex subproblem, while the second approximates its solution via a single gradient ascent step.~Convergence guarantees for both algorithms to a stationary point were obtained. The upshot of the proposed approach is that it is amenable to federated learning from unreliable datasets across multiple workers. The novel DRFL algorithm ensures data privacy and integrity, while offering robustness with minimal computational and communication overhead. Numerical tests for classifying standard real images showcased the merits of the proposed algorithms against distributional uncertainties and adversaries.~This work also opens up several interesting directions for future research, including distributionally robust deep reinforcement learning.

\chapter{Distributionally Robust Semi-Supervised Learning Over Graphs}\label{chap:sslovergraphs}
%%%%%%%%%%%%%%%%%%%%%%%%%%%%%%%%%%%%%%%%%%%%%%%%%%%%%%%%%%%%%%%%%%%%%%%%%%%%%%%%

\section{Introduction}
\label{sec:intro}

Building upon but going well beyond the scope of previous robust learning paradigms, 
the present Chapter puts forth a novel iterative semi-supervised learning (SSL) over graphs framework.  Relations among data in real world applications can often be captured by graphs, for instance the analysis and inference tasks for social, brain, communication, biological, transportation, and sensor networks \cite{shuman2013mag, kolaczyk2014statistical}. In practice however, the data is only available for a subset of nodes, due to for example the cost, and computational or privacy constraints. Most of these applications however, deal with inference of processes across all the network nodes. Such semi-supervised learning (SSL) tasks over networks can be addressed by exploiting the underlying graph topology \cite{sslbook, adadif2019berb, qin20120icassp}.

Graph neural networks (GNNs) are parametric models 
%capable of representing nonlinear functions over graph-structured data . By 
that combine graph-filters and topology information with point-wise nonlinearities, to form nested architectures to easily express the functions defined over graphs \cite{gnn2018survey}. By exploiting the underlying irregular structure of network data, the GNNs enjoy lower computational complexity, less parameters for training, and improved generalization capabilities relative to traditional deep neural networks (DNNs), making them appealing for learning over graphs \cite{gnn2018survey, gnn2020comprehensivesurvey, antonio2019tsp}.
%, and for SSL tasks [??]. 

Similar to other DNN models, GNNs are also susceptible to adversarial manipulated input data  or, distributional uncertainties, such as mismatches between training and testing data distributions. For instance small perturbations to input data would significantly deteriorate the regression performance, or result in classification error \cite{gnn2020adversarial, gnn2020adversarialsurvey}, just to name a few. Hence, it is critical to develop principled methods that can endow GNNs with robustness, especially in safety-critical applications, such as robotics \cite{multirobot2020learning}, and transportation \cite{gnn2020transport}.

\textbf{Contributions.} This Chapter endows SSL over graphs using GNNs with \textit{robustness} against distributional uncertainties and possibly adversarial perturbations. Assuming the data distribution lies inside a Wasserstein ball centered at empirical data distribution, we robustify the model by minimizing the worst expected loss over the considered ball, which is challenging to solve. Invoking recently developed strong duality results, we develop an equivalent unconstrained and tractable learning problem. \footnote{Results of this Chapter are published in \cite{sadeghi2021distributionally}}.

\section{Problem formulation}
Consider a SSL task over a graph $\mathcal{G} := \left\{\mathcal{V}, \mathbf{W} \right\}$ with $N$ nodes, where $\mathcal {V}:= \{1, \ldots, N\}$ denotes the vertex set, and $\mathbf{W}$ represents the $N\times N$ weighted adjacency matrix capturing node connectivity. The associated unnormalized graph Laplacian matrix of the undirected graph $\mathcal{G}$ is $\mathbf{L} := \mathbf{D} - \mathbf{A}$, where $\mathbf{D} := \textrm{diag}\{\mathbf{W} \mathbf{1}_N\}$, with $\mathbf{1}_N$ denoting the $N \times 1$ all-one column vector. Denote by matrix $\mathbf{X}_s \in \mathbb{R}^{N \times F}$ the nodal feature vectors sampled at instances $s = 1, 2, \cdots $, with $n$-th row $\mathbf{x}^\top_{n,s} := \left[\mathbf{X}_s\right]_{n:}$ representing a feature vector of length $F$ associated with node $n \in \mathcal{V}$, and $\top$ stands for transposition. 
In the given graph, the labels $\{y_{n,s}\}_{n \in \mathcal{O}_s}$ are given for \emph{only a small subset} of nodes,
where $\mathcal O_s$ represents the index set of \textit{observed} nodes sampled at~$s$, and $\mathcal{U}_s$ the index set of \textit{unobserved} nodes. 

Given $\{\mathbf{X}_s, \mathbf{y}_s\}$, where $\mathbf{y}_s$ is the vector of observed labels, the goal is to find the labels of unobserved nodes $\{y_{n,s}\}_{n \in \mathcal{U}_s}$. 
To this aim our objective is to learn a functional mapping $f(\mathbf{X}_s; \mathbf{W})$ that can infer the missing labels based on available information. 
Such a function can be learned by solving the following optimization problem (see e.g., \cite{kipf2016semi} for more details)
\begin{align}
\label{eq:loss}
\min_{f \in{\mathcal{F}}}  \mathbb{E}  \, \Big[   \overbrace{\underset{n \in \mathcal{O}_s}{\sum}  \|f(\mathbf{x}_{n}; \mathbf{W}) - y_{n}\|^2}^{\mathcal{L}_0} +  \lambda \overbrace{\underset{{n,n'}}{\sum} \mathbf{W}_{nn'} \| f(\mathbf{x}_n; \mathbf{W}) - f(\mathbf{x}_{n'}; \mathbf{W}) \|^2}^{\mathcal{L}_{\rm{reg}}} \Big],
\end{align}
where $\mathcal{L}_0$ represents the supervised loss w.r.t. the observed part of the graph, $\mathcal{L}_{\rm{reg}}$ represents the Laplacian regularization term, $\mathcal{F}$ denotes the feasible set of functions that we can learn, and $\lambda \ge 0$ is a hyper parameter. The regularization term relies on the premise that connected nodes in the graph are likely to share similar labels. The expectation here is taken with respect to (w.r.t) the feature and label data generating distribution.

In this work, we first encode the graph structure using a GNN model denoted by $f(\mathbf{X}; \boldsymbol{\theta}, \mathbf{W})$, where $\boldsymbol{\theta}$ represents the model parameters. Such a parametric representation enables bypassing explicit graph-based
regularization $\mathcal{L}_{\rm{reg}}$ represented in  \ref{eq:loss}. The GNN model of $f(\cdot)$ relies on the weighted adjacency $\mathbf{W}$ and therefore can easily propagate information from observed nodes $\mathcal{O}_s$ to unobserved ones $\mathcal{U}_s$. In a nutshell, objective is to learn a parametric model by solving the following problem
\begin{equation}
\underset{\boldsymbol{\theta} \in {\Theta}}{\min} \;\; \mathbb{E}_{\small{{\mathbf{\{X, y\}}\sim P_0}}} \; {\mathcal{L}_0\big(f(\mathbf{X}, \boldsymbol{\theta};\mathbf{W}), \mathbf{y}\big)} 
\end{equation} 
where $\Theta$ is a feasible set, and $P_0$ is the feature and label data generating distribution. Despite restricting the modeling capacity through parameterizing $f(\cdot)$ with GNNs, we may infuse additional prior information into the sought formulation through exploiting the weighted adjacency matrix $\mathbf{W}$, which does not necessarily encode node similarities. 

In practice, $P_0$ is typically unknown, instead some data samples $\{\mathbf{X}_s, \mathbf{y}_s\}_{s=1}^S$  are given. Upon replacing the nominal distribution with an empirical one, we arrive at the empirical loss minimization problem, that is $
{\min}_{\boldsymbol{\theta} \in {\Theta}} \;S^{-1} \sum_{s=1}^{S} \; {\mathcal{L}_0 \big(f(\mathbf{X}_s, \boldsymbol{\theta};\mathbf{W}), \mathbf{y}_s\big)}$. The model obtained by solving empirical risk minimization does not exhibit any robustness in practice, specifically if there is any mismatch between the training and testing data distributions. To endow robustness, we reformulate this learning problem in a fresh manner as described in ensuing section.

\section{Distributionally robust learning}
\label{gen_inst}
To endow robustness, we consider the following optimization problem
\begin{equation} \label{eq:robusterm}
\underset{\boldsymbol{\theta} \in {\Theta}}{\min} \; \underset{P \in \mathcal{P}}{\sup} \; \; \mathbb{E}_{\small{{\mathbf{(X, y)}\sim P}}} \; {\mathcal{L}_0\big(f(\mathbf{X}, \boldsymbol{\theta};\mathbf{W}), \mathbf{y}\big)} 
\end{equation} 
where $\mathcal P$ is a set of distributions centered around the \textit{empirical data distribution} $\widehat{P}_0$. This novel reformulation in \ref{eq:robusterm} yields a model that performs reasonably well among a continuum of distributions. Various ambiguity sets $\mathcal{P}$ can be considered in practice, and they lead to different robustness guarantees with different computational requirements. For instance momentum, KL divergence,  statistical test, and Wasserstein distance-based sets are popular in practice; see also \cite{blanchet2019quantifying, sinha2017certifying, blanchet2017data} and references therein. Among possible choices, we utilize the optimal transport theory and the Wasserstein distance to characterize the ambiguity set $\mathcal{P}$. As a result, we can offer a tractable solution for this problem, as delineated next.

%Optimal transport theory is a mathematical framework to quantify the distance between two probability distributions. First introduced by Monge \cite{villani2008optimal}, optimal transport theory deals 
%with a transportation problem. In its original form, there exists piles of
%sand and some holes with exactly the same total volume of sand. The sand and piles are
%randomly distributed over a geographical area. The goal is to find the optimal moves to transport the entire piles
%to the holes with the \textit{minimum} possible transportation cost, which obviously depends on the distance between piles and holes. 

To formalize our framework, let us first define the Wasserstein distance between two probability measures. To this aim, consider probability measures $P$ and $\widehat{P}$ supported on some set $\mathcal{X}$, and let $\Pi(P,\widehat{P})$ denote the set of joint measures (a.k.a coupling) defined over $\mathcal{X} \times \mathcal{X}$, with marginals $P$ and $\widehat{P}$, and let $c: \mathcal{X} \times \mathcal{X}  \rightarrow [0, \infty)$ measure the transportation cost for a unit of mass from $\mathbf{X \in \mathcal{X}}$ in $P$ to $\mathbf{X}' \in \mathcal{X}$ in $\widehat{P}$. The so-called optimal transport problem is concerned with the minimum cost associated with transporting all the mass from $P$ to $\widehat{P}$ through finding the optimal coupling, i.e., $
W_c(P,\widehat{P}) := \; \underset{\pi \in \Pi}{\inf} \, \mathbb{E}_\pi [ c(\mathbf {X},\mathbf{X}')]
$. If $c(\cdot,\cdot)$ satisfies the axioms of distance, then $W_c$ defines a distance on the space of probability measures. For instance, if $P$ and $\widehat{P}$ are defined over a Polish space equipped with metric $d$, then fixing $c(\mathbf X, \mathbf X') = d^p(\mathbf X, \mathbf X')$ for some $p\in [1, \infty)$ asserts that $W_c^{1/p}(P,\widehat{P})$ is the well-known Wasserstein distance of order $p$ between $P$ and $\widehat{P}$. 

Using the Wasserstein distance, let us define the uncertainty set $\mathcal{P} := \{P| W_c(P,\widehat{P}_0) \le \rho\}$ to include all probability distribution functions (pdfs) having at most $\rho$-distance from $\widehat{P}_0$. Incorporating this ambiguity set into \ref{eq:robusterm}, the following robust surrogate is considered in this work
\begin{equation}
\label{eq:robform}
\underset{\boldsymbol{\theta} \in {\Theta}}{\min}  \; \underset{P \in \mathcal{P}}{\sup} \quad \mathbb{E}_{\small{{\mathbf{(X, y)}\sim P}}} \; {\mathcal{L}_0\big(f(\mathbf{X}, \boldsymbol{\theta};\mathbf{W}), \mathbf{y}\big)}, \quad 
{\rm where}\; \mathcal{P}:= \left\{P\, | W_c(P,\widehat{P}_0) \le \rho \right\}. 
\end{equation}
The inner supremum here goes after pdfs characterized by $\mathcal{P}$. Solving this optimization directly over the infinite-dimensional space of distribution functions raises practical challenges. Fortunately, under some mild conditions over losses as well as transport costs, the inner maximization satisfies a strong duality condition (see \cite{blanchet2017data} for a detailed discussions), which means the optimal objective of this inner maximization and its Lagrangian dual are equal. Enticingly, the dual reformulation involves optimization over only one-dimensional dual variable. These properties make it practically appealing to solve \ref{eq:robform} directly in the dual domain. The following proposition highlights the strong duality result, whose proofs can be found in \cite{blanchet2019quantifying}.  

\begin{proposition} \label{prop1}
	Under some mild conditions over the loss $\mathcal{L}_0(\cdot)$ and cost $c(\cdot)$, it holds that
	\begin{align} \label{eq:strongduality}
	\sup_{P\in\mathcal{P}} \, \mathbb{E}_{P} \; {\mathcal{L}_0\big(f(\mathbf{X}, \boldsymbol{\theta};\mathbf{W}), \mathbf{y}\big)}
	= \inf_{\gamma \ge 0}   \frac{1}{S} \sum_{s=1}^{S}    \sup_{\boldsymbol{\xi} \in {\mathcal X}}   \{ {\mathcal{L}_0\big(f(\bm{\xi}, \boldsymbol{\theta};\mathbf{W}), \mathbf{y}_s\big)} + \gamma \, (\rho - c(\mathbf X_s, \boldsymbol{\xi}) ) \}   
	\end{align}
	where $\mathcal{P}:= \left\{P\, | W_c(P,\widehat{P}_0) \le \rho \right\}$.
\end{proposition}

The right-hand side in \ref{eq:strongduality} simply is the univariate dual reformulation of the primal problem represented in the left-hand side. Furthermore, different from the primal formulation, the expectation in the dual domain is replaced with the summation over available training data, rather than any $P \in \mathcal{P}$ that needs to be obtained by solving for the optimal $\pi \in \Pi$ to form $\mathcal{P}$. Because of these two properties, solving the dual problem is practically more appealing. Thus, hinging on Proposition \ref{prop1}, the following distributionally robust surrogate is considered in this work 
\begin{align} \label{eq:robustdual}
\min_{\boldsymbol{\theta}\in \Theta} \; \inf_{\gamma \ge 0}  \frac{1}{S} \sum_{s=1}^{S}   \sup_{\boldsymbol{\xi} \in {\mathcal X}}  \left\{ 
{\mathcal{L}_0\big(f(\boldsymbol{\xi}, \boldsymbol{\theta};\mathbf{W}), \mathbf{y}_s \big)}
+  \gamma (\rho - c(\mathbf X_s, \boldsymbol{\xi}))\right\} \!\! 
\end{align}

This problem requires the supremum to be solved separately for each sample $\mathbf{X}_s$, which cannot be handled through existing methods. 
%A relaxed (hence suboptimal) version of this problem with a fixed $\gamma$ value  has been studied in \cite{sinha2017certify}. Unfortunately, one has to select an appropriate $\gamma$ using cross validation over a grid, which incurs a prohibitive computational cost. Instead, we advocate algorithms that optimize $\gamma$ and $\mathbf{\theta}$ simultaneously. 
Our approach to address this relies on the structure of this problem to iteratively update parameters $\bar{\boldsymbol{\theta}}:=[{\boldsymbol \theta}^\top, \gamma]^\top$ and $\boldsymbol{\xi}$. Specifically, we rely on Danskin's theorem to first maximize over $\boldsymbol{\xi}$, which results in a differentiable function of $\bar{\boldsymbol{\theta}}$, and then minimize the objective w.r.t. $\bar{\boldsymbol{\theta}}$ using gradient descent. However, to guarantee convergence to a stationary point and utilize Danskin's theorem, we need to make sure the inner maximization admits a unique solution (singleton). By choosing a strongly convex transportation cost such as $c(\mathbf{X}, \boldsymbol{\xi}) :=\| \mathbf{X} - \boldsymbol{\xi} \|_F^2$, and by selecting $\gamma \in \Gamma :=\{\gamma| \gamma >\gamma_0\}$ with a large enough $\gamma_0$, we arrive at a strongly concave objective function for the maximization over $\boldsymbol \xi$.
Since $\gamma$ is the dual variable associated with the constraint in \ref{eq:robform}, having $\gamma \in \Gamma$ is tantamount to tuning $\rho$, which in turn \emph{controls} the level of robustness. Replacing $\gamma \ge 0$ in \ref{eq:robustdual} with $\gamma \in \Gamma$, our \emph{robust model} can be obtained as the solution of 
\begin{align}
\min_{\boldsymbol{\theta}\in \Theta}~\inf_{\gamma \in \Gamma}~ \frac{1}{S} \sum_{s=1}^{S}  \sup_{\boldsymbol{\xi} \in\mathcal{X}} \psi(\mathbf{\bar{\boldsymbol{{\theta}}}}, \boldsymbol{\xi}; \mathbf{X}_s)   
\label{eq:objective}
\end{align}
where $ \psi(\mathbf{\bar{\boldsymbol{{\theta}}}}, \boldsymbol{\xi}; \mathbf{X}_s) = {\mathcal{L}_0(f(\boldsymbol{\xi}, \boldsymbol{\theta};\mathbf{W}), \mathbf{y}_s)} +\gamma (\rho - c(\mathbf X_s, \boldsymbol{\xi}))$.
Intuitively, input $\mathbf{X}_s$ in \ref{eq:objective} is pre-processed by maximizing $\psi(\cdot)$ accounting for a perturbation. We iteratively solve \ref{eq:objective}, where after sampling a mini-batch of data, we first pre-process them by maximizing the function $\psi(\cdot)$. Then, we use a simple gradient descent to update $\bar{\boldsymbol{{\theta}}}$. Notice that the $\bm{\theta}$ inside function $\psi(\cdot)$, represents the weights of our considered GNN, whose details are provided next.

\section{Graph neural networks}
GNNs are parametric models to represent functional  relationship for graph structured data. Specifically, the input to a GNN is a data matrix $\mathbf X$. Upon multiplying the input $\mathbf X$ by $\mathbf W$, features will diffuse over the graph, giving a new graph signal $ \check{\mathbf{Y}}=\mathbf{W X}$. To model feature propagation, one can also replace $\mathbf{W}$ with the (normalized) graph Laplacian or random walk Laplacian, since they will also preserve dependencies among nodal attributes.

During the diffusion process, the feature vector of each node is updated by a linear combination of its neighbors. Take the $n$-th node as an example, the shifted $f$-th feature $[\check{\mathbf Y}]_{nf}$ is obtained by $ [\check{\mathbf{Y}}]_{nf} = \sum_{i=1}^{N}[\mathbf{W}]_{ni}[\mathbf{X}]_{if}=\sum_{i \in \mathcal{N}_{n}} w_{ni} x_{i}^{f} $, 
where ${\mathcal{N}}_{n}$ denotes the set of neighboring nodes for node $n$. 
The so-called convolution operation in GNNs utilizes topology to combine features, namely
\begin{equation}
\label{eq:gc}
[\mathbf{Y}]_{nd}:=[\mathcal{H} \star \mathbf{X}; \mathbf{W}]_{nd}:=\sum_{k=0}^{K-1}[\mathbf{W}^{k} \mathbf{X}]_{n:} [\mathbf{H}_k]_{:d} 
\end{equation}
where $\mathcal{H}:=[\mathbf{H}_0~ \cdots~\mathbf{H}_{K-1}]$ with ${\mathbf H}_{k} \in \mathbb{R}^{F\times D}$ as filter coefficients;~$\mathbf Y \in \mathbb R^{N\times D}$ the intermediate (hidden) matrix with $D$ features per node;~and $\mathbf{W}^{k} \mathbf{X}$ as the linearly combined features of nodes within the $k$-hop neighborhood.

To construct a GNN with $L$ hidden layers, first let us denote by $\mathbf{X}_{l-1}$ the output of the $(l-1)$-th layer, which is also the $l$-th layer input for $l=1, \ldots, L$, and $\mathbf{X}_0 = \mathbf{X}$ to represent the input matrix. The hidden $\mathbf{Y}_{l} \in \mathbb{R}^{N\times D_{l}}$ with $D_l$ features is obtained by applying the graph convolution operation \ref{eq:gc} at layer $l$, i.e., $[\mathbf{Y}_l]_{nd} =\sum_{k=0}^{K_l-1}[\mathbf{W}^{k} \mathbf{X}_{l-1}]_{n:} [\mathbf{H}_{lk}]_{:g}$, 
where $\mathbf{H}_{lk} \in \mathbb{R}^{F_{l-1}\times F_{l}}$ is the convolution coefficients for $k=0, \ldots, K_l-1$. The output at layer $l$ is constructed by applying a graph convolution followed by a point-wise nonlinear operation $\sigma_{l}(\cdot)$. The input-output relationship at layer $l$ can be represented succinctly by $
\mathbf{X}_{l} =\sigma_{l}(\mathbf Y_{l})=\sigma_{l}\!\left(\sum_{k=0}^{K_l-1} \mathbf{W}^{k} \mathbf{X}_{l-1} \mathbf{H}_{l k}\right).
$ 
Using this mapping, GNNs use a nested architecture to represent nonlinear functional operator $\mathbf{X}_L={f}(\mathbf{X}_{0} ; \bm{\theta}, \mathbf{W})$ that maps the GNN input $\mathbf{X}_{0}$ to label estimates by taking into account the graph structure through $\mathbf W$. Specifically, in a compact representation we have that
\begin{align}  \label{eq:GNNbasepsse} 
f(\mathbf{X}_{0} ; \bm{\theta}, \mathbf{W}) :=  
\sigma_{L}\!\left(\sum_{k=0}^{K_{L }-1} \mathbf{W}^{k} \!\left(\ldots \!\left(\sigma_{1}\!\left(\sum_{k=0}^{K_1-1} \mathbf{W}^{k} \mathbf{X}_{0} \mathbf{H}_{1 k}\right) \ldots\right)\right)\mathbf{H}_{L k}\right)  
\end{align}
where the parameter set $\bm \theta$ contains all the \textit{trainable} filter weights $\{\mathbf{H}_{lk}, \forall l, k\}$.

\section{Experiments}
The performance of our novel distributionally robust GNN-based SSL is tested in a regression task using real 
load consumption data from the 2012 Global Energy Forecasting Competition (GEFC). Our objective here is to estimate only the amplitudes of voltages across all the nodes in a standard IEEE $118$-bus network.  Utilizing this data set, the training and testing data are prepared by solving the so-called AC power ﬂow equations using the MATPOWER toolbox \cite{matpower}. 

The measurements $\mathbf{X}$ used include all active and reactive power injections, corrupted by small additive white Gaussian noise. Using MATPOWER we generated $1,000$ pairs of measurements and ground-truth voltages. We used 80\% of this data for training and the remaining  for testing. Throughout the training, the Adam optimizer with a fixed learning rate $10^{-3}$ was employed to minimize the H\"uber loss. Furthermore, the batch size was set to $32$ during all $100$ epochs. 

To compare our method we employed $3$ different benchmarks, namely: i) the prox-linear network introduced in \cite{liang2019}; ii) a 6-layer vanilla feed-forward neural network (FNN); and, iii) an 8-layer FNN. Our considered GNN uses $K=2$ with $D=8$ hidden units with ReLU activation. 

\begin{figure}[t]
	\begin{subfigure} 
		\centering
		\includegraphics[width=0.48\textwidth]{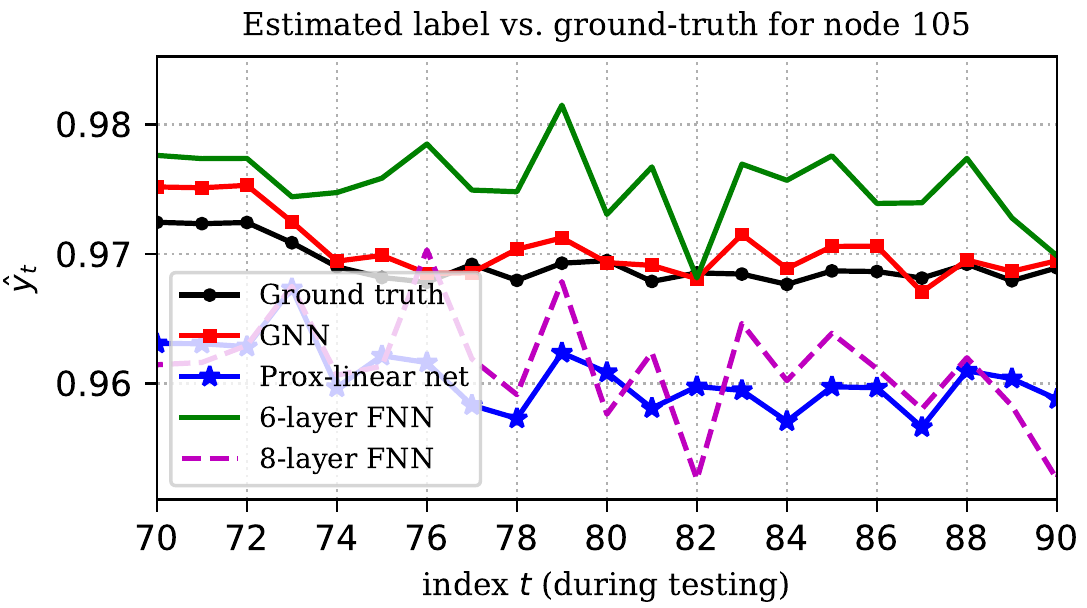}
	\end{subfigure}
	\hfill
	\begin{subfigure} 
		\centering
		\includegraphics[width=0.48\textwidth]{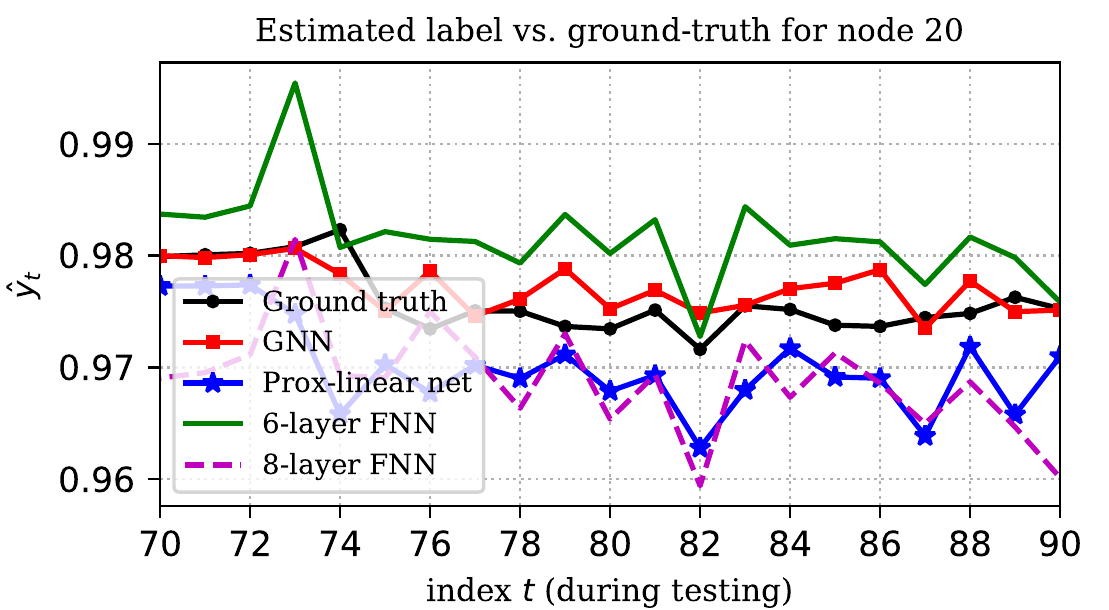}
	\end{subfigure}
	\hfill
	%	\begin{subfigure}[t]{0.2\textwidth}
	%		\centering
	%		\includegraphics[width=\textwidth]{figure_3}
	%		\caption{ }
	%		\label{fig:}
	%	\end{subfigure}
	%	\caption{Performance during testing for \textit{normal} input features.}
	\hfill
	\begin{subfigure}
		\centering
		\includegraphics[width=0.48\textwidth]{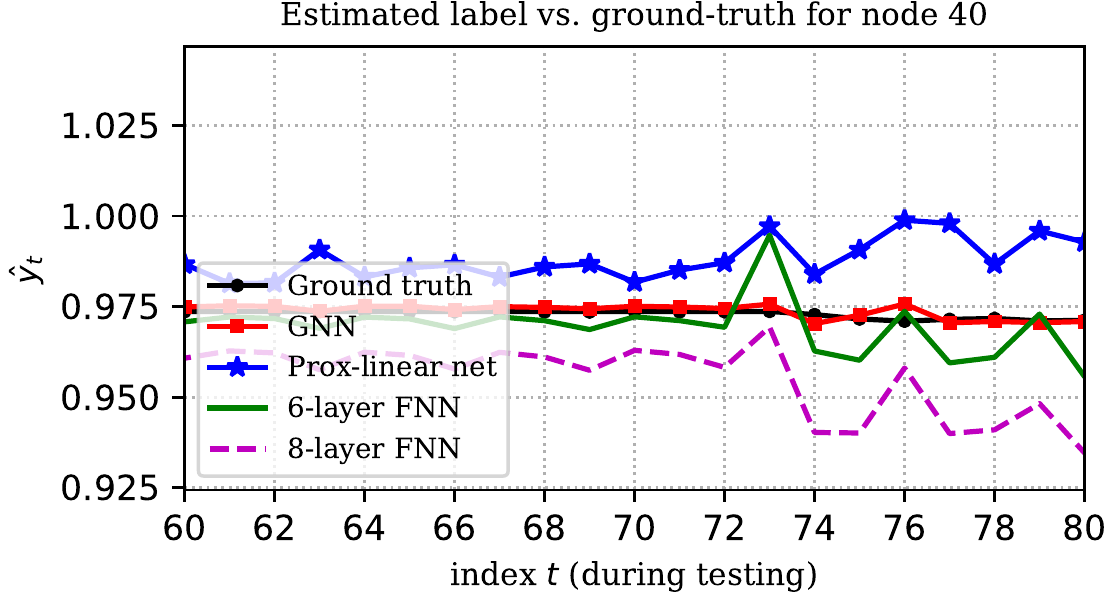}
	\end{subfigure}
	%	\hfill
	%	\begin{subfigure}[t]{0.2\textwidth}
	%		\centering
	%		\includegraphics[width=\textwidth]{adv_2}
	%		\caption{ }
	%		\label{fig:}
	%	\end{subfigure}
	\hfill
	\begin{subfigure}
		\centering
		\includegraphics[width=0.48\textwidth]{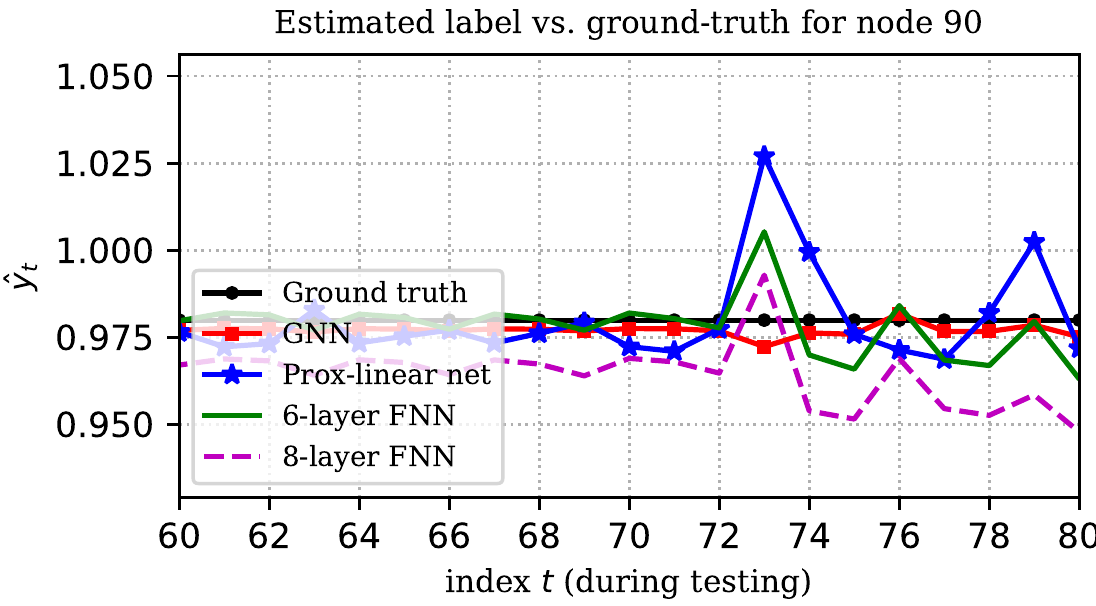}
	\end{subfigure}
	\caption{Performance during testing for both normal (a) - (b), and perturbed input features (c) - (d).}
	\label{fig1}
\end{figure}

The first set of tests are carried out using normal (not-corrupted) data, where the results are depicted in Fig. \ref{fig1}. Here we show the estimated (normalized) voltage amplitudes at different nodes, namely $105$, and $20$ during the given time course. The black curve represents the ground truth signal to be estimated. Clearly our GNN-based method outperforms alternative methods. 

The second set of experiments are carried out over corrupted input signals, and the results are reported in Fig. \ref{fig1}. Specifically the training samples were generated according to $P_0$, but during testing samples were perturbed to satisfy the constraint $P \in \mathcal{P}$, that would yield the worst expected loss. Fig. \ref{fig1} depicts  the estimated signals across nodes $40$ and $90$. Here we fixed $\rho = 10$ and related hyper-parameters are tuned using grid search. As the plots showcase, the our proposed GNN-based robust method outperforms competing alternatives with corrupted inputs.   

\section{conclusions}
This Chapter dealt with semi-supervised learning over graphs using GNNs. To account for uncertainties associated with data distributions, or adversarially manipulated input data, a principled robust learning framework was developed. Using the parametric models, we were able to reconstruct the unobserved nodal values. Experiments corroborated the outstanding performance of the novel method when the input data are corrupted.

%\begin{figure}[t]
%	\centering
%	\begin{subfigure}
%		\centering
%		% 		\includegraphics[width=.47\textwidth]{rratenew.pdf}
%		\includegraphics[scale=0.49]{figs/raf/rratenew.pdf}
%	\end{subfigure}
%	\begin{subfigure}
%		\centering
%		% 		\includegraphics[width=.47\textwidth]{mse.pdf}
%		\includegraphics[scale=0.481]{figs/raf/mse.pdf}
%	\end{subfigure}
%
%	\caption{ Real Gaussian model.
%%		 with $\bm{x}\in\mathbb{R}^{1,000}$. 
%		 Left: Empirical success rate; Right: NMSE vs. SNR.
%	}
%	\label{fig3:ireal}
%\end{figure}

%Software Chapters
%%%%%%%%%%%%%%%%%%%%%%%%%%%%%%%%%%%%%%%%%%%%%%%%%%%%%%%%%%%%%%%%%%%%%%%%%%%%%%%
% intro.tex: Introduction to the thesis
%%%%%%%%%%%%%%%%%%%%%%%%%%%%%%%%%%%%%%%%%%%%%%%%%%%%%%%%%%%%%%%%%%%%%%%%%%%%%%%%
\chapter{Deep and Reinforced Learning for Network Resource Management}\label{chap:saf}
%%%%%%%%%%%%%%%%%%%%%%%%%%%%%%%%%%%%%%%%%%%%%%%%%%%%%%%%%%%%%%%%%%%%%%%%%%%%%%%%

\section{Introduction}

The advent of smart phones, tablets, mobile routers, and a massive number of devices connected through the Internet of Things (IoT) have led to an unprecedented growth in data traffic. Increased number of users trending towards video streams, web browsing, social networking and online gaming, have urged  providers to pursue new service technologies that offer acceptable quality of experience (QoE). One such technology entails network densification by deploying small pico- and femto-cells, each serviced by a low-power, low-coverage, small basestation (SB). In this infrastructure,  referred to as  heterogeneous network (HetNet), SBs are connected to the backbone by a cheap `backhaul' link.  While boosting the network density by substantial reuse of scarce resources, e.g., frequency, the HetNet architecture is restrained by its low-rate, unreliable, and relatively slow backhaul links~\cite{Femtocell_Andrews_2012}.   

During peak traffic periods specially when electricity prices are also high, weak backhaul links can easily become congested--an effect lowering the QoE for end users. One approach  to mitigate this limitation is to shift the excess load from peak periods to off-peak periods. Caching realizes this shift by fetching the ``anticipated'' popular contents, e.g., reusable video streams, during off-peak periods, storing this data in SBs equipped with memory units, and reusing them during peak traffic hours~\cite{Wireless_caching_Paschos_2016,Femrocaching_Golrezaei_2013, Cacheinair}. 
In order to utilize the caching capacity intelligently, a content-agnostic SB must rely on available observations to learn what and when to cache. To this end, machine learning tools can provide 5G cellular networks with efficient caching, in which a ``smart'' caching control unit (CCU) can learn, track, and possibly adapt to the space-time popularities of reusable contents~\cite{Wireless_caching_Paschos_2016,What5Gbe_Andrews_2014}. 

%\subsection{Related Work}
\noindent{\bf Prior work}. Existing efforts in 5G caching  have focused on enabling SBs to learn unknown time-invariant  content popularity profiles, and cache the most popular ones accordingly. A multi-armed bandit approach is reported in \cite{MAB_Gunduz_2014}, where a reward is received when user requests are served via cache; see also \cite{LearnDist_Sengupta_2014} for a distributed, coded, and convexified reformulation. 
%
%
%
%version of this problem is formulated in ,
%however, certain  assumptions such as  neglecting out-of-cache requests for an improved estimation of popularities impede the implementation of the proposed approach in a realistic scenario. 
A belief propagation-based approach for distributed and collaborative caching is also investigated in \cite{DistBeliefprop}. Beyond \cite{MAB_Gunduz_2014}, \cite{LearnDist_Sengupta_2014}, and \cite{DistBeliefprop} that deal with deterministic caching, \cite{JointOptimal} and \cite{OptimalGeographic} introduce probabilistic alternatives. Caching, routing and video encoding are jointly pursued in \cite{VideoDeliveryOpt} with users having different QoE requirements. However, a limiting assumption in \cite{MAB_Gunduz_2014,LearnDist_Sengupta_2014,DistBeliefprop,JointOptimal,
	OptimalGeographic,VideoDeliveryOpt} pertains to space-time invariant modeling of  popularities, which can only serve as a crude approximation for real-world requests. Indeed, temporal dynamics of local requests are prevalent due to  user mobility, as well as emergence of new contents, or, aging of older ones.
To accommodate dynamics,   Ornstein-Uhlenbeck processes and Poisson shot noise models are utilized in  \cite{Dynamic_Debbah_2017} and \cite{leconte2016}, respectively, while context- and trend-aware caching approaches are investigated in \cite{Context_aware_Schaar_2017} and \cite{Trend_aware_Schaar_2016}.

Another practical consideration for 5G caching is driven by the fact that a relatively small number of users request contents during a caching period. This along with the small size of cells can challenge SBs from estimating accurately the underlying content popularities.
To address this issue, a  transfer-learning approach  is advocated in  \cite{Transferlearning_Debbah_2015}, \cite{Learning_Poor_2016} and  \cite{leconte2016}, to improve the time-invariant popularity profile estimates by leveraging prior information obtained from a surrogate (source) domain, such as social networks.

%In \cite{leconte2016} centralized caching scheme combined with distributed counterpart is adopted where learning the content popularities happen in a global node by aggregating all the requests from local caches. This globally coordinated method will modify a threshold aged based caching mechanism in the local caches, which in turn makes the local caching performance more reliable.}}  

Finally, recent studies have investigated the role of coding for enhancing performance in cache-enabled networks \cite{Maddahalifundamentals,Maddahalidecent,MaddahaliOnline}; see also \cite{CaireFundamentalsD2D}, \cite{CaireWD2Dcaching}, and \cite{DtoD}, where device-to-device ``structureless'' caching approaches are envisioned.

\section{Our Contribution}
The present chapter introduces a novel approach to account for space-time popularity of user requests by casting the caching task in a reinforcement learning (RL) framework. The CCU of the local SB is equipped with storage and processing units for solving  the emerging RL optimization in an online fashion. Adopting a Markov model for the popularity dynamics, a Q-learning caching algorithm is developed to learn the optimal policy even when the underlying transition probabilities are unknown.   

Given the geographical and temporal variability of cellular traffic, global popularity profiles may not always be representative of local demands. To capture this, the proposed framework entails estimation of the popularity profiles  both at the local as well as at the global scale. Specifically, each SB estimates its local vector of popularity profiles based on limited observations, and transmits it to the network operator, where an estimate of the global profile is obtained by  aggregating the local ones. The estimate of the global popularity vector is then  sent back to the SBs. The SBs can adjust the cost (reward) to  trade-off tracking global trends versus serving local requests.
%local cache for  complete, partial, or disabled tracking of globally popular demands  in local caching decisions. Furthermore, the  cache-refreshing costs considers possible costs of deleting a content and fetching new ones from the network backbone.

To obtain a scalable caching scheme, a novel approximation of the proposed Q-learning algorithm is also developed. Furthermore, despite the stationarity assumption on the popularity Markov models, proper selection of stepsizes broadens the scope of the proposed algorithms for  tracking  demands even in non-stationary settings. \footnote{Results of this Chapter are published in \cite{RL1, RL3, sadeghi2020reinforcement, sadeghi2018optimal, sadeghi2019reinforcement, sadeghi2018reinforcement, sadeghi2019deep, sadeghi2020hierarchical, sadeghi2019distributed}.}

\section{Modeling and problem statement}\label{sec:model}
Consider a local section of a HetNet with  a single SB connected to the backbone network through a low-bandwidth, high-delay, unreliable backhaul link. Suppose further that the SB is equipped with $M$ units to store contents (files) that are assumed for simplicity to have unit size; see Fig.~1. Caching will be carried out in a slotted fashion over slots $t=1,2,\ldots$, where at the beginning of each slot $t$, the CCU-enabled SB selects ``intelligently'' $M$ files from the total of $F \gg M$ available ones at the backbone, and prefetches them for possible use in subsequent slots. The slots may not be of  equal length, as the starting times  may be set a priori, for example at 3~AM, 11~AM, or 4~PM, when the network load is low; or, slot intervals may be dictated to CCU by the network operator on the fly. Generally, a slot starts when the network is at an off-peak period, and its duration coincides with the peak traffic time when pertinent costs of serving users are high.

During slot $t$, each user locally requests a subset of  files from the set ${\cal F}:=\left\{1,2,\ldots,F\right\}$. If a requested file has been stored in the cache, it will be simply served locally, thus  incurring (almost) zero cost.  Conversely, if the requested file is not available in the cache, the SB must fetch it from  the cloud  through its cheap backhaul link, thus incurring a  considerable cost due to possible electricity price surges, processing cost, or the sizable delay resulting  in low QoE and user dissatisfaction.  The CCU wishes to intelligently select the cache contents so that costly services from the cloud be avoided as often as possible.

Let ${\bf a} (t) \in \mathcal{A}$ denote the $F \times 1$ binary \emph{caching action vector} at slot $t$, where $\mathcal{A}:=\{\mathbf{a}| \mathbf{a} \in \{0,  1\}^F, \mathbf{a}^\top \mathbf{1}=M\}$ is the set of all feasible actions; that is,  $[{{\bf a}(t)}]_f  = 1$ indicates that file $f$ is cached for the duration of slot $t$, and  $[{\bf a}(t)]_f  = 0$ otherwise.

%\begin{figure}[t]\label{fig:sys1}
%	\centering
%	\includegraphics[width=0.9\columnwidth]{Systemmodel}
%	\vspace{0.2 cm}
%	\caption{Local section of a HetNet.}
%\end{figure} 

Depending on the received requests from locally connected users, the CCU computes the $F \times 1$-vector of \emph{local popularity profile} ${\bf p}_L (t)$ per slot $t$, whose $f$-th entry indicates  the expected local demand for file $f$, defined  as 
\begin{align}
\nonumber
\bigg[\mathbf{p}_{\text{L}} (t)\bigg]_f :=\dfrac{\text{ Number of local requests for } f \; {\text {at slot} } \;t }{\text{Number of all local requests at slot}\; t}.
\end{align}
Similarly, suppose that the backbone network estimates  the $F \times 1$ \emph{global popularity profile} vector ${\bf p}_G (t)$, and transmits it to~all~CCUs. 

Having observed the local and global user requests by the end of slot $t$, our overall system state is 
\begin{equation}
{\bf s} (t):= \left[{\bf p}^{\top}_G (t),{\bf p}^{\top}_L (t),{\bf a}^{\top}(t)\right]^{\top}.
\label{eq.1}
\end{equation}
Being at slot $t-1$, our \emph{objective} is to leverage historical observations of states, $\left\{{\bf s} (\tau)\right\}_{\tau = 0}^{t-1}$, and pertinent costs in order to learn the optimal action for the next slot, namely ${\bf a}^{\ast}(t)$. Explicit expression of the incurred costs, and analytical formulation of the objective will be elaborated in the ensuing subsections.

\subsection{Cost functions and caching strategies}
\label{Regorously_Problem_formulation}
{Efficiency of a caching strategy will be measured by how well it utilizes the available storage of the local SB to keep the most popular files, versus how often local user requests are met via fetching through the more expensive backhaul link. The overall cost incurred will be modeled as the superposition of three types of costs.

	\begin{subequations}	
		The first type $c_{1,t}$ corresponds to the cost of refreshing the cache contents. In its  general form, $c_{1,t} (\cdot)$ is a function of the upcoming action $\mathbf{a}(t)$, and available contents at the cache according to current caching action $\mathbf{a}(t-1)$, where the subscript $t$ captures the possibility of a time-varying cost for refreshing the cache. A reasonable choice of $c_{1,t}(\cdot)$ is
		\begin{equation}
		c_{1,t}(\mathbf{a}(t),\mathbf{a}(t-1)) := \lambda_{1,t} \mathbf{a}^\top(t) \left[\mathbf{1}- \mathbf{a}(t-1)\right] 
		\label{subcost1}
		\end{equation}
		which upon recalling that the action vectors $\mathbf{a}(t-1)$ and $\mathbf{a}(t)$ have binary $\left\{0,1\right\}$ entries, implies that $c_{1,t}$ counts the number of those files to be fetched and cached  prior to slot $t$, which were not stored according to action $\mathbf{a}(t-1)$.

		The second type of cost is incurred during the operational phase of slot $t$ to satisfy user requests.
		With $c_{2,t}(\mathbf{s}(t))$ denoting this type of cost, a prudent choice must: i)~penalize requests for files already cached much less than requests for files not stored; and, ii) be a non-decreasing function of popularities $[\mathbf{p}_L]_f$. Here for simplicity, we assume that the transmission cost of cached files is relatively negligible, and choose 
		\begin{equation}
		c_{2,t}(\mathbf{s}(t)):= \lambda_{2,t} \left[\mathbf{1}-\mathbf{a}(t)\right]^\top \mathbf{p}_L(t)
		\label{subcost2}
		\end{equation}
		which solely penalizes the non-cached files in descending order of their local {popularities}.
		
		The third type of  cost captures the  ``mismatch'' between caching action $\mathbf{a}(t)$, and the global popularity profile $\mathbf{p}_{\text{G}}(t)$. Indeed, it is  reasonable to consider the global popularity of files as an acceptable representative of what the local profiles will look like in the near future; thus, keeping the caching action close to $\mathbf{p}_{\text{G}}(t)$ may reduce future possible costs. Note also that a relatively small number of local requests may only provide a crude estimate of local popularities, while the global popularity profile can serve as side information in tracking the evolution of content popularities over the network. This has prompted the advocation of transfer learning approaches, where content popularities in a surrogate domain are utilized for improving estimates of popularity; see, e.g., \cite{Transferlearning_Debbah_2015} and \cite{Learning_Poor_2016}. However, this approach is limited by the degree the surrogate (source) domain, e.g., Facebook or Twitter, is a good representative of the target domain requests. When it is not, techniques will  misguide caching decisions, while imposing excess  processing overhead to the network operator or to the SB. 
		
		To account for this issue, we introduce the third type of cost as
		\begin{equation}
		c_{3,t}(\mathbf{s}(t)):= \lambda_{3,t} \left[\mathbf{1}-\mathbf{a}(t)\right]^\top \mathbf{p}_G(t)
		\label{subcost3}
		\end{equation}
		penalizing the files not cached according to the global popularity profile  ${\bf p}_G (\cdot)$ provided by the  network operator, thus promoting adaptation of caching policies close to global demand~trends.

	\end{subequations}
	
	All in all, upon taking action $\mathbf{a}(t)$ at slot $t$, the \emph{aggregate cost conditioned} on the popularity vectors revealed, can be expressed as (cf. \eqref{subcost1}-\eqref{subcost3}) 
	\begin{align}\label{Overall_Cost}
	& C_t \Big({ {\bf s} (t-1), {\bf a} (t) \Big| {\mathbf{p}_{\text{G}}(t)},{\mathbf{p}_{\text{L}}(t)}} \Big) \\ & \nonumber  \hspace{1cm} :=   c_{1,t}\left({\bf a} (t), {\bf a} (t-1)\right) + c_{2,t}\left({\bf s} (t)\right) + c_{3,t}(\mathbf{s}(t)) \\ \nonumber &  \hspace{1.1cm}= \lambda_{1,t} \mathbf{a}^\top(t) (\mathbf{1}- \mathbf{a}(t-1)) +\lambda_{2,t} (\mathbf{1}-\mathbf{a}(t))^\top \mathbf{p}_L(t)  \\& \hspace{1.1cm}+ \lambda_{3,t} (\mathbf{1}-\mathbf{a}(t))^\top \mathbf{p}_G(t).\nonumber
	\end{align} 
	Weights $\lambda_{1,t}$, $\lambda_{2,t}$, and $\lambda_{3,t}$ control the relative significance of  the corresponding summands, whose tuning influences the optimal caching policy at the CCU. As asserted earlier, the cache-refreshing cost at off-peak periods is considered to be less than fetching the contents during  slots, which justifies the choice $\lambda_{1,t} \ll \lambda_{2,t}$. In addition, setting $\lambda_{3,t} \ll \lambda_{2,t}$ is of interest when  the local popularity profiles are of acceptable accuracy, or, if tracking local popularities is of higher importance. In particular, setting $\lambda_{3,t} = 0$ corresponds to the special case where the caching cost is  decoupled from the global popularity profile evolution. On the other hand, setting $\lambda_{2,t} \ll \lambda_{3,t}$  is desirable in networks where globally popular files are of high importance, for instance when
	users have high mobility and may change SBs rapidly, or, when a few local requests prevent the SB from estimating accurately the local popularity profiles. Fig.~\ref{fig:sys2} depicts the evolution of popularity and action vectors along with the aggregate conditional costs across slots.

\begin{figure}[t] 
	\centering
	{\includegraphics[width=0.6\columnwidth]{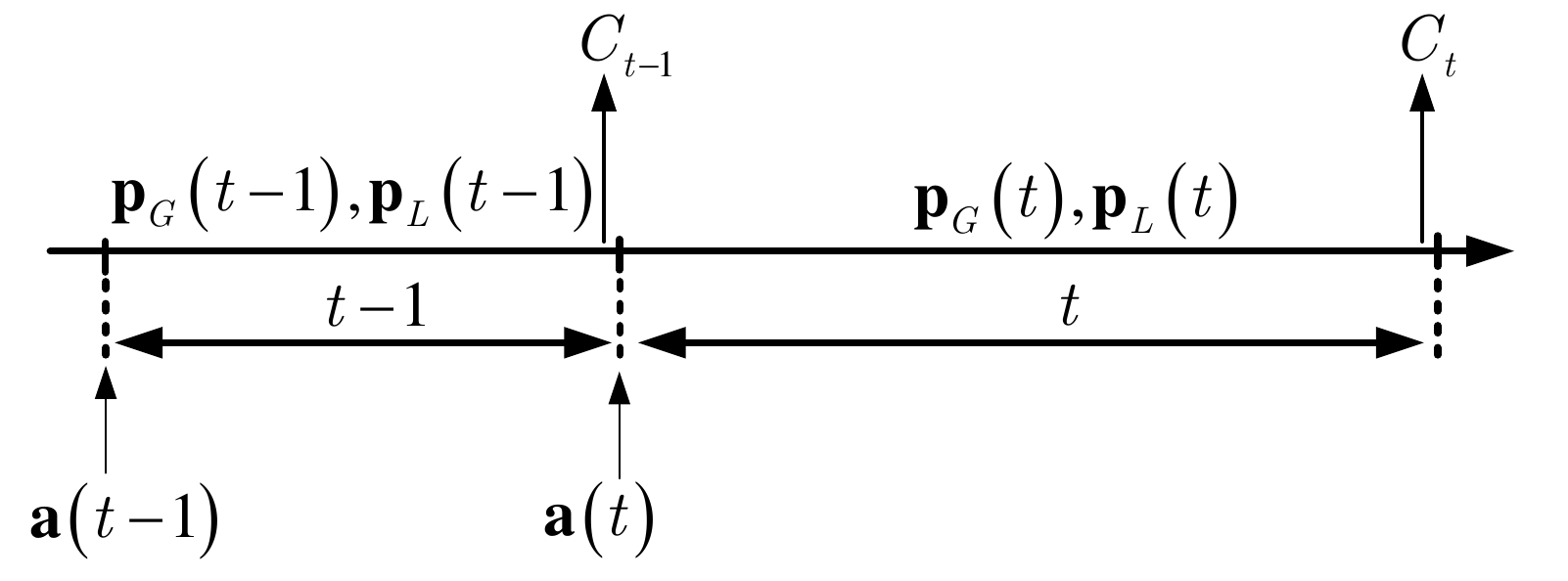}}
	\caption{A schematic depicting the evolution of key quantities across time slots. Duration of slots can be unequal.} 
	\label{fig:sys2} 	
\end{figure}

\noindent{\bf Remark 1.} As with slot sizes, proper selection of $ \lambda_{1,t}, \lambda_{2,t}$, and $\lambda_{3,t}$ is a design choice. Depending on how centralized or decentralized the network operation is desired to be, these parameters may be selected autonomously by the CCUs or provided by the network operator in a centralized fashion. However, the overall approach requires the network service provider and the SBs  to inter-operate by exchanging relevant information. On the one hand,  estimating global popularities  requires SBs to transmit their locally obtained ${\bf p}_L(t)$  to the network operator at the end of each slot. On the other hand, the network operator informs the CCUs of the global popularity ${\bf p}_G(t)$, and possibly weights $\lambda_{1,t}, \lambda_{2,t}$, and $\lambda_{3,t}$.  By providing the network operator with means of parameter selection, a
``master-slave'' hierarchy emerges, which enables the network operator (master)  to influence SBs (slaves) caching decisions, leading to a  centrally controlled adjustment of caching policies. Interestingly, these  few bytes of information exchanges occur once per slot and at  off-peak instances, thus imposing  negligible overhead to the system, while enabling a simple, yet practical and powerful optimal semi-distributed caching process; see Fig.~3.

\subsection{Popularity profile dynamics}
\label{Dynamic evolution of user demands}
As depicted in Fig.~3, we will model user requests (and thus popularities) at both global and local scales using Markov chains. Specifically,  global popularity profiles will be assumed generated by an underlying Markov process with $|{\cal P}_G|$ states collected in the set  \linebreak
${\cal P}_G :=\left\{ {\bf p}_{G}^{1}, \ldots, {\bf p}_{G}^{|{\cal P}_G|} \right\}$; and likewise for the set of all local popularity profiles 
${\cal P}_L:=\left\{ {\bf p}_{L}^{1}, \ldots, {\bf p}_{L}^{|{\cal P}_L|} \right\}$. Although ${\cal P}_G$ and ${\cal P}_L$ are known, the underlying transition probabilities of the two Markov processes are considered unknown. 

Given ${\cal P}_G$ and ${\cal P}_L$ as well as feasible caching decisions in set $\cal A$, the overall set of states in the network is 
\begin{align}
\nonumber {\cal S} := \left\{{\bf s}| {\bf s}= [{\bf p}^{\top}_G, {\bf p}^{\top}_L, {\bf a}^{\top}]^{\top}, {\bf p}_G \in {\cal P}_G  \textrm{ , } {\bf p}_L \in {\cal P}_L, {\bf a} \in {\cal A} \right\}.
\end{align} 

The lack of knowledge on transition probabilities of the underlying Markov chains  motivates well our ensuing  RL-based approach, where the learner seeks the optimal policy by interactively making sequential decisions, and observing the corresponding costs. The caching task is formulated in the following subsection,  and an efficient solver is developed to cope with the ``curse of dimensionality'' typically emerging with RL problems \cite{Sutton1998reinforcement}.

\subsection{Reinforcement learning formulation}	

As showing in Fig.~2, the CCU takes caching action $\mathbf{a}(t)$, at the beginning of slot $t$, and by the end of slot $t$, the profiles $\mathbf{p}_G(t)$ and $\mathbf{p}_L(t)$ become available, so that the system state is updated to $\mathbf{s}(t)$, and the conditional cost $C_t \Big( { {\bf s} (t-1), {\bf a} (t) {\Big |} {\mathbf{p}_{\text{G}}(t)},{\mathbf{p}_{\text{L}}(t)}} \Big)$ is revealed.	
Given the random nature of user requests locally and globally, $C_t$ in \eqref{Overall_Cost} is a random variable with mean	
\begin{align}\label{mean_Cost}
& {\overline C}_t \left( {\bf s} (t-1), {\bf a} (t) \right)   \\
& :=\mathbb{E}_{{\mathbf{p}_{\text{G}}(t)},{\mathbf{p}_{\text{L}}(t)}} \Big[ C_t \left( { {\bf s} (t-1), {\bf a} (t) {\Big |} {\mathbf{p}_{\text{G}}(t)},{\mathbf{p}_{\text{L}}(t)}} \right) \Big] \nonumber \\
&  = \lambda_1 \mathbf{a}^\top(t) \left[\mathbf{1}- \mathbf{a}(t-1)\right] +\lambda_2 \mathbb{E} \left[ (\mathbf{1}-\mathbf{a}(t))^\top \mathbf{p}_L(t)\right] \nonumber \\  & + \lambda_3 \mathbb{E} \left[(\mathbf{1}-\mathbf{a}(t))^\top \mathbf{p}_G(t)\right] \nonumber
\end{align}
where the expectation is taken with respect to (wrt)  ${\bf p}_L (t)$ and ${\bf p}_G (t)$, while the weights are selected as $\lambda_{1,t} = \lambda_1$, $\lambda_{2,t} = \lambda_2$, and $\lambda_{3,t} = \lambda_3$ for simplicity.% and previous actions $\{{\bf a}\left[t-1\right]\}_{t=1}^\infty$, which are implicitly coupled with popularity profiles.

Let us now define the policy function $\pi: \mathcal{S} \rightarrow \mathcal{A}$, which maps any state $\mathbf{s} \in  \mathcal{S}$ to the action set. Under policy $\pi(\cdot)$, for the current state ${\bf s}(t)$,  caching is carried out via action $\mathbf{a}(t+1)=\pi(\mathbf{s}(t))$  dictating what files to be stored for the $(t+1)$-st slot. Caching performance is measured through the so-termed state value function 
\begin{align}\label{eq_Value_function}
V_{\pi} \left({\bf s} (t)\right) :=   \lim\limits_{T \rightarrow \infty} \mathbb{E}  \left[\sum\limits_{\tau = t}^{T} \gamma^{\tau-t} {\overline C}\left({\bf s} \left[\tau\right], {\pi} \left(\mathbf{s}\left[\tau\right]\right) \right)\right] 
\end{align}
which is the total average cost incurred over an infinite time horizon, with future terms discounted by factor  $\gamma \in \left[0,1\right)$. Since taking action ${\bf a}(t)$ influences the SB state in future slots, future costs are always affected by past and present actions. Discount factor $\gamma$ captures this effect, whose tuning trades  off current versus future costs. Moreover, $\gamma$ also accounts for modeling uncertainties, as well as imperfections, or dynamics. For instance, if there is ambiguity about future costs, or if the system changes very fast, setting  $\gamma$ to a small value enables one to prioritize current costs, whereas in a stationary setting one may prefer to demote future costs through a larger $\gamma$. 

The objective of this paper is to find the optimal policy $\pi^*$ such that  the average cost of any state ${\bf s}$ is minimized  (cf. \eqref{eq_Value_function}) 
\begin{equation}
\label{eq_Opt_policy}
{\pi^{\ast}} = \arg \min \limits_{\pi \in \Pi} V_{\pi} \left( {\bf s} \right), \quad \forall {\bf s} \in {\cal S}
\end{equation} where $\Pi$ denotes the set of all feasible policies.

The  optimization in  \eqref{eq_Opt_policy} is a sequential decision making problem. In the ensuing section, we present optimality conditions (known as Bellman equations) for our problem, and  introduce a~{Q-learning} approach for solving \eqref{eq_Opt_policy}.}

\section{Optimality conditions}
\label{Bellman}
Bellman equations, also known as dynamic programming equations, provide necessary conditions for optimality of a policy in a sequential decision making problem. Being at the $(t-1)$st slot, let  $[{\bm P}^a]_{{\bf s}{\bf s}'}$ denote the transition probability of going from the current state ${\bf s}$ to the next state ${\bf s}'$ under action ${\bf a}$; that is, \[ [{\bm P}^a]_{{\bf s}{\bf s}'} := {\rm{Pr}} \Big \{{{\bf s}(t) = {\bf s}'}  {\Big |} {\bf s}(t-1) = {\bf s}, \pi(\mathbf{s}(t-1)) = \mathbf{a}\Big \}. \]
Bellman equations express the state value function by \eqref{eq_Value_function} in a recursive fashion as \cite[pg. 47]{Sutton1998reinforcement}
\begin{equation}\label{eq_Value_function_recurring}
V_{\pi} \left({\bf s} \right) = { \overline{C}\left({\bf s}, \pi(\mathbf{s})  \right)  }+ \gamma  \sum_{{\bf s}' \in \mathcal{S}} [{\bm P}^{\pi({\bf s})}]_{{\bf s}{\bf s}'}   V_{\pi} \left({\bf s}' \right)\;, \forall {\bf s,s}'
\end{equation}
which amounts to the superposition of $\overline C$ plus a discounted version of future state value functions under a given policy $\pi$. Specifically, after dropping the current slot index $t-1$ and indicating with prime quantities of the next slot $t$,  ${ \overline{C}}$ in \eqref{mean_Cost} can be written as
\begin{align*}
\label{cbardef}
{ \overline{C}\left({\bf s}, \pi(\mathbf{s})  \right)  } = \sum_{\bf s'  := [\bf p_G',\bf p_L',\bf a']\in \mathcal{S}}  [{\bm P}^{\pi({\bf s})}]_{{\bf s}{\bf s}'}  C\Big({\bf s}, \pi(\mathbf{s}) \Big| \bf p_G',  \bf p_L'   \Big)  
\end{align*}
where $ C\Big({\bf s}, \pi(\mathbf{s}) \Big| \bf p_G',  \bf p_L'   \Big)$ is found as in \eqref{Overall_Cost}.
It turns out that, with $[{\bm P}^a]_{{\bf s}{\bf s}'}$ given $\forall {\bf s,s}'$, one can readily obtain $\left\{V_{\pi} ({\bf s}), \forall {\bf s} \right\}$ by solving \eqref{eq_Value_function_recurring}, and eventually the optimal policy $\pi^{\ast}$ in \eqref{pi*} using the so-termed policy iteration algorithm \cite[pg. 79]{Sutton1998reinforcement}. To outline how this algorithm works in our context, define the state-action value function that we will rely on under policy $\pi$ \cite[pg. 62]{Sutton1998reinforcement}
\begin{equation}
\label{Q_pi}
Q_{\pi} \left({\bf s} , {{\bf a}'}\right) :=    { \overline{C}\left({\bf s},  {\bf a'}  \right)}  +  \gamma  \sum_{{\bf s}' \in \mathcal{S}} [{\bm P}^{{\bf a'}}]_{{\bf s}{\bf s}'}   V_{\pi} \left({\bf s}' \right). 
\end{equation}
Commonly referred to as the ``Q-function,'' $Q_{\pi}({\bf s},{\boldsymbol \alpha})$ basically captures the expected current cost of taking action ${\boldsymbol \alpha}$ when the system is in state $\bf s$, followed by the discounted value of the future states, provided that the future actions are taken according to policy~$\pi$.

In our setting, the policy iteration algorithm initialized with $\pi_0$, proceeds with the following updates at the  $i$th iteration.
\begin{itemize}
	\item \textbf{Policy evaluation}: Determine $V_{\pi_i}(\bf s)$ for all states ${\bf s} \in {\cal S}$ under the current (fixed) policy $\pi_i$, by solving the system of linear equations in \eqref{eq_Value_function_recurring} $\forall {\bf s}$.
	\item \textbf{Policy update}: Update the policy using
	\[ \pi_{i+1} ({\bf s}) := \arg \max_{{\boldsymbol \alpha}} Q_{\pi_{i}} ({\bf s},{\boldsymbol \alpha}), \quad \forall {\bf s} \in {\cal S}. \]
\end{itemize}

The policy evaluation step is of complexity~$\mathcal{O}(|\mathcal{S}|^3)$, since it requires  matrix inversion for solving  the linear system of equations in \eqref{eq_Value_function_recurring}. Furthermore, given $V_{\pi_i}(\bf s) \; \forall \bf s$, the complexity of the policy update step is  $\mathcal{O}(|{\cal A}||\mathcal{S}|^2)$, since the Q-values must be updated per state-action pair, each subject to~$|\cal S|$ operations;  see also \eqref{Q_pi}.
Thus, the per iteration complexity of the policy iteration algorithm is $\mathcal{O}(|\mathcal{S}|^3+|{\cal A}||\mathcal{S}|^2)$. Iterations proceed until convergence, i.e., $\pi_{i+1}({\bf s})={\pi_i({\bf s})}, \, \forall {\bf s}\in {\cal S}$.

Clearly, the policy iteration algorithm relies on knowing $[{\bm P}^{\bf a}]_{{\bf s}{\bf s}'}$, which is typically not available in practice. This motivates the use of adaptive dynamic programming (ADP) that learn $[{\bm P}^{\bf a}]_{{\bf s}{\bf s}'}$ for all ${\bf s, s}' \in \mathcal{S}$, and ${\bf a} \in \mathcal{A}$, as iterations proceed \cite[pg. 834]{AIModernapproach}. Unfortunately, ADP algorithms are often very slow and impractical, as they must estimate $|\mathcal{S} |^2 \times |\mathcal{A}|$ probabilities. In contrast, the Q-learning algorithm elaborated next finds the optimal $\pi^{*}$ as well as $V_{\pi}({\bf s})$, while circumventing the need to estimate $[{\bm P}^a]_{{\bf s}{\bf s}'}, \forall {\bf s,s}'$; see e.g., \cite[pg.~140]{Sutton1998reinforcement}. 

\subsection{Optimal caching via Q-learning}
\label{Q-learning}
Q-learning is an online RL scheme to jointly infer the optimal policy $\pi^{\ast}$, and estimate the optimal state-action value function  $Q^*(\mathbf{s,a}')  := Q_{\pi^{\ast}}(\mathbf{s,a}') \quad \forall {\bf s},{\bf a}'$. Utilizing \eqref{eq_Value_function_recurring} for the optimal policy $\pi^{\ast}$, it can be shown that \cite[pg. 67]{Sutton1998reinforcement} \linebreak
\begin{equation}\label{pi*}
\pi^*({\bf s}) = \arg\min_{{\boldsymbol \alpha}}~Q^{*}({\bf s}, {\boldsymbol \alpha}), \quad \forall {\bf s} \in {\cal S}.
\end{equation}
The Q-function and  $V(\cdot)$ under $\pi^{\ast}$ are related by
\begin{equation}\label{v*}
V^{*}({\bf s}):= V_{\pi^*}({\bf s}) =\min_{{\boldsymbol \alpha}} Q^{*}({\bf s}, {\boldsymbol \alpha})
\end{equation}
which in turn yields 
\begin{equation}
\label{eq_Q_function2}
Q^* \left({\bf s} , {\bf a}'\right) =  \overline{C}\left({\bf s}, {\bf a}'  \right) +  \gamma  \sum_{{\bf s}' \in \mathcal{S}} [{\bm P}^{\bf a}]_{{\bf s}{\bf s}'}  \min_{{\boldsymbol \alpha}\in {\cal A}} Q^* \left({\bf s'} , {\boldsymbol \alpha}\right).
\end{equation} 

Capitalizing on the optimality conditions \eqref{pi*}-\eqref{eq_Q_function2}, an online Q-learning scheme for caching is listed under  Alg.~1. 
In this algorithm, the agent updates its estimated $\hat{Q}(\mathbf{s}(t-1),\mathbf{a}(t))$ as  $C\Big( {{\bf s} (t-1), {\bf a} (t)} \Big |  {\mathbf{p}_{\text{G}}(t)},{\mathbf{p}_{\text{L}}(t)} \Big)$ is observed. That is, given ${\bf s}(t-1)$, Q-learning takes action ${\mathbf{a}}(t)$, and  upon observing ${\mathbf{s}}(t)$, it incurs cost $C\Big( {{\bf s} (t-1), {\bf a} (t)} \Big|  {\mathbf{p}_{\text{G}}(t)},{\mathbf{p}_{\text{L}}(t)} \Big)$.  Based on the instantaneous error 
\begin{align}\label{error}
\varepsilon \left({\bf s}(t-1), {\bf a}(t)\right) :=  & \frac{1}{2} \Big( C\left({\bf s}(t-1),{\bf a}(t)\right) + \gamma \min \limits_{{\boldsymbol \alpha}}^{} {\widehat Q} \left({\bf s}(t), {\boldsymbol \alpha}\right)  \nonumber\\ & \hspace{0.4cm} - {\widehat Q} \left({\bf s}(t-1),{\bf a}(t)\right) \Big)^2 
\end{align} 
the Q-function is updated using stochastic gradient descent as 
\begin{align}
&\hat{Q}_t\left({\bf s}(t-1),{\bf a}(t)\right) = (1-\beta_t) \hat{Q}_{t-1}\left({\bf s}(t-1),{\bf a}(t)\right)   \nonumber+  \\ 
\nonumber
&  \beta_t  \Big[C\left( {{\bf s} (t-1), {\bf a} (t)} {\Big |}  {\mathbf{p}_{\text{G}}(t)},{\mathbf{p}_{\text{L}}(t)} \right) + \gamma \min_{{\boldsymbol \alpha}} \hat{Q}_{t-1}\left({\bf s}(t),{{\boldsymbol \alpha}}\right) \Big] \nonumber 
\end{align}
while keeping the rest of the entries in $\hat{Q}_t(\cdot,\cdot)$ unchanged.
\begin{algorithm}[t]
	\caption{Caching via Q-learning at CCU}
	\label{alg:QR}
		%% 	\SetKwData{Left}{left}\SetKwData{This}{this}\SetKwData{Up}{up}
		% %	\SetKwFunction{Union}{Union}\SetKwFunction{FindCompress}{FindCompress}
		% %	\SetKwInOut{Input}{input}\SetKwInOut{Output}{output}
		% %	\BlankLine
		{{\bf Initialize}  $\mathbf{s}(0)$ randomly and $\hat{Q}_0(\mathbf{s,a}) = 0 \; \forall \mathbf{s,a}$} 

	    {For $t = 1,2,... $}
		
		{Take action ${\bf a}(t) $ chosen probabilistically by \[{\bf a}(t)  = \left\{
			\begin{array}{ll}
			\arg \min \limits_{\bf a} {\hat Q}_{t-1}\left({\bf s} (t-1),{\bf a}\right) & \textrm{w.p. }\;\; 1-\epsilon_t \\
			\textrm{random } \mathbf{a} \in \mathcal{A}  & \textrm{w.p.} \;\; \; \epsilon_t
			\end{array}
			\right. \]  }
		
		 ${\bf p}_L (t)$ and ${\bf p}_G (t)$ are revealed based on user requests
		 \newline
			Set ${\bf s} (t) := \left[{\bf p}_G^{\top} (t), {\bf p}_L^{\top} (t) , {\bf a}(t)^{\top}\right]^{\top}$
		
			Incur cost $C\Big({ {\bf s} (t-1), {\bf a} (t) {\Big |} {\mathbf{p}_{\text{G}}(t)},{\mathbf{p}_{\text{L}}(t)}}\Big)$ 
			Update \hfill
		\begin{align}
		\hspace{0.6cm}	&\hat{Q}_t\left({\bf s}(t-1),{\bf a}(t)\right) = (1-\beta_t) \hat{Q}_{t-1}\left({\bf s}(t-1),{\bf a}(t)\right)   \nonumber \\ 
		\nonumber
		&\hspace{2cm} +  \beta_t  \Big[C\left( {{\bf s} (t-1), {\bf a} (t)} {\Big |}  {\mathbf{p}_{\text{G}}(t)},{\mathbf{p}_{\text{L}}(t)} \right) \\ &\hspace{4cm} + \gamma \min_{{\boldsymbol \alpha}} \hat{Q}_{t-1}\left({\bf s}(t),{{\boldsymbol \alpha}}\right) \Big]
		\end{align}
\end{algorithm}

Regarding convergence of the Q-learning algorithm, a necessary condition ensuring ${\hat Q}_{t}\left(\cdot,\cdot\right) \rightarrow {Q}^{*}\left(\cdot,\cdot\right)$, is  that  all state-action pairs must be continuously updated. Under this and the usual stochastic approximation conditions that will be specified later, ${\hat Q}_t \left(\cdot,\cdot\right)$ converges to ${Q}^{*} \left(\cdot,\cdot\right)$ with probability~$1$; see~\cite{Stochastic_app_and_reinforcement_learning} for a detailed description.

To meet the requirement for continuous updates, Q-learning utilizes a probabilistic exploration-exploitation approach to selecting actions. At slot $t$, exploitation happens with probability $1-\epsilon_t$ through the action ${\bf a}(t) = \arg\min_{\boldsymbol \alpha \in {\cal A}} \hat{Q}_{t-1}(\mathbf{s}(t-1),{\boldsymbol \alpha})$, while the exploration happens with probability $\epsilon_t$ through a random action $\mathbf{a}\in \mathcal{A}$.  Parameter $\epsilon_t$ trades off exploration for exploitation, and its proper selection guarantees a necessary condition for convergence. During initial iterations or when the CCU observes considerable shifts in content popularities, setting $\epsilon_t$ high promotes exploration in order to learn the underlying dynamics. On the other hand, in stationary settings and once ``enough'' observations are made, small values of $\epsilon_t$ promote exploiting the learned $\hat{Q}_{t-1}(\mathbf{\cdot,\cdot})$ by taking the estimated optimal action  $\arg\min _{\boldsymbol \alpha} {\hat Q}_{t-1}\left({\bf s} (t),{\boldsymbol \alpha}\right)$. 

Regarding stochastic approximation conditions,
the stepsize sequence $\{\beta_t\}_{t=1}^\infty$ must obey $\sum_{t=1}^{\infty} \beta_t = \infty$ and $\sum_{t=1}^{\infty} \beta_t^2 < \infty$~\cite{Stochastic_app_and_reinforcement_learning}, both of which are satisfied by e.g., $\beta_t = 1 \mathbin{/} t$. 
However, with a selection  of constant stepsize $\beta_t=\beta$, the mean-square error (MSE) of ${\hat Q}_{t+1}(\cdot,\cdot)$ is bounded as (cf. \cite{Stochastic_app_and_reinforcement_learning})
\begin{equation}
\label{eq_Qhat_Q}
\mathbb{E} \left[ \left\|{\hat Q_{t+1}}-Q^{*} \right\|_{F}^2 \left| {\hat Q_0} \right. \right] \le \varphi_1\left(\beta\right) +  \varphi_2({\hat Q}_0) \exp\left(-2\beta t\right) 
\end{equation}
where $\varphi_1 \left( \beta \right)$ is a positive and increasing function of $\beta$; while the second term denotes the initialization error, which decays exponentially as the iterations proceed.

Although selection of a constant stepsize prevents the algorithm from exact convergence to  $Q^*$ in stationary settings, it enables CCU adaptation to the underlying non-stationary Markov processes in dynamic scenaria. Furthermore, the optimal policy in practice can be obtained from the Q-function values before convergence is achieved \cite[pg. 79]{Sutton1998reinforcement}.

However, the main practical limitation of the Q-learning algorithm is its slow convergence, which is a consequence of independent updates of the Q-function values. Indeed, Q-function values are related, and leveraging these relationships can lead to multiple updates per observation as well as faster convergence. In the ensuing section, the structure of the problem at hand will be exploited  to develop a linear function approximation of the Q-function, which in turn  will endow our algorithm not only with fast convergence, but also with scalability.

\section{Scalable caching}
\label{Linear Function approximation}
Despite simplicity of the updates as well as optimality guarantees of the Q-learning algorithm, its applicability over real networks faces practical challenges. Specifically, the Q-table is of size $|\mathcal{P}_G| |\mathcal{P}_L| |\mathcal{A}|^2$, where  $|\mathcal{A}|={F \choose M} $ encompasses all possible selections of $M$ from $F$ files. Thus, the Q-table size grows prohibitively with $F$, rendering convergence of the table entries, as well as the policy iterates unacceptably slow. Furthermore, action selection in $\min_{{\boldsymbol \alpha} \in {\cal A}} Q({\bf s},{\bf a})$ entails an expensive exhaustive search over the feasible action set $\mathcal{A}$. 

Linear function approximation is  a popular scheme for rendering Q-learning applicable to real-world settings \cite{geramifard2013tutorial,mahadevan2009learning,AIModernapproach}.  % to learn the optimal policy in a Markovian setting, 
A linear approximation for $Q(\mathbf{s},\mathbf{a})$ in our setup is inspired by the additive form of the instantaneous costs in \eqref{Overall_Cost}. Specifically, we propose to approximate $Q(\mathbf{s},\mathbf{a}')$ as  
\begin{equation}
\label{approximation}
Q(\mathbf{s},\mathbf{a}')\simeq Q_G(\mathbf{s},\mathbf{a}')+Q_L(\mathbf{s},\mathbf{a}')+Q_R(\mathbf{s},\mathbf{a}')
\end{equation}
where $Q_G$, $Q_L$, and $Q_R$ correspond to global and local popularity mismatch, and cache-refreshing costs, respectively.

Recall that the state vector $\mathbf{s}$ consists of three subvectors, namely $\mathbf{s} := [{\mathbf p}^{\top}_G, {\mathbf p}^{\top}_L,\mathbf{a}^{\top}]^{\top}$. Corresponding to the global popularity subvector, our first term of the approximation in \eqref{approximation} is 
\begin{align}
\label{Q_G_est_1}
Q_G(\mathbf{s},\mathbf{a}'):= \sum_{i=1}^{|\mathcal{P}_G|}\sum_{f=1}^F \theta^{G}_{i,f} \bf{1}_{\left\{\mathbf{p}_G={\bf p}^{i}_{G}\right\}} \bf{1}_{\left\{[{\bf a}']_f=0\right\}}
\end{align}	
where the sums are over all possible global popularity profiles as well as  files, and the indicator function ${\bf{1}}_{\left\{ \cdot \right\}}$ takes value $1$ if its argument holds, and $0$ otherwise; while $\theta^{G}_{i,f}$ captures the average ``overall'' cost if the system is in global state ${\bf p}^{i}_G$, and the CCU decides not to cache the $f$th content. By defining the  $|{\cal P}_G| \times |\cal F|$ matrix with $(i,f)$-th entry $\left[{\boldsymbol \Theta}^G\right]_{i,f} := \theta^{G}_{i,f}$, one can rewrite \eqref{Q_G_est_1} as 
\begin{equation}
\label{Matrix_G}
Q_G(\mathbf{s},\mathbf{a}')= \boldsymbol{\delta}_G^{\top}({\bf p_G}) \boldsymbol{\Theta}^G (\mathbf{1}-{\bf a}')
\end{equation}  
where 
\[ {\boldsymbol \delta}_G ({\bf p}_G) := \left[\delta({\bf p}_G-{\bf p}^{1}_{G}), \ldots, \delta({\bf p}_G-{\bf p}^{|{\cal P}_G|}_{G})\right]^{\top}.\]

Similarly, we advocated the second summand in the approximation  \eqref{approximation} to be
\begin{align}
\nonumber
Q_L(\mathbf{s},\mathbf{a}')&:= \sum_{i=1}^{|\mathcal{P}_L|}\sum_{f=1}^F \theta^{L}_{i,f} \bf{1}_{\left\{\mathbf{p}_L={\bf p}^{i}_{L}\right\}} \bf{1}_{\left\{[{\bf a}']_f=0\right\}} \\ \label{Q_L_est_2} &   = \boldsymbol{\delta}^\top_{L} ({{\bf p}_L}) \boldsymbol{\Theta}^L (\mathbf{1}-{\bf a}')
\end{align}
where $\left[{\boldsymbol \Theta}^L\right]_{i,f} := \theta^{L}_{i,f}$, and  
\[ {\boldsymbol \delta}_L ({\bf p}_L) := \left[\delta({\bf p}_L-{\bf p}^{1}_{L}), \ldots, \delta({\bf p}_L-{\bf p}^{|{\cal P}_L|}_{L})\right]^{\top}\]
with $\theta^{L}_{i,f}$ modeling the average overall cost for not caching file $f$ when the local popularity is in state ${\bf p}^i_L$.

Finally, our third summand in \eqref{approximation} corresponds to the cache-refreshing cost
\begin{align}
Q_R(\mathbf{s},\mathbf{a}'):&= \sum_{f=1}^F \theta^{R} \bf{1}_{\left\{[{\bf a}']_f=1\right\}} \bf{1}_{\left\{[\bf a]_f=0\right\}} \\ & = \theta^{R} {\bf a}'^{\top} \left(1-{\bf a}\right)  \nonumber \\ &= \theta^{R} \left[ {\bf a}'^{\top} \left(1-{\bf a}\right) +{\bf a}^{\top} {\bf 1} - {\bf a}'^{\top} {\bf 1}\right] \nonumber
\\ &   = \theta^{R} {\bf a}^{\top} (\mathbf{1}-{\bf a}') \nonumber 
\label{eq.Qr}
\end{align}
where $\theta^R$ models average cache-refreshing cost per content. The constraint {$\mathbf{a}^\top \mathbf{1}= \mathbf{a}'^\top \mathbf{1} = M$},
is utilized to factor out the term $\bf{1-a}'$, which will become useful later.

Upon defining the set of parameters  $\Lambda := \{\boldsymbol{\Theta}^G,\boldsymbol{\Theta}^L, \theta^a\}$, the Q-function is readily approximated (cf. \eqref{approximation}) 
\begin{equation}
\label{eq.app}
{\widehat Q}_{\Lambda}(\bf s,{\bf a}'):= {\Big( \boldsymbol{\delta}_G^{\top}({\bf p_G}) \boldsymbol{\Theta}^G +  \boldsymbol{\delta}_L^{\top}({\bf p_L}) \boldsymbol{\Theta}^L    + \theta^{R} {\bf a}^{\top} \Big)} (\mathbf{1}-{{\bf a}'}).
\end{equation}
Thus, the original task of learning $|{\cal P}_G| |{\cal P}_L| |{\cal A}|^2$ parameters  in Alg.~1 is now reduced to learning $\Lambda$ containing $\left( \left|{\cal P}_G\right| + \left|{\cal P}_L\right| \right) \left|{\cal F}\right|  +1$ parameters.

\subsection{Learning $\Lambda$}
Given the current parameter estimates  $\{\widehat{\boldsymbol{\Theta}}_{t-1}^G,\widehat{\boldsymbol{\Theta}}_{t-1}^L, \hat{\theta}_{t-1}^R\}$ at the end of slot $t$, the instantaneous error is given by  
\begin{align}\label{error}
\widehat\varepsilon & \left({\bf s}(t-1), {\bf a}(t)\right) :=  \frac{1}{2} \Big( C\left({\bf s}(t-1),{\bf a}(t)\right)  \\ & \hspace{+.7cm }+ \gamma \min \limits_{{\bf a}'}^{} {\widehat Q}_{{\Lambda_{t-1}}} \left({\bf s}(t), {\bf a}'\right) - {\widehat Q}_{{\Lambda_{t-1}}} \left({\bf s}(t-1),{\bf a}(t)\right) \Big)^2. \nonumber 
\end{align} 

Based on this error form, the parameter update rules are obtained using stochastic gradient descent iterations as 
\begin{align}
\label{UpdatethetaG}
& \hat{\boldsymbol{\Theta}}_t^G = \hat{\boldsymbol{\Theta}}_{t-1}^G - \alpha_G \nabla_{{\boldsymbol{\Theta}}^G} \widehat\varepsilon \left({\bf s}(t-1), {\bf a}(t)\right) 
\\ & = \hat{\boldsymbol{\Theta}}_{t-1}^G + \alpha_G \sqrt{\widehat\varepsilon \left({\bf s}(t-1), {\bf a}(t)\right) } \; \;  \nabla_{{\boldsymbol \Theta}^{G}} {\widehat Q}_{{\Lambda_{t-1}}} ({\bf s}(t-1),{\bf a}(t))\nonumber \\  &
= \hat{\boldsymbol{\Theta}}_{t-1}^G + \alpha_G \sqrt{\widehat\varepsilon \left({\bf s}(t-1), {\bf a}(t)\right)} \; \;   \boldsymbol{\delta }_G ({\bf p_G(t-1)}) (\mathbf{1}-\bf a(t))^\top \nonumber
\end{align} 
\begin{align}
\label{UpdatethetaL}
& \hat{\boldsymbol{\Theta}}_t^L = \hat{\boldsymbol{\Theta}}_{t-1}^L - \alpha_L \nabla_{{\boldsymbol{\Theta}}^L} \widehat\varepsilon \left({\bf s}(t-1), {\bf a}(t)\right) 
\\ & = \hat{\boldsymbol{\Theta}}_{t-1}^L + \alpha_L \sqrt{\widehat\varepsilon \left({\bf s}(t-1), {\bf a}(t)\right) } \; \;  \nabla_{{\boldsymbol \Theta}^{L}} {\widehat Q}_{{\Lambda_{t-1}}} ({\bf s}(t-1),{\bf a}(t))\nonumber \\  
& = \hat{\boldsymbol{\Theta}}_{t-1}^L + \alpha_L \sqrt{\widehat\varepsilon \left({\bf s}(t-1), {\bf a}(t)\right)} \; \;   \boldsymbol{\delta }_L ({\bf p_L(t-1)}) (\mathbf{1}-\bf a(t))^\top \nonumber
\end{align} 
and 
\begin{align}
\label{UpdatethetaR}
\hat{{\theta}}_t^R & = \hat{{\theta}}_{t-1}^R - \alpha_R \nabla_{{{\theta}}^R} \widehat\varepsilon \left({\bf s}[t-1], {\bf a}[t]\right)
\\ &= \hat{{\theta}}_{t-1}^R + \alpha_R \sqrt{\widehat\varepsilon \left({\bf s}[t-1], {\bf a}[t]\right)} \; \;  \nabla_{{\theta}^R} {\widehat Q}_{\Lambda_{t-1}} ({\bf s}\left[t-1\right],{\bf a}\left[t\right])\nonumber \\  
&= \hat{{\theta}}_{t-1}^R + \alpha_R \sqrt{\widehat\varepsilon\left({\bf s}[t-1], {\bf a}[t]\right)} \; \;  {\bf a}^{\top}[t-1] (\mathbf{1}-{\bf a}[t]) \nonumber.
\end{align}

The pseudocode for this scalable approximation of the Q-learning scheme is tabulated in Alg.~2.

\begin{algorithm}[t]
	\caption{Scalable Q-learning}
	\label{LFAQ}
		{{\bf Initialize}  $\mathbf{s}(0)$ randomly, ${\widehat{\boldsymbol \Theta}_0^G} = {\bf 0}$, ${\widehat{\boldsymbol \Theta}_0^L} = {\bf 0}$, ${\hat \theta}_0^R = 0$, and thus $\widehat{\boldsymbol{\psi}}(\bf s) = {\bf 0}$} 
		\newline
		%% 	\SetKwData{Left}{left}\SetKwData{This}{this}\SetKwData{Up}{up}
		% %	\SetKwFunction{Union}{Union}\SetKwFunction{FindCompress}{FindCompress}
		% %	\SetKwInOut{Input}{input}\SetKwInOut{Output}{output}
		% %	\BlankLine
%		\State	{\bf Initialize}  $\mathbf{s}(0)$ randomly, ${\widehat{\boldsymbol \Theta}_0^G} = {\bf 0}$, ${\widehat{\boldsymbol \Theta}_0^L} = {\bf 0}$, ${\hat \theta}_0^R = 0$, and thus $\widehat{\boldsymbol{\psi}}(\bf s) = {\bf 0}$ \\
	   {For $t=1,2,...$}
		\newline
		{Take action ${\bf a}(t) $ chosen probabilistically by \[{\bf a}(t)  = \left\{
			\begin{array}{ll}
			\text {$M$ best files via ${\widehat{\boldsymbol \psi} \left({\bf s}(t-1)\right)}$} & \textrm{w.p. }\;\; 1-\epsilon_t \\
			\textrm{random } \mathbf{a} \in \mathcal{A}  & \textrm{w.p.} \;\; \; \epsilon_t
			\end{array}
			\right. \]  }
		
		 ${\bf p}_G (t)$ and ${\bf p}_L (t)$ are revealed based on user requests
		 \newline
		Set \quad  \hspace{0.9 cm} ${\bf s} (t) := \left[{\bf p}_G^{\top} (t), {\bf p}_L^{\top} (t) , {\bf a}(t)^{\top}\right]^{\top}$
		
		Incur cost \quad $C\Big({ {\bf s} (t-1), {\bf a} (t) {\Big |} {\mathbf{p}_{\text{G}}(t)},{\mathbf{p}_{\text{L}}(t)}}\Big)$ 
		\newline
		 Find \hspace{1 cm} $\widehat \varepsilon \left({\bf s}(t-1),{\bf a}(t)\right)$
		Update \quad \hspace{0.3cm} ${\widehat{\boldsymbol{\Theta}}}_t^G$, ${\widehat{\boldsymbol{\Theta}}}_t^L$ and ${\hat \theta}_t^R$ based on \eqref{UpdatethetaG}-\eqref{UpdatethetaR}
\end{algorithm}                

The upshot of this scalable scheme is three-fold. 
\begin{itemize}
	\item The large state-action  space in the Q-learning algorithm is handled by reducing the number of parameters from $|{\cal P}_G| |{\cal P}_L| |{\cal A}|^2$ to  $\left( \left|{\cal P}_G\right| + \left|{\cal P}_L\right| \right) \left|{\cal F}\right| +1$.
	\item In contrast to single-entry updates in the exact Q-learning Alg.~1, $F-M$ entries in  $\widehat{\boldsymbol\Theta}^G$ and $\widehat{\boldsymbol\Theta}^L$ as well as $\theta^R$, are updated per observation using \eqref{UpdatethetaG}-\eqref{UpdatethetaR}, which leads to a much faster convergence.
	\item The exhaustive search in $ \min \limits_{{\bf a} \in {\cal A}} Q \left({\bf s},{\bf a}\right)$ required in exploitation; and also in the error evaluation \eqref{error}, is circumvented. Specifically, it holds that (cf.~\eqref{eq.app}) \begin{equation}
	\label{a}
	\min \limits_{{\bf a}' \in {\cal A}} Q({\bf s},{\bf a}') \approx \min \limits_{{\bf a}' \in {\cal A}} {\boldsymbol \psi}^{\top} ({\bf s}) \left({\bf 1- a}' \right) = \max \limits_{{\bf a}'\in {\cal A}} {\boldsymbol \psi}^{\top}({\bf s}) \, \, {\bf a}'
	\end{equation}
	where \[{\boldsymbol{ \psi} ({\bf s})} := \boldsymbol{\delta}_G^{\top}({\bf p_G}) \boldsymbol{\Theta}^G +  \boldsymbol{\delta}_L^{\top}({\bf p_L}) \boldsymbol{\Theta}^L    + \theta^{R} {\bf a}^{\top}. \] 
	The solution of \eqref{a} is readily given by $[{\bf a}]_{\nu_i} = 1$ \linebreak for $i=1, \ldots, M$, and $[{\bf a}]_{\nu_i} = 0$ for $i>M$, where $\left[{{\boldsymbol{ \psi} ({\bf s})}}\right]_{\nu_F} \le \cdots \le  \left[{{\boldsymbol{ \psi} ({\bf s})}}\right]_{\nu_1} $ are sorted entries of ${\boldsymbol{ \psi} ({\bf s})}$.  
\end{itemize}

\noindent{\bf Remark 2.}
In the model of Sec. II-B, the state-space cardinality of the popularity vectors is finite. These vectors can be viewed as centroids of quantization regions partitioning a state space of infinite cardinality. Clearly, such a partitioning  inherently bears a complexity-accuracy trade off, motivating  optimal designs to achieve a desirable accuracy for a given affordable complexity. This is one of our future research directions for the  problem at hand.

Simulation based evaluation  of the proposed algorithms for RL-based caching is now in order.

%\noindent{\bf Remark 2.}
%As it is considered in our modeling, the local and global states are predefined and known at both network operator and CCU sides. However, as time passes and new user demands are revealed, if the network operator observes a new and popularity profile that cannot be well represented by the currently available ones, it can add this new profile to its states, and then inform the CCUs.  The CCUs will be informed with this newly emerged profile  via the message passing procedure happing during off-peak periods. This approach will enable expanding or changing the representing popularity profiles of the network, and is of paramount important. 

\section{Numerical tests}

\begin{figure}[t]
	\centering
	\begin{subfigure}
		\centering
		\includegraphics[width=0.42\columnwidth]{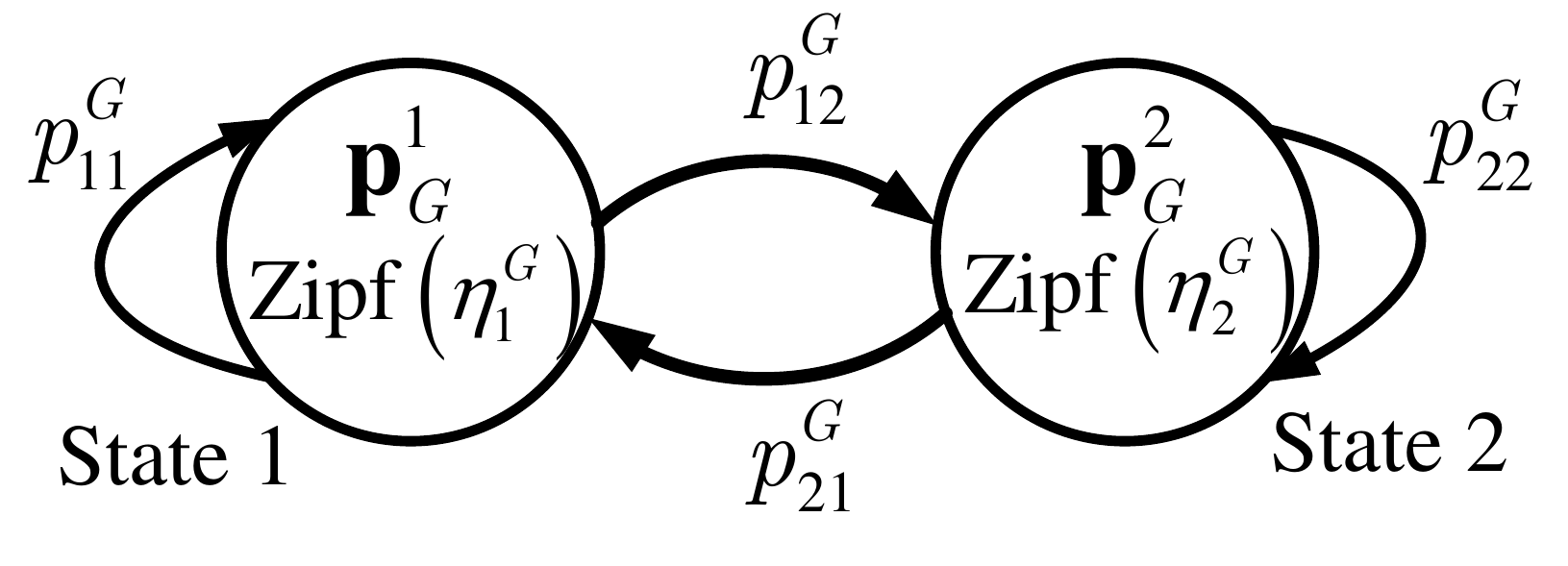}
		\caption{Global popularity Markov chain.}
	\end{subfigure} 
	\begin{subfigure}
		\centering
		\includegraphics[width=0.42\columnwidth]{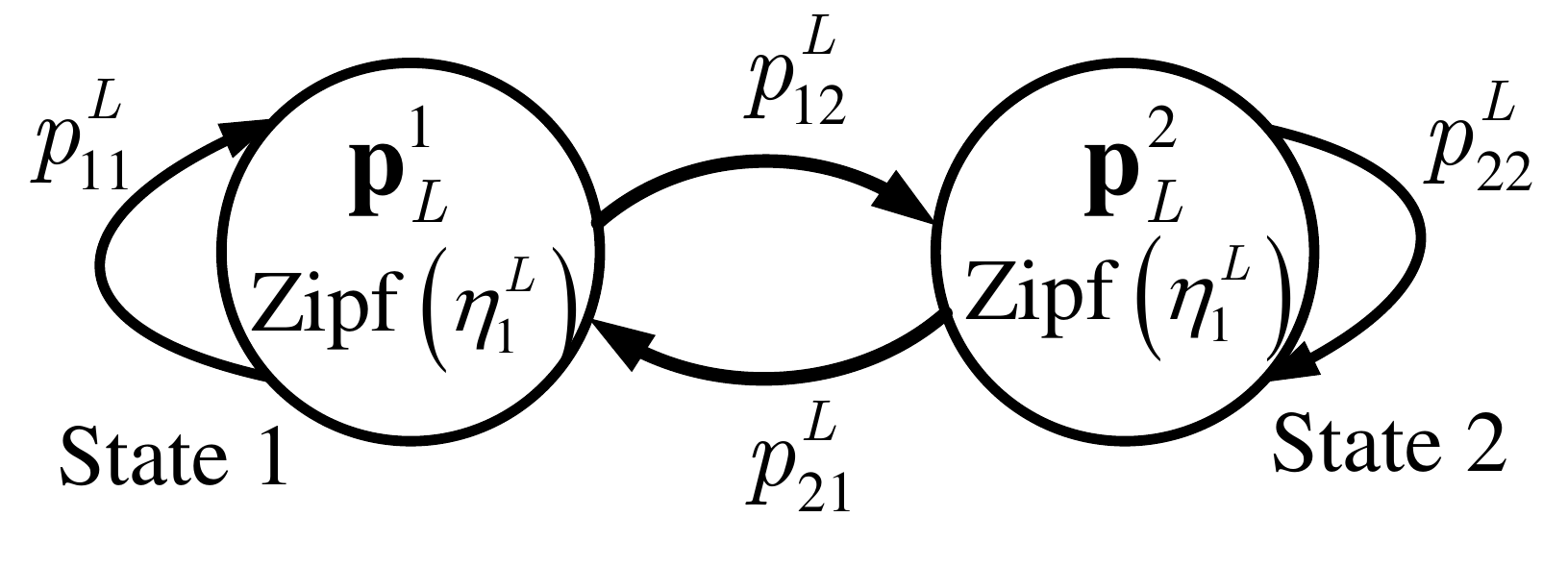}
		\caption{Local popularity Markov chain.}
	\end{subfigure}
	\caption{Popularity profiles Markov chains.}
	\label{Markovchains}
\end{figure}

In this section, performance of the proposed Q-learning algorithm and its scalable approximation is tested. To compare the proposed algorithms with the optimal offline caching policy, we first simulated a small network with $F = 10$ contents, and caching capacity $M = 2$ at the local SB. Global popularity profile is modeled by a two-state Markov chain with states ${\bf p}^{1}_G$ and
${\bf p}^{2}_G$,that are drawn from Zipf distributions having parameters $\eta_1^G=1$ and $\eta_2^G=1.5$, respectively \cite{breslau1999web}; see also Fig.~\ref{Markovchains}.   That is, for state $i \in \left\{1,2\right\}$, the $F$ contents are assigned a random ordering of popularities, and then sorted accordingly in a descending order. Given this ordering and the Zipf distribution parameter $\eta_i^G$, the popularity of the $f$-th  content is set to  
\begin{equation}
\nonumber
\bigg[\mathbf{p}_{\text{G}}^{i} \bigg]_f = \frac{1}{f^{\eta_i} \sum \limits_{l=1}^{F}   1 \mathbin{/} l^{\eta_i^G}} \;\quad \text{{for}} \;\;i=1,2 \end{equation} 
where the summation normalizes the components to follow a valid probability mass function, while $\eta_i^G \geq 0$ controls the skewness of popularities. Specifically, $\eta_i^G=0$ yields a  uniform spread of popularity among contents, while a large value of $\eta_i$ generates more skewed popularities. 
Furthermore, state transition probabilities  of the Markov chain modeling  global popularity profiles are 
\begin{equation}
\nonumber
{\bm P}^{G}:= \left[ {\begin{array}{*{20}{c}}
	{{p^{G}_{11}}}&{{p^{G}_{12}}}\\{{p^{G}_{21}}}&{{p^{G}_{22}}}
	\end{array}} \right] = \left[ {\begin{array}{*{20}{c}}
	{{0.8}}&{{0.2}}\\ {0.75}&{0.25}
	\end{array}} \right].
\end{equation} 

Similarly, local popularities are modeled by a two-state Markov chain, with states $\mathbf{p}_L^{1}$ and  $\mathbf{p}_L^{2}$, whose entries are drawn from Zipf distributions with parameters $\eta^{L}_1 = 0.7$ and $\eta^{L}_2 = 2.5$, respectively. The transition probabilites of the local popularity Markov chain are  
\begin{equation}
\nonumber
{\bm P}^{L}:= \left[ {\begin{array}{*{20}{c}}
	{{p^{L}_{11}}}&{{p^{L}_{12}}}\\{{p^{L}_{21}}}&{{p^{L}_{22}}}
	\end{array}} \right] = \left[ {\begin{array}{*{20}{c}}
	{{0.6}}&{{0.4}}\\ {0.2}&{0.8}
	\end{array}} \right].
\end{equation}

Caching performance is assessed under two cost-parameter settings: (s1)   $\lambda_1 = 10, \lambda_2 = 600, \lambda_3 = 1000$; and, (s2) $\lambda_1 = 600, \lambda_2 = 10, \lambda_3 = 1000$. For both (s1) and (s2), the optimal offline caching policy is found by utilizing the policy iteration with known transition probabilities. In addition, Q-learning in Alg. 1 and its scalable approximation in Alg. 2 are run with $\beta_t = 0.8$, $\alpha_{G} = \alpha_{L} = \alpha_{R} = 0.005$, and $\epsilon_t = 1 \mathbin{/} {\textrm{iteration index}}$, thus promoting exploration in the early iterations, and exploitation in later iterations.

\begin{figure}[t]
	\centering
	\includegraphics[width=.6\columnwidth]{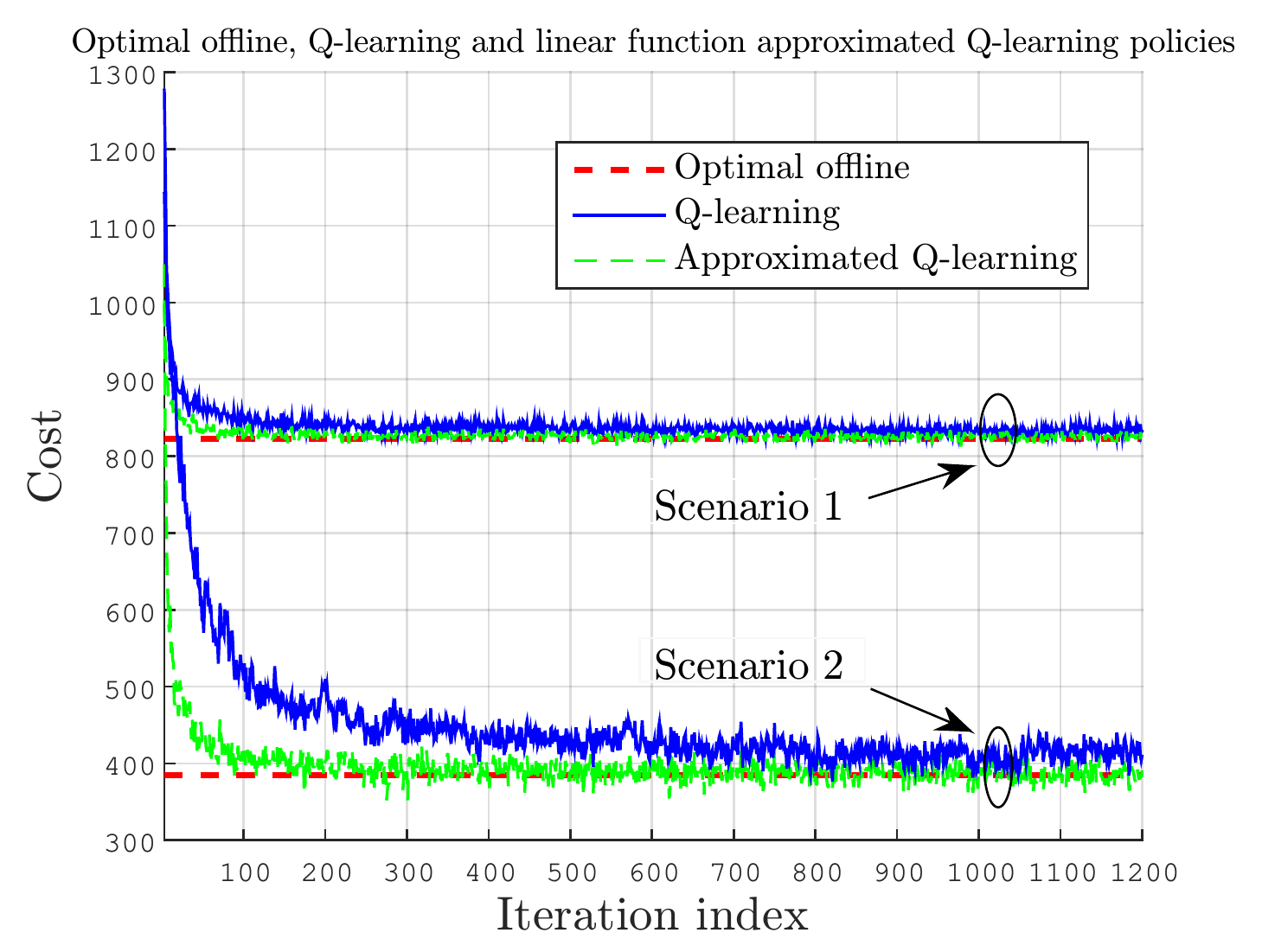}
	\caption{Performance of the proposed algorithms.}
	\label{Cost}
\end{figure}

Fig.~\ref{Cost} depicts the observed cost versus iteration (time) index averaged over 100 realizations. It is seen that the cashing cost via Q-learning, and through its scalable approximation converge to that of the  optimal offline policy. As anticipated,  even for the small size of this network, namely $|{\cal P}_G| = |{\cal P}_L| = 2$ and $|{\cal A}| = 45$, the Q-learning algorithm converges slowly to the optimal policy, especially under s2, while its scalable approximation exhibits faster convergence.

\begin{figure}[ht]
	\centering
	\includegraphics[width=.6\columnwidth]{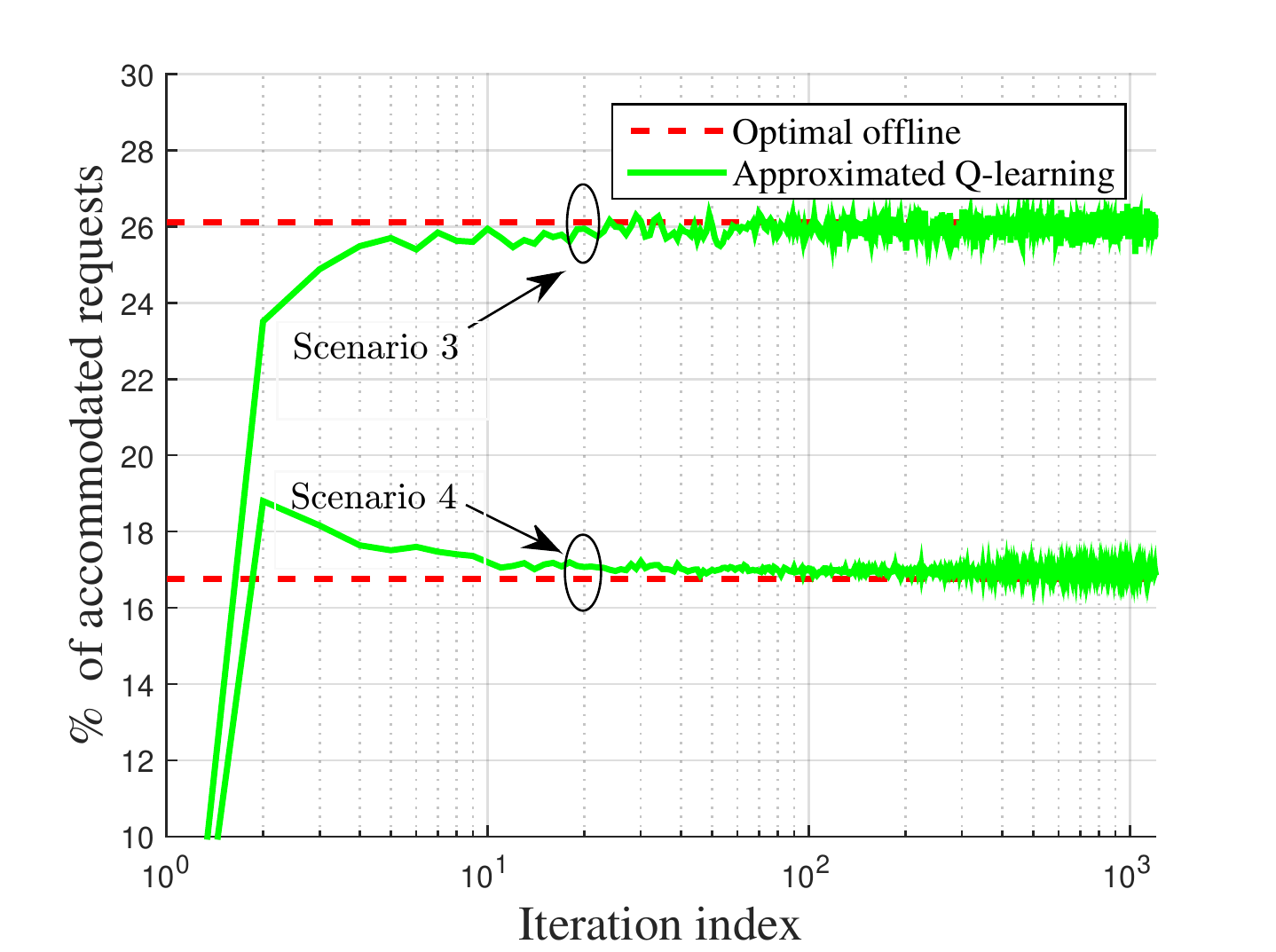}
	\caption{Percentage of accommodated requests via cache.}
	\label{percent}
\end{figure}

In order to highlight the trade-off between global and local popularity mismatches, 
the percentage of accommodated requests  via cache is depicted in Fig. \ref{percent} for settings (s3)  $\lambda_1 = \lambda_3 = 0, \lambda_2 = 1,000$, and (s4) $ \lambda_1 = \lambda_2 = 0, \lambda_3 = 1,000$. 
Observe that penalizing local popularity-mismatch in (s3) forces  the caching policy to  adapt to local request dynamics, thus accommodating a higher percentage of requests via cache, while (s4) prioritizes tracking global popularities, leading  to a lower cache-hit in this setting. Due to slow convergence of the exact Q-learning under (s3) and (s4), only the performance of the scalable solver is presented here.

\begin{figure}[ht]
	\centering
	\includegraphics[width=.6\columnwidth]{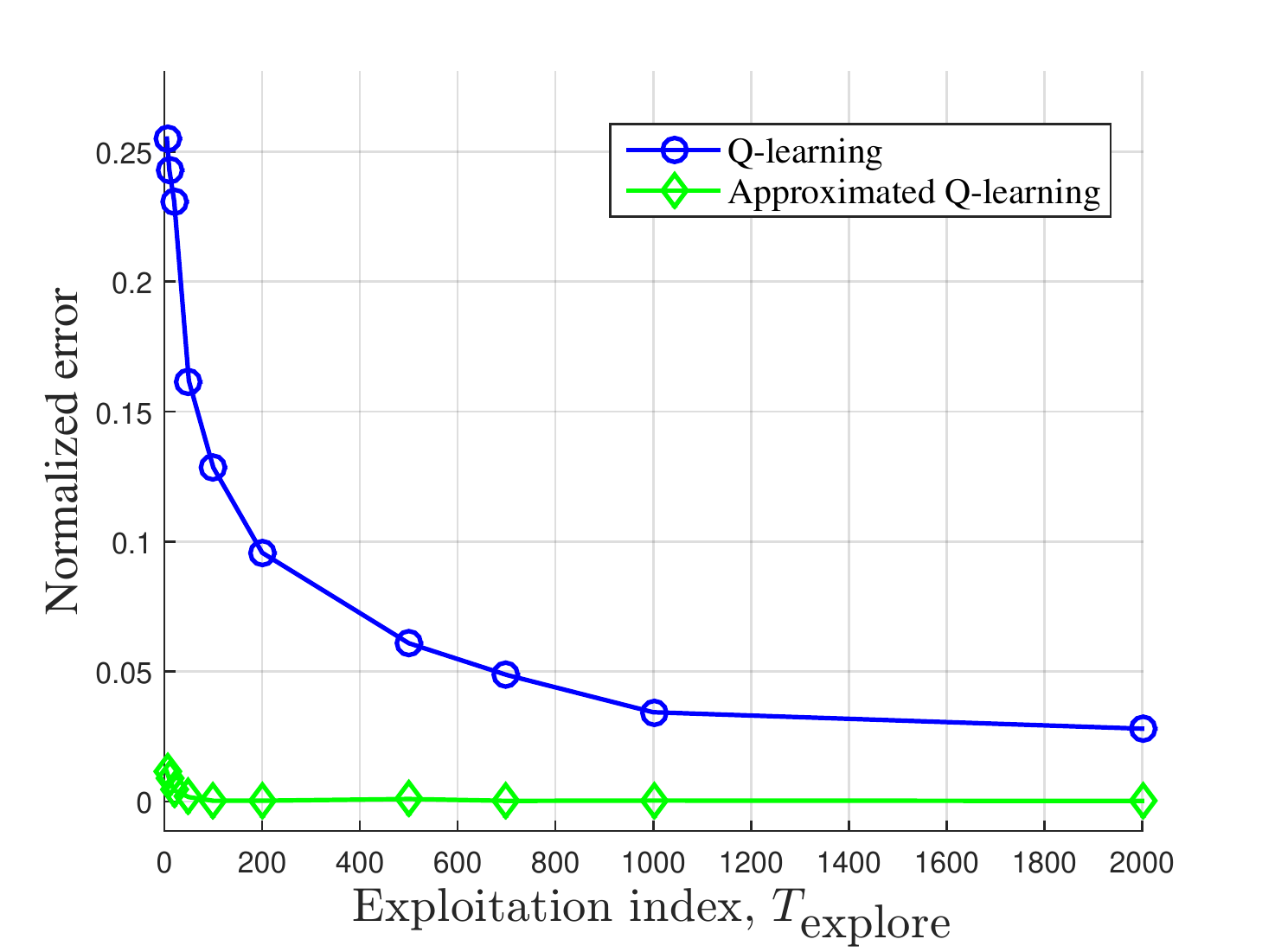}
	\caption{Convergence rate of the exact and scalable Q-learning.}
	\label{Convergence}
\end{figure}

Furthermore, the convergence rate of Algs.~1 and 2 is illustrated in Fig.~\ref{Convergence}, where average normalized error is evaluated in terms of the ``exploitation index.'' Specifically,  a pure exploration is taken for the first $T_{\rm{explore}}$ iterations of the algorithms, i.e., $\epsilon_t = 1$ for $t=1,2,\ldots,T_{\rm{explore}}$; and a pure exploitation  with $\epsilon_t = 0$ is adopted afterwards. 
We have set $\alpha=0.005$, and selected  $\beta_t=\beta \in (0,1)$ so that the fastest convergence is achieved. As the plot demonstrates, the exact Q-learning Alg. 1 exhibits slower convergence, whereas just a few iterations suffice for the scalable Alg. 2 to converge to the optimal solution, thanks to the reduced dimension of the problem as well as the multiple updates that can be afforded per iteration.

Having established the accuracy and efficiency  of the Alg. 2, we next simulated a larger network with $F = 1,000$ available files, and a cache capacity of $M = 10$, giving rise to a total of ${{1000}\choose{10} }\simeq 2 \times 10^{23}$ feasible caching actions. In addition, we set the local and global popularity Markov chains to have $|{\cal P}_L| = 40$  and $|{\cal P}_G| = 50$ states, for which the underlying state transition probabilities  are drawn randomly, and  Zipf parameters are drawn uniformly over the interval $(2,4)$. 

\begin{figure}[ht]
	\centering
	\includegraphics[width=.6\columnwidth]{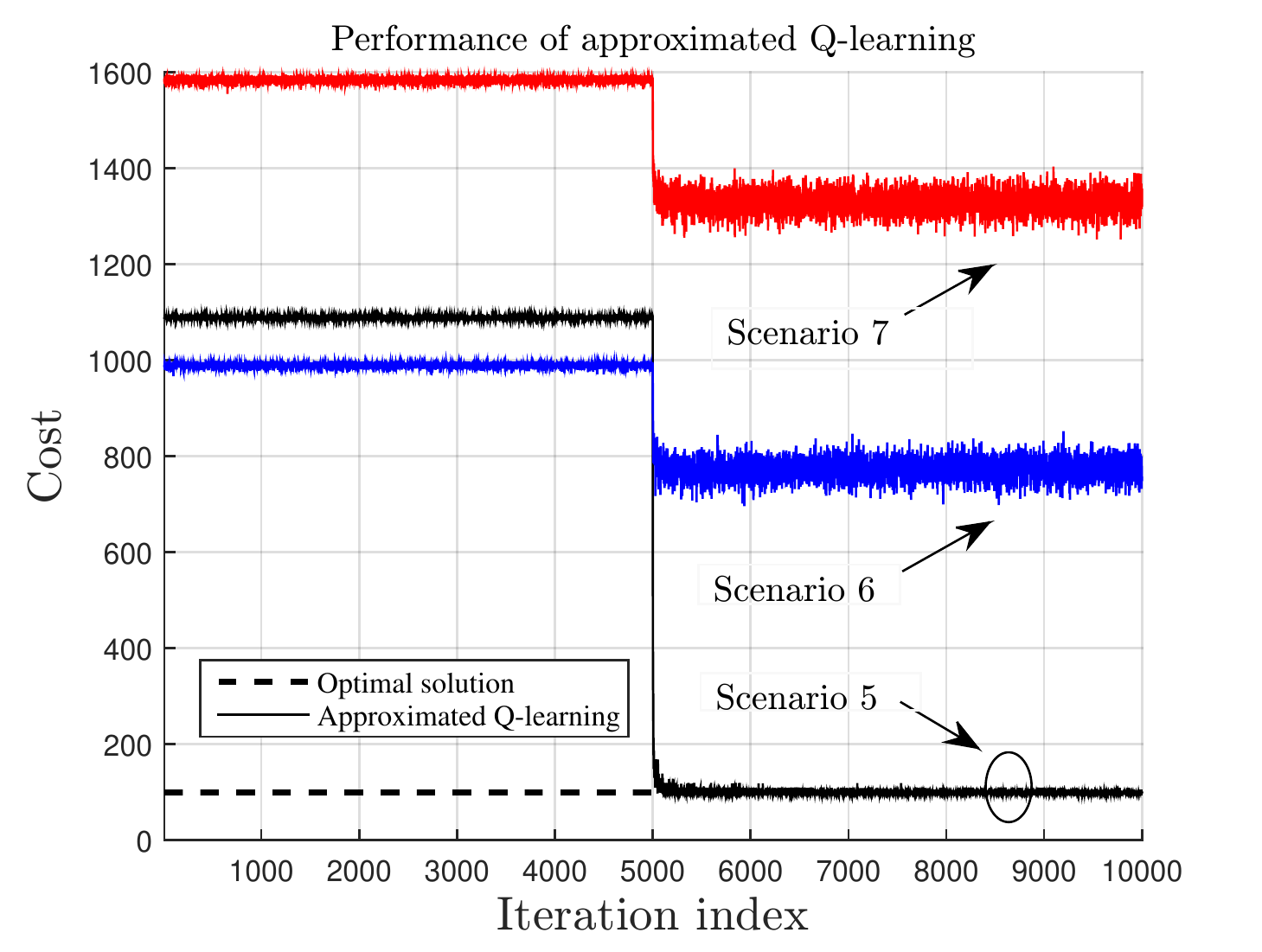}
	\caption{Performance  in large state-action space scenaria.}
	\label{appr}
\end{figure}

Fig.~\ref{appr} plots the performance of Alg. 2 under (s5) $\lambda_1 = 100$, $\lambda_2 =20$, $\lambda_3 =20$, (s6) $\lambda_1 = 0$, $\lambda_2 = 0$, $\lambda_3 = 1,000$, and (s7) $\lambda_1 = 0$, $\lambda_2 = 1,000$, $\lambda_3 = 600$. Exploration-exploitation parameter is set to  $\epsilon_t=1 $ for $t=1,2,\ldots,5,000$, in order to greedily explore the entire state-action space in initial iterations, and $\epsilon_t = 1 \mathbin{/} {\textrm{iteration index}}$ for $t>5,000$.  
Finding the optimal offline policy in (s6) and (s7) requires prohibitively sizable memory as well as extremely high computational complexity, and it is thus unaffordable for this network. However, having large cache-refreshing cost with $\lambda_1 \gg \lambda_2,\lambda_3$ in (s5) forces the optimal caching policy to freeze its cache contents, making the optimal caching policy predictable in this setting. Despite the very limited storage capacity, of $10\mathbin{/}1,000=0.01$ of available files, utilization of RL-enabled caching offers a considerable reduction in incurred costs, while the proposed approximated Q-learning endows the approach with scalability and light-weight updates.

\section{Conclusions}
The present Chapter addressed caching in  5G cellular networks, where space-time popularity of requested files is modeled via local and global Markov chains. By considering local and global popularity mismatches as well as cache-refreshing costs, 5G caching is cast as a reinforcement-learning task. A Q-learning algorithm is developed for finding the optimal caching policy in an online fashion, and its linear approximation is provided to offer scalability over large networks. The novel RL-based caching offers an asynchronous and semi-distributed caching scheme, where adaptive tuning of parameters can readily bring about policy adjustments to space-time variability of file requests via light-weight updates.

\section{Deep Reinforcement Learning for Adaptive Caching  in Hierarchical Content Delivery Networks}

\section{Introduction}
\label{Sec:Intro}
%In light of the tremendous growth of data traffic over both wireline and wireless communications, next-generation networks including future Internet architectures, content delivery infrastructure, and cellular networks stand in need of emerging technologies to meet the ever-increasing data demand. Recognized as an appealing solution is \textit{caching}, which amounts to storing reusable contents in geographically distributed storage-enabled network entities so that future requests for those contents can be served faster. The rationale is that unfavorable shocks of peak traffic periods can be smoothed by proactively storing `anticipated' highly popular contents at those storage devices and during off-peak periods~\cite{Paschos18, Bastug14}. Caching popular content is envisioned to achieve major savings in terms of energy, bandwidth, and cost, in addition to user satisfaction~\cite{Paschos18}.

%To fully unleash its potential, a content-agnostic caching entity has to rely on available observations to learn what and {when} to cache. Toward this goal, contemporary machine learning and artificial intelligence tools hold the promise to empower next-generation networks with `smart' caching control units, that can learn, track, and adapt to unknown dynamic environments, including space-time evolution of content popularities and network topology, as well as entity-specific caching policies.

Deep neural networks (DNNs) have lately boosted the notion of ``learning from data'' with field-changing performance improvements reported in diverse  artificial intelligence tasks \cite{goodfellow2016deep}. DNNs can cope with the `curse of dimensionality' by providing compact low-dimensional representations of high-dimensional data~\cite{pami2013bengio}. Combining deep learning with RL, deep (D) RL has created the first artificial agents to achieve human-level performance across many challenging domains~\cite{minh2015,survey2019}. As another example, a DNN system was built to operate Google's data centers, and shown able to consistently achieve a 40\% reduction in  energy consumption for cooling \cite{2014datacenter}. This system provides a general-purpose framework to understand complex dynamics, which has also been applied to address other challenges including e.g., dynamic spectrum access \cite{DRLSA}, multiple access and handover control \cite{DRLMA}, \cite{DRLHO}, as well as resource allocation in fog-radio access networks \cite{DRLMS,wc2018dyj} or software-defined networks \cite{2018gzh,jsac2019gzh}. 

%Early approaches to caching include the least recently used (LRU), least frequently used (LFU), first in first out (FIFO), random replacement (RR) policies, and their variants. Albeit simple, these schemes cannot deal with the dynamics of content popularities and network topologies.	Recent efforts have gradually shifted toward developing learning and optimization based approaches that can `intelligently' manage the cache resources. For unknown but time-invariant content popularities, multi-armed bandit online learning was pursued in~\cite{multiarm2014}. Yet, these methods are generally not amenable to online implementation. To serve delay-sensitive requests, a learning approach was developed in~\cite{DNN_NonCVX} using a pre-trained DNN to handle a non-convex problem reformulation.

In realistic networks, popularities exhibit dynamics, which motivate well the so-termed \emph{dynamic} caching. A Poisson shot noise model was adopted to approximate the evolution of popularities in \cite{PSN1}, for which an age-based caching solution was developed in~\cite{PSN2}. RL based methods have been pursued in \cite{RL1, RL2, RL3, CacheIA}. Specifically, a {Q}-learning based caching scheme was developed in \cite{RL1} to model global and local content popularities as Markovian processes. Considering Poisson shot noise popularity dynamics, a policy gradient RL based caching scheme was devised in \cite{RL2}. Assuming stationary file popularities and service costs, a dual-decomposition based Q-learning approach was pursued in~\cite{RL3}. Albeit reasonable for discrete states, these approaches cannot deal with large continuous state-action spaces. To cope with such spaces, DRL approaches have been considered for content caching in e.g., \cite{CacheIA, DRL_AC, DRL_Vehic, DRL_Smartcity,survey2019, DRL4edgecaching}. Encompassing finite-state time-varying Markov channels, a deep Q-network approach was devised in~\cite{CacheIA}. An actor-critic method with deep deterministic policy gradient updates was used in~\cite{DRL_AC}. Boosted network performance using DRL was documented in several other applications, such as connected vehicular networks~\cite{DRL_Vehic}, and smart cities \cite{DRL_Smartcity}.

The aforementioned works focus on devising caching policies for a \emph{single} caching entity. A more common setting in next-generation networks however, involves a network of interconnected caching nodes. It has been shown that considering a network of connected caches jointly can further improve performance \cite{Maddahali2014, Distributed2010}. For instance, leveraging network topology and the broadcast nature of links, the coded caching strategy in~\cite{Maddahali2014} further reduces data traffic over a network. This idea has been extended in \cite{Online2015} to an online setting, where popularities are modeled Markov processes. Collaborative and distributed online learning approaches have been pursued \cite{collaborative2012,Distributed2010,decentralized2018}. {\color{black} Indeed, today's content delivery networks such as Akamai \cite{nygren2010akamai}, have tree network structures. Accounting for the hierarchy of caches has become a common practice in recent works; see also \cite{ JointDehgan17,JointShukla18, tong2016hierarchical}. Joint routing and in-network content caching in a hierarchical cache network was formulated in \cite{JointDehgan17}, for which greedy schemes with provable performance guarantees can be found in~\cite{JointShukla18}. }

We identify the following challenges that need to be addressed when designing practical caching methods for network of caches.
\begin{enumerate}
	\item[\bf{c1)}] \emph{Networked caching.} Caching decisions of a node, in a network of caches, influences decisions of all other nodes. Thus, a desired caching policy must adapt to the network topology and policies of neighboring nodes.
	
	\item[\bf{c2)}] \emph{Complex dynamics.} Content popularities are random and exhibit unknown space-time, heterogeneous, and often non-stationary dynamics over the entire network. 
	
	\item[\bf{c3)}] \emph{Large continuous state space.} Due to the shear size of available content, caching nodes, and possible realizations of content requests, the decision space is huge. 
	
\end{enumerate} 

\subsection{This section} 
Prompted by the recent interest in hierarchical caching, here we focus on a two-level network caching, where a parent node is connected to multiple leaf nodes to serve end-user file requests. Such a two-level network constitutes the building block of the popular tree hierarchical cache networks in e.g., \cite{nygren2010akamai}. To model the interaction between caching decisions of parent and leaf nodes along with the space-time evolution of file requests, a scalable DRL approach  based on hyper deep Q-networks (DQNs) is developed. As corroborated by extensive numerical tests, the novel caching policy for the parent node can adapt itself to local policies of leaf nodes and space-time evolution of file requests. Moreover, our approach is simple-to-implement, and performs close to the optimal policy. 	

\begin{figure}[t]
	\centering
	\includegraphics[width =0.3 \columnwidth]{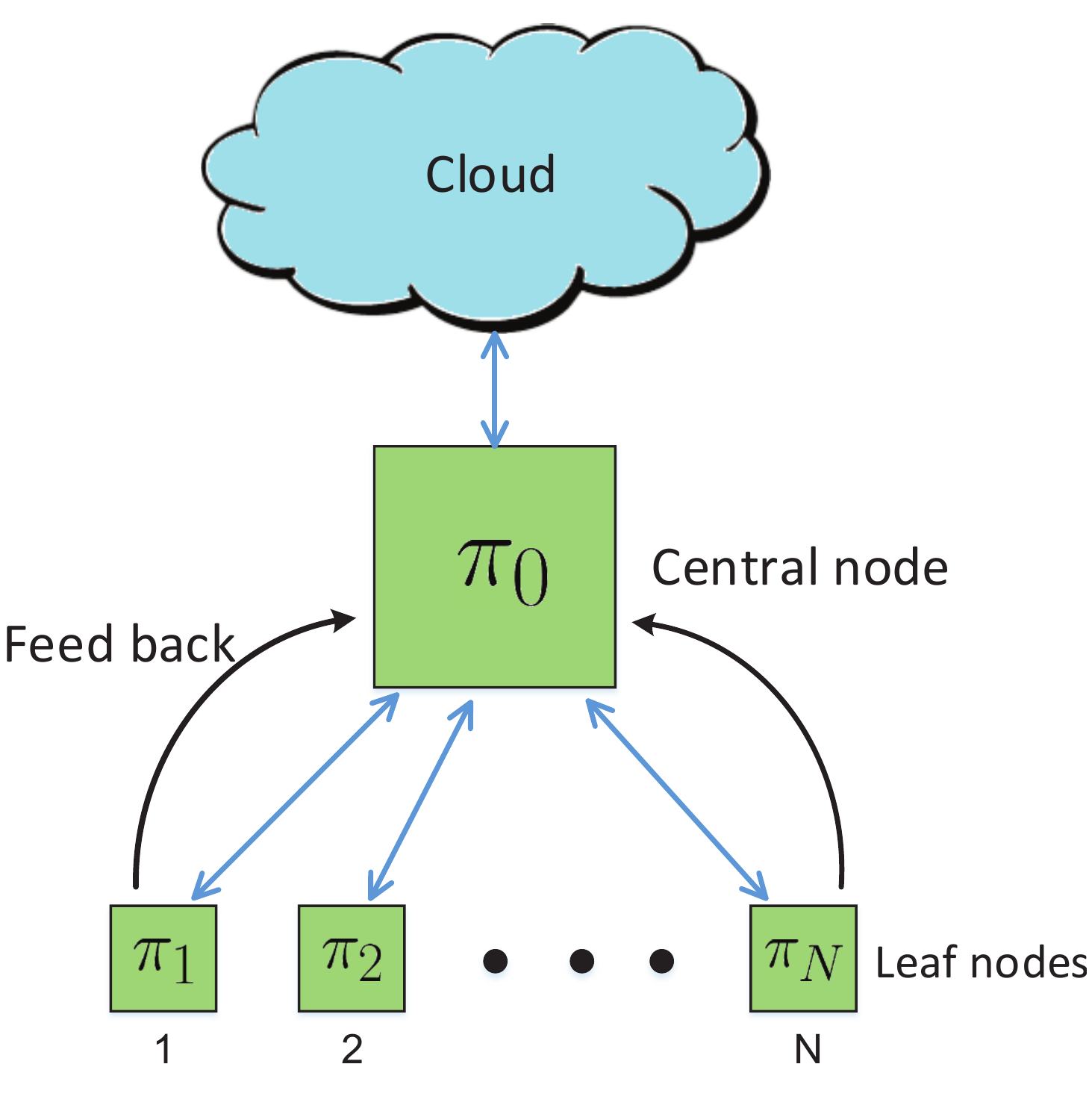}
	\caption{A network of caching nodes.}
	\label{fig:model}
	\vspace{+.2 cm}
	\centering \includegraphics[width =0.3\textwidth]{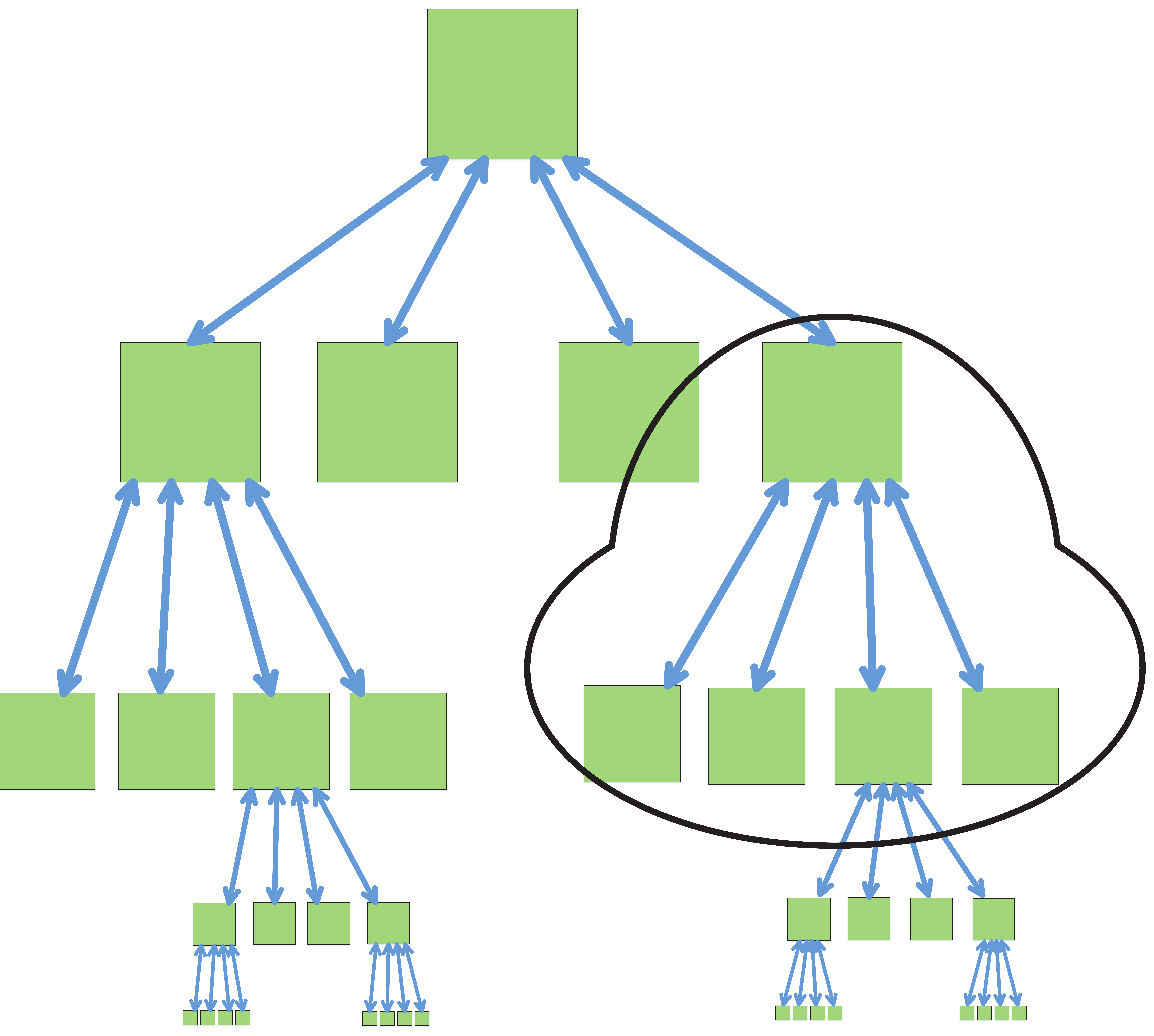}
	\vspace{+.2 cm}
	\caption{A hierarchical tree network cache system.} 
	\label{fig:higherarchy}
\end{figure}

%\begin{figure}[t] 
%	\centering
%	{\includegraphics[width=1\columnwidth]{figs/saf/Systemmodel_3.pdf}}
%	\caption{A schematic depicting the evolution of key quantities across time slots. Duration of slots can be unequal.} 
%	\label{fig:sys2} 	
%\end{figure} 

\section{Modeling and Problem Statement}\label{sec:modeldrlcach}

Consider a two-level network of interconnected caching nodes, where a parent node is connected to $N$ leaf nodes, indexed by $n\in\mathcal{N}:=\{1,\ldots,N\}$. The parent node is connected to the cloud through a (typically congested) back-haul link; see Fig.~\ref{fig:model}. One could consider this network as a part of a large hierarchical caching system, where the parent node is connected to a higher level caching node instead of the cloud; see Fig.~\ref{fig:higherarchy}. In a content delivery network for instance, edge servers (a.k.a. points of presence or PoPs) are the leaf nodes, and a fog server acts as the parent node. Likewise, (small) base stations in a 5G cellular network are the leaf nodes, while a serving gate way (S-GW) may be considered as the parent node; see also~\cite[p.~110]{LTE}.   

All nodes in this network store files to serve file requests. Every leaf node serves its locally connected end users, by providing their requested files. If a requested content is locally available at a leaf node, the content will be served immediately at no cost. If it is not locally available due to limited caching capacity, the content will be fetched from its parent node, at a certain cost. Similarly, if the file is available at the parent node, it will be served to the leaf at no cost; otherwise, the file must be fetched from the cloud at a higher cost.

To mitigate the burden with local requests on the network, each leaf node stores `anticipated' locally popular files. In addition, this paper considers that each parent node stores files to serve requests that are \emph{not} locally served by leaf nodes. Since leaf nodes are closer to end users, they frequently receive file requests that exhibit rapid temporal evolution at a \emph{fast} timescale. The parent node on the other hand, observes aggregate requests over a large number of users served by the $N$ leaf nodes, which naturally exhibit smaller fluctuations and thus evolve at a \emph{slow} timescale. 

\begin{figure}
	\centering
	\includegraphics[width =0.6 \textwidth]{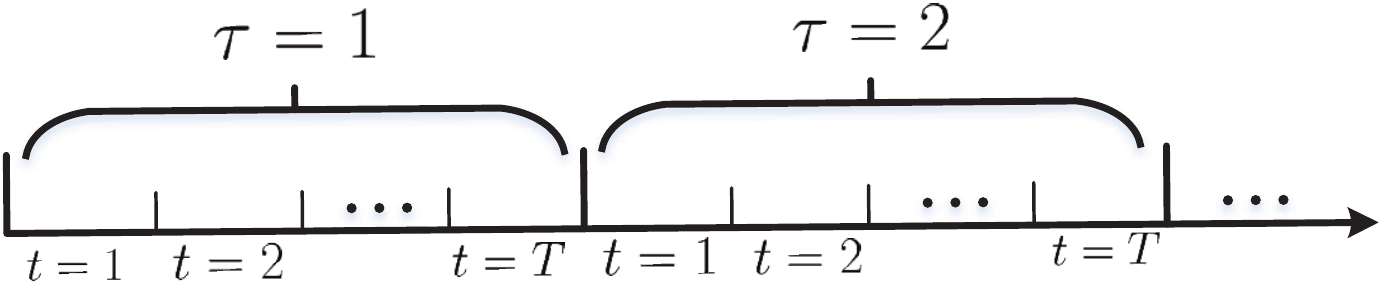}
	\caption{Slow and fast time slots.}
	\label{fig:Timescales}
	\vspace{-5pt}
\end{figure}

This motivated us to pursue a two-timescale approach to managing such a network of caching nodes. To that end, let $\tau=1,2,\ldots$ denote the slow time intervals, each of which is further divided into $T$ fast time slots indexed by $t=1,\ldots,T$; see Fig.~\ref{fig:Timescales} for an illustration. Each fast time slot may be e.g., $1$-$2$ minutes depending on the dynamics of local requests, while each slow time interval is a period of say $4$-$5$ minutes. We assume that the network state remains unchanged during each fast time slot $t$, but can change from $t$ to $t+1$.

Consider a total of $F$ files in the cloud, which are collected in the set ${\mathcal F} = \{1,  \ldots, F\}$. At the beginning of each slot $t$, 	every leaf node $n$ selects a subset of files in $\mathcal{F}$ to prefetch and store for possible use in this slot.
To determine which files to store, every leaf node relies on a local caching policy function denoted by $\pi_n$, to take (cache or no-cache) action ${\pmb a}_n ( t+1, \tau  ) = \pi_n (\pmb s_n (t, \tau)) $ at the beginning of slot $t+1$, based on its \emph{state} vector ${\pmb s}_n$ at the end of slot $t$. We assume this action takes a negligible amount of time relative to the slot duration; and  define the state vector $\pmb s_n (t, \tau)\! :=\! \pmb r_n(t,\tau) := [r_n^1(t,\tau) \cdots r_n^{F}(t,\tau)]^\top$ to collect the number of requests received at leaf node $n$ for individual files over the duration of slot $t$ on interval $\tau$. Likewise, to serve file requests that have not been served by leaf nodes, the parent node takes action $\pmb a_0 ( \tau )$ to store files at the beginning of every interval $\tau$, according to a certain policy $\pi_0$. Again, as aggregation smooths out request fluctuations, the parent node observes slowly varying file requests, and can thus make caching decisions at a relatively slow timescale. In the next section, we present a two-timescale approach to managing such a network of caching nodes.

\section{Two-timescale Problem Formulation}\label{sec:two}
File transmission over any network link consumes resources, including e.g., energy, time, and bandwidth. Hence, serving any requested file that is not locally stored at a node, incurs a cost. Among possible choices, the present paper considers the following cost for node $n\in\mathcal{N}$, at slot $t+1$ of interval $\tau$
\begin{multline}
\hspace{-0.3 cm} \pmb c_{n}(\pi_{n} (\pmb s_n ( t,\tau ) ), \pmb r_{n} (t+1,\tau), \pmb a_0 (	\tau ) )  \! :=  \! \pmb r_n (t+1, \tau) \odot  ( {\bf 1}\! - \pmb a_{0} (\tau))  \\  \!\odot \! ( {\bf 1} \!- \pmb a_{n} (t\!+\!1,\tau) )\! + \pmb r_n (t\!+\!1, \tau) \odot ( {\bf 1} \!- \pmb a_n(t\!+\!1,\tau) )
\label{eq:nodalcost}
\end{multline}
%\normalsize
where $\pmb c_{n} (\cdot):=[c_n^1(\cdot)~\cdots~c_n^F(\cdot)]^{\top}$ concatenates the cost for serving individual files per node $n$; symbol $\odot$ denotes entry-wise vector multiplication; entries of $\pmb a_0$ and $\pmb a_n$ are either $1$ (cache, hence no need to fetch), or, $0$ (no-cache, hence fetch); and $\bf 1$ stands for the all-one vector. Specifically, the second summand in \eqref{eq:nodalcost} captures the cost of the {\color{black} leaf} node fetching files for end users, while the first summand corresponds to that of the parent fetching files from the cloud. 

We model user file requests as Markov processes with unknown transition probabilities~\cite{RL1}. Per interval $\tau$, a reasonable caching scheme for leaf node $n\in\mathcal{N}$ could entail minimizing the expected cumulative cost; that is, 
\begin{align} 
\pi^{ \ast}_{n,\tau} \!\!:= \underset{\pi_n \in \Pi_n}{\arg \min} ~ {\mathbb{E}}\Big[ \!\sum_{t=1}^{T}\! {\bf 1}^{\!\top} \!\pmb c_{n} (\pi_{n} (\pmb s_n ( t,\tau ) ), \pmb r_{n} (t\!+\!1,\tau), \pmb a_0 (\tau ))\Big] 
\label{eq:leaf_node}
\end{align} 
where $\Pi_n$ represents the set of all feasible policies for node~$n$. Although solving \eqref{eq:leaf_node} is in general challenging, efficient near-optimal solutions have been introduced in several recent contributions; see e.g.,~\cite{RL1,RL2,multiarm2014}, and references therein. In particular, a RL based approach using tabular $Q$-learning was pursued in our precursor \cite{RL1}, which can be employed here to tackle this fast timescale optimization. The remainder of this paper will thus be on designing the caching policy $\pi_0$ for the parent node, that can learn, track, and adapt to the leaf node policies as well as user file requests.   

\begin{figure}
	\centering
	\includegraphics[width =0.6 \textwidth]{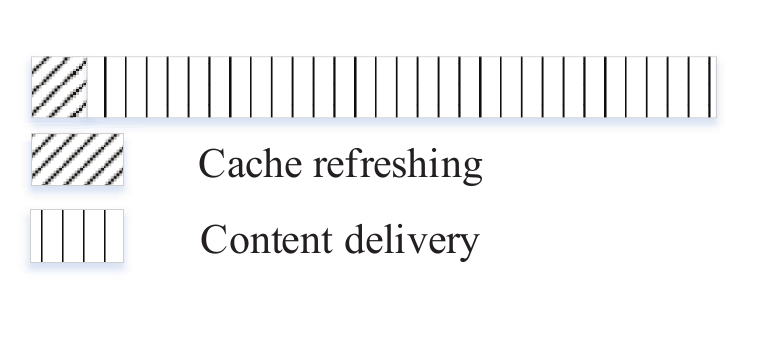}
	\vspace{-5pt}
	\caption{Structure of slots and intervals.}
	\label{fig:slotstructure}
\end{figure}

\section{Reinforcement Learning for Adaptive Caching with Dynamic Storage Pricing}
\section{Introduction}

To target different objectives such as content-access latency, energy, storage or bandwidth utilization, corresponding deterministic cost parameters are defined, and the aggregated cost is minimized in~\cite{Cacherent, OnlineOpt}. Deterministic cost parameters, however, may be inaccurate in modeling practical settings, as spatio-temporal popularity evolutions, network resources such as bandwidth and cache capacity are \textit{random} and subject to change over time and space, due to e.g., time-varying data traffic over links, previous cache decisions, or channel fluctuations. Therefore, this  necessitates modeling the caching problem from a \textit{stochastic optimization} perspective, while accounting for the inherently random nature of available resources and file requests.

\textit{Contributions:} 
	This Section aspires to fill this gap by relying on dual decomposition techniques which transform the limits on the available resources in the original (primal) optimization into stochastic prices in the dual problem.
	Building on this approach, the goal is to design more flexible caching schemes by introducing a generic dynamic pricing formulation, while enabling SBs to learn the optimal fetching-caching decisions using low-complexity techniques. Our contributions are listed as follows.

	\begin{enumerate}
		\item[1)]  A general formulation of the caching problem by introducing time-varying and stochastic costs is presented, in which the fetching and caching decisions are found through a constrained optimization with the objective of reducing the overall cost, aggregated across files and time instants (Section~\ref{Sec_Formulation}).  
		
		\item[2)] Since the caching decision in a given time slot not only affects the instantaneous cost, but also influences the cache availability in the future, the problem is indeed a dynamic programming (DP), and therefore can be effectively solved by reinforcement learning-based approaches. By assuming known and stationary distributions for the costs and popularities, and upon relaxing the limited cache capacity constraint, the proposed generic optimization problem is shown to become separable across files, and thus can be efficiently solved using the value-iteration algorithm~(Section~\ref{Sec:DP_Formulation}).

		\item[3)] Subsequently, it is shown that the particular case where the cache capacity is limited and the distribution of the pertinent parameters are unknown can be handled by the proposed generic formulation. Thus, in order to address these issues, a dual-decomposition technique is developed to cope with the coupling constraint associated with the storage limitation. Finally, an online low complexity ($Q$-function based) reinforcement learning solver is put forth for learning the optimal fetch-cache decisions on-the-fly (Section~\ref{Sec_limited_storage_and_coms}). 
		
		\item[4)] The separability of the objective across files together with the use of marginalized value functions \cite{Luismi2014DP_CRs} enable the decomposition of the original problem into smaller-dimension sub-problems. This in turn leads to circumventing the so-called curse of dimensionality, which commonly arises in reinforcement learning problems (Sections~\ref{Sec:DP_Formulation} and \ref{Sec_limited_storage_and_coms}).   	
	\end{enumerate} 
	
The effectiveness of the proposed scheme in terms of efficiency as well as scalability is corroborated by various numerical tests.
Although our proposed approach enjoys theoretical guarantees in learning the optimal fetch-cache decisions in stationary settings,  numerical tests also corroborate its merits in non-stationary scenarios.

%The rest of this section is organized as follows. Section~\ref{Sec_Formulation} provides a generic formulation of the problem, where solvers adopted from reinforcement learning are developed in Section~\ref{Sec:DP_Formulation}. Limited storage and back-haul transmission rate settings are discussed in Section~\ref{Sec_limited_storage_and_coms}. Section~\ref{Sec_results} reports numerical results, and finally section~\ref{Sec_Conclusion} provides concluding remarks.

\section{Operating conditions and costs}
\label{Sec_Formulation}
Consider a memory-enabled SB responsible for serving file (content) requests denoted by $f=1,2,\ldots,F$ across time. The requested contents are transmitted to users either by fetching through a (costly)  back-haul transmission link connecting the SB to the cloud, or, by utilizing the local storage unit in the SB where popular contents have been proactively cached ahead of time. The system is considered to operate in a slotted fashion with $t = 1,2, \ldots$ denoting time. 

During slot $t$ and given the available cache contents, the SB receives a number of file requests whose provision incurs certain costs. Specifically, for a requested file $f$, fetching it from the cloud through the back-haul link gives rise to scheduling, routing and transmission costs, whereas its availability at the cache storage in the SB will eliminate such expenses. However, local caching also incurs a number of (instantaneous) costs corresponding to memory or energy consumption. This gives rise to an inherent  caching-versus-fetching trade-off, where one is promoted over the other depending on their relative costs. The objective here is to propose a simple yet sufficiently general framework to minimize the sum-average cost over time by optimizing fetch-cache decisions while adhering to the constraints inherent to the operation of the system at hand, and user-specific requirements. The variables, constraints, and costs involved in  this optimization are described in the ensuing subsections.

\subsection{Variables and constraints}
Consider the system at time slot $t$, where the binary variable $r^f_t$ represents the incoming request for file $f$; that is, $r^f_t = 1$ if  the file $f$ is requested during slot $t$, and $r^f_t = 0$, otherwise. Here, we assume that $r^f_t = 1$ necessitates serving the file to the user and dropping requests is not allowed; thus, requests must be carried out either by fetching the file from the cloud or by utilizing the content currently available in the cache. Furthermore, at the end of each slot, the SB will decide if content $f$ should be stored in the cache for its possible reuse in a subsequent slot.

To formalize this, let us define the ``fetching'' \textit{decision} variable $w^f_t \in \{0,1\}$ along the ``caching'' \textit{decision} variable $a^f_t \in \{0,1\}$. Setting $w^f_t = 1$ implies ``fetching'' file $f$ at time $t$, while $w^f_t=0$ means ``no-fetching.'' Similarly, $a^f_t = 1$ implies that content $f$ will be stored in  cache at the end of slot $t$ for the next slot, while $a^f_t = 0$ implies that it will not. Furthermore, let the storage \textit{state} variable $s_t^f\in\{0,1\}$ account for the availability of files at the local cache. In particular, $s_t^f=1$ if file $f$ is available in the cache at the beginning of slot $t$, and $s_t^f=0$ otherwise. 
Since the availability of file $f$ directly depends on the caching decision at time $t-1$, we have 
\begin{equation}
\label{c1}
%\textrm {(C$1$)}
{\textrm {C1:}}   \quad s_t^f = a^f_{t-1},   \quad \forall f,t,
\end{equation}
which will be incorporated into our optimization as constraints.

Moreover, since having  $r_t^f=1$ implies transmission of file $f$ to the user(s), it requires either having the file in cache ($s_t^f=1$) or fetching it from the cloud ($w_t^f=1$), giving rise to the second set of constraints 
\begin{equation}
\label{c2}
\textrm {C2:} \quad r^f_t \le w^f_t+s^f_t, \quad  \forall f,t.
\end{equation}  
Finally, the caching decision $a_t^f$ can be set to $1$ only when the content $f$ is available at time $t$; that is, only if either fetching is carried out ($w_t^f=1$) or the current cache state is  $s_t^f=1$. This in turn implies the third set of constraints as    
\begin{equation}
\label{c3}
\textrm {C3:} \quad a^f_t \le s^f_t + w^f_t, \quad  \forall f,t.
\end{equation}

\subsection{Prices and aggregated costs}
To account for the caching and fetching costs, let $\rho_t^f$ and $\lambda_t^f$ denote the (generic) costs associated with $a_t^f=1$ and $w_t^f=1$, respectively. Focusing for now on the caching cost and with $\sigma_f$ denoting the size of content $f$, a simple form for $\rho_t^f$ is 
\begin{equation}\label{eq_generic_form_rho_intro}
\rho_t^f=\sigma_f({\rho'}_t+{\rho'}_t^f) + ({\rho''}_t+{\rho''}_t^f),
\end{equation}
where the first term is proportional to the file size $\sigma_f$, while the second one is constant. Note also that we consider file-dependent costs (via variables ${\rho'}_t^f$ and ${\rho''}_t^f$), as well as cost contributions which are common across files (via ${\rho'}_t$ and ${\rho''}_t$). In most practical setups, the latter will dominate over the former. For example, the caching cost per bit is likely to be the same regardless of the particular type of content, so that ${\rho'}_t^f={\rho''}_t^f=0$. From a modeling perspective, variables $\rho_t^f$ can correspond to actual prices paid to an external entity (e.g., if associated with energy consumption costs), marginal utility or cost functions, congestion indicators, Lagrange multipliers associated with constraints, or linear combinations of those (see, e.g., \cite{Luismi2014DP_CRs,Neely2016ReourceAllocationTutorialBook,Tianyi2017DistributedCloudNets,tps2015wang} and Section \ref{Sec_limited_storage_and_coms}). Accordingly, the corresponding form for the fetching cost is
\begin{equation}\label{eq_generic_form_lambda_intro}
\lambda_t^f=\sigma_f({\lambda'}_t+{\lambda'}_t^f) + ({\lambda''}_t+{\lambda''}_t^f).
\end{equation}
As before, if the transmission link from the cloud to the SB is the same for all contents, the prices ${\lambda'}_t$ and ${\lambda''}_t$ are expected to dominate their file-dependent counterparts ${\lambda'}_t^f$ and ${\lambda''}_t^f$.

\begin{figure} 
	\centering	
	\includegraphics[width=0.47 \textwidth]{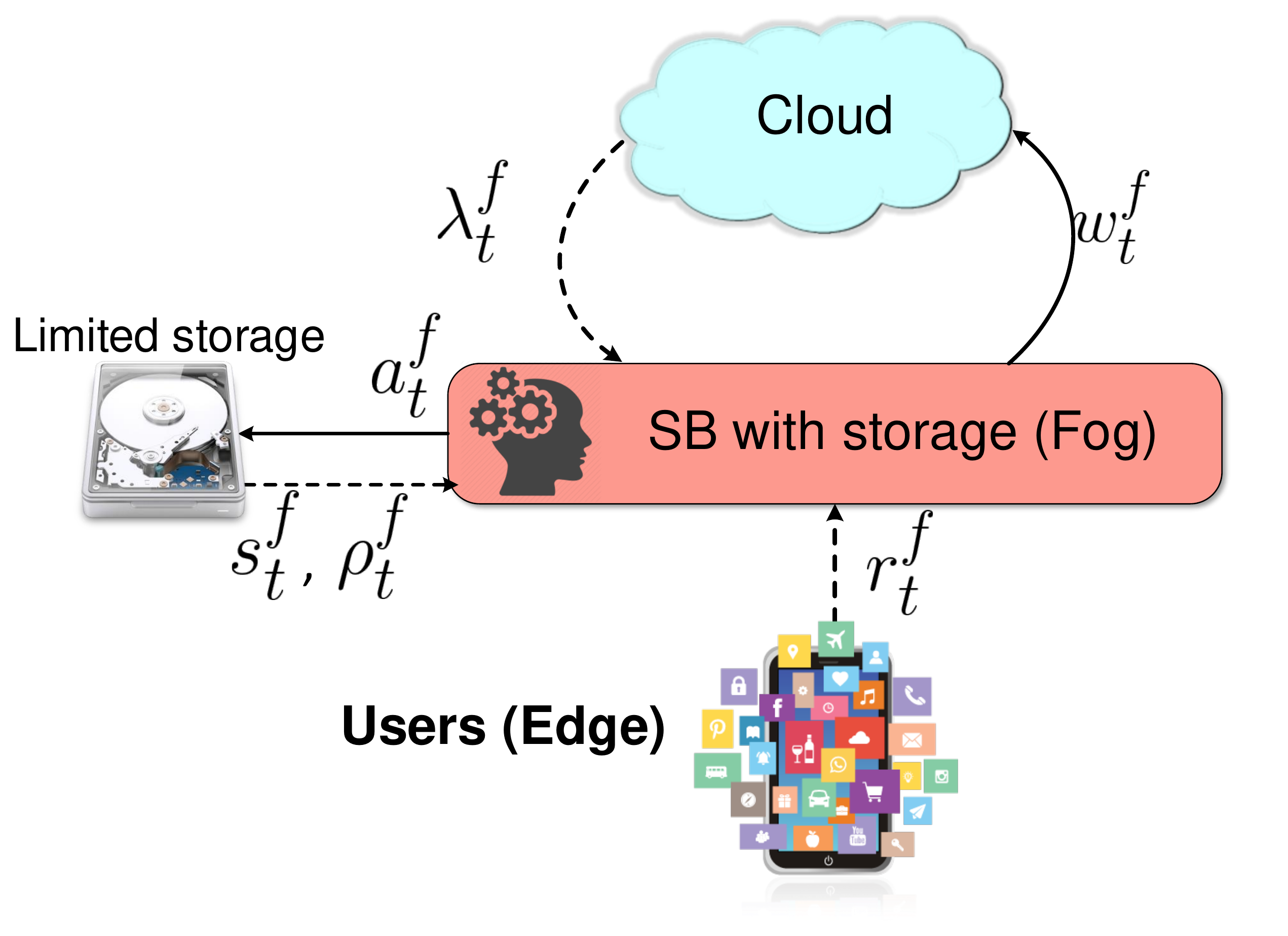}	\caption{System model and main notation. The state variables (dashed lines) are the storage indicator $s_t^f$ and the content request $r_t^f$, as well as the dynamic caching and fetching prices $\rho_t^f$ and $\lambda_t^f$. The optimization variables (solid lines) are the caching and fetching decisions $a_t^f$ and $w_t^f$. The instantaneous per-file cost is $c_t^f=\rho_t^fa_t^f+\lambda_t^fw_t^f$. Per slot $t$, the SB collects the state variables $\{s_t^f,r_t^f;\rho_t^f,\lambda_t^f\}_{f=1}^F$, and decides the values of $\{a_t^f,w_t^f\}_{f=1}^F$ considering not only the cost at time $t$ but also the cost at time instants $t'>t$. }	
	\label{Model} 
\end{figure}

Upon defining the corresponding cost for a given file as $c^f_t (a^f_t,w^f_t;\rho_t^f,\lambda_t^f)=\rho_t^fa^f_t+\lambda_t^f w^f_t $, the aggregate cost at time $t$ is given by 
\begin{equation}
\label{Sum_cost}
c_t := \sum_{f=1}^F c^f_t (a^f_t,w^f_t;\rho_t^f,\lambda_t^f)=\sum_{f=1}^F \rho_t^fa^f_t + \lambda_t^f w^f_t, 
\end{equation}
which is the basis for the DP formulated in the next section. For future reference, Fig.~\ref{Model} shows a schematic of the system model and the notation introduced in this section. 

\section{Optimal caching with time-varying costs}\label{Sec:DP_Formulation}
Since decisions are coupled across time [cf. constraint \eqref{c1}], and the future values of prices as well as state variables are inherently random, our goal is to sequentially make fetch-cache decisions to minimize the long-term average discounted aggregate cost
\begin{align}
\label{eq.11}
{\cal {\bar C}} := {\mathbb E} \; \left[\sum _{t=0}^{\infty} \sum  _{f=1}^{F} \gamma ^{t} c^f_t \left(a^f_t,w^f_t;\rho^f_t,\lambda^f_t\right)\right]
\end{align}
where the expectation is taken with respect to (w.r.t.) the random variables $\boldsymbol \theta_t^f := \{r_t^f,\lambda_t^f,\rho_t^f\}$, and $0<\gamma<1$ is the discounting factor whose tuning trades off current versus more uncertain future costs.  To address the optimization, the following assumptions are considered:
	
	\vspace{+.15 cm} 
	\begin{enumerate}
		\item[AS1)] The values of $\boldsymbol \theta_t^f$ are drawn from a stationary distribution. 
		\item[AS2)] The distribution of $\boldsymbol \theta_t^f$ is known.
		\item[AS3)] The drawn value of $\boldsymbol \theta_t^f$  is revealed at the beginning of each slot $t$, before fetch-cache decisions are made.\end{enumerate} 
	\vspace{+.15 cm} 
	
	AS1 and AS2 allow finding the expectations in this section, and will be relaxed in Section III-E to further generalize our approach to settings where the distributions are unknown. In practice, one may estimate these distributions through e.g., historical data.

The ultimate goal here is to take \textit{real-time} fetch-cache decisions by minimizing the expected \textit{current plus future cost} while adhering to operational constraints, giving rise to the following optimization 
% at each the prices are revealed at All in all, the ultimate objective is to sequentially make feasible fetch-cache decisions $(w_t^f,a_t^f)$ such that ${\bar {\cal C}}$ is minimized, giving rise to the following  constrained optimization 
\begin{align}
{\textrm {(P$1$)} }  \min  \limits_{ \{(w^f_{k},a^f_{k}) \}_{f,k\geq t}}    &{\cal {\bar C}}_t:=\sum \limits_{k=t}^{\infty} \sum  \limits_{f=1}^{F} \gamma ^{k-t} {\mathbb E}  \left[c^f_k \left(a^f_k,w^f_k;\rho^f_k,\lambda^f_k\right)\right]  \nonumber\\
\mathrm{s.t.}\;\;\;\; &(w^f_{k},a^f_{k}) \in \mathcal{X}( r^f_{k}, a^f_{k-1}),\;\;\;\; \forall f,\,\,k\geq t \nonumber
\end{align}
where 
\begin{align}
\nonumber
\mathcal{X}( r^f_k, a^f_{k-1})&:= \Big\{(w,a)\;\Big|\; w\in\{0,1\}, \,a\in\{0,1\}, \\ \nonumber & s_k^f = a^f_{k-1} , \;  r^f_{k} \le w + s^f_{k},\;\; a \le s^f_{k} + w\Big\},
\end{align}
and the expectation is taken w.r.t.  $\{ \boldsymbol \theta^f_k \}_{\forall k\ge t+1}$.

% In contrast to many resource allocation problems where, after introducing pertinent prices I don't think this is clear here (Lagrange multipliers), the optimization decouples across time \cite{Neely2016ReourceAllocationTutorialBook,Tianyi2017DistributedCloudNets}, %
The presence of the set $\mathcal{X}( r^f_k, a^f_{k-1})$ in the constraints demonstrates that the cache state at a given time depends on previous cache decisions, thus coupling the optimization variables across time. It also implies that any instantaneous decision will influence the optimization problem in subsequent slots, having a long-standing influence on future costs. The coupling of the optimization variables across time indeed necessitates utilization of DP tools, motivating the implementation to reinforcement learning algorithms to design efficient solvers.
	
	To find the solution of the DP in (P$1$) we implement the following steps: a) identifying the current and expected future aggregate costs (the latter gives rise to the so-called value functions); b) expressing the corresponding Bellman equations over the value functions; and c) proposing a method to estimate the value functions accordingly. This is the subject of the ensuing subsections, which start by further exploiting the structure of our problem to reduce the complexity of the proposed solution.

\subsection{Bellman equations for the per-content problem}
Focusing on (P$1$), one can readily deduce that: (i) consideration of the content-dependent prices renders the objective in (P$1$) separable across $f$, and (ii) the constraints in (P$1$) are also separable across $f$. Furthermore,  the decisions $a_t^f$ and $w_t^f$ for a given $f$, do not affect the values (distribution) of $ \boldsymbol \theta_{k'}^{f'}$ for files $f'\neq f$ and for times $t'>t$. Thus, (P$1$) naturally gives rise to the per-file optimization 
\begin{align}
{\textrm {(P$2$)} }  \min  \limits_{ \{(w^f_{k},a^f_{k}) \}_{k\geq t}}    &{\cal {\bar C}}_t^f:=\sum \limits_{k=t}^{\infty}  \gamma ^{k-t} {\mathbb E}  \left[c^f_k \left(a^f_k,w^f_k;\rho^f_k,\lambda^f_k\right)\right]  \nonumber\\
\mathrm{s.t.}\;\;\;\; &(w^f_{k},a^f_{k}) \in \mathcal{X}( r^f_{k}, a^f_{k-1}),\;\;\;\; k\geq t \nonumber
\end{align}
which must be solved for $f=1,...,F$. Indeed, the aggregate cost associated with (P$2$) will not depend on variables corresponding to files $f'\neq f$ \cite{Luismi2014DP_CRs}. This is the case if, for instance, the involved variables are independent of each other (which is the setup considered here), or when the focus is on a large system where the contribution of an individual variable to the aggregate network behavior is practically negligible.

\begin{figure*}[t!]
	\tiny
	\begin{align} 
	\label{long1} \left(w_t^{f\ast}, a_t^{f\ast}\right) :=& \!\!\!\!\!\underset{(w,a) \in \mathcal{X}( r^f_t, { a^f_{t-1}})}{\arg\mathop{\min}} \left \{ {\mathbb E}_{\boldsymbol \theta_k^f} \left[\underset{(w_k,a_k)\in {\cal X}(r^f_k,{ a^f_{k-1}})}{\min} \left\{\sum \limits_{k=t}^{\infty} \gamma^{k-t} \left[c^f_k(a^f_k,w^f_k;\rho^f_k,\lambda^f_k) \Big | a^f_t \!=\! a, w^f_t \!=\! w, {\boldsymbol \theta^f_t = {\boldsymbol \theta}^f_0} \right]\right\}\right] \right\}\\ 
	\label{long2}  = &\!\!\!\!\! \underset{(w,a) \in \mathcal{X}( r^f_t, { a^f_{t-1}})}{\arg\mathop{\min}} \left\{c^f_t(a,w;\rho_t^f,\lambda_t^f)  +   {\mathbb {E}}_{\boldsymbol \theta_k^f} \left[ \underset{(w_k,a_k)\in {\cal X}(r^f_k,{ a^f_{k-1}})}{\min}\sum \limits_{k=t+1}^{\infty} \gamma^{k-t}  \left[c^f_k(a^f_k,w^f_k;\rho^f_k,\lambda^f_k) \Big | s^f_{t+1} = a \right]  \right] \right \}
	\end{align} 
	\begin{align}
	& \label{value_function}  V^f\left(s^f,r^f;\rho^f,\lambda^f\right) := \!\!\!\!\!\!\!\!\underset{(w,a) \in \mathcal{X}( r^f_t, { a^f_{t-1}})}{\mathop{\min}} \left \{ {\mathbb E}_{\boldsymbol \theta_k^f} \left[\underset{(w_k,a_k)\in {\cal X}(r^f_k,{ a^f_{k-1})}}{\min} \left\{\sum \limits_{k=t}^{\infty} \gamma^{k-t} \left[c^f_k(a^f_k,w^f_k;\rho^f_k,\lambda^f_k) \Big | a^f_t \!=\! a, w^f_t \!=\! w, {\boldsymbol \theta^f_t = {\boldsymbol \theta}^f} \right]\right\}\right] \right\} 
	\end{align} 
	\begin{align}
	\nonumber 
	{\bar V}^f(s^f) :=& % {\mathbb E}_{r^f, \rho^f,\lambda^f} V^f\left(s^f,r^f;\rho^f,\lambda^f\right)  \\ 		& \nonumber\hspace{2.1 cm} = 
	{\mathbb E}_{\boldsymbol \theta^f} \left[        \underset{(w,a) \in \mathcal{X}( r^f_t, { a^f_{t-1}})}{\mathop{\min}} \left \{ {\mathbb E}_{\boldsymbol \theta_k^f} \left[\underset{(w_k,a_k)\in {\cal X}(r^f_k,{ a^f_{k-1})}}{\min} \left\{\sum \limits_{k=t}^{\infty} \gamma^{k-t} \left[c^f_k(a^f_k,w^f_k;\rho^f_k,\lambda^f_k) \Big | a^f_t \!=\! a, w^f_t \!=\! w, {\boldsymbol \theta^f_t = {\boldsymbol \theta}^f} \right]\right\}\right] \right\}\right] \\
	\label{bellman} =  &{\mathbb E}_{\boldsymbol \theta^f}  \mathop {\min} \limits_{(w,a) \in \mathcal{X}( r^f, s^f)} \left\{ c^f_0(a,w;\rho^f,\lambda^f)   + \gamma {\bar V^f}(a)  \right\}
	%		&   \label{optimal_aw}\hspace{0.9 cm} \small{\left(a_t^{f\ast}, w^{f\ast}_t\right) = \arg \mathop {\min} \limits_{\scriptstyle{{ \hspace{0.8cm}(\alpha, \omega) \in \left\{0,1\right \}^2}}\hfill\atop
	%		\scriptstyle{s.t.} {\small\begin{array}{l}
	%					s^f_t = s^f, r_t^f = r^f\\
	%					\text{\eqref{c1}-\eqref{c3}}
	%					\end{array}} \hfill} \left\{  c^f_t(\alpha,\omega;\rho^f_t,\lambda^f_t)   + \gamma {\bar V}^f \left(\alpha\right) \right \}}
	\\ \nonumber
	\end{align} 
	\hrulefill
\end{figure*}

\noindent \textit{Bellman equations and value function:} The DP in (P$2$) can be solved with the corresponding Bellman equations, which require finding the associated value functions. To this end, consider the system at  time $t$, where the cache state as well as the file requests and cost parameters are all given, so that we can write $s^f_t = s^f_{0}$ and $\boldsymbol \theta^f_t = \boldsymbol \theta^f_0$. Then, the optimal fetch-cache decision  $(w^{f \ast}_t,a^{f \ast}_t)$ is readily expressible  as the  solution to \eqref{long1}. The objective in \eqref{long1} is rewritten in \eqref{long2} as the summation of current and discounted average future costs. The form of \eqref{long2} is testament to the fact that problem (P$2$) is a DP and the caching decision $a$ influences not only the current cost $c_t^f(\cdot)$, but also future costs through the second term as well. Bellman equations can be leveraged for tackling such a DP. Under the stationarity assumption for variables $r_{t}^{f}$, $\rho_{t}^{f}$ and $\lambda_{t}^{f}$, the term accounting for the future cost can be rewritten in terms of the \textit{ stationary value function} $V^f\left(s^f,r^f;\rho^f,\lambda^f\right)$. This function, formally defined in \eqref{value_function}, captures the minimum sum average cost for the ``state'' $(s^f,r^f)$, parametrized by $(\lambda^f,\rho^f)$, where for notational convenience, we define  $\boldsymbol \theta^f := [r^f,\rho^f,\lambda^f]$. 

\subsection{Marginalized value-function}
If one further assumes that price parameters and requests are i.i.d. across time, it can be shown  that the  optimal solution to (P$2$) can be expressed in terms of the  \textit{reduced value function}~\cite{Luismi2014DP_CRs}
\begin{equation}
{\bar V}^f \left(s^f\right) := {\mathbb {E} _ {\boldsymbol \theta^f}} \left[V^f\left(s^f,r^f;\rho^f, \lambda^f\right)\right],
\label{eq:marginalzied_V}
\end{equation}
where the expectation is w.r.t ${\boldsymbol \theta^f}$. Marginalization of the value function is important not only because it captures the average future cost  of file $f$ for cache state $s^f\in\{0,1\}$, but also because ${\bar V}^f (\cdot)$ is a function of a binary variable,  and  therefore its estimation requires only estimating two values. This is in contrast with the original four-dimensional value function in \eqref{value_function}, whose estimation is more difficult due to its continuous arguments. 

By rewriting the proposed alternative value function ${\bar V}^f (\cdot)$ in a recursive fashion as the summation of instantaneous cost and discounted future values ${\bar V}^f (\cdot)$,  one readily arrives at the Bellman equation form provided in \eqref{bellman}. Thus, the problem reduces to finding $\bar{V}^f(0)$ and $\bar{V}^f(1)$ for all $f$, after which the optimal fetch-cache decisions $(w_t^{f\ast}, a^{f\ast}_t)$ are easily found as the solution to 

\begin{align}
{\textrm {(P$3$)} }\quad  \min \limits_{(w,a)} \;\;\;\;&  c^f_t(a,w;\rho^f_t,\lambda^f_t)   + \gamma {\bar V}^f \left(a\right)  \nonumber\\
\mathrm{s.t.}\;\;\;\; &(w,a) \in \mathcal{X}( r^f_t, a^f_{t-1}). \nonumber
\end{align}

If the value-function is known, so that we have access to  ${\bar V}^f(0)$ and ${\bar V}^f(1)$, the corresponding optimal (Bellman) decisions can be found as
%\begin{subequations} 
%\begin{align} 
%\!\!r_t^f=0,~s_t^f=0 & \quad w_t^f=a_t^f,~a_t^f = \mathbb{I}_{\{\Delta {\bar V}^f_\gamma \geq \lambda_t^f + \rho_t^f \}} \label{E:Bellman_input00}\\
%\!\!r_t^f=0,~s_t^f=1 & \quad w_t^f=0,~a_t^f = \mathbb{I}_{\{\Delta {\bar V}^f_\gamma \geq \rho_t^f \}} \\
%\!\!r_t^f=1,~s_t^f=0 & \quad w_t^f=1,~a_t^f = \mathbb{I}_{\{\Delta {\bar V}^f_\gamma \geq \rho_t^f \}} \\
%\!\!r_t^f=1,~s_t^f=1 & \quad w_t^f=0,~a_t^f = \mathbb{I}_{\{\Delta {\bar V}^f_\gamma \geq  \rho_t^f \}},\label{E:Bellman_input11}
%\end{align}
%\end{subequations}
\begin{subequations} 
	\begin{align} 
	\!\!&w_t^f=a_t^f,~a_t^f = \mathbb{I}_{\{\Delta {\bar V}^f_\gamma \geq \lambda_t^f + \rho_t^f \}} &  \text{if}~ (r_t^f,s_t^f)=(0,0) \label{E:Bellman_input00}\\
	\!\!&w_t^f=0,~a_t^f = \mathbb{I}_{\{\Delta {\bar V}^f_\gamma \geq \rho_t^f \}} &  \text{if}~ (r_t^f,s_t^f)=(0,1)\\
	\!\!&w_t^f=1,~a_t^f = \mathbb{I}_{\{\Delta {\bar V}^f_\gamma \geq \rho_t^f \}} &  \text{if}~ (r_t^f,s_t^f)=(1,0)\\
	\!\!&w_t^f=0,~a_t^f = \mathbb{I}_{\{\Delta {\bar V}^f_\gamma \geq  \rho_t^f \}}&  \text{if}~ (r_t^f,s_t^f)=(1,1)\label{E:Bellman_input11}
	\end{align}
\end{subequations}
where $\Delta {\bar V}^f_\gamma $ represents the \textit{future} marginal cost, which is obtained as $\Delta {\bar V}^f_\gamma = \gamma (  {\bar V}^f(1) -  {\bar V}^f(0))$, and $\mathbb{I}_{\{ \cdot \}}$ is an indicator function that yields value one if the condition in the argument holds, and zero otherwise. 

The next subsection discusses how ${\bar V}^f(0)$ and ${\bar V}^f(1)$ can be calculated, but first a remark is in order.

\vspace{.15cm}
\noindent {\bf{Remark 1 (Augmented value functions)}}. The value function ${\bar V}^f (s^f)$ can be redefined to account for extra information on $r_t^f$, $\rho_t^f$ or $\lambda_t^f$, if available. For instance, consider the case where  the distribution of $r_t^f$ can be parametrized by $p^f$, which measures content ``popularity'' \cite{breslau1999web}. In such cases,  the value function can incorporate the popularity parameter as an additional input to yield ${\bar V}^f (s^f,p^f)$. Consequently, the optimal decisions will depend not only on the current requests and prices, but also on the (current) popularity $p^f$. This indeed broadens the scope of the proposed approach, as certain types of \textit{non-stationarity} in the distribution of $r_t^f$ can be handled by allowing $p^f$ to (slowly) vary with time.

\subsection{Value function in closed form}\label{Subsec:value_function_stationary}

\noindent 
For notational brevity, we have removed the superscript $f$ in this subsection, and use $\bar V_0$ and $\bar V_1$ in lieu of ${\bar V}(0)$, and ${\bar V}(1)$. Denoting the \textit{long-term} popularity of the content as $p:={\mathbb E} [r_t]$, using the expressions for the optimal actions in  \eqref{E:Bellman_input00}-\eqref{E:Bellman_input11}, and leveraging the independence among $r_t$, $\lambda_t$, and $\rho_t$, the expected cost-to-go function can be readily derived as in \eqref{E:Eq_V1_a}-\eqref{E:Eq_V0_b}. The expectation in \eqref{E:Eq_V1_b_0} is w.r.t. $\rho$, while that in~\eqref{E:Eq_V0_b_0} is w.r.t. both~$\lambda$~and~$\rho$. 
\begin{figure*}[t!]
	\tiny
	\begin{eqnarray}
	\label{E:Eq_V1_a}   
	\bar V_1 & =& (1-p) \Big({\mathbb E} \min_{a \in \left\{0,1\right\}} \left[\gamma \bar V_0 (1-a) + ( \rho +\gamma \bar V_1)a~ \Big|s = 1, r = 0 \right] \Big) \label{E:Eq_V1_b_0}
	+ p \Big({\mathbb E} {\min_{a\in \left\{0,1\right\}}} \left[ \gamma \bar V_0 (1-a) + ( \rho +\gamma \bar V_1)a~\Big|s=1, r= 1 \right] \Big) \nonumber \\ \nonumber \\
	&=&\gamma \bar V_0 \Pr \big( \rho \geq \Delta \bar V_{\gamma}\big) + {\mathbb E} \Big( \rho  + \gamma \bar V_1 \Big|\rho < \Delta \bar V_{\gamma} \Big)  \Pr \big( \rho < \Delta \bar V_{\gamma}\big)\label{E:Eq_V1_b}
	\end{eqnarray}
	
	\begin{eqnarray}
	\label{E:Eq_V0_b_0} \bar V_0 &=& (1-p) \Big({\mathbb E} {\min_{a\in\left\{0,1\right\}}} \left[\gamma \bar V_0 (1-a) + (\lambda + \rho +\gamma \bar V_1)a~\Big|s=0, r=0 \right] \Big)  
	\\ \nonumber &+& p \Big({\mathbb E} {\min_{a \in \left\{0,1\right\}} } \left[(\lambda + \gamma \bar V_0) (1-a) + (\lambda + \rho +\gamma \bar V_1)a~\Big|s=0,r=1] \right] \Big) \label{E:Eq_V0_b} \\ 
	&=& (1-p) \Big( \gamma \bar V_0 \Pr \left(\lambda + \rho \geq \Delta \bar V_{\gamma}\right) +  {\mathbb E} \left( \lambda + \rho +\gamma \bar V_1~\Big|\lambda + \rho < \Delta \bar V_{\gamma}\right)   \Pr \left( \lambda + \rho < \Delta \bar V_{\gamma} \right) \Big)
	\\ \nonumber &+& p \Big({\mathbb E} [\lambda ] + \gamma \bar V_0 \Pr \left(\rho \ge \Delta \bar V_\gamma \right) +{\mathbb E} \left(\rho + \gamma \bar V_1~\Big|\rho \le \Delta \bar V_\gamma \right) \Pr \left(\rho \le \Delta \bar V_\gamma \right) \Big)
	\end{eqnarray}
	\hrulefill
\end{figure*}

Solving the system of equations in \eqref{E:Eq_V1_a}-\eqref{E:Eq_V0_b} yields the optimal values for $\bar V_1$ and $\bar V_0$. A simple solver would be to perform exhaustive search over the range of these values since it is only a two-dimensional search space.  However, a better alternative to solving the given system of equations is to rely on the well known \textit{value iteration} algorithm. In short, this is an offline algorithm, which per iteration $i$ updates the  estimates $\{\bar{V}^{i+1}_0 , \bar{V}^{i+1}_1\}$ by computing the expected cost using $\{\bar{V}^{i}_0 , \bar{V}^{i}_1\}$, until the desired accuracy is achieved. This scheme is tabulated in detail in Algorithm~1, for which the distributions of $r, \rho, \lambda$ are assumed to be known.  

\begin{algorithm}[t]
	{{\bf Initialize}  $\gamma<1$, probability density function of $\rho,\lambda$ and $r$,  precision $\epsilon$, in order to stop} 
	\newline 
	{{\bf Initialize} $\bar V_0$, $\bar V_1$} 
	\newline
%	\SetKwInOut{Input}{Input}
%	\SetKwInOut{Output}{Output}
%	\underline{Set} $\bar V^0_0 = \bar V^0_1 = 0$ \;
%	
%	\Input{$\gamma<1$, probability density function of $\rho,\lambda$ and $r$,  precision $\epsilon$, in order to stop}
%	\Output{$\bar V_0$, $\bar V_1$}
	 {While $|\bar V^i_s-\bar V^{i+1}_s| < \epsilon; s \in \left\{0,1\right\}$} 
	 \newline
	 {For \;\; {$s = 0,1$} 
	 \newline
	 {$\bar{V}^{i+1}_s = \mathbb{E}_{r,\rho,\lambda}\mathop {\min} \limits_{(w,a) \in \mathcal{X}( r, s)} \left\{  c(a,w;\rho,\lambda) + \gamma {\bar V^i_a}\right\}$}
		$i = i+1$}    
	\caption{Value iteration  for finding $\bar V \left(\cdot\right)$}
\end{algorithm}

\noindent \textbf{Remark 2 (Finite-horizon approximate policies).} In the proposed algorithms, namely exhaustive search as well as Algorithm~1, the solver is required to compute an expectation, which can be burdensome in setups with limited computational resources. For such scenarios,  the class of finite-horizon policies emerges as a computationally affordable suboptimal alternative. The idea behind such policies is to truncate the infinite summation in the objective of (P$1$); thus, only considering the impact of the current decision on a few number of future time instants denoted by $h$, typically referred to as the \textit{horizon}. The extreme case of a finite-horizon policy is that of a \textit{myopic policy} with $h=0$, which ignores any future impact of current decision,  a.k.a. zero-horizon policy, thus taking the action which minimizes the instantaneous cost. This is equivalent to setting the future marginal cost to zero, hence solving \eqref{E:Bellman_input00}-\eqref{E:Bellman_input11} with  $\Delta {\bar V}_\gamma=\Delta {\bar V}_\gamma^{h=0} = 0$. 

Another commonly used alternative is to consider the impact of the current decision for only the next time instant, which corresponds to the so-called horizon-1 policy. This  entails setting the future cost at $h=1$ as $\Delta {\bar V}_\gamma^{h=1} = \gamma (\bar V_1^{h=0} - \bar V_0^{h=0})$ with
\begin{eqnarray}
\!\! \bar V_0^{h=0} \!\!& = & \!\!\! (1-p) {\mathbb E}[\lambda w^{h=0}+\rho a^{h=0}|s=0,r=0] \nonumber \\
\!\!\!\!& + &\!\!\! p {\mathbb E}[\lambda w^{h=0}\!+\!\rho a^{h=0}|s=0,r=1]=p{\mathbb E}[\lambda] \label{eq_value_function_finite_horizon0}\\
\!\!\bar V_1^{h=0} \!\!& = & \!\!\!(1-p) {\mathbb E}[\lambda w^{h=0}+\rho a^{h=0}|s=1,r=0] \nonumber \\ 
\!\!\!\!&+& \!\!\!p {\mathbb E}[\lambda w^{h=0}+\rho a^{h=0}|s=1,r=1]=0,\label{eq_value_function_finite_horizon1}
\end{eqnarray}
which are then substituted into \eqref{E:Bellman_input00}-\eqref{E:Bellman_input11} to yield the actions $ w^{h=1}$ and  $a^{h=1}$. The notation $ w^{h=0}$ and  $a^{h=0}$ in \eqref{eq_value_function_finite_horizon0} and \eqref{eq_value_function_finite_horizon1} is used to denote the actions obtained when \eqref{E:Bellman_input00}-\eqref{E:Bellman_input11} are solved using the future marginal cost at horizon zero $\Delta {\bar V}_\gamma^{h=0}$, which as already mentioned, is zero; that is, under the myopic policy in lieu of the original optimal solution. Following an inductive argument, the future marginal cost at $h=2$ is obtained as $\Delta {\bar V}_\gamma^{h=2} = \gamma (\bar V_1^{h=1} - \bar V_0^{h=1})$ with
\begin{eqnarray}
\nonumber
\!\! \bar V_0^{h=1} \!\!& = & \!\!\! (1-p) {\mathbb E}[\lambda w^{h=1}+\rho a^{h=1}+ \gamma \bar{V}_a^{h=0}|s=0,r=0] \\\nonumber 
\!\!\!\!& + &\!\!\! p {\mathbb E}[\lambda w^{h=1}\!+\!\rho a^{h=1}+ \gamma \bar{V}_a^{h=0}|s=0,r=1], \\ 	\nonumber
\!\!\bar V_1^{h=1} \!\!& = & \!\!\!(1-p) {\mathbb E}[\lambda w^{h=1}+\rho a^{h=1}+\gamma \bar{V}_a^{h=0}|s=1,r=0] \\  \nonumber
\!\!\!\!&+& \!\!\!p {\mathbb E}[\lambda w^{h=1}+\rho a^{h=1}+\gamma \bar{V}_a^{h=0}|s=1,r=1],
\end{eqnarray}
which will allow to obtain the actions $ w^{h=2}$ and  $a^{h=2}$. While increasing horizons can be used, as $h$ grows large, solving the associated equations becomes more difficult and computation of the optimal stationary policies, is preferable. 

\subsection{State-action value function ($Q$-function):}\label{Subsec:q_function_stationary}
In many practical scenarios, knowing the underlying distributions for $\rho_t$, $\lambda_t$ and $r_t$ may not be possible, which motivates the introduction of online solvers that can learn the parameters on-the-fly. As clarified in the ensuing sections, in such scenarios, the so-called $Q$-function (or state-action value function) becomes helpful, since there are rigorous theoretical guarantees on the convergence of its stochastic estimates; see~\cite{tsitsiklis} and \cite{watkins1992q}. 
Motivated by this fact, instead of formulating our dynamic program using the value (cost-to-go) function, we can alternatively formulate it using the $Q$-function. 
Aiming at an online solver, let us tackle the DP through the estimation (learning) of the $Q$-function. Equation\footnote{Equations \eqref{Q_function}-\eqref{marg_q}, and \eqref{eq_Q_factors_esemble} are shown at the top of page 7.} \eqref{Q_function} defines the $Q$-function for a specific file under a given state $\left(s_t,r_t\right)$, parametrized by cost parameters $\left(\rho_t,\lambda_t\right)$. Under stationarity distribution assumption for $\left\{ \rho_t, \lambda_t,r_t \right\}$, the $Q$-function $ Q\left(s_t,r_t, w_t,a_t;\rho_t,\lambda_t\right)$ accounts for the minimum average aggregate cost at state $\left(s_t,r_t\right)$, and taking specific fetch-cache decision $(w_t,a_t)$ as for the first decision, while followed by the best possible decisions in next slots. This function is parametrized by $\left(\rho_t,\lambda_t\right)$ since while making  the current cache-fetch decision, the current values for these cost parameters are assumed to be known. The original $Q$-function in \eqref{Q_function} needs to be learned over all values of $\left\{s_t,r_t, w_t,a_t,\rho_t, \lambda_t, r_t\right\}$, thus suffering from the curse of dimensionality, especially due to the fact that $\rho_t$ and $\lambda_t$ are continuous variables. 

\begin{figure*}[t!]
	\tiny
	\begin{align} 
	\label{Q_function}  Q\left(s_t,r_t, {w_t,a_t};\rho_t,\lambda_t\right) & := {\mathbb E} \left[\underset{\left\{(w_k,a_k)\in {\cal X}(r_k,{ a_{k-1}})\right\}_{k=t+1}^{\infty}}{\min} \left\{\sum \limits_{k=t}^{\infty} \gamma^{k-t} \left[c_k(a_k,w_k;\rho_k,\lambda_k) \Big |a_t, w_t, {\boldsymbol \theta_t = {\boldsymbol \theta}} \right]\right\}\right] \\   & \hspace{-2cm} = \underbrace{c_{ t}(a_t,w_t;\rho_t,\lambda_t)}_{\textrm{Immediate cost}} + \gamma \underbrace{{\mathbb E} \left[\underset{\left\{(w_k,a_k)\in {\cal X}(r_k,{ a_{k-1}})\right\}_{k=t+1}^{\infty}}{\min} \left\{\sum \limits_{k=t+1}^{\infty} \gamma^{k-{(t+1)}} \left[c_k(a_k,w_k;\rho_k,\lambda_k) \Big | s_{t+1} = a_t \right]\right\}\right]}_{\textrm{Average minimum future cost}}
	\end{align} 
	\begin{align}
	\label{marg_q}
	{\bar Q}_{r_t,s_t}^{w_t,a_t}:= &\; {{\mathbb E}}_{ \rho_t, \lambda_t} \; \left[ Q\left(s_t,r_t, { w_t,a_t };\rho_t,\lambda_t\right) \right], \quad \forall (w_t,a_t) \in { \mathcal X}(r_t,{{a_{t-1}}}) \\ = &\; {\mathbb E}_{ \rho_t, \lambda_t} \left[c_{ t}(a_t,w_t;\rho_t,\lambda_t) \right]  + \gamma \left[ {\mathbb E}_{{\boldsymbol \theta}_{t+1}} \left[ Q\left(s_{t+1},r_{t+1}, { w_{t+1}^{\ast},a_{t+1}^{\ast}};\rho_{t+1},\lambda_{t+1}\right) \Big| {{\boldsymbol \theta}_{t+1}}, s_{t+1} = a_t \right] \right]. 
	\nonumber
	\end{align}
	\hrulefill
	\begin{align}\label{eq_Q_factors_esemble}
	\bar Q_{r,s}^{w,a} =  {\mathbb E} [\lambda] w +  {\mathbb E} [\rho] a +  \gamma (1-p) \hspace{-.75 cm}  \sum_{\forall (z_1,z_2){ \in \mathcal{X}(0,a)}} \hspace{-0.5 cm} \bar Q_{0,a}^{z_1,z_2} \Pr \Big((w_{t+1}^*,a_{t+1}^*)=(z_1,z_2)|(s_{t+1},r_{t+1})=(a,0)\Big)\nonumber && \\   +\gamma p  \sum_{\forall (z_1,z_2) { \in \mathcal{X}(1,a)}} \bar Q_{1,a}^{z_1,z_2} \Pr \Big((w_{t+1}^*,a_{t+1}^*)=(z_1,z_2)|(s_{t+1},r_{t+1})=(a,1)\Big). && \end{align}
	\hrulefill
\end{figure*}
To alleviate this burden, we define the \textit{marginalized} $Q$-function $ Q(s_t,r_t, w_t,a_t)$ in \eqref{marg_q}. By changing the notation for clarity of exposition, the marginalized $Q$-function, $\bar Q_{r_t,s_t}^{w_t,a_t}$, can be rewritten in a more compact form as 
%$$Q_{r_t,s_t}^{w_t,a_t} = \lambda_t w_t + \rho_t a_t + \gamma {\mathbb E} \big[ Q_{r_{t+1},a_t}^{w_{t+1}^*,a_{t+1}^*} \big].$$
\begin{equation}
\label{M_Q}
\bar Q_{r_t,s_t}^{w_t,a_t} = {\mathbb E} \Big[\lambda_t w_t + \rho_t a_t + \gamma  \bar Q_{r_{t+1},a_t}^{w_{t+1}^*,a_{t+1}^*} \Big] {\; \forall (w_t,\!a_t) \!\in\! { \mathcal {X}}(r_t,{a_{t-1}})}.
\end{equation}
Note that, while the marginalized value-function is only a function of the state, the marginalized $Q$-function depends on both the state $\left(r,s\right)$ and the immediate action $\left(w,a\right)$. The main reason one prefers to learn the value-function rather than the $Q$-function is that the latter is computationally more complex. To see this, note that the input space of $\bar Q_{r_t,s_t}^{w_t,a_t}$ is a four-dimensional binary space, hence the function has $2^4=16$ different inputs and one must estimate the corresponding $16$ outputs. Each of these possible values are called $Q$-factors, and under the stationarity assumption, they can be found using \eqref{eq_Q_factors_esemble} 
defined for all $(r,s,w,a)$. In this expression, we have $(z_1,z_2)\in\{0,1\}^2$ and the term $\Pr \left( \left(w_{t+1}^\ast a^{\ast}_{t+1}\right) = (z_1,z_2)\right)$ stands for the probability of specific action $(z_1, z_2)$ to be optimal at slot $t+1$. This action is random because the optimal decision at $t+1$ depends on $\rho_{t+1}$, $\lambda_{t+1}$ and $r_{t+1}$, which are not known at slot $t$.  Although not critical for the discussion, if needed, one can show that half of the 16 $Q$-factors can be discarded, either for being infeasible -- recall that $(w_t,a_t) \!\in\! { \mathcal {X}}(r_t,{a_{t-1}})$ -- or suboptimal. This means that \eqref{eq_Q_factors_esemble} needs to be computed only for $8$ of the $Q$-factors.

From the point of view of offline estimation, working with the $Q$-function is more challenging than working with the $V$-function, since more parameters need to be estimated. In several realistic scenarios however, the distributions of the state variables are unknown, and one has to resort to stochastic schemes in order to learn the parameters on-the-fly. In such scenarios, the $Q$-function based approach  is preferable, because it enables learning the optimal decisions in an online fashion even when the underlying distributions are unknown. 

\subsection{Stochastic policies: Reinforcement learning}\label{Subsec:value_q_function_stochastic}
As discussed in Section \ref{Subsec:value_function_stationary}, there are scenarios where obtaining the optimal value function (and, hence, the optimal stationary policy associated with it) is not computationally feasible. The closing remark in that section discussed policies which, upon replacing the optimal value function with approximations easier to compute, trade reduced complexity for loss in optimality. However, such reduced-complexity methods still require knowledge of the state distribution [cf. \eqref{eq_value_function_finite_horizon0} and \eqref{eq_value_function_finite_horizon1}]. In this section, we discuss stochastic schemes to approximate the value function under unknown distributions, thus relaxing assumption AS2 made earlier. The policies resulting from such stochastic methods offer a number of advantages since they: (a) incur a reduced complexity; (b) do not require knowledge of the underlying state distribution; (c) are able to handle some non-stationary environments; and in some cases, (d) they come with asymptotic optimality guarantees. To introduce this scheme, we first start by considering a simple method that updates stochastic estimates of the value function itself, and then proceed to a more advanced method which tracks the value of the $Q$-function. Specifically, the presented method is an instance of the celebrated $Q$-learning algorithm \cite{watkins1989learning}, which is the workhorse of stochastic approximation in DP.

\vspace{.1cm}
\subsubsection{Stochastic value function estimates}
The first method relies on current stochastic estimates of $\bar V_0$ and $\bar V_1$, denoted by $\hat{\bar V}_0(t)$ and $\hat{\bar V}_1(t)$ at time $t$ (to be defined rigorously later). Given $\hat{\bar V}_0(t)$ and $\hat{\bar V}_1(t)$ at time $t$, the (stochastic) actions $\hat{w}_t$ and $\hat{a}_t$  are taken via solving \eqref{E:Bellman_input00}-\eqref{E:Bellman_input11} with $\Delta {\bar V}_\gamma=\gamma(\hat{\bar V}_0(t)-\hat{\bar V}_1(t))$. Then, stochastic estimates of the value functions $\hat{\bar V}_0(t)$ and $\hat{\bar V}_1(t)$ are updated as
\begin{itemize}
	\item If $s_t=0$, then $\hat{\bar V}_1(t+1)=\hat{\bar V}_1(t)$ and 
	$\hat{\bar V}_0(t+1)= (1-{ \beta_t}) \hat{\bar V}_0(t) + { \beta_t}(\hat{w}_t\lambda_t + \hat{a}_t\rho_t  + \gamma \hat{\bar V}_{\hat{a}_t}(t) )$;
	\item If $s_t=1$, then $ \hat{\bar V}_0({t+1})=\hat{\bar V}_0({t})$ and 
	$\hat{V}_1(t+1)= (1-{ \beta_t}) \hat{\bar V}_1(t) + { \beta_t}(\hat{w}_t\lambda_t + \hat{a}_t\rho_t  + \gamma \hat{\bar V}_{\hat{a}_t}(t) ) $;
\end{itemize}
where ${\beta_t}>0$ denotes the stepsize. While easy to implement (only two recursions are required), this algorithm has no optimality guarantees.

\vspace{.1cm}
\subsubsection{\textit{Q}-learning algorithm} Alternatively, one can run a stochastic approximation algorithm on the $Q$-function. This entails replacing the $Q$-factors $\bar Q_{r,s}^{w,a}$ with stochastic estimates $\hat{\bar Q}_{r,s}^{w,a}(t)$. To describe the algorithm, suppose for now that at time $t$, the estimates $\hat{\bar Q}_{r,s}^{w,a}(t)$ are known for all $(r,s,w,a)$. 
Then, in a given slot $t$ with $(r_{t}, s_{t})$,  action $ \left(\hat{w}^\ast_t,\hat{a}^\ast_t\right)$ is obtained via either an exploration or an exploitation step. When exploring, which happens with a small probability $\epsilon_t$, a random and feasible action $ \left(\hat{w}^\ast_t,\hat{a}^\ast_t\right) \in {\mathcal X} \left(r_t,a_{t-1}\right)$ is taken. In contrast, in the exploitation mode, which happens with a probability $1 - \epsilon_t$,  the optimal action according to the current estimate of $\hat{\bar Q}_{r,s}^{w,a}(t)$ is 
\begin{equation} 
\setcounter{equation}{\value{equation}+1}
\left(\hat{w}^\ast_t,\hat{a}^\ast_t\right) := \underset{(w,a) \in \mathcal{X}( r_{t}, a_{t-1})}{\arg \min} \;  w\lambda_t + a \rho_t + \gamma \hat{\bar Q}_{r_t,s_t}^{w,a}(t) 
\label{margQ_solver}.
\end{equation} 
After taking this action, going to next slot $t+1$, and observing $\rho_{t+1}, \lambda_{t+1}$, and $r_{t+1}$, the $Q$-function estimate is updated as
\begin{flalign}
\label{Q_update}
&\hat{\bar Q}_{r,s}^{w,a}(t+1) = \nonumber \\
&\begin{cases}
\!\hat{\bar Q}_{r,s}^{w,a}(t) \quad {\text{ if} }\quad  (r,s,w,a)\neq(r_t,s_t,\hat{w}^\ast_t,\hat{a}^\ast_t) \\ \\
\!\!(1\!-\!{ \beta_t}) \hat{\bar Q}_{r_t,s_t}^{\hat w^{\ast}_t,\hat a^{\ast}_t}(t)  +  { \beta_t}\Big(\hat{w}_t^{\ast} \lambda_t + \hat{a}_t^{\ast} \rho_t  + {\gamma \hat{\bar Q}_{r_{t+1},\hat a_t^\ast}^{\hat w_{t+1}^{\ast}, \hat a_{t+1}^{\ast}}(t)\Big)} \;\text{o.w.,} 
\end{cases}
\end{flalign}
where ``o.w.'' stands for ``otherwise'', $(\hat w^{\ast}_{t+1},\hat a^{\ast}_{t+1})$ is the optimal action for the next slot and, if needed, the stepsize $\beta_t$ can be adapted for each particular state-action pair. %and $\hat{\bar Q}_{r,s}^{w,a}(t+1)=\hat{\bar Q}_{r,s}^{w,a}(t)$ if $(r,s,w,a)\neq (r_t,s_t,\hat{w}^\ast_t,\hat{a}^\ast_t)$. 
This update rule describes one of the possible implementations of the $Q$-learning algorithm, which was originally introduced in \cite{watkins1989learning}. This online algorithm enables making sequential decisions in an unknown environment, and is guaranteed to learn optimal decision-making rules under certain conditions specified next~\cite{watkins1992q}.

Regarding convergence of the Q-learning algorithm, the following necessary conditions should hold~\cite{ watkins1992q,ODE}: 	(c1) all feasible state (${r,s}$) and action (${w,a}$) pairs should be continuously updated; and, (c2) the learning rate ${\beta_t}$ should be a diminishing step size. 		
	Under these conditions, the factors $\hat{\bar Q}_{r,s}^{w,a}$ converge to their optimal value ${{\bar Q}^{\ast w,a}}_{\;\;r,s}$ with probability 1; see \cite{ODE} for details. To satisfy (c1), various exploration-exploitation algorithms have been proposed~\cite[p.~839]{russell2016artificial}. Particularly, any such scheme needs to be \emph{greedy in the limit of infinite exploration}, or GLIE~\cite[p.~840]{russell2016artificial}. A common choice to meet this property is the $\epsilon$-greedy approach,  as considered in this work, with $\epsilon_t = 1/t$, which provides guaranteed yet slow convergence. In practice however, $\epsilon_t$ can be set to a small value for faster convergence~\cite{ODE}, \cite{tsitsiklis}. To satisfy the diminishing step size rule in (c2), let us define ${t^{w,a}_{r,s}}$ as the index of the $t$-th time when the state-action pair $(r,s)$ and $(w,a)$ is visited, and updated with the corresponding learning rate $\beta_{t^{w,a}_{r,s}}$. Condition (c2)  requires   $\sum_{t=1}^{\infty} \beta_{t^{w,a}_{r,s}} = \infty,$ and $ {\sum}_{t=1}^{\infty} \beta_{t^{w,a}_{r,s}}^2 < \infty $
	to hold for all feasible state-action pairs, a typical choice for which is setting $\beta_{t^{w,a}_{r,s}} = 1/t$. Similar to $\epsilon_t$, a constant but small learning rate  is preferred in practice as it endows the algorithm to adapt to possible changes of pertinent parameters in dynamic settings.

The resultant algorithm for the problem at hand is tabulated in Algorithm~\ref{Q_learning}. It is important to stress that in our particular case, we expect the algorithm to converge fast. That is the case because, under the decomposition approach followed in this paper as well as the introduction of the marginalized $Q$-function, the state-action space of the resultant $Q$-function has very low dimension and hence, only a small number of $Q$-factors need to be estimated.  

\begin{algorithm}[t]
	\caption{$Q$-learning algorithm to estimate ${\bar Q}_{r,s}^{w,a}$ for a given file $f$}
	\label{Q_learning}
	{{\bf Initialize}  $\hat {\bar Q}_{r,s}^{w,a}(1) = 0$, $s_1 = 0$, {$\{r_0, \rho_0, \lambda_0\}$ are revealed}	} 
	\newline
	{{\bf Output}  $ \hat {\bar Q}_{r,s}^{w,a}(t+1)$} 
	\newline 
%	\SetKwInOut{Input}{Input}
%	\SetKwInOut{Output}{Output}
%	\Input{$0< \gamma, {\beta_t}< 1$}
%	\Output{$ \hat {\bar Q}_{r,s}^{w,a}(t+1)$}
%	{\textbf{Initialize} $\hat {\bar Q}_{r,s}^{w,a}(1) = 0$, $s_1 = 0$, {$\{r_0, \rho_0, \lambda_0\}$ are revealed}	\\
		{For \quad $t = 1 , 2, \ldots$}
		\newline
		{\textrm	For the current state $(r_t,s_t)$, choose $(\hat w_t^{\ast},\hat a_t^{\ast})$ }
		 \begin{eqnarray}
		 (\hat {w}_t^{\ast},\hat {a}_t^{\ast})  = \left\{
			\begin{array}{ll}
			{\textrm {Solve \eqref{margQ_solver}}} 
			 \textrm{w.p. } 1-\epsilon_t 
			\\
			\textrm{random } (w ,a ) \in {\mathcal X}_t(r_t,s_t)  
			 \textrm{w.p. }  \epsilon_t
			\end{array}
			\right. 
		  \end{eqnarray}
		\newline 
		{Update state $s_{t+1} = \hat a_t^{\ast}$} 
		\newline 
			{Request and cost parameters, $ \boldsymbol \theta_{t+1}$, are revealed} 
			\newline
			{Update $Q$ factor by \eqref{Q_update}}
\end{algorithm}

\section{Limited storage and back-haul transmission rate via dynamic pricing}\
\label{Sec_limited_storage_and_coms}

So far, we have considered that the prices $\{\rho_t^f,\lambda_t^f\}$ are provided by the system, and we have not assumed any explicit limits (bounds) neither on the capacity of the local storage nor on the back-haul transmission link between the SB and the cloud. 	
In this section, we discuss such limitations, and describe how by leveraging dual decomposition techniques, one can redefine the prices $\{\rho_t^f,\lambda_t^f\}$ to account for capacity constraints. 

\subsection{Limiting the instantaneous storage rate}

In this subsection, practical limitations on the cache storage capacity are explored. Suppose that the SB is equipped with a \textit{single} memory device that can store $M$ files. Clearly, the cache decisions should then satisfy the following constraint per time slot 	
\begin{equation}\nonumber
{\textrm {C$4$:}}\quad \sum \limits_{f = 1}^{F} a_t^f \sigma^f \le M, \quad t = 1,2, \ldots 
\label{eq20}
\end{equation}	
In order to respect such hard capacity limits, the original optimization problem in (P$1$) can be simply augmented with C$4$, giving rise to a new optimization problem which we will refer to as (P$4$).  
%\begin{align}
%{\textrm {(P$4$)} }  \min  \limits_{ \{(w^f_{k},a^f_{k}) \}_{f,k\geq t}}    &{\cal {\bar C}}_t:=\sum \limits_{k=t}^{\infty} \sum  \limits_{f=1}^{F} \gamma ^{k-t} {\mathbb E}  \left[c^f_k \left(a^f_k,w^f_k;\rho^f_k,\lambda^f_k\right)\right]  \nonumber\\
%\mathrm{s.t.}\;\;\;\; &(w^f_{k},a^f_{k}) \in \mathcal{X}( r^f_{k}, s^f_{k}),\;\;\;\; \forall f,\,\,k\geq t \nonumber \\ \nonumber
%\textrm{C$4$:} \quad & \sum \limits_{f = 1}^{F} a_k^f \sigma^f \le M, \quad \forall k = t, t+1 \ldots 
%\end{align}
Solving (P$4$) is more challenging than (P$1$), since the constraints in C$4$ must be enforced at each time instant, which subsequently couples the optimization across files. In order to deal with this, one can dualize C$4$ by augmenting the cost with the primal-dual term $\mu_t(\sum_{f=1}^F\sigma_fa_t^f-M)$, where $\mu_t$ denotes the Lagrange multiplier associated with the capacity constraint C$4$. The resultant problem is separable across files, but requires finding $\mu_t^*$, the optimal value of the Lagrange multiplier, at each and every time instant. 

If the solution to the original unconstrained problem (P$1$) does satisfy C$4$, then  $\mu_t^* = 0 $ due to complementary slackness. On the other hand, if the storage limit is violated, then the constraint is active, the Lagrange multiplier satisfies $\mu_t^*>0$, and its exact value must be found using an iterative algorithm. Once the value of the multiplier is known, the optimal actions associated with (P$4$) can be found using the expressions for the optimal solution to (P$1$) provided that the original storage price $\rho_t^f$ is replaced with the new storage price $\rho_{t,aug}^f=\rho_t^f+\mu_t^*\sigma_f$ {[cf. \eqref{eq_generic_form_rho_intro}]}. The reason for this will be explained in detail in the following subsection, after introducing the ensemble counterpart of C$4$.  

\subsection{Limiting the long-term storage rate}\label{subsec_long-term_storage_rate}
Consider now the following constraint [cf. C$4$] 
\begin{equation}
\textrm{C$5$:} \quad  \sum \limits_{k=t}^{\infty} \gamma^{k-t} {\mathbb {E}}\left[  \sum \limits_{f = 1}^{F} a_k^f \sigma^f  \right]  \le \sum \limits_{k=t}^{\infty} \gamma^{k-t} M'
\label{relaxedconstraint}
\end{equation}
where the expectation is taken w.r.t.  all state variables. 
%$r_k^f, \rho_k^f,$ and $ \lambda_k^f, \forall f = 1, \ldots, F$. 
%Note that caching decision per slot, $a_k^f$, is indeed influenced by realizations of $\left\{\rho_k^f, \lambda_k^f,r_k^f\right\}$ (c.f. (P$1$) or (P$2$)). 
By setting $M'=M$, one can view C$5$ as a relaxed version of C$4$. That is, while C$4$ enforces the limit to be respected at every time instant, C$5$ only requires it to be respected \textit{on average}. From a computational perspective, dealing with C$5$ is easier than  its instantaneous counterpart, since in the former only one constraint is enforced and, hence, only one Lagrange multiplier, denoted by $\mu$, must be found. This comes at the  price that guaranteeing C$5$ with $M'=M$ does not imply that C$4$ will always be satisfied. Alternatively, enforcing C$5$ with $M'<M$, will increase the probability of satisfying C$4$, since the solution will guarantee that ``on average'' there exists free space on the cache memory. A more formal discussion on this issue will be provided in the remark closing the subsection.

To describe in detail how accounting for C$5$ changes the optimal schemes, let (P$5$) be the problem obtained after augmenting (P$1$) with C$5$. Suppose now that to solve (P$5$) we dualize the single constraint in C$5$. Rearranging terms, the augmented objective associated with (P$5$) is given by
\begin{align}\label{eq_aumented_objective_storage_multiplier}
\sum \limits_{k=t}^{\infty} \sum  \limits_{f=1}^{F} \gamma ^{k-t} {\mathbb E}  
\left[c^f_k \left(a^f_k,w^f_k;\rho^f_k,\lambda^f_k\right)+ \mu a_k^f \sigma^f \right]  
- \sum \limits_{k = t}^{\infty} \gamma^{k-t}  M'.
\end{align}
Equation \eqref{eq_aumented_objective_storage_multiplier} demonstrates that after dualization and provided that the multiplier $\mu$ is known, decisions can be optimized separately across files. To be more precise, note that the term $\sum_{k = t}^{\infty} \gamma^{k-t}  M'$ in the objective is constant, so that it can be ignored, and define the modified instantaneous cost as
\begin{align}
\nonumber 
{\check{c}_k^f} := & \; c^f_k \left(a^f_k,w^f_k;\rho^f_k,\lambda^f_k\right)  + \mu \sigma^f a_k^f \\
= & \; \left(\rho_k^f + \mu \sigma^f \right) a^f_k+\lambda_k^f w^f_k.\label{eq_instantaneous_cost_augmented_storage_mult}
\end{align}
%
%\begin{align}
%{\check{c}_k^f} := c^f_k \left(a^f_k,w^f_k;\rho^f_k,\lambda^f_k\right)  + \mu a_k^f = \left(\rho_t^f + \mu \sigma^f \right) a^f_t+\lambda_t^f w^f_t,
%\end{align}
%
The last equation not only reflects that the dualization indeed facilitates separate per-file optimization, but it also reveals that term $\mu \sigma^f$ can be interpreted as an additional storage cost associated with the long-term caching constraint. More importantly, by defining the modified (augmented) prices ${\rho^f_{t, \textrm {aug}}} := \rho_t^f + \mu \sigma^f$ for all $t$ and $f$, the optimization of  \eqref{eq_instantaneous_cost_augmented_storage_mult} can be carried out with the schemes presented in the previous sections, provided that $\rho_t^f$ is replaced with ${\rho^f_{t, \textrm {aug}}}$. 

Note however that in order to run the optimal allocation algorithm, the value of $\mu$ needs to be known. Since the dual problem is always convex, one option is to use an iterative dual subgradient method,  which computes the satisfaction/violation of the constraint C$5$ per iteration~\cite{palomar2006tutorial}, \cite[p.223]{boydconvex}. Clearly, this requires knowledge of the state distribution, since the constraint involves an expectation. When such knowledge is not available, or when the computational complexity to carry out the expectations cannot be afforded, stochastic schemes are worth considering. For the particular case of estimating Lagrange multipliers associated with long-term constraints, a simple but powerful alternative is to resort to \textit{stochastic dual} subgradient schemes \cite{palomar2006tutorial}, \cite{boydconvex},  which for the problem at hand, estimate the value of the multiplier $\mu$ at every time instant $t$ using the update rule
\begin{align}
\hat{\mu}_{t+1}  =& \left[ \hat{\mu}_t + \zeta  \left( \sum \limits_{f = 1}^{F} \hat{a}^{f\ast}_t \sigma^f - M' \right) \right]^{+}.
\label{dual_cache}
\end{align}
In the last expression, $\zeta>0$ is a (small) positive constant, the update multiplied by $\zeta$ corresponds to the violation of the constraint after removing the expectation, the notation $[\cdot]^+$ stands for the $\max\{0,\cdot\}$, and $\hat{a}^{f\ast}_t$ denotes the optimal caching actions obtained with the policies described in Section \ref{Sec:DP_Formulation} provided that $\rho_t^f$ is replaced by $\hat{\rho}^f_{t, \textrm {aug}} = \rho_t^f + \hat{\mu}_t \sigma^f$. 

We next introduce another long-term constraint that can be considered to limit the storage rate. This constraint is useful not only because it gives rise to alternative novel caching-fetching schemes, but also because it will allow us to establish connections with well-known algorithms in the area of congestion control and queue management. To start, define the variables $\alpha_{in,t}^f:=[a_t^f-s_t^f]^+$ and $\alpha_{out,t}^f:=[s_t^f-a_t^f]^+$ for all $f$ and $t$. Clearly, if $\alpha_{in,t}^f=1$, then content $f$ that was not in the local cache at time $t-1$, has been stored at time $t$; and as a result, less storage space is available. On the other hand, if $\alpha_{out,t}^f=1$, then content $f$ was removed from the cache at time $t$, thus freeing up new storage space. With this notation at hand, we can consider the long term constraint
\begin{equation}
\textrm{C$6$:} \,  \sum \limits_{k=t}^{\infty} \gamma^{k-t} {\mathbb {E}}\left[  \sum \limits_{f = 1}^{F} \alpha_{in,k}^f \sigma^f  \right]  \le \sum \limits_{k=t}^{\infty} \gamma^{k-t} {\mathbb {E}}\left[  \sum \limits_{f = 1}^{F} \alpha_{out,k}^f \sigma^f  \right],
\label{relaxedconstraint}
\end{equation}
which basically ensures the long-term stability of the local-storage. That is, the amount of data stored in the local memory is no larger than that taken out from the memory, guaranteeing that in the long term stored data does not grow unbounded. 

To deal with C$6$ we can follow an approach similar to that of C$5$, under which we first dualize C$6$ and then use a stochastic dual method to estimate the associated dual variable. With a slight abuse of notation, supposing that the Lagrange multiplier associated with stability is by also denoted $\mu$, the counterpart of \eqref{dual_cache} for the constraint C$6$ is
\begin{align}
\hat{\mu}_{t+1}  = \left[ \hat{\mu}_t + \zeta \sum \limits_{f = 1}^{F}   [\hat{a}^{f\ast}_t - s^f_t]^+  - [s^f_t - \hat{a}^{f\ast}_t]^+ \right]^{+}.
\label{dualupdate_long_term_storage_as_queue}
\end{align}
Note that the update term in the last iteration follows after removing the expectations in C$6$ and replacing $\alpha_{in,t}^f$, and $\alpha_{out,t}^f$ with their corresponding definitions. The modifications that the expressions for the optimal policies require to account for this constraint are a bit more intricate. If $s_t^f=0$, the problem structure is similar to that of the previous constraints, and we just need to replace $\rho_t^f$ with $\hat{\rho}^f_{t, \textrm {aug}} = \rho_t^f + \hat{\mu}_t \sigma^f$. However,  if $s_t^f=1$, it turns out that: i) deciding ${\hat a}_t^{ f \ast}=1$ does not require modifying the caching price, but ii) deciding ${\hat a}_t^{f \ast}=0$ requires considering the \textit{negative} caching price $-\hat{\mu}_t \sigma^f$. In other words, while our formulation in Section \ref{Sec:DP_Formulation} only considers incurring a cost when $a_t^f=1$ (and assumes that the instantaneous cost is zero for $a_t^f=0$), to fully account for C$6$, we would need to modify our original formulation so that costs can be associated with the decision $a_t^f=0$ as well. This can be done either by considering a new cost term or, simply by replacing $\gamma \bar{V}^f(0)$ by $\gamma \bar{V}^f(0) -\hat{\mu}_t \sigma^f$ in \eqref{E:Bellman_input00}-\eqref{E:Bellman_input11}, which are Bellman's equations describing the optimal policies.

\vspace{.15cm} 
\noindent \textbf{Remark 3 (Role of the stochastic multipliers).} It is well-established that the Lagrange multipliers can be interpreted as the marginal price that the system must pay to \linebreak (over-)satisfy the constraint they are associated with \cite[p.241]{boydconvex}. When using stochastic methods for estimating the multipliers, further insights on the role of the multipliers can be obtained \cite{Neely2016ReourceAllocationTutorialBook,Marques_Queues12,Tianyi2017DistributedCloudNets}. Consider for example the update in \eqref{dual_cache}. The associated constraint C$5$ establishes that the long-term storage rate cannot exceed $M'$. To guarantee so, the stochastic scheme updates the estimated price in a way that, if the constraint for time $t$ is oversatisfied, the price goes down, while if the constraint is violated, the price goes up. Intuitively, if the price estimate $\hat{\mu}_t$ is far from its optimal value and the constraint is violated for several consecutive time instants, the price will keep increasing, and eventually will take a value sufficiently high so that storage decisions are penalized/avoided. How quickly the system reacts to this violation can be controlled via the constant $\zeta$. Interestingly, by tuning the values of $M'$ and $\zeta$, and assuming some regularity properties on the distribution of the state variables, conditions under which deterministic short-term limits as those in C$4$ are satisfied can be rigorously derived; see, e.g., \cite{Tianyi2017DistributedCloudNets} for a related problem in the context of distributed cloud networks. A similar analysis can be carried out for the update in \eqref{dualupdate_long_term_storage_as_queue} and its associated constraint C$6$. Every time the instantaneous version of the constraint is violated because the amount of data stored in the memory exceeds the amount exiting the memory, the corresponding price $\hat{\mu}_t$ increases, thus rendering future storage decisions more costly. In fact, if we initialize the multiplier at $\hat{\mu}_t=0$ and set $\zeta=1$, then the corresponding price is the total amount of information stored at time $t$ in the local memory. In other words, the update in \eqref{dualupdate_long_term_storage_as_queue} exemplifies how the dynamic prices considered in this paper can be used to account for the actual state of the caching storage. Clearly, additional mappings from the instantaneous storage level to the instantaneous storage price can be considered. The connections between stochastic Lagrange multipliers and storing devices have been thoroughly explored in the context of demand response, queuing management and congestion control. We refer the interested readers to, e.g., \cite{Neely2016ReourceAllocationTutorialBook,Marques_Queues12}.    

\subsection{Limits on the  back-haul transmission rate}\label{subsec_long-term_comm_rate}
The previous two subsections dealt with limited caching storage, and how some of those limitations could be accounted for by modifying the caching price $\rho_t^f$. This section addresses limitations on the back-haul transmission rate between the SB and the cloud as well as their impact on the fetching price $\lambda_t^f$.

While our focus has been on optimizing the decisions at the SB, contemporary networks must be designed following a holistic (cross-layer) approach that accounts for the impact of local decisions on the rest of the network. Decomposition techniques (including those presented in this paper) are essential to that end \cite{palomar2006tutorial}. For the system at hand, suppose that $\mathbf{x}_{CD}$ includes all variables at the cloud network, $\bar{C}_{CD}(\mathbf{x}_{CD})$ denotes the associated cost, and the feasible set $\mathcal{X}_{CD}$ accounts for the constraints that cloud variables $\mathbf{x}_{CD}$ must satisfy. Similarly, let $\mathbf{x}_{SB}$, $\bar{C}_{SB}(\mathbf{x}_{SB})$, and $\mathcal{X}_{SB}$ denote the corresponding counterparts for the SB optimization analyzed in this paper. Clearly, the fetching actions $w_t^f$ are included in $\mathbf{x}_{SB}$, while the variable $b_t$ representing back-haul transmission rate (capacity) of the connecting link between the cloud and the SB, is included in $\mathbf{x}_{CD}$. This transmission rate will depend on the resources that the cloud chooses to allocate to that particular link, and will control the communication rate (and hence the cost of fetching requests) between the SB and the cloud. As in the previous section, one could consider two types of capacity constraints
\begin{subequations}
	\begin{flalign}\label{eq_constraint_com_limits_short_term}
	{\rm C}7a: \quad	&\sum_{f=1}^{F} w^f_t \sigma^f \le b_t, \quad  t = 1, \ldots, \\
	{\rm C}7b: \quad \label{eq_constraint_com_limits_long_term}
	&\sum_{k=t}^\infty \gamma^{k-t} \sum_{f=1}^{F} {\mathbb E} [w^f_t \sigma^f] \le \sum_{k=t}^\infty \gamma^{k-t} {\mathbb E} [b_k],
	\end{flalign}
\end{subequations}
depending on whether the limit is imposed in the short term or in the long term.	

With these notational conventions, one could then consider the \textit{joint} resource allocation problem
\begin{align}
\min  \limits_{ \mathbf{x}_{CD}, \mathbf{x}_{SB}}    & \bar{C}_{CD}(\mathbf{x}_{CD}) + \bar{C}_{SB}(\mathbf{x}_{SB})  \nonumber\\
\mathrm{s.t.}\;\;\;\; &\mathbf{x}_{CD}\in \mathcal{X}_{CD},\;\;\mathbf{x}_{SB}\in \mathcal{X}_{SB},\;\;({\rm C}7)
\end{align}
where the constraint C$7$ -- either the instantaneous one in C$7a$ or the lon-term version in C$7b$ -- couples both optimizations. It is then clear that if one dualizes  C$7$, and the value of the Lagrange multiplier associated with C$7$ is known, then two separate optimizations can be run: one focusing on the cloud network and the other one on the SB. For this second optimization, consider for simplicity that the average constraint in \eqref{eq_constraint_com_limits_long_term} is selected and let $\nu$ denote the Lagrange multiplier associated with such a constraint. The optimization corresponding to the SB is then      
\begin{align}
\min  \limits_{ \mathbf{x}_{SB}}    \; \bar{C}_{SB}(\mathbf{x}_{SB}) + \sum_{k=t}^\infty \gamma^{k-t} \sum_{f=1}^{F} {\mathbb E} [w^f_t \nu \sigma^f]\;\;\;\mathrm{s.t.}\;\; \mathbf{x}_{SB}\in \mathcal{X}_{SB}.
\end{align}

Clearly, solving this problem is equivalent to solving the original problem in Section \ref{Sec:DP_Formulation}, provided that the original cost is augmented with the primal-dual term associated with the coupling constraint. To address the modified optimization, we will follow steps similar to those in Section \ref{subsec_long-term_storage_rate}, defining first a stochastic estimate of the Lagrange multiplier as 
\begin{align}
\label{dual_backhaul}
\hat{\nu}_{t+1}  = \left[ \hat{\nu}_t + \zeta  \left( \sum \limits_{f = 1}^{F} \hat{w}^{f\ast}_t \sigma^f - b_t \right) \right]^{+},
\end{align}
and then obtaining the optimal caching-fetching decisions running the schemes in Section \ref{Sec:DP_Formulation} after replacing the original fetching cost $\lambda_t^f$ with the augmented one $\lambda^f_{t,{\textrm {aug}}}=\lambda_t^f+\hat{\nu}_t\sigma_f$.

For simplicity, in this section we will limit our discussion to the case where $\hat{\nu}_t$ corresponds to the value of a Lagrange multiplier corresponding to a communication constraint. However, from a more general point of view, $\hat{\nu}_t$ represents the marginal price that the cloud network has to pay to transmit the information requested by the SB. In that sense, there exists a broad range of options to set the value of $\hat{\nu}_t$, including the congestion level at the cloud network (which is also represented by a Lagrange multiplier), or the rate (power) cost associated with the back-haul link. While a detailed discussion on those options is of interest, it goes beyond the scope of the present work. 

\begin{algorithm}[t]
	\caption{Modified $Q$-learning for online caching}
	\label{MQ_learning}
	{{\bf Initialize}  $0 < \gamma, \; {\beta_t}< 1, \; \hat \mu_0, \zeta, \epsilon_t, \; M$} 
	\newline
	{{\bf Output} $\hat {\bar Q}_{r^f,s^f}^{w^f,a^f} \left(t+1\right)$ } 
	\newline
	
	%	\SetKwInOut{Input}{Input}
	%	\SetKwInOut{Output}{Output}
	%	\Input{$0 < \gamma, \; {\beta_t}< 1, \; \hat \mu_0, \zeta, \epsilon_t, \; M$}  
	%	\Output{ $\hat {\bar Q}_{r^f,s^f}^{w^f,a^f} \left(t+1\right)$ }
	{Set $ \hat {\bar Q}_{r^f,s^f}^{w^f,a^f} (1) = 0$ for all factors  
		\newline Set $s_0^f = 0$ and variables ${\boldsymbol \theta}^f_{0}=\{r^f_0, \rho^f_0, \lambda^f_0\}$ are revealed}  
	\newline 	
	{For \quad $t = 0,1 \ldots$} 
	\textrm{For the current state $(r^f_t,s^f_t)$, choose $ ({\breve w}_t^{f \ast},{\breve a}_t^{f \ast})$ }
	{ \[ ({\breve w}_t^{f \ast},{\breve a}_t^{f \ast}) \! = \! \left\{
		\begin{array}{ll}
		{\textrm {Solve \eqref{margQ_solver}}} &  \textrm{w.p. } 1-\epsilon_t \\
		\textrm{random }  (w,a) \! \in \! { \mathcal X}_t^f(r^f_t,s^f_t)  & \textrm{w.p. }  \epsilon_t
		\end{array}
		\right. \]}
	\newline
	{Update dual variable \[\hat \mu_{t+1} = \left[\hat \mu_{t} + \zeta \left( \sum \limits_{f = 1}^{F} {{ \breve{a}}}_t^{f \ast} \sigma^f - M \right) \right]^{+}\]}
	\newline
	{Incur cost \quad $\check{c}^f_t := c^f_t ({ {\breve a}_t^{f \ast}},{\breve w}_t^{f \ast};\rho^f_t,\lambda^f_t) + \hat \mu_t {\breve  a}^{f\ast}_t \sigma^f$}
	\newline
	{\textrm Apply  $\Pi_{\rm C4}(\cdot)$ to guarantee C$4$ {(if required)} {\hspace{+2.5 cm} \[ \Pi_{\rm C4} \left[ \left\{  ({\breve w}_t^{f \ast},{\breve a}_t^{f \ast}) \right\}_f \right] \rightarrow \left\{{w}^{f\ast}_t,{ {a}^{f\ast}_t}\right\}_f \] }}
	\newline
	{Update state  $s^f_{t+1} = { {a}^{f\ast}_t}$}
	\newline
	{Request and cost parameters,  $ \boldsymbol \theta^f_{t+1}$, are revealed} 
	\newline
	{Update all ${\hat {\bar Q}}$  factors as}
	\newline
	\[\hat {\bar Q}_{r_{t}^f,s^f_{t}}^{w^{f \ast}_{t},a^{f \ast}_{t}}(t+1) = (1-{\beta_t})  \; \hat {\bar Q}_{r_{t}^f,s^f_{t}}^{w^{f \ast}_{t},a^{f \ast}_{t}}(t) \; \] \[ \hspace{2 cm} + \; {\beta_t}  \left[ \check{c}_{t}^f + \nonumber \gamma  \underset{{ (w^f,a^f)} \in { \mathcal X}^f_{t+1} } {\min}  \hat {\bar Q}^{{w^f,a^f}}_{r^f_{t+1},s^f_{t+1}}(t) \right] \]   
\end{algorithm}
\subsection{Modified online solver based on $Q$-learning}
\label{Qlearning_solver}
We close this section by providing an online reinforcement-learning algorithm that modifies the one introduced in Section \ref{Sec:DP_Formulation} to account for the multipliers introduced in Section \ref{Sec_limited_storage_and_coms}.  

By defining per file cost $\hat c_k^f$ as
\begin{align}
\nonumber \hat c_k^f \left(w_k^f,a_k^f;\rho_k^f,\lambda_k^f,{ \hat \mu_k},{ \hat \nu_k}\right) := &  \\ \left(\rho_k^f + \hat \mu_k \sigma^f \right) & a^f_k  + \left( \lambda_k^f + \hat \nu_k \sigma^f\right) w^f_k
\end{align}
the problem of caching under limited cache capacity and back-haul link reduces to per file optimization as follows 
\begin{align}
\nonumber ({\textrm P}8) \min \limits_{ \{(w^f_{k},a^f_{k}) \}_{k\geq t}}  & \; \sum \limits_{k=t}^{\infty} \gamma ^{k-t} {\mathbb E}  \left[ {\hat{c}_k^f}\left(a^f_k,w^f_k;\rho^f_k,\lambda^f_k,{ \hat \mu_k},{ \hat \nu_k}\right) \right]   \nonumber \\ \nonumber \mathrm{s.t.}\;\;\;\; &(w^f_{k},a^f_{k}) \in \mathcal{X}( r^f_{k}, { a^f_{k-1}}),\;\;\;\; \forall f,\,\,k\geq t
\end{align}
where the updated dual variables $\hat \mu_k$ and $\hat \nu_k$ are obtained respectively by iteration \eqref{dual_cache} and \eqref{dual_backhaul}. If we plug $\hat{c}_k^f$ instead of $c_k^f$ into the marginalized $Q$-function in \eqref{marg_q}, then the solution for (P8) in current iteration $k$ for a given file $f$ can readily be found by solving 
\begin{equation} 
\underset{(w,a) \in \mathcal{X}( r_{t}, {  a_{t-1}})}{\arg \min} \; {\bar Q}_{r_t,s_t}^{w,a} + w (\lambda_t +\hat \nu_t \sigma^f) + a (\rho_t+\hat \mu_t \sigma^f).
\label{light_wieght_solver}
\end{equation}
Thus, it suffices to form a marginalized $Q$-function for each file and solve \eqref{light_wieght_solver}, which can be easily accomplished through exhaustive search over $8$ possible cache-fetch decisions~$ (w,a) \in \mathcal{X} (r_{t}, {  a_{t-1}})$. 

To simplify notation and exposition, we focus on the \textit{limited caching capacity} constraint, and suppose that the back-haul is capable of serving any requests, thus $\hat \nu_t = 0, \; \forall t $. Modifications to account also for  $\hat \nu_t \neq  0$ are straightforward. 

The modified $Q$-learning (MQ-learning) algorithm, tabulated in Algorithm~\ref{MQ_learning}, essentially learns to make optimal fetch-cache decisions while accounting for the limited caching capacity constraint in C$4$ and/or C$5$. In particular, to provide a computationally efficient solver the stochastic updates corresponding to C$5$ are used. Subsequently, if C$4$ needs to be enforced, the obtained solution is projected into the feasible set through projection algorithm $\Pi_{\rm C4}(\cdot)$.  The projection $\Pi_{\rm C4}(.)$ takes the obtained solution $ \{{  \breve w ^{f\ast}_t, {\breve a}^{f\ast}_t } \}_{\forall f}$, the file sizes, as well as the marginalized $Q$-functions as input, and generates a feasible solution $\{{w}^{f\ast}_t,{a}^{f\ast}_t \}_{\forall f}$ satisfying C$4$ as follows: it sorts the files with $ {\breve a}^{f\ast}_t = 1$ in ascending $Q$-function order, and caches the files with the lowest $Q$-values until the cache capacity is reached. Overall, our modified algorithm performs a ``double'' learning: i) by using reinforcement schemes it learns the optimal policies that map states to actions, and ii) by using a stochastic dual approach it learns the mechanism that adapt the prices to the saturation and congestion conditions in the cache. Given the operating conditions and the design approach considered in the paper, the proposed algorithm has moderate complexity, and thanks to the reduced input dimensionality, it also converges in a moderate number of iterations.

\section{Numerical tests}
\label{Sec_results}
\begin{figure}
	\centering
	\includegraphics[width=0.48 \textwidth]{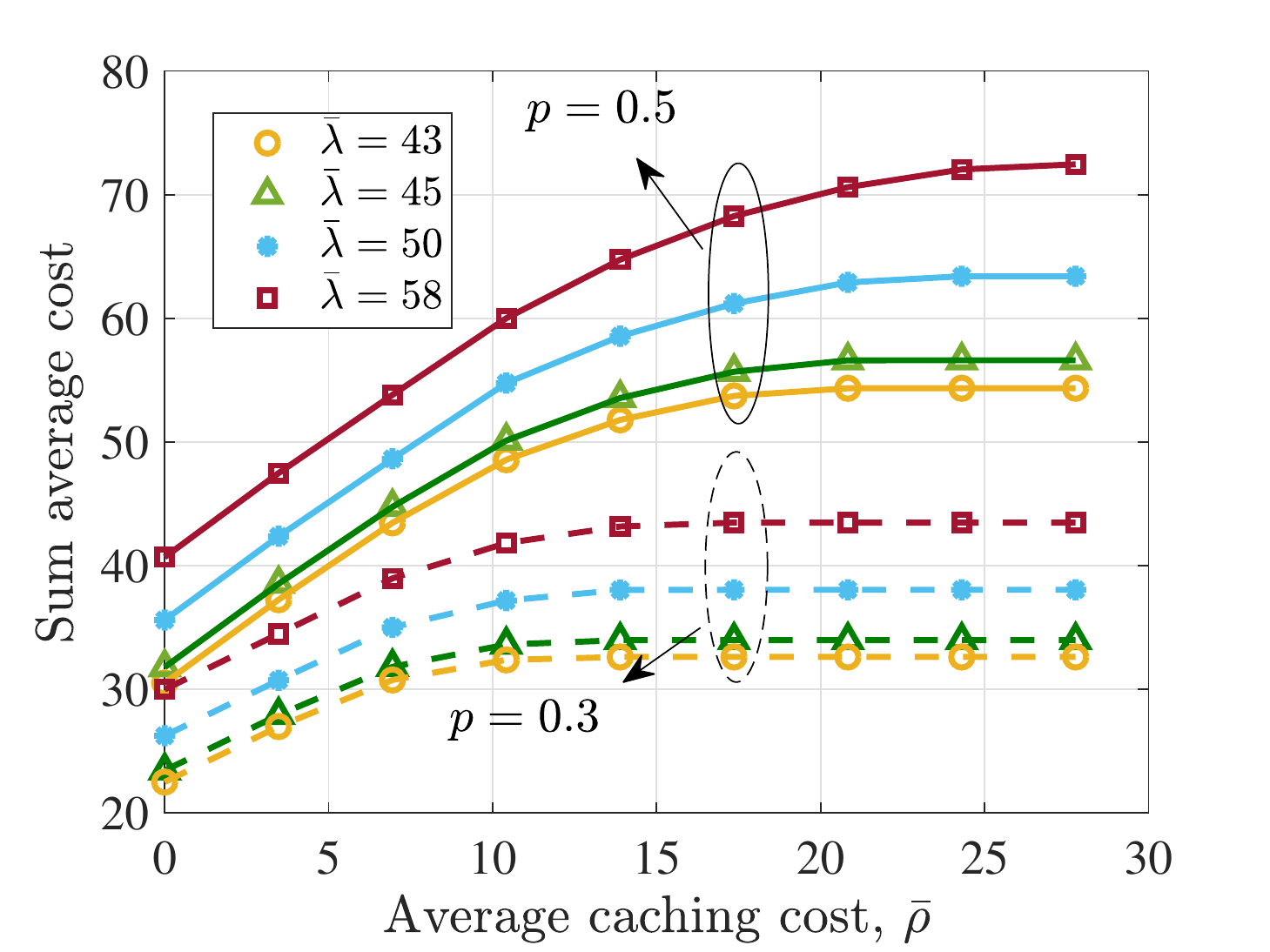}
	\caption{Average cost versus $\bar \rho$ for different values of $p, \bar \lambda$.}
	%\vspace{-0.05in}
	\label{fig:result1}
\end{figure}

\begin{figure}
	\centering
	\includegraphics[width=0.48 \textwidth]{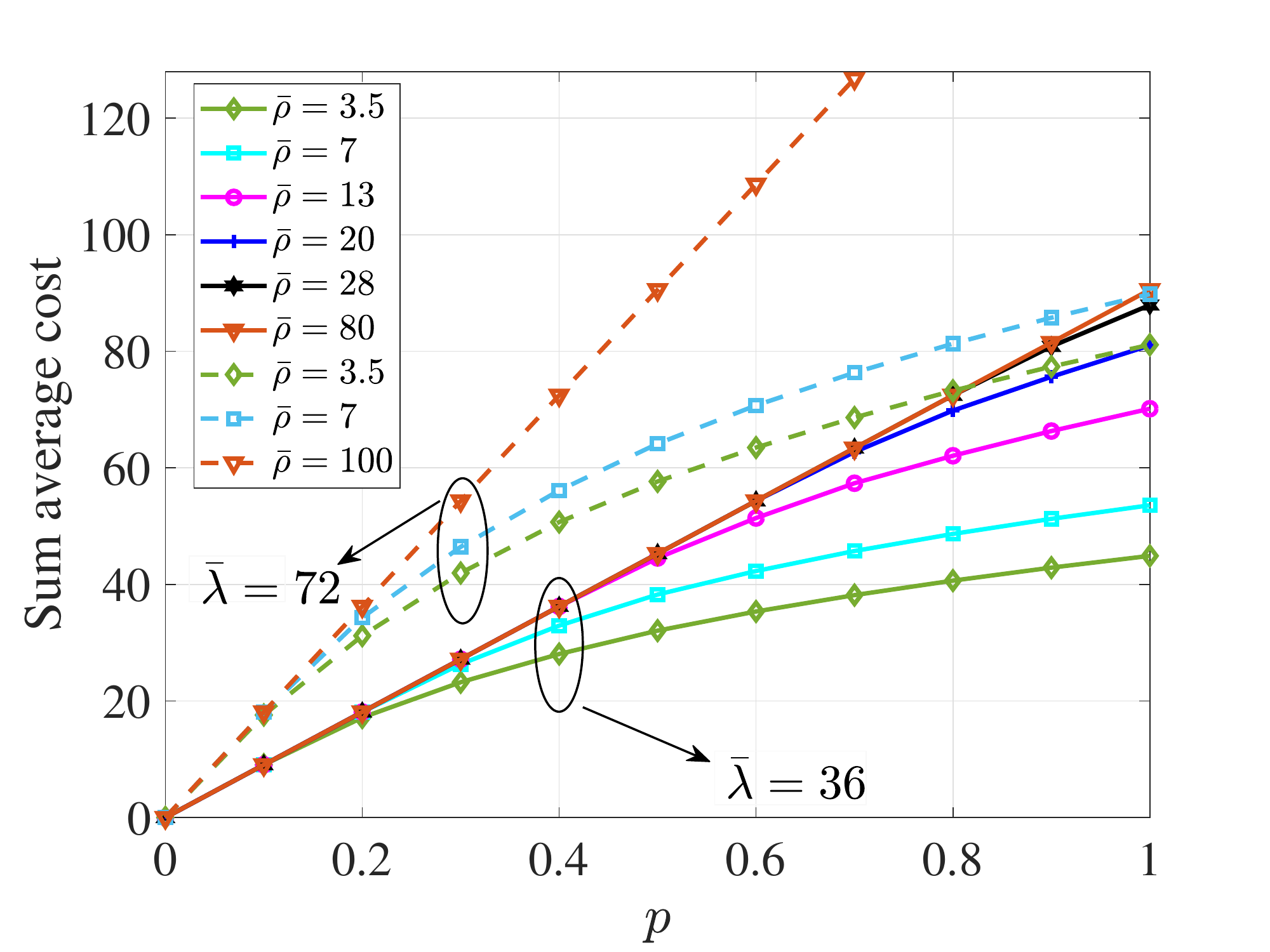}
	\caption{Average cost versus $p$ for different values of $\bar \lambda, \bar \rho$.}
	%\vspace{-0.05in}
	\label{fig:result2}
\end{figure}

\begin{figure}
	\centering
	\includegraphics[width=0.48 \textwidth]{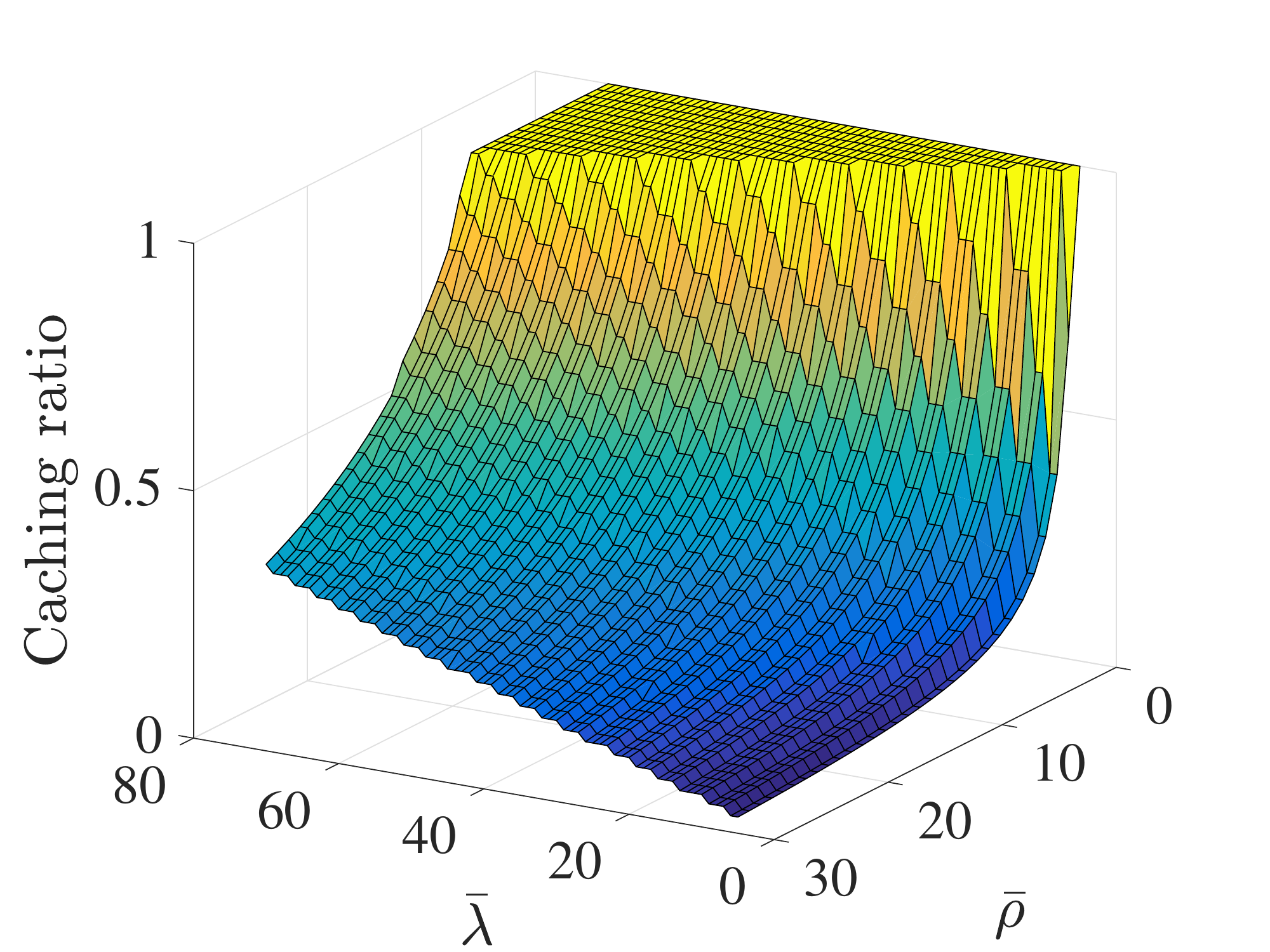}
	\caption{Caching ratio vs. $\bar \rho$ and $\bar \lambda$ for $p = 0.5$ and $s = r = 1$.}
	\label{fig:result3}
\end{figure}

\begin{figure}	
	\centering
	\includegraphics[width=0.45 \textwidth]{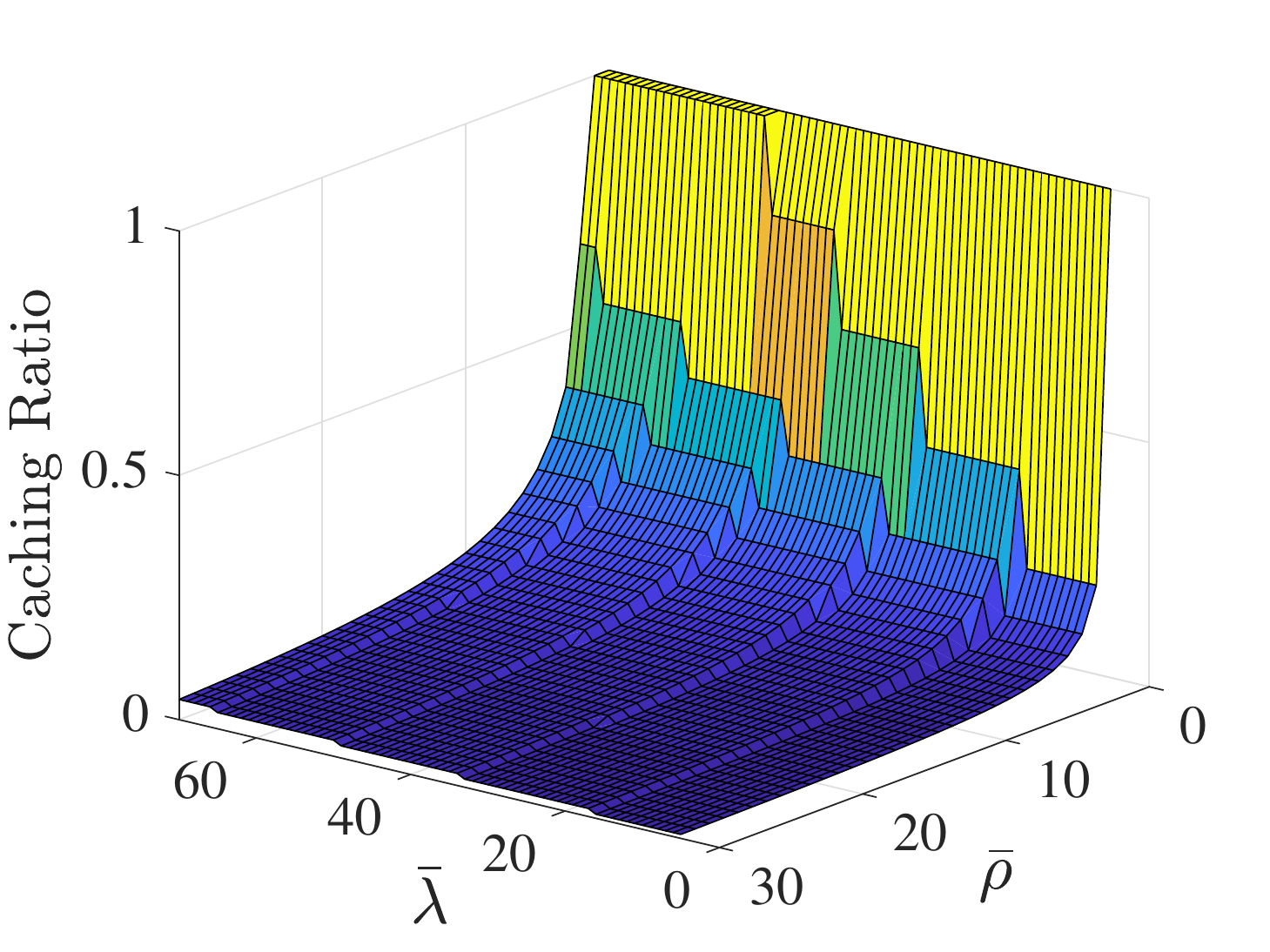}
	\caption{{Caching ratio vs. $\bar \rho$ and $\bar \lambda$ for $p = 0.05$ and $s = r = 1$.}}
	%\vspace{-0.05in}
	\label{result11}
\end{figure}

\begin{figure}
	\centering
	\includegraphics[width=0.48 \textwidth]{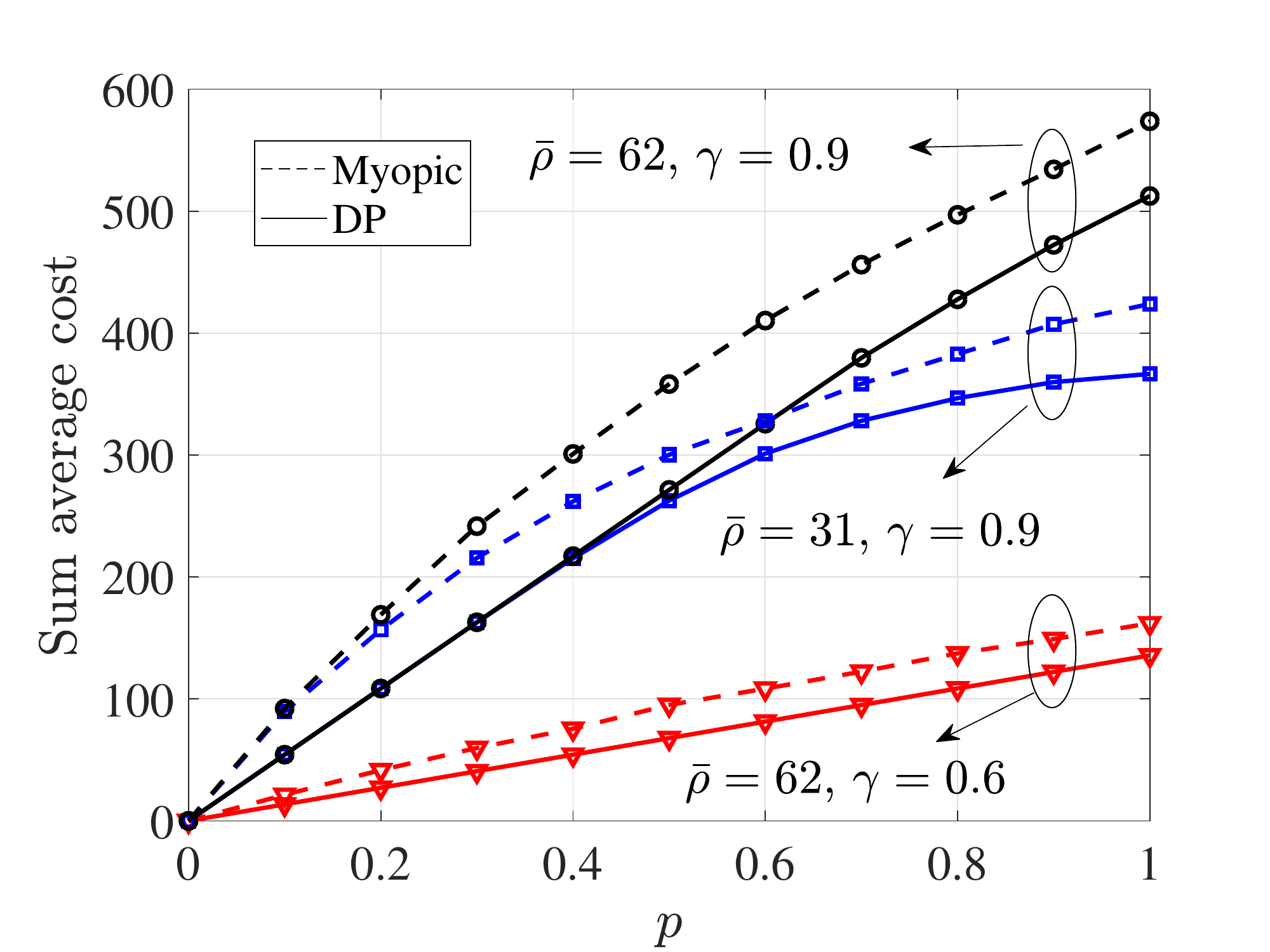}
	\caption{Performance of DP versus myopic caching for $\bar \lambda =53$. }
	%\vspace{-0.05in}
	\label{fig:result4}
\end{figure}

\begin{figure}
	\centering
	\includegraphics[width=0.5 \textwidth]{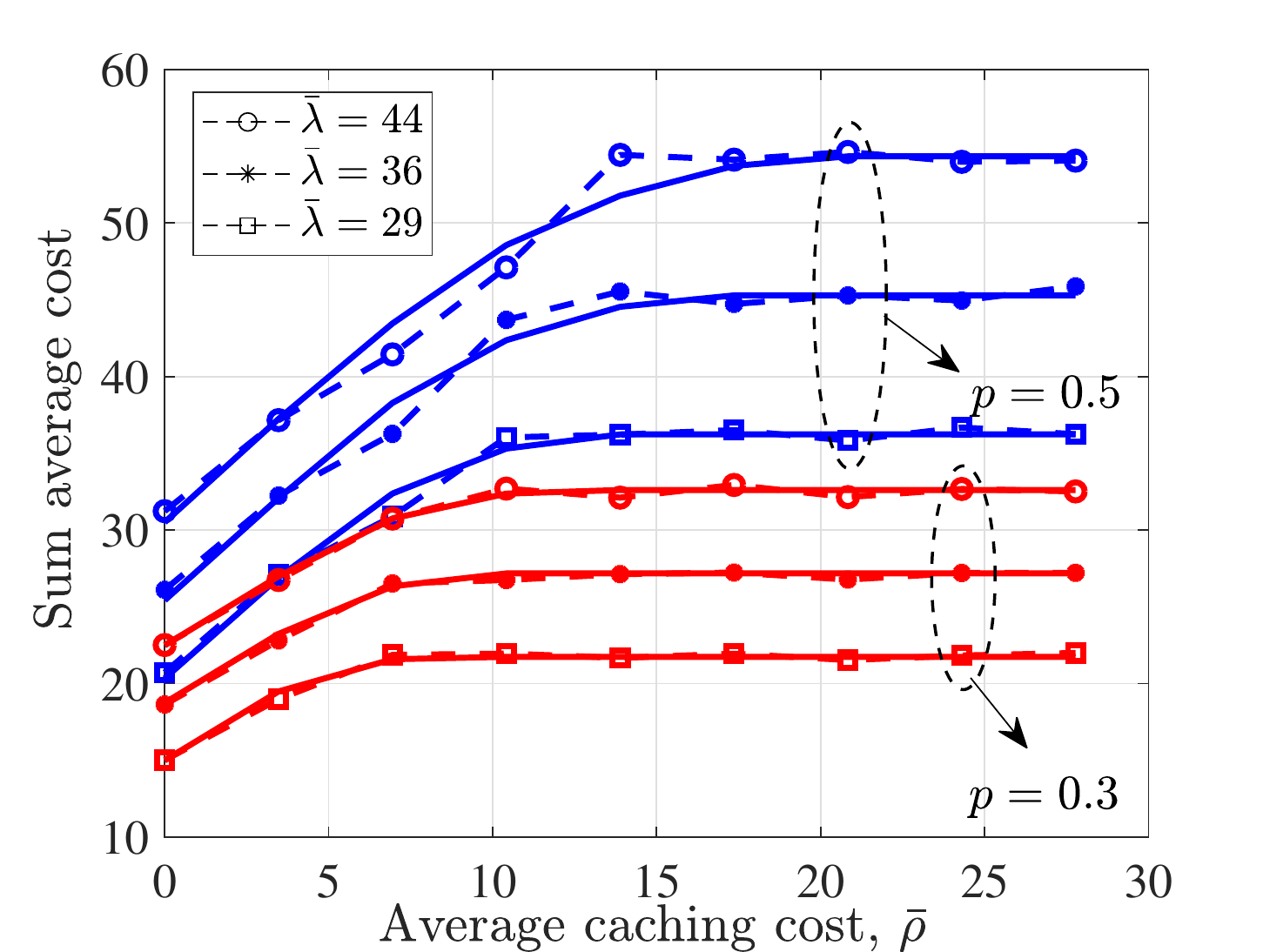}
	\caption{Average cost versus ${\bar \rho}$ for different values of $\bar \lambda, p$. Solid line is for value iteration while dashed lines are for $Q$-learning based solver.}
	%\vspace{-0.05in}
	\label{result6}
\end{figure}

\begin{figure}
	\centering
	\includegraphics[width=0.5\textwidth]{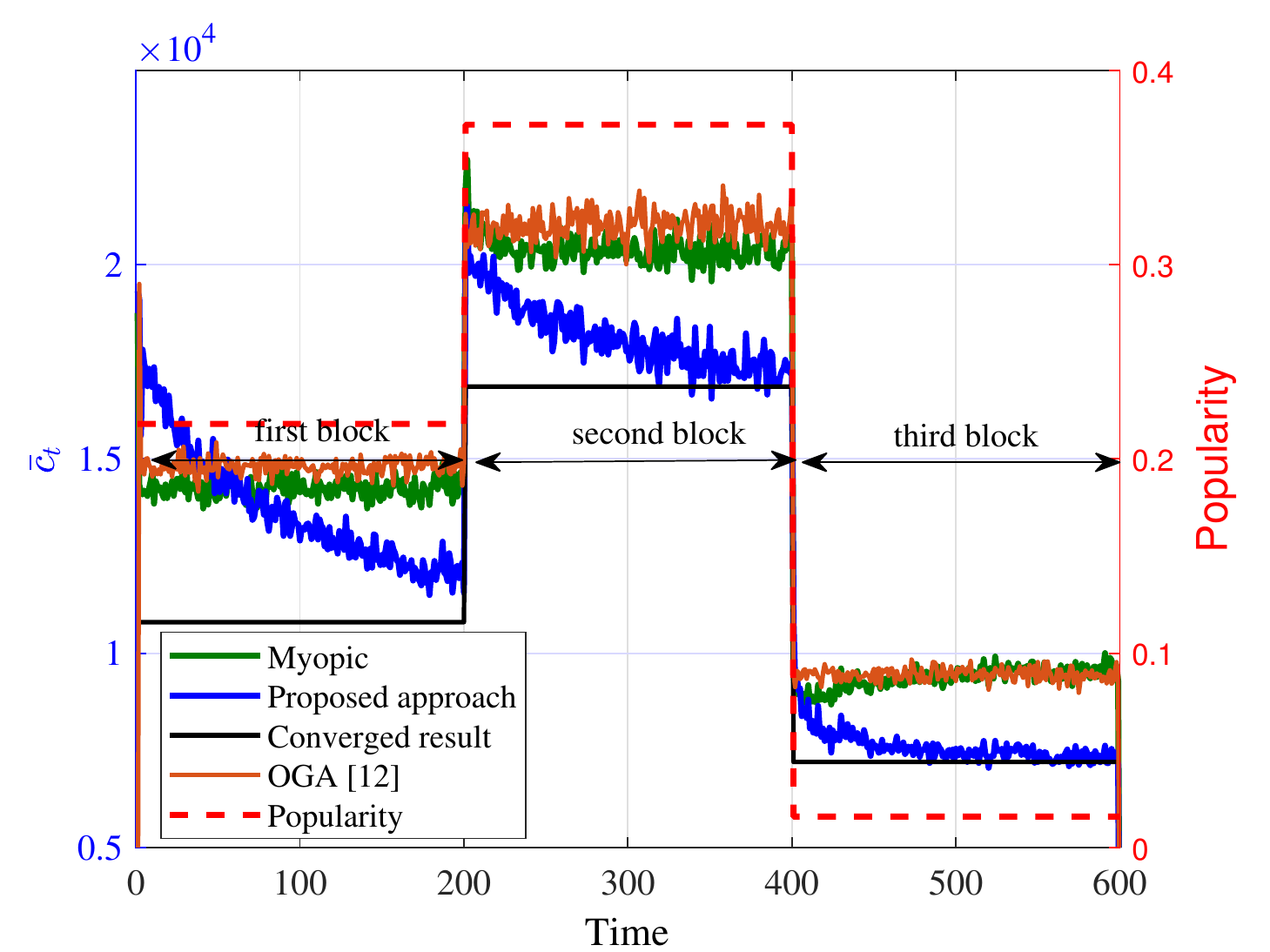}
	\caption{Averaged immediate cost over $1000$ realizations in a non-stationary setting, and a sample from popularities.}
	\label{result7}
\end{figure}

In this section, we numerically assess the  performance of the proposed approaches for learning optimal fetch-cache decisions. Two sets of numerical tests are provided. In the first set, summarized in Figs~\ref{fig:result1}-\ref{fig:result4}, the performance of the value iteration-based scheme in Alg. 1 is evaluated, and in the second set,  summarized in Figs.~\ref{result6}-\ref{result7},  the performance of the $Q$-learning solver is investigated. In both sets, the cache and fetch cost parameters are drawn with equal probability from a finite number of values, where the mean is $\bar{\rho}^f$ and  $\bar{\lambda}^f$, respectively. Furthermore, the request variable $r^f$ is modeled as a Bernoulli random variable with mean $p^f$, whose value indicates the popularity of file $f$. 

In the first set, it is assumed that $ p^f$ as well as the distribution of $\rho^f, \lambda^f$, are known a priori. Simulations are carried out for a content of unit size, and can be readily extended to files of different sizes. To help readability, we drop the superscript $f$ in this section.

Fig.~\ref{fig:result1} plots the sum average cost $\bar{\mathcal{C}}$ versus $\bar{\rho}$ for different values of $\bar{\lambda}$ and $p$. The fetching cost is set to $\bar{\lambda} \in \left\{43, 45, 50, 58\right\}$ for two different values of popularity $p  \in \{ 0.3,0.5\}$.  As depicted, higher values of $\bar{\rho},\bar{\lambda},p$ generally lead to a higher average cost. In particular, when $\bar{\rho}\ll \bar{\lambda}$, caching is considerably cheaper than fetching, thus setting $a_t=1$ is optimal for most $t$. As a consequence, the total cost linearly increases with $\bar{\rho}$ as most requests are met via cached contents rather than fetching. Interestingly, if $\bar{\rho}$ keeps increasing, the aggregate cost gradually saturates and does not grow anymore. The reason behind this  observation is the fact that, for very high values of $\bar{\rho}$, fetching becomes the optimal decision for meeting most file requests and, hence, the aggregate cost no longer depends on $\bar{\rho}$. While this behavior occurs for the two values of $p$, we observe that for the smallest one, the saturation is more abrupt and takes place at a lower  $\bar{\rho}$. The intuition in this case is that for lower popularity values, the file is requested less frequently, thus the caching cost aggregated over a (long) period of time often exceeds the ``reward'' obtained when (infrequent) requests are served by the local cache. As a consequence, fetching in the infrequent case of $r_t=1$ incurs less cost than the caching cost aggregated over time.

To corroborate these findings, Fig.~\ref{fig:result2} depicts the sum average cost versus $p$ for different values of  $\bar \rho$ and $\bar \lambda$. The results show that for large values of $\bar{\rho}$, fetching is the optimal action, resulting in a linear increase in the total cost as $p$ increases. In contrast, for small values of $\bar{\rho}$, caching is chosen more frequently, resulting in a sub-linear cost growth.
\hspace{0.1cm}
\hspace{0.1cm}
\vspace{0.1cm}
\vspace{0.1in}

To investigate the  caching-versus-fetching trade-off for a broader range of $\bar{\rho}$ and $\bar{\lambda}$, let us define the \textit{caching ratio} as the aggregated number of positive caching decisions (those for which $a_t=1$) divided by  the total number of decisions. Fig. \ref{fig:result3} plots this ratio for different values of $(\bar{\rho},\bar\lambda)$ and fixed $p=0.5$. As the plot demonstrates, when $\bar \rho$ is small and $\bar \lambda$ is large, files are cached almost all the time, with the caching ratio decreasing (non-symmetrically) as $\bar{\rho}$ increases and $\bar{\lambda}$ decreases. Similarly, the caching ratio is plotted by setting $p = 0.05$ in Fig.~\ref{result11}, in which fetching is mostly preferred over a wide range of storage costs due to the small value of $p$. Interestingly this is true despite high fetching costs as well, and can be intuitively explained as follows: due to low popularity, deciding to cache may result in idle storing of the file in cache, thus entailing an unnecessary aggregated caching cost before the stored file can be utilized to meet user request, rendering caching suboptimal. The comparison between Fig. \ref{fig:result3} and \ref{result11} clearly demonstrates the effect of different values of $p$ on the performance of the cache-fetch decisions, while the proposed approach automatically adjusts to the underlying  popularities.

Finally,   Fig.~\ref{fig:result4} compares the performance of the proposed DP-based strategy with that of a myopic one.  The myopic policy sets $a_t\!=\!1$ if $\lambda_t\!>\!\rho_t$ and the content is locally available (either because $w_t\!=\!1$ or because $s_t\!=\!1$), and sets $a_t\!=\!0$ otherwise.
The results indicate that the proposed strategy outperforms the myopic one for all values of $\bar{\rho},\bar{\lambda},p$ and $\gamma$.

In the second set of tests, the performance  of the online Q-learning solvers is investigated. As explained in Section \ref{Sec:DP_Formulation}, under the assumption that the underlying distributions are stationary, the performance of the Q-learning solver should converge to the optimal one found through the value iteration algorithm. Corroborating this statement, Fig.~\ref{result6} plots the sum average cost $\bar {\mathcal C}$ versus $\bar \rho$ of both the marginalized value iteration and the Q-learning solver, with $\bar \lambda \in \left\{29,36,44\right\}$ and $p \in \left\{0.3,0.5\right\}$. The solid lines are obtained when assuming a priori knowledge of the distributions and then running the marginalized value iteration algorithm; the results and analysis are similar to the ones reported for Fig.~\ref{fig:result1}. The dashed curves however, are found by assuming unknown distributions and running the 
Q-learning solver. Sum average cost is reported after first $1000$ iterations. As the plot suggests, despite the lack of a priori knowledge on the distributions, the Q-learning solver is able to  find the optimal decision making rule. As a result, it yields the same sum average cost as that of value-iteration under known distributions. 	

The last experiment investigates the impact of the instantaneous cache capacity constraint in C$4$ as well as non-stationary distributions for popularities and costs.  To this end, 1,000 different realizations (trajectories) of the random state processes are drawn, each of length $T=600$. For every realization, the cost $c_t$ [cf. \eqref{Sum_cost}] at each and every time instant is found, and the cost trajectory is averaged across the 1,000 realizations. Specifically, let $c_t^i$ denote the $i$th realization cost at time $t$, and define the averaged cost trajectory as ${\bar c}_t := \frac{1}{1000} \sum_{i=1}^{1000} c^i_t$. Fig.~\ref{result7} reports the average trajectory of ${\bar c}_t$ in a setup where the total number of files is set to $F = 500$, the file sizes are drawn uniformly at random from the interval $[1,100]$, and the total cache capacity is set to  $40\%$ of the aggregate file size. Adopted parameters for the MQ-learning solver are set to  ${\beta_t}= 0.3,$ and $ \epsilon = 0.01$. Three blocks of iterations are shown in the figure, where in each block a specific distribution of popularities and costs are considered. For instance, the dashed line shows the popularity of a specific file in one of the realizations, where in the fist block $p = 0.23$, in the second block $p = 0.37$, and in the third one  $p = 0.01$. The cost parameters have means $\bar \lambda = 44, \bar \rho = 2$,  $\bar \lambda = 40,\bar \rho = 5$, and $\bar \lambda = 38,\bar \rho = 2$ in the consecutive blocks, respectively.

As Fig. \ref{result7}, the proposed MQ-learning algorithm incurs large costs during the first few iterations. Then, it gradually adapts to the file popularities and cost distributions, and learns how to make optimal fetch-cache decisions, decreasing progressively the cost in each of the blocks. To better understand the behavior of the algorithm and assess its effectiveness, we compare it with that of Online Gradient Ascent (OGA)~\cite{Infocom_Paschos} as a representative state-of-the-art method among the class of online expert algorithms, the myopic policy and the stationary policy serving as the benchmark, respectively. In contrast to the OGA method, our decision variables are not continuous, but binary. Hence, caching decisions in OGA are projected into the binary feasible set for fair comparison. In general, since OGA and the myopic caching only use blue the current state and requests, their performance is inferior to that of our proposed method, where knowledge of the underlying  request and price distributions is carefully utilized. During the first iterations however, when the MQ-learning algorithm has not adapted to the distribution of pertinent parameters, OGA and the myopic policy perform better; on the other hand, as the learning proceeds, the MQ-learning starts to make more precise decisions and, remarkably, in a couple of hundreds of iterations it is able to perform very close to the optimal policy. 

Furthermore, to investigate the scalability of our proposed approach, Tables~\ref{tab:run_DP}, \ref{tab:run_OGA}, and \ref{tab:run_MYO_2} report the run-time (in seconds\footnote{We run these simulations in parallel with 4 pools of workers, utilizing a machine with Intel(R) Core(TM) i7-4770 CPU @ 3.4 GHz specifications.}) versus the number of files $F$ as well as the storage capacity $M$, set as a ratio of the total aggregated file sizes. Although the proposed approach has slightly higher run-time due to the utilized dual-decomposition technique and the solution of the arising integer DP, all methods scale gracefully (linearly) as the number of files increases from 1K to 10K.

\begin{table}[t!]
		\centering
		\begin{tabular}{|c|c|c|c|c|c|c|c|c|c|c|} 
			\hline
			{$\%M / F$} & $1K$ &$2K$& $3K$& $4K$&$6K$&$8K$&$10K$
			\\
			\hline
			$10$ \% & $ 148  $ & $ 240  $ & $ 337 $ &$ 444  $ & $ 662  $ & $ 870  $ & $ 1089  $ \\ \hline
			$20$ \% & $ 141  $ & $ 229  $ & $ 327 $ & $ 435  $ & $ 625  $ & $ 858  $ & $ 1052 $ \\ \hline
			$40$ \% & $139 $ & $232 $ & $326$ & $422 $ & $610 $ & $815 $ & $980 $  \\ \hline
			$60$ \% & $ 149  $ & $ 251  $ & $ 372 $ & $ 497 $ & $ 699 $ & $ 949 $ & $ 1086  $ \\ \hline
			
		\end{tabular}
		\caption{Run-time of the proposed caching.}
		\label{tab:run_DP}
\end{table}

\begin{table}[t!]
	\centering
		\begin{tabular}{|c|c|c|c|c|c|c|c|c|c|c|} 
			\hline
			{$\% M / F $} & $1K$ &$2K$&$3K$&$4K$&$6K$&$8K$&$10K$ 
			\\
			\hline
			$10$ \% & $ 70 $ &  $ 120 $ & $ 176$ & $ 241 $ & $ 411 $ & $ 622$ & $ 808$ \\ \hline
			$20$ \% & $ 73 $ & $ 123 $ & $ 183 $ & $ 250 $ & $ 389 $ & $ 569$ & $ 796 $ \\ \hline
			$40$ \% & $70$ & $122$ & $182$ & $272$ & $406$ & $585$ & $779$ \\ \hline
			$60$ \% & $ 92 $ & $ 170$ & $ 254 $ & $ 356$ & $ 551 $ & $ 736 $ & $ 908 $ \\ \hline
			
		\end{tabular}
		\bigskip
		\caption{Run-time of OGA caching.}
		\label{tab:run_OGA}
\end{table}

\begin{table}[t!]
	\centering
		\begin{tabular}{|c|c|c|c|c|c|c|c|c|c|c|} 
			\hline
			{$\% M / F $} & $1K$ &$2K$&$3K$&$4K$&$6K$&$8K$&$10K$ 
			\\
			\hline
			$10$ \%  & $70$  & $126$ & $205$ & $286$ & $491$ & $721$ & $969$ \\ \hline
			$20$ \%  & $84$  & $153$ & $232$ & $317$ & $507$ & $736$ & $1025$ \\ \hline
			$40$ \%  &  $88$  & $157$ & $236$ & $328$ & $575$ & $822$ & $1105$ \\ \hline
			$60$ \%  &  $87$  & $161$ & $240$ & $336$ & $563$ & $834$ & $1141$  \\ \hline
		\end{tabular}
		\bigskip
		\caption{Run-time  of Myopic caching.}
		\label{tab:run_MYO_2}
\end{table}

\section{Conclusions}
\label{Sec_Conclusion}
A generic setup where a caching unit makes sequential fetch-cache decisions based on dynamic prices and user requests was investigated. Critical constraints were identified, the aggregated cost across files and time instants was formed, and the optimal adaptive caching was then formulated as a stochastic optimization problem.  Due to the effects of the current cache decisions on future costs, the problem was cast as a dynamic program. To address the inherent functional estimation problem that arises in this type of programs, while leveraging the underlying problem structure, several computationally efficient algorithms were developed, including off-line (batch) approaches, as well as online (stochastic) approaches based on Q-learning. The last part of the paper was devoted to dynamic pricing mechanisms that allowed handling constraints both in the storage capacity of the cache memory, as well as on the back-haul transmission link connecting the caching unit with the cloud.

%Analysis Chapters
%%%%%%%%%%%%%%%%%%%%%%%%%%%%%%%%%%%%%%%%%%%%%%%%%%%%%%%%%%%%%%%%%%%%%%%%%%%%%%%
% intro.tex: Introduction to the thesis
%%%%%%%%%%%%%%%%%%%%%%%%%%%%%%%%%%%%%%%%%%%%%%%%%%%%%%%%%%%%%%%%%%%%%%%%%%%%%%%%
\chapter{Data-driven, Reinforced, and Robust Learning Approaches for \\ a Smarter Power Grid}\label{chap:sparse}
%%%%%%%%%%%%%%%%%%%%%%%%%%%%%%%%%%%%%%%%%%%%%%%%%%%%%%%%%%%%%%%%%%%%%%%%%%%%%%%%
	\section{Introduction} \label{Sec:Intro}
	
	Frequent and sizable voltage fluctuations caused by the growing deployment of electric vehicles, demand response programs, and renewable energy sources, challenge modern distribution grids. Electric utilities are currently experiencing major issues related to the unprecedented levels of load peaks as well as renewable penetration. For instance, a solar farm connected at the end of a long distribution feeder in a rural area can cause voltage excursions along the feeder, while the apparent power capability of a substation transformer is strained by frequent reverse power flows. Moreover, over-voltage happens during midday when photovoltaic (PV) generation peaks and load demand is relatively low; whereas voltage sags occur mostly overnight due to low PV generation even when load demand is high~\cite{carvalho2008distributed}. This motivates why voltage regulation, the task of maintaining bus voltage magnitudes within desirable ranges, is critical in modern distribution grids.
	
	Early approaches to regulating the voltages at a residential level have mainly relied on utility-owned devices, including load-tap-changing transformers, voltage regulators, and capacitor banks, to name a few. %~\cite{kersting2006distribution}. 
	They offer a convenient means of controlling reactive power, through which the voltage profile at their terminal buses as well as at other buses can be regulated~\cite[p. 678]{kundur1994power}. 
	Obtaining the optimal configuration for these devices entails solving mixed-integer programs, which are NP-hard in general. To optimize the tap positions, a semi-definite relaxation heuristic was used in~\cite{robbins2016optimal,bazrafshan2018optimal}. 
	%, where they split the power line into two sections. By regulating the voltage on one section, the transformer regulated the voltage along the other section~\cite{masters2002voltage}. 
	%A major challenge in here entails 
	%, where the nonconvexity in coupling capacitor voltage transformers stems from the discrete nature of configuration values, which have been addressed in the models using magnetic core representations 
	%Heuristics based on magnetic core representations 
	Control rules based on heuristics were developed 
	%to configure capacitor units or adjust voltage regulators 
	in \cite{tziouvaras2000mathematical,carvalho2008distributed}. 
	However, these approaches can be computationally demanding, and do not guarantee optimal performance. A batch reinforcement learning (RL) scheme based on linear function approximation was lately advocated in~\cite{xu2018optimal}. \footnote{Results of this Chapter are reported in \cite{yang2019statistical, yang2020deep, yang2020power, yang2020powersmartgrid, yang2019twotrans, yang2019twosmartgrid}.}
	
	%reinforcement learning paper on IEEE PES     ~\cite{vj2004reinforcement}  
	% cybernetics      ~\cite{xu2012multiagent}

	Another characteristic inherent to utility-owned equipment is their limited life cycle, which prompts control on a daily or even monthly basis.
	Such configurations have been effective in traditional distribution grids without (or with low) renewable generation, and with slowly varying load. Yet, as distributed generation grows in residential networks nowadays \cite{su2014stochastic},  \cite{ipakchi2009grid}, %\cite{liang16scalable},
	%renewable energy~\cite{painuly2001barriers}, intrinsically stochastic renewable energy generation~\cite{su2014stochastic}, and more recently emerging of electric vehicles~\cite{liang16scalable},
	rapid voltage fluctuations occur frequently. %~\cite{turitsyn2011options}.
	According to a recent landmark bill, California mandated $50\%$ of its electricity to be powered by renewable resources by $2025$ and $60\%$ by $2030$.
	The power generated by a solar panel can vary by 15\% of its nameplate rating within one-minute intervals~\cite{wang2016ergodic}. Voltage control would entail more frequent switching actions, and further installation of control devices. 
	
	Smart power inverters on the other hand, come with contemporary distributed generation units, such as PV panels, and wind turbines. Embedded with computing and communication units, these can be commanded to adjust reactive power output within seconds, and in a continuously-valued fashion. Indeed, engaging smart inverters in reactive power control
	has recently emerged as a promising solution \cite{kekatos2015stochastic}.
	Computing the optimal setpoints for inverters' reactive power output is an instance of the optimal power flow task, which is non-convex~\cite{farivar2011inverter}.  %; see \cite{molzahn2019survey} for a recent survey of convex relaxation solutions. 
	To deal with the renewable uncertainty as well as other communication issues (e.g., delay and packet loss),  
	stochastic, online, decentralized, and localized reactive control schemes have been advocated~\cite{kekatos2015stochastic,zhu2016fast,kekatos2016voltage,wang2016ergodic,fitee2019wgcs,lin2018real,zhang2018distributed}.   % , 
	%		zamzam2016beyond,
	%		tac2018zhou}.
	
	RL refers to a collection of tools for solving Markovian decision processes (MDPs),
	especially when the underlying transition mechanism is unknown~\cite{RLbook}.
	%	A number of stochastic control problems can be naturally formulated as Markov decision processes (MDPs). Reinforcement learning (RL) is a collection of techniques for solving MDPs, when the underlying transition mechanism is unknown \cite{RLbook}. 
	In settings involving high-dimensional, continuous action and/or state spaces however, it is well known that conventional RL approaches suffer from the so-called `curse of dimensionality,' which limits their impact in practice~\cite{mnih2015human}.  
	Deep neural networks (DNNs) can address the curse of dimensionality in the high-dimensional and continuous state space by providing compact low-dimensional representations of high-dimensional inputs \cite{goodfellow2016deep}. Wedding deep learning with RL (using a DNN to approximate the action-value function), deep (D) RL has offered artificial agents with human-level performance across diverse application domains~\cite{mnih2015human,sadeghi2019optimal}. 
	(D)RL algorithms have also shown great potential in several challenging power systems control and monitoring tasks~\cite{diao2019autonomous,ernst2004power,xu2018optimal,zamzam2019energy,yan2018data, lu2019demand}, and load control \cite{tsg2017cvr,duan2018qlearning}.
	A batch RL scheme using linear function approximation was developed for voltage regulation in distribution systems~\cite{xu2018optimal}. For voltage control of transmission networks, DRL was recently investigated to adjust generator voltage setpoints \cite{diao2019autonomous}.	
	A shortcoming of the mentioned (D)RL voltage control schemes is their inability to cope with the curse of dimensionality in action space.  Moreover, \textit{joint control} of both utility-owned devices and emerging power inverters has not been fully investigated. In addition, the discrete variables describing the on-off operation of capacitors and slow timescale associated with changing capacitor statuses, compared with those of fast-responding inverters further challenges voltage regulation. As a consequence, current capacitor decisions have a long-standing influence on future inverter setpoints. The other way around, current inverter setpoints also affect future commitment of capacitors through the aggregate cost. Indeed, this two-way long-term interaction is difficult to model and cope with.
	
	In this context, voltage control is dealt with in the present Chapter using shunt capacitors and smart inverters. Preliminary results were presented in \cite{yang2019smartgridcomm}.
	A novel two-timescale solution combining first principles based on physical models and data-driven advances is put forth. On the slow timescale (e.g., hourly or daily basis), the optimal configuration (corresponding to the discrete on-off commitment) of capacitors is formulated as a Markov decision process, by carefully defining state, action, and cost according to the available control variables in the grid. The solution of this MDP is approached by means of a DRL algorithm. This framework leverages the merits of the so-termed \textit{target network} and \textit{experience replay}, which can remove the correlation among the sequence of observations, to make the DRL stable and tractable. On the other hand, the setpoints of the inverters' reactive power output, are computed by minimizing the instantaneous voltage deviation using the exact or approximate grid models on the fast timescale (e.g., every few seconds).
	%	 \textcolor{cyan} {This purple text will be deleted.} \textcolor{purple} {Evidently, capacitor decisions have a long-standing impact on the bus voltage profile, and are thus intertwined with the inverter optimization. }
	
	%	\textcolor{purple}  {Besides this \textit{long-term} dependency dealing with discrete actions, the \textit{unknown dynamic} evolution of load demand and solar generation, motivates well RL solutions that can learn the optimal control policy from data. Indeed, RL has shown great potential in several challenging power engineering tasks~\cite{ernst2004power,xu2018optimal}. In settings involving high-dimensional or continuous state spaces however, conventional RL approaches suffer from the so-called `curse of dimensionality,' that discourages their  employment~\cite{zamzam2019energy}.} 
	
	Compared with past works, our contributions can be summarized as follows.
	\begin{itemize}
		\item[\bf c1)] \textit{Joint control of two types of assets.} A hybrid data- and physics-driven approach to  managing both utility-owned equipment as well as smart inverters;
		\item[\bf c2)] \textit{Slow-timescale learning.} Modeling demand and generation as Markovian processes, optimal capacitor settings are learned from data using DRL; 
		\item[\bf c3)] \textit{Fast-timescale optimization.} Using exact or approximate grid models, the optimal setpoints for inverters are found relying on the most recent slow-timescale solution; and,
		{  \item[\bf c4)] \textit{Curse of dimensionality in action space.} Introducing hyper deep $Q$-network to handle the curse of dimensionality emerging due to large number of capacitors. }
	\end{itemize}

	\section{Voltage Control in Two Timescales}\label{sec:problem}

	In this section, we describe the system model, and formulate the two-timescale voltage regulation problem. 
	
	\subsection{System model}\label{subsec:syst}
	Consider a distribution grid of $N+1$ buses rooted at the substation bus indexed by $i = 0$,
	%			 modelled as a graph $\mathcal{G}:=(\mathcal{N}_0,\mathcal{L})$, 
	whose buses are collected into $\mathcal{N}_0:=\{0\} \cup \mathcal{N}$, and lines into $\mathcal{L}:=\{1,\ldots,N\}$. 
	%The grid is typically operated radially as a tree and served by the substation (a.k.a. the root) indexed by $n=0$, whose squared voltage magnitude $v_0$ is regulated to some constant (e.g., 1). 
	%	All buses excluding the root comprise $\mathcal{N}:=\{1,...,N\}$. 
	{ For all $i\in\mathcal{N}$ (i.e., without substation bus), let $v_{i}$ denote their squared voltage magnitude, and $p_i + j q_i$ their complex power injected. 
		%, where $p_{i}:=p_{i}^g-p_{i}^c$ and $q_{i}:=q_{i}^g-q_{i}^c$
		%with superscript $g$ ($c$) denoting generation (consumption).
		% are found as the surplus between the corresponding generation and consumption. 
		%With further $\ell_i$ denoting the squared current magnitude on line $i\in \mathcal L$, the so-called \emph{branch flow model} is dictated by the following equations \cite{baran1989optimal,low2014convex} 
		%\begin{subequations}\label{eq:ac}
		%	\begin{align}
		%	~&~	p_i=\sum_{j\in\chi_i}P_j  - (P_i-r_i \ell_i)  \label{eq:acp}\\
		%	~&~	q_i=\sum_{j\in\chi_i}Q_j  - (Q_i-x_i \ell_i)\label{eq:acq}\\
		%	~&~	v_i={v_{\pi_i}}- 2(r_iP_i+x_iQ_i)+(r_i^2+x_i^2)\ell_i \label{eq:acv}\\
		%	~&~\ell_i=\frac{P^2_i+Q^2_i}{v_{\pi_i}}\label{eq:acl}
		%	\end{align}
		%\end{subequations}
		%for all $i\in\mathcal{N}$, where $\chi_i$ denotes the set of all children buses for bus~$i$. 
		For brevity, collect all nodal quantities into column vectors $\pmb{v}$, $\pmb{p}$, $\pmb{q}$. %, $\pmb{p}^g$, $\pmb{q}^g$, $\pmb{p}^c$, and $\pmb{q}^c$. 
		Active power injection is split into its generation $p_{i}^g$ and consumption $p_{i}^c$ as $p_{i}:=p_{i}^g-p_{i}^c$; likewise, reactive power injection is $q_{i}:=q_{i}^g-q_{i}^c$. In distribution grids, it holds that $p_{i}^g = p_{i}^c = q_{i}^c = 0$ and $q_{i}^g >0$ 
		if bus $i$ has a capacitor; while $p_{i}^g = q_{i}^g =0$ if bus $i$ is a purely load bus; and $p_{i}^c \geq 0$, $q_{i}^c \geq 0$, $p_{i}^g \geq 0$ if bus $i$ is equipped with a DG. Let us stack generation and consumption components into vectors $\pmb{p}^g$, $\pmb{q}^g$, $\pmb{p}^c$, and $\pmb{q}^c$ accordingly.
		Predictions of active power consumption and solar generation $(\pmb{p}^c,\pmb{q}^c,\pmb{p}^g)$ can be obtained through the hourly and real-time market (see e.g., \cite{kekatos2015stochastic}), or by running load demand (solar generation) prediction algorithms \cite{tsp2019psse}.}
	
\begin{figure}
		\centering
		\includegraphics[width =0.7 \textwidth]{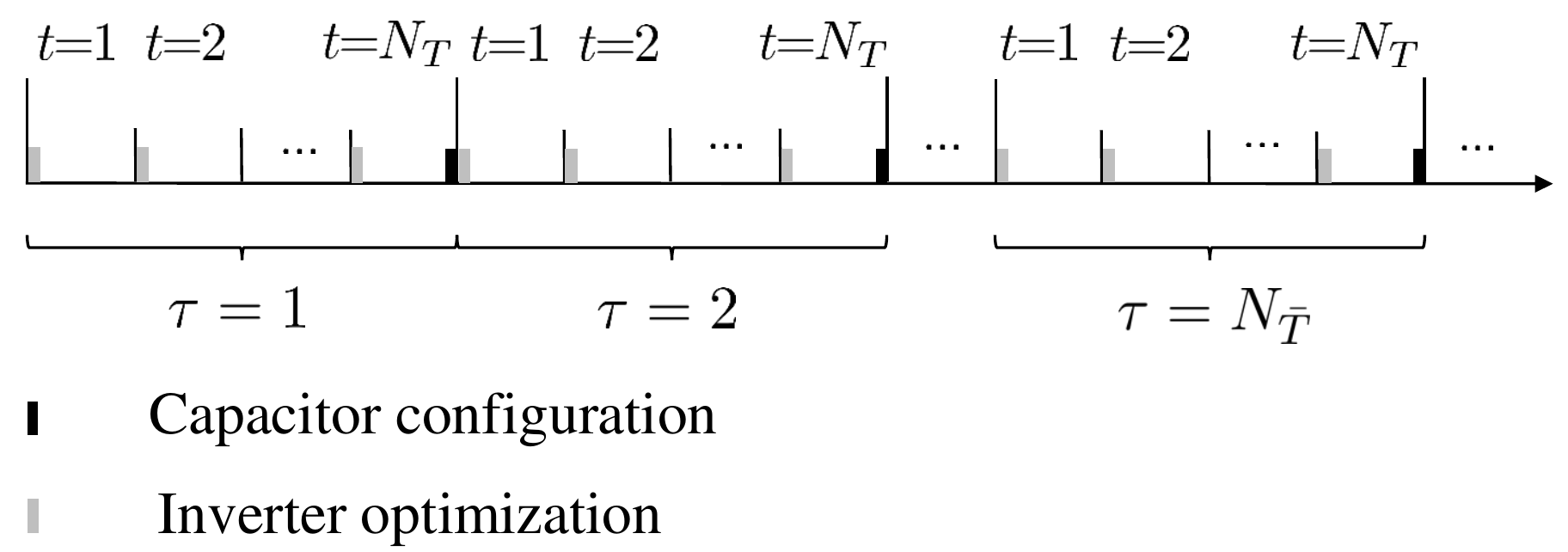}
		\caption{Two-timescale partitioning of a day for joint capacitor and inverter control.}
		\label{fig:twotimescale}
	\end{figure}

	As mentioned earlier, there are two types of assets in modern distribution grids that can be engaged in reactive power control; that is, utility-owned equipment featuring discrete actions and limited lifespan, as well as smart inverters controllable within seconds and in a continuously-valued fashion. As the aggregate load varies in a relatively slow way, traditional devices have been sufficient for providing voltage support; while fast-responding solutions using inverters become indispensable with the increase of uncertain renewable penetration. 
	In this context, the present work focuses on voltage regulation by capitalizing on the reactive control capabilities of both capacitors and inverters, while our framework can also account for other reactive power control devices. To this end, we divide every day into $N_{\bar T}$ intervals indexed by $\tau=1,\ldots,N_{\bar T}$. Each of these $N_{\bar T}$ intervals is further partitioned into $N_T$ time slots which are indexed by $t=1,\ldots,N_T$, as illustrated in Fig.~\ref{fig:twotimescale}. To match the slow load variations, the on-off decisions of capacitors are made (at the end of) every interval $\tau$, which can be chosen
	to be e.g., an hour; yet, to accommodate the rapidly changing renewable generation, the inverter output is adjusted (at the beginning of) every slot $t$, taken
	to be e.g., a minute. 
	We assume that quantities 
	$\pmb{p}^g(\tau,t)$, 
	%$\pmb{q}^g(T,t)$,
	$\pmb{p}^c(\tau,t)$, and $\pmb{q}^c(\tau,t)$ remain the same within each $t$-slot, but may change from slot $t$ to $t+1$. 
	
	Suppose there are $N_a$ shunt capacitors installed in the grid, whose bus indices are collected in $\mathcal{N}_a$, and are in one-to-one correspondence with entries of $\mathcal{K}:=\{1,\ldots,N_a \}$ (a simple renumbering). Assume that every bus is equipped with either a shunt capacitor or a smart inverter, but not both. The remaining buses, after removing entries in $\mathcal{N}_a$ from $\mathcal{N}$, collected in $\mathcal{N}_r$, are assumed equipped with inverters. This assumption is made without loss of generality as one can simply set the upper and lower bounds on the reactive output to zero at buses having no inverters installed. 
	
	As capacitor configuration is performed on a slow timescale (every  $\tau$), the reactive compensation $q_{i}^g(\tau,t)$ provided by capacitor $k_i\in\mathcal{K}$ (i.e., capacitor at bus $i$) is represented by
	\begin{equation}
	q_{i}^{g}(\tau,t) = \hat y_{k_i}(\tau) q_{a,k_i}^{g}, \quad \forall i \in \mathcal N_a, \tau,t
	%~ k_i\in\mathcal{K},
	\end{equation}
	where $\hat y_{k_i}(\tau)\in \{0,1\}$ is the on-off commitment of capacitor $k_i$ for the entire interval $\tau$. Clearly, if $\hat y_{k_i}(\tau)=1$, a constant amount (nameplate value) of reactive power $q_{a,k_i}^{g}$ is injected in the grid during this interval, and $0$ otherwise. For convenience, the on-off decisions of capacitor units at interval $\tau$ are collected in a column vector $\hat {\pmb y}(\tau)$.

	On the other hand, the reactive power $q_{r,i}^{g}(\tau,t)$ generated by inverter $i$ is adjusted on the fast timescale (every $t$), and it is constrained by
	$|q_{r,i}^{g}(\tau,t)|\leq \sqrt{(\bar s_i)^2-( p^g_i(\tau, t))^2}$, where $\bar s_i$ is the power capability of inverter $i$. Traditionally, inverter $i $ is designed as $ \bar s_i = \bar p^g_i$, where $\bar p^g_i$ is the active power capacity of the renewable generation unit installed at bus $i$. However, when maximum output is reached, i.e., $p^g_i(\tau, t) = \bar p^g_i$, no reactive power can be provided. To address this, oversized inverters' nameplate capacity has been advocated such that $ \bar s_i > \bar p^g_i$ \cite{kekatos2015stochastic}.
	For instance, choosing $ \bar s_i = 1.08 \bar p^g_i$ and limiting $q_{r,i}^{g}(\tau,t)$ to $ \sqrt{(\bar s_i)^2-( \bar p^g_i)^2}$ instead of $\sqrt{(\bar s_i)^2-( p^g_i(\tau, t))^2}$, the reactive power compensation provided by inverter $i$ is $|q_{r,i}^{g}(\tau,t)|\leq 0.4 \bar p^g_i$, regardless of the instantaneous PV output $p^g_i(\tau, t)$ \cite{kekatos2015stochastic}. As such, $q_{r,i}^{g}(\tau,t)$ generated by inverter $i$ is constrained as 
	\begin{equation}
	|q_{r,i}^{g}(\tau,t)|\leq \bar q^g_i:= \sqrt{(\bar s_i)^2-(\bar p^g_i)^2}, \quad \forall i \in \mathcal N_r ,t.
	\end{equation}
	
	%	On the other hand, the reactive power $q_{r,i}^{g}(\tau,t)$ generated by inverter $i$ is adjusted on the fast timescale (every $t$), which is constrained as
	%	\begin{equation}
	%	|q_{r,i}^{g}(\tau,t)|\leq \bar q^g_i:= \sqrt{(\bar s^g_i)^2-(\bar p^g_i)^2}, ~\forall i \in \mathcal N_r ,t
	%	\end{equation}
	%	where $\bar s^g_i$ and $\bar p^g_i$ are the nameplate values of apparent power and active power of inverter $i$, respectively; see e.g., \cite{kekatos2015stochastic}. 
	
	\subsection{Two-timescale voltage regulation formulation}
	Given two-timescale load consumption and generation that we model as Markovian processes \cite{carta2009review}, the task of voltage regulation is to find the optimal reactive power support per slot by configuring capacitors in every interval and adjusting inverter outputs in every slot, such that the \emph{long-term} average voltage deviation is minimized. As voltage magnitudes $\pmb{v}(\tau,t)$ depend solely on the control variables $\pmb{q}^g(\tau,t)$, they are expressed as implicit functions of $\pmb{q}^g(\tau,t)$, yielding $\pmb{v}_{\tau,t}(\pmb{q}^g(\tau,t))$, whose actual function forms for postulated grid models will be given Section \ref{sec:fast}. The novel two-timescale voltage control scheme entails solving the following stochastic optimization problem
	\begin{subequations}\label{eq:twotimescale}
		\begin{align}
		\underset{\{\pmb{q}_r^g(\tau,t)\}\atop
			\left\{ \pmb{y}(\tau)\in \{0,1\}^{N_a}  \right \}}{\minimize}
		~~&{\mathbb{E}}\! \left[\sum_{\tau=1}^\infty \sum_{t=1}^{N_T}\gamma^\tau
		\left\|\pmb v_{\tau,t}(\pmb q^g(\tau,t))	- v_0\pmb{1}\right\|^2\right]\\
		{\rm subject\;to}\quad \;&q^g_i(\tau,t) =\hat y_{k_i}(\tau) q_{a,k_i}^{g}, 
		~\,\quad 
		\forall i\in\mathcal N_a, \tau, t\label{eq:const1} \\
		&q^g_i(\tau,t) = q_{r,i}^g(\tau,t),~\,\quad \quad
		\forall i\in \mathcal{N}_r, \tau, t\\
		&|q_{r,i}^{g}(\tau,t)|\leq \bar q^g_i,\qquad \qquad \forall i \in \mathcal N_r ,   \tau,t \label{eq:trad3} 
		\end{align}
	\end{subequations}
	%where we have neglected the set of power flow constraints \eqref{eq:ac};
	for some discount factor $\gamma \in (0,1)$, where the expectation is taken over the joint distribution of $(\pmb{p}^c(\tau,t),\pmb{q}^c(\tau,t),\pmb{p}^g(\tau,t))$ across all intervals and slots. Clearly, the optimization problem \eqref{eq:twotimescale} involves infinitely many
	variables $ \{\pmb{q}^g_r(\tau,t)\}$ and $\{\hat {\pmb{y}}(\tau) \}$, which are coupled across time via the cost function and the constraint \eqref{eq:const1}. 
	Moreover, discrete variables $\hat{\pmb{y}}(\tau)\in\{ 0,1\}^{N_a} $ render problem \eqref{eq:twotimescale} nonconvex and generally \emph{NP-hard}. Last but not least, it is a multi-stage optimization, whose decisions are not all made at the same stage, and must also account for the power variability during real-time operation. In words, tackling \eqref{eq:twotimescale} exactly is challenging. 
	
	Instead, our goal is to design algorithms that sequentially observe predictions \linebreak $\{(\pmb{p}^c(\tau,t),\pmb{q}^c(\tau,t)),\pmb{q}^g(\tau,t)\}$, and solve near optimally problem \eqref{eq:twotimescale}. The assumption is that,  although no distributional knowledge of those stochastic processes involved is given, their realizations can be made available in real time, by means of e.g., accurate forecasting methods \cite{tsp2019psse}. In this sense, the physics governing the electric power system will be utilized together with data to solve \eqref{eq:twotimescale} in real time. Specifically, on the slow timescale, say at the end of each interval $\tau-1$, the optimal on-off capacitor decisions $\pmb y(\tau)$ will be set through a DRL algorithm that can learn from the predictions collected within the current interval $\tau-1$; while, on the fast timescale, namely at the beginning of each slot $t$ within interval $\tau$, our two-stage control scheme will compute the optimal setpoints for inverters, by minimizing the instantaneous bus voltage deviations while respecting physical constraints, given the current on-off commitment of capacitor units $\hat{\pmb y}(\tau)$ found at the very end of interval $(\tau -1)$. These two timescales are detailed in Sections~\ref{sec:fast} and \ref{sec:slow}, respectively.
	
	%%%%%%%%%%%%%%%%%%%%%%%%%%%%%%%%%%%%%%%%%%%%%%%%%%%%%%%%%%%%%%%%%%%%%%%
	\section{Fast-timescale Optimization of Inverters}\label{sec:fast}
	As alluded earlier, the actual forms of $\pmb{v}_{\tau,t}(\pmb{q}^g(\tau,t))$ will be specified in this section, relying on the exact AC model or a linearized approximant of it.
	Leveraging convex relaxation to deal with the nonconvexity, the considered AC model yields a second-order cone program (SOCP), whereas the linearized one leads to a linearly constrained quadratic program. In contrast, the latter offers an approximate yet computationally more affordable alternative to the former. Selecting between these two models relies on affordable computational capabilities. 
	
	\subsection{Branch flow model}\label{subsec:branch}
	
	\begin{figure}
		\centering
		\includegraphics[width =0.45 \textwidth]{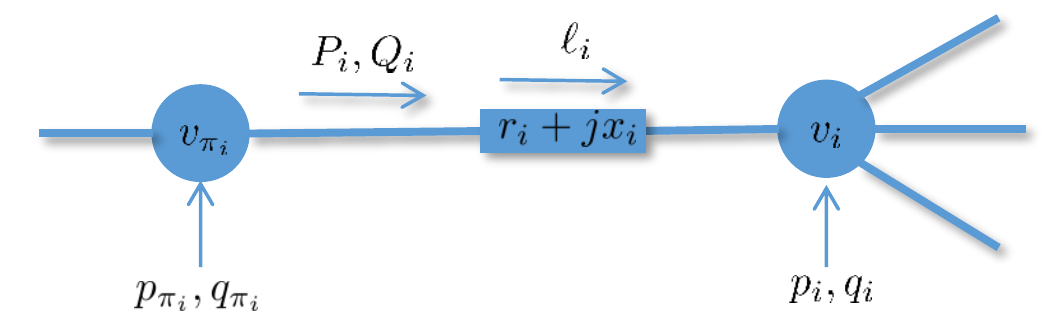}
		\caption{Bus $i$ is connected to its unique parent $\pi_i$ via line $i$.}
		\label{fig:lineardiagram}
	\end{figure}
	
	Due to the radial structure of distribution grids, every non-root bus $i\in\mathcal{N}$ 
	has a unique parent bus termed $\pi_i$. The two are joined through the $i$-th distribution line represented by $(\pi_i,i) \in\mathcal{L}$ having impedance $r_i+jx_i$. Let $P_i(\tau, t)+jQ_i(\tau, t)$ stand for the complex power flowing from buses $\pi_i$ to $i$ seen at the `front' end at time slot $t$ of interval $\tau$, as depicted in Fig.~\ref{fig:lineardiagram}. Throughout this section, the interval index $\tau$ will be dropped when it is clear from the context.  
	
	With further $\ell_i$ denoting the squared current magnitude on line $i\in \mathcal L$, the celebrated \emph{branch flow model} is described by the following equations for all buses $ i \in \mathcal N$, and for all $ t$ within every interval $\tau$ \cite{baran1989optimal,low2014convex} 
	\begin{subequations}\label{eq:nonlinear1}
		\begin{align}
		p_i(t)&=\sum_{j\in\chi_i}P_j(t)  -( P_i(t) -r_i \ell_i(t)) \label{eq:nonp}\\
		q_i(t)&=\sum_{j\in\chi_i}Q_j(t)  - (Q_i(t)-x_i \ell_i(t))\label{eq:nonq}\\
		v_i(t)&={v_{\pi_i}(t)}\!- 2(r_iP_i(t)\!+x_iQ_i(t))\!+(r_i^2\!+x_i^2)\ell_i(t)  \label{eq:nonv}\\
		\ell_i(t)&=\frac{P^2_i(t)+Q^2_i(t)}{v_{\pi_i}(t)}\label{eq:ml}
		\end{align}
	\end{subequations}
	% (but the index $\tau$ is ignored for brevity), 
	where we have ignored the dependence on $\tau$ for brevity, and $\chi_i$ denotes the set of all children buses for bus~$i$.
	%	Likewise, we collect $\{r_i\}$, $\{P_i(t)\}$, $\{Q_i(t)\}$, and $\{\ell_i(t)\}$ into vectors $\pmb{r}$, $\pmb{P}(t)$, $\pmb{Q}(t)$ and $\pmb{\ell}(t)$, accordingly.
	
	%	Equations \eqref{eq:nonp}-\eqref{eq:nonv} are linear in variables $\pmb{P}(t)$, $\pmb{Q}(t)$, $\pmb{v}(t)$, and $\pmb{\ell}(t)$. 	Nonetheless,
	Clearly, the set of equations in \eqref{eq:ml} is quadratic in ${P}_i(t)$ and ${Q}_i(t)$, yielding a nonconvex set. To address this challenge, consider relaxing the equalities \eqref{eq:ml} into inequalities (a.k.a. hyperbolic relaxation, see e.g., \cite{farivar2011inverter}) 
	\begin{align}\label{eq:inequality}
	P^2_i(t)+Q^2_i(t)\le {v_{\pi_i}(t)}\ell_i(t), \quad \forall i \in \mathcal N,  t
	\end{align}
	which can be equivalently rewritten as the following second-order cone constraints
	\begin{align}\label{eq:soc}
	\!\left\|\begin{array}{c}
	2P_i(t)\\
	2Q_i(t)\\
	\ell_i(t)-v_{\pi_i}(t)\end{array}\right\|\le {v_{\pi_i}(t)}+\ell_i(t), \quad \forall i \in \mathcal N.
	\end{align}
	Equations \eqref{eq:nonp}-\eqref{eq:nonv} and \eqref{eq:soc} now define a convex feasible set. 
	%Recent efforts have leveraged this relaxed set (instead of the original nonconvex one) to study several key grid management tasks; see e.g., \cite{low2014convex, molzahn2019survey} for recent surveys. 
	The procedure of leveraging this relaxed set (instead of the nonconvex one) is known as SOCP relaxation \cite{low2014convex}. Interestingly, it has been shown that under certain conditions, SOCP relaxation is exact in the sense that the set of inequalties \eqref{eq:soc} holds with equalities at the optimum \cite{gan2015exact}. 
	
	Given the capacitor configuration $\hat{\pmb y}(\tau)$ found at the end of the last interval $\tau-1$, under the aforementioned relaxed grid model, the voltage regulation on the fast timescale based on the exact AC model can be described as follows
	\begin{subequations}\label{eq:nonlinear}
		\begin{align}
		\underset{\pmb{v}(t),\pmb{q}^g_r(t),
			\pmb{P}(t),\pmb{Q}(t)}{\minimize}~&~ \|\pmb v(t)
		- v_0\pmb{1}\|^2\\
		{\rm subject\;to}~~~\,
		~&~ \eqref{eq:nonp}-\eqref{eq:ml}\notag\\
		~&~q^g_i(t) = \hat y_{k_i}(\tau) q_{a,k_i}^{g},
		~\,\,\forall i\in\mathcal N_a \label{eq:trad11}\\
		~&~q^g_i(t) = q_{r,i}^g(t),~\,\quad \,~~~
		\forall i\in \mathcal{N}_r \label{eq:trad12}\\
		~&~|q_{r,i}^{g}(t)|\leq \bar q^g_i,\quad\quad\,\,\,\,\,\,\,\, \forall i \in \mathcal N_r  \label{eq:trad13} 
		\end{align}
	\end{subequations}
	which is readily a convex SOCP and can be efficiently solved by off-the-shelf convex programming toolboxes. 
	%Problem \eqref{eq:nonlinear} can be reformulated to an SOCP.
	%Note that the relaxed problem \eqref{eq:nonlinear} in the distribution grid is not only convex, which results from the convexity of constraint~\eqref{eq:soc}, but also exact, that is, the optimal solution of \eqref{eq:nonlinear} attains \eqref{eq:soc} with equality; see~\cite{low2014convex} and reference therein.   
	The optimal setpoints of smart inverters for the exact AC model are found as the $\bm{q}_r^g$-minimizer of~$\eqref{eq:nonlinear}$.
	
	However, solving SOCPs could be computationally demanding when dealing with relatively large-scale distribution grids, say of several hundred buses. Trading off modeling accuracy for computational efficiency, our next instantiation of the fast-timescale voltage control relies on an approximate grid model.
	
	\subsection{Linearized power flow model}\label{sec:fastlinear}
	%	The linearized distribution flow model can be obtained as follows.
	As line current magnitudes $\{ \ell_i\}$ are relatively small compared to line flows, the last term in \eqref{eq:nonp}-\eqref{eq:nonv} can be ignored yielding the next set of linear equations for all $i,t$ \cite{baran1989network}
	\begin{subequations}\label{eq:linearpowerflow}
		\begin{align}
		~&p_i(t)=\sum_{j\in\chi_i}P_j(t)- P_i(t)\label{eq:linp}\\
		~&q_i(t)=\sum_{j\in\chi_i}Q_j(t)  - Q_i(t)\label{eq:linq}\\
		~&v_i(t) =v_{\pi_i}(t)- 2(r_i P_i(t)+x_i Q_i(t))\label{eq:linv}
		\end{align}
	\end{subequations}
	which is known as the linearized distribution flow model. 
	In this fashion, all squared voltage magnitudes $\pmb{v}(t)$ can be  expressed as linear functions of $\pmb q^g( t)$. 
	
	Adopting the approximate model \eqref{eq:linearpowerflow}, the optimal setpoints of inverters can be found by solving the following optimization problem per slot $ t $ in interval $\tau$, provided $\hat {\pmb y} (\tau)$ is available from the last interval on the slow timescale
	\begin{subequations}\label{eq:linear}
		\begin{align}
		\underset{\pmb{v}(t),\pmb{q}^g_r(t),
			\pmb{P}(t),\pmb{Q}(t)}{\minimize} ~&~ \|\pmb v(t)
		- v_0\pmb{1}\|^2\\
		{\rm subject~to}~~~~~&~\eqref{eq:linp}-\eqref{eq:linv} \notag\\
		~&~q^g_i(t) = \hat y_{k_i}(\tau) q_{a,k_i}^{g},	~\,\,\forall i\in\mathcal N_a \label{eq:trad21}\\
		~&~q^g_i(t) = q_{r,i}^g(t),~~~\,\quad \,\,\,\forall i\in \mathcal{N}_r\\
		~&~|q_{r,i}^{g}(t)|\leq \bar q^g_i,\quad\,\,\,\,\,\,\,\,\,~\,~ \forall i \in \mathcal N_r . \label{eq:trad23} 
		\end{align}
	\end{subequations}
	
	As all constraints are linear and the cost is quadratic, \eqref{eq:linear} constitutes a standard convex quadratic program. As such, it can be solved efficiently by e.g.,  primal-dual algorithms, or off-the-shelf convex programming solvers, whose implementation details are skipped due to space limitations. 
	
	%%%%%%%%%%%%%%%%%%%%%%%%%%%%%%%%%%%%%%%%%%%%%%%%%%%%%%%%%%%%%%%%%%%%%%%
	\section{Slow-timescale Capacitor Reconfiguration}\label{sec:slow}
	
	Here we deal with reconfiguration of shunt capacitors on the slow timescale. This amounts to  determining their on-off status for the ensuing interval. Past approaches to solving the resultant integer-valued optimization were heuristic, or, relied on semidefinite programming relaxation. They do not guarantee optimality, while they also incur high computational and storage complexities. We take a different route by drawing from advances in artificial intelligence, to develop data-driven solutions that could near optimally learn, track, as well as adapt to unknown generation and consumption dynamics.

\subsection{A data-driven solution} \label{subsec:datadriven}
%\qiu{There exists a two-way interactive influence between capacitor on/off decisions and optimal setpoints of smart inverters. The capacitor configuration has a long-standing influence on voltage profiles and therefore on smart inverters, and on the other hand, the aggregated performance of smart inverters affects the decisions of capacitor configuration. This interactive long-standing influence well motivates our RL formulation. 
%RL deals with learning action-taking policy functions in an environment with action-dependent dynamically evolving states and costs. 
%By interacting with the environment (through successive actions
%and observed states and costs), RL seeks a policy function (of
%states) to draw actions from, in order to minimize the average
%cumulative cost~\cite{RLbook}.}

\begin{figure}
	\centering
	\includegraphics[width =0.7 \textwidth]{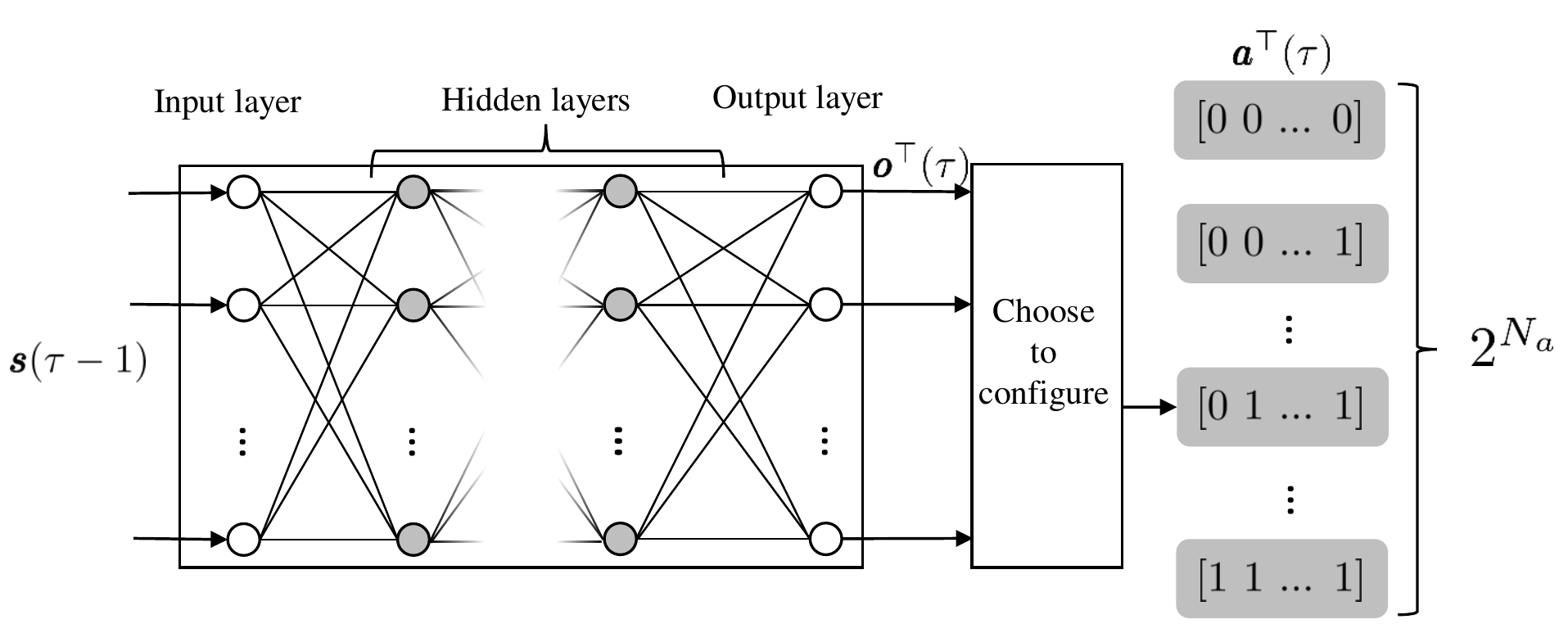}
	\caption{Deep $Q$-network}
	\label{fig:deepneuralnetwork}
\end{figure}

Clearly from \eqref{eq:trad11}--\eqref{eq:trad21}, the capacitor decisions  $ \hat {\pmb y}(\tau)$ made at the end of interval $\tau-1$ (slow-timescale learning) influence inverters' setpoints during the entire interval $\tau$ (fast-timescale optimization).  The other way around, inverters' regulation on voltages influences the capacitor commitment for the next interval. This two-way between the capacitor configuration and the optimal setpoints of inverters motivates our RL formulation.
% RL deals with learning action-taking policy functions in an environment with action-dependent dynamically evolving states and costs. 
%By interacting with the environment (through successive actions as well as observed states and costs), 
Dealing with learning 
%	action-taking 
policy functions in an environment with action-dependent dynamically evolving states and costs, 
RL seeks a policy function (of states) to draw actions from, in order to minimize the average cumulative cost~\cite{RLbook}.

%\qiu{In the distribution grid, the capacitor on/off decisions have a long-standing influence on voltage profiles and therefore on smart inverters. On the other hand, the aggregated performance of smart inverters affects the capacitor configuration. This two-way interactive long-standing influence between the capacitor configuration and the optimal setpoints of smart inverters well motivates our RL formulation. RL deals with learning action-taking policy functions in an environment with action-dependent dynamically evolving states and costs. By interacting with the environment (through successive actionsand observed states and costs), RL seeks a policy function (of states) to draw actions from, in order to minimize the average cumulative cost~\cite{RLbook}.  }

Modeling load demand and renewable generation as Markovian processes, the optimal configuration of capacitors can be formulated as an MDP, which can be efficiently solved through RL algorithms.
An MDP is defined as a 5-tuple $(\mathcal{S},\mathcal{A}, \mathcal{P}, c, \gamma)$, where $\mathcal {S}$
is a set of states; $\mathcal {A}$ is a set of actions; $\mathcal {P}$ is a set of transition matrices;  $c:\mathcal{S}\times \mathcal{A}\mapsto {\mathbb{R}}$ is a cost function such that, for $\pmb s \in \mathcal{S}$ and $\pmb a \in \mathcal{A}$, $c=(c(\pmb s, \pmb a))_{\pmb s\in\mathcal{S},\pmb a\in\mathcal{A}}$ are the real-valued instantaneous costs after the system operator takes an action $\pmb a$ at state $\pmb s$; and $\gamma\in [0,1)$ is the discount factor.
These components are defined next before introducing our voltage regulation scheme.

\textit{Action space $\mathcal{A}$}. Each action corresponds to one possible 
on-off commitment of capacitors $1$ to $N_a$, giving rise to an action vector $\pmb a(\tau)=\pmb y(\tau)$ per interval $\tau$.
%Let action $\pmb a (T-1)=[a_{1}(T-1)~\cdots~a_{N_c}(T-1)]^\top $ collect on-off commitments of capacitors $1$ to $N_c$ over the entire interval $T$.
%Note that the action is discrete in the capacitor configuration problem.
The set of binary action vectors constitutes the action space $\mathcal{A}$, whose cardinality is exponential in the number of capacitors, meaning $|\mathcal{A}|=2^{N_a}$.

\textit{State space $\mathcal{S}$}. This includes per interval $\tau$ the average active power at all buses except for the substation, along with the current capacitor configurations; that is, $\pmb s(\tau) :=[{ {\,\bar{\pmb p}} }^{\top}(\tau), { \hat{\pmb y}}^{\top}(\tau)]^{\top}$, which contains both continuous and discrete variables. Clearly, it holds that $\mathcal{S}\subseteq \mathbb R^N \times 2^{N_a}$. 

The action is decided according to the configuration policy $\pi$ that is a function of the most recent state $\pmb s(\tau-1)$, given as
\begin{align}
\pmb a(\tau) = \pi(\pmb s(\tau-1)).
\end{align}

\textit{Cost function $c$}. 
%The objective of voltage regulation is to minimize the voltage deviation at each bus from its nominal value. Therefore, the cost function is defined as 
The cost on the slow timescale is 
\begin{align}\label{eq:rlcost}
c(\pmb s(\tau-1), \pmb a(\tau)) = 	
\sum_{t=1}^{N_T}\!\left\|\pmb v_{\tau,t}(\pmb q^g(\tau,t)) - v_0\pmb{1}\right\|^2.
\end{align}
%where $\pmb q^g_c(T)$ is the reactive power generation in capacitor buses with $\pmb q^g_c(T) = \pmb y(T) \pmb q_c$.

\textit{Set of transition probability matrices $\mathcal P$.} While being at a state $\pmb s \in {\mathcal S}$ upon taking an action $\pmb a$, the system moves to a new state $\pmb s ' \in {\mathcal S}$ probabilistically. Let $P^{\pmb a}_{\pmb s \pmb s'}$ denote the transition probability matrix from state $\pmb s$ to the next state $\pmb s'$ under a given action $\pmb a$. Evidently, it holds that ${\mathcal P} :=\left\{P^{\pmb a}_{\pmb s \pmb s'}| \forall {\pmb a} \in {\mathcal A} \right\}$. 

\textit{Discount factor $\gamma$.} The discount factor $\gamma \in [0,1)$, trades off the current versus future costs. The smaller $\gamma$ is, the more weight the current cost has in the overall cost. 

Given the current state and action, the so-termed action-value function under the control policy $\pi$ is defined as
\begin{align} \label{eq:Qdef}
&	Q_{\pi} (\pmb s(\tau-1), \pmb a(\tau))  :=  \notag\\ &{\mathbb{E}}\! \left[\sum_{\tau' = \tau}^{\infty}  \gamma^{\tau' - \tau} c(\pmb s(\tau'-1), \pmb a(\tau'))\Big | \pi , \pmb s(\tau-1),  \pmb a(\tau)\right]
\end{align}
%where the expectation ${\mathbb{E}}$ is taken with respect to all randomness.
where the expectation ${\mathbb{E}}$ is taken with respect to all sources of randomness.  

To find the optimal capacitor configuration policy $\pi^{\ast}$, that minimizes the average voltage deviation in the long run, we resort to the Bellman optimality equations; see e.g., \cite{RLbook}. Solving those yields the action-value function under the optimal policy $\pi ^*$ on the fly, given by 
\begin{equation}
\label{eq:Qoptimal}
Q_{\pi^*} \! \left(\pmb s, \pmb a\right) = {\mathbb{E}} \!\left[ c(\pmb s, \pmb a) \right] + \gamma \sum_{\pmb s' \in \mathcal{S}} P_{\pmb s \pmb s'}^{\pmb a} \min_{\pmb a \in \mathcal{A}} {Q_{\pi^*}} (\pmb s', \pmb a').
\end{equation}
%where %$\gamma$ is discount factor with $\gamma \in [0,1]$; 
%$P^{\pmb a}_{\pmb s \pmb s'}$ is a state transition probability from any state $\pmb s \in\mathcal{S}$ to its next state $\pmb s'\in\mathcal{S}$ after taking an action $\pmb a$. 
With $Q_{\pi^{\ast}}(\pmb s, \pmb a)$ obtained, the optimal capacitor configuration policy can be found as 
\begin{equation}
\label{eq:optimal} 
\pi^{\ast} (\pmb s) = \arg \min_{\pmb a}\, Q_{\pi^{\ast}} (\pmb s, \pmb a ).
\end{equation}

It is clear from \eqref{eq:Qoptimal} that if all transition probabilities $\{P^{\pmb a}_{\pmb s \pmb s'}\}$ were available, we can derive $Q_{\pi^{\ast}}(\pmb s, \pmb a)$, 
and subsequently the optimal policy $\pi^{\ast}$ from \eqref{eq:optimal}. Nonetheless, obtaining those transition probabilities is impractical in practical distribution systems. This calls for approaches that aim directly at $\pi^*$, without assuming any knowledge of $\{P^{\pmb a}_{\pmb s \pmb s'}\}$. 

One celebrated approach of this kind is Q-learning, which can learn $\pi^{\ast}$ by approximating $Q_{\pi^{\ast}}(\pmb s, \pmb a)$ `on-the-fly'~\cite[p. 107]{RLbook}. Due to its high-dimensional continuous state space $\mathcal{S}$ however, $Q$-learning is not applicable for the problem at hand.
This motivates function approximation based $Q$-learning schemes that can deal with continuous state domains.

%~\cite{sadeghi2019optimal,Zamzam2019energy}. 
\subsection{A deep reinforcement learning approach} 
\label{subsec:DRL}
%	Parameterizing the $Q$-function with a DNN has lately been demonstrated to be effective in dealing with high-dimensional and/or continuous state spaces \cite{mnih2015human}. Praised as the first artificial agents to achieve human-level performance across diverse challenging domains, deep RL based on the so-called deep $Q$-networks (DQN) was introduced.

%	\textcolor{blue}
%	{It is widely known that the input of deep neural network is the state-action pair $(\pmb s, \pmb a)$, while the output is a scalar variable. The cardinality of the sought Q-function is therefore $|\mathcal S \times \mathcal A| = |\mathcal S||\mathcal A|$. In our proposed method, a parametric DQN takes as input the state vector $\pmb s$ excluding $\pmb a$, reducing the computation complexity from $|\mathcal S||\mathcal A|$ to $|\mathcal S|$.}
%\textcolor{blue}{Praised as the first artificial agents to achieve human-level performance across diverse challenging domains, RL based on the so-called deep $Q$-network (DQN) has lately been proven effective in dealing with high-dimensional and/or continuous state spaces \cite{mnih2015human}.}
DQN offers a NN function approximator of the $Q$-function, chosen to be e.g., a fully connected feed-forward NN, or a convolutional NN, depending on the application \cite{mnih2015human}. It takes as input the state vector, to generate at its output $Q$-values for all possible actions (one for each). As demonstrated in \cite{mnih2015human}, such a NN indeed enables learning the $Q$-values of \textit{all} state-action pairs, from just a few observations obtained by interacting with the environment. Hence, it effectively addresses the challenge brought by the `curse of dimensionality'~\cite{mnih2015human}. Inspired by this, we employ a feed-forward NN to approximate the $Q$-function in our setting. Specifically, our DNN consists of $L$ fully connected hidden layers with ReLU activation functions, depicted in Fig.~\ref{fig:deepneuralnetwork}. At the input layer, each neuron is fed with one entry of the state vector $\pmb s(\tau-1)$, which, after passing through $L$ ReLU layers, outputs a vector $\pmb o(\tau)\in\mathbb{R}^{2^{N_a}}$, whose elements predict
the $Q$-values for all possible actions (i.e., capacitor configurations).  
Since each output unit corresponds to a particular configuration of all $N_a$ capacitors, there is a total of $2^{N_a}$ neurons at the output layer. 
For ease of exposition, let us collect all weight parameters of this DQN into a vector $\pmb \theta$ which parameterizes the input-output relationship as $\pmb o(\tau) = Q_{\pi}(\pmb s(\tau-1), \pmb a(\tau); \pmb \theta)$
(c.f.~\eqref{eq:Qdef}). 
At the end of a given interval $\tau-1$, upon passing the state vector $\pmb s(\tau-1)$ through this DQN, the corresponding predicted $Q$-values $\pmb o(\tau)$ for all possible actions become available at the output. 
Based on these predicted values, the system operator selects the action having the smallest predicted $Q$-value to be in effect over the next interval.

Intuitively, the weights $\pmb \theta$ should be chosen such that the DQN outputs match well the actual $Q$-values with input any state vector. Toward this objective, the popular stochastic gradient descent (SGD) method is employed to update $\pmb \theta$ `on the fly' \cite{mnih2015human}. 
%	\textcolor{blue}{In particular, while being in a state $\pmb s(\tau-1)$, action $\pmb a(\tau)$ is taken by $\epsilon$ greedy algorithm, that is, the action $\pmb a(\tau)$ is either randomly taken with probability (w.p.) $\epsilon_\tau$ or is an exploitation w.p. $1- \epsilon_\tau$. This exploration-exploitation is a crucial step to gather more information about environment, while obtaining a reasonable performance before converging to the optimal Q-function. % Denoting $\hat Q_{T} (\cdot,\cdot)$ as the estimate of $ Q_\pi^* (\cdot,\cdot)$, 
%		The $\epsilon$ greedy algorithm for taking action $\pmb a(\tau)$ is as follows
%		\begin{equation}
%		\label{eq:actcapacitor}
%		\pmb a(\tau)=\!\left\{\!\!\begin{array}{ll}
%		{\rm { random }~\;~\pmb a \in {\mathcal A}},&{{\rm w.p.} \;\; \epsilon_\tau}\\
%		{\arg \min}_{\pmb a'}\;~ Q (\pmb s(\tau-1), \pmb a'; \pmb \theta_{\tau} ),& {\rm w.p.~}{1-\epsilon_\tau.}
%		\end{array}
%		\right.
%		\end{equation}
%	}
At the end of a given interval $\tau$, precisely when i)  the system operator has made decision $\pmb a (\tau)$, ii) the grid has completed the transition from the state $\pmb s(\tau-1)$ to a new state $\pmb s(\tau)$, and, (iii) the network has incurred and revealed cost $c(\pmb s(\tau-1), \pmb a(\tau))$, we perform a SGD update based on the current estimate $\pmb \theta_\tau$ to yield $\pmb \theta_{\tau +1}$. The so-termed temporal-difference learning \cite{RLbook} confirms that a sample approximation of the optimal cost-to-go from interval $\tau$ is given by $c(\pmb s(\tau-1), \pmb a(\tau) ) + \gamma \min \limits_{\pmb a' \in \mathcal{A}} Q_{\pi}(\pmb s(\tau), \pmb a'; \pmb \theta_{\tau})$, where $c(\pmb s(\tau-1), \pmb a(\tau) )$ is the instantaneous cost observed, and $\min \limits_{\pmb a '}Q_{\pi}(\pmb s(\tau), \pmb a'; \pmb \theta_{\tau})$ represents the smallest possible predicted cost-to-go from state $\pmb s(\tau)$, which can be computed through our DQN with weights $\pmb \theta_{\tau}$, and is discounted by factor $\gamma$. In words, the target value  $c(\pmb s(\tau-1), \pmb a(\tau) ) + \gamma \min \limits_{\pmb a' \in \mathcal{A}} Q_{\pi}(\pmb s(\tau), \pmb a'; \pmb \theta_{\tau})$ is readily available at the end of interval $\tau-1$. 
%However, our DQN can provide the predicted cost-to-go from state $\pmb s(\tau-1)$ through $Q_{\pi}(\pmb s(\tau-1), \pmb a(\tau); \pmb \theta_{\tau})$, therefore a loss can be defined to capture prediction error. 
Adopting the $\ell_2$-norm error criterion, a meaningful approach to tuning the weights $\pmb \theta$ entails minimizing the following loss function  
\begin{align}
\label{eq:errorDef}
{\mathcal L} (\pmb \theta)  := \Big[c(\pmb s(\tau-1), \pmb a(\tau) ) + \gamma \min \limits_{\pmb a'\in \mathcal{A}} Q_{\pi}(\pmb s (\tau), \pmb a'; \pmb \theta_{\tau}) \notag  - Q_{\pi}(\pmb s(\tau-1), \pmb a(\tau); \pmb \theta)\Big]^2
&	\end{align}
for which the SGD update is given by
\begin{equation}
\label{eq:SGD}
\pmb \theta_{\tau+1} = \pmb \theta_\tau - \beta_\tau \nabla {\mathcal L}(\pmb \theta)|_{{\pmb \theta}_\tau}
\end{equation} 
where $\beta_\tau>0$ is a preselected learning rate, and $\nabla {\mathcal L}(\pmb \theta)$ denotes the (sub-)gradient. 

\begin{algorithm}[t]
	\caption{Two-timescale voltage regulation scheme.}
	\label{Alg_a}
		{{\bf Initialize}    
		$\pmb \theta_0$ randomly; weight of the target network $\pmb \theta^{\rm Tar}_0=\pmb \theta_0$; replay buffer $\mathcal R$; and the initial state $\pmb s(0)$.}
	\newline	
		{For \quad $\tau=1,2,... $ } 
		\newline
		{Take action $\pmb a(\tau)$ through exploration-exploitation
			\begin{equation}
			\label{eq:actcapacitor}\notag
			\hspace{+0.4 cm} \pmb a(\tau)=\!\left\{\!\!\begin{array}{ll}
			\!{\rm { random }~\;~\pmb a \in {\mathcal A}}\!\!\!\!\!\!&{{\rm w.p.} \;\; \epsilon_\tau}\\
			\!	{\arg \min}_{\pmb a'}~ Q (\pmb s(\tau-1), \pmb a'; \pmb \theta_{\tau} )\!\!\!\!& {\rm w.p.~}{1\!-\!\epsilon_\tau}
			\end{array}
			\right.
			\end{equation}
			\hspace{+6.37 cm} {where $\epsilon_\tau    = {\rm max} \big \{ 1 - 0.1  \times \lfloor \tau / 50 \rfloor, \, 0\big\}.$}}
		\newline
		Evaluate $\pmb c(\pmb s(\tau-1),\pmb a (\tau))$ using \eqref{eq:rlcost}.
		\newline
		{For \quad $t=1,2,...,N_T$} 
		\newline
		Compute $\pmb q^g(\tau,t)$ using \eqref{eq:nonlinear} or \eqref{eq:linear}.
		\newline
		Update $\pmb s(\tau)$.
		\newline
		 Save $(\pmb s(\!\tau-\!1),\pmb a (\tau),c(\pmb s(\tau-1),\pmb a (\tau)),\pmb s(\tau))$ into $\mathcal R(\tau)$.
		 \newline
		Randomly sample $M_\tau$ experiences from $\mathcal R(\tau)$.  
		\newline
		Form the mini-batch loss ${\mathcal{L}^{\rm Tar}}(\pmb \theta_\tau;\mathcal M_\tau)$ using \eqref{eq:minilosstar2}. 
		\newline
		Update $\pmb \theta_{\tau+1}$ using \eqref{eq:miniSGD}.
		\newline
	    {If mod$(\tau,B)=0$}
		\newline 
		Update the target network $\pmb \theta^{\rm Tar}_\tau=\pmb \theta_\tau$.
\end{algorithm}

%	\textcolor{blue}{Since the dimensionality of $\pmb \theta$ can be much smaller than $|\mathcal S||\mathcal A|$, the DQN is efficiently trained with few experiences, and generalizes to unseen state vectors.} 
However, due to the compositional structure of DNNs, the update \eqref{eq:SGD} does not work well in practice. In fact, the resultant DQN oftentimes does not provide a stable result; see e.g.,~\cite{wu2016google}. To bypass these hurdles, several modifications have been introduced. In this work, we adopt the \textit{target network} and \textit{experience replay} \cite{mnih2015human}. 
To this aim, let us define an experience $e (\tau'):=(\pmb s(\tau'-1),\pmb a (\tau')),c(\pmb s(\tau'-1),\pmb a (\tau')),\pmb s(\tau'))$, to be a tuple of state, action, cost, and the next state.  Consider also having a replay buffer $\mathcal{R}(\tau)$ on-the-fly, which stores the most recent $R>0$ experiences visited by the agent. For instance, the replay buffer at any interval $\tau\ge R$ is $\mathcal{R}(\tau):=\{e(\tau-R+1), 
%	e(\tau-R+2),
\ldots, e(\tau)\}$. Furthermore, as another effective remedy to stabilizing the DQN updates, we replicate the DQN to create a second DNN, commonly referred to as the \textit{target} network, whose weight parameters are concatenated in the vector $\pmb \theta^{\rm Tar}$. It is worth highlighting that this target network is not trained, but its parameters $\pmb \theta^{\rm Tar}$ are only periodically reset to estimates of $\pmb \theta$, say every $B$ training iterations of the DQN.  
Consider now the temporal-difference loss for some randomly drawn experience $e (\tau')$ from $\mathcal{R}(\tau)$ at interval $\tau$ 
\begin{align}
\label{eq:minilosstar}
&{\mathcal{L}^{\rm Tar}}(\pmb \theta_\tau; e(\tau')) :=\frac{1}{ 2} \Big[{ c(\pmb s(\tau'-1),\pmb a(\tau'))}\notag \\
&+ \gamma  \!\min_{\pmb a'}  Q^{\rm Tar} (\pmb s(\tau), \pmb a';\pmb \theta^{\rm Tar}_{\tau'}) - Q(\pmb s(\tau'-1),\pmb a(\tau'); \pmb \theta_{\tau})\Big]^2.
\end{align}
Upon taking expectation with respect to all sources of randomness generating this experience, we arrive at 
\begin{equation}
\label{eq:minilosstar}
{\mathcal{L}^{\rm Tar}}(\pmb \theta_\tau; \mathcal R(\tau))) :=  \mathbb{E}_{e(\tau')} \, \mathcal{L}^{\rm Tar}(\pmb \theta_\tau;e(\tau')).
\end{equation}
In practice however, the underlying transition probabilities are unknown, which challenges evaluating and hence minimizing ${\mathcal{L}^{\rm Tar}}(\pmb \theta_\tau; \mathcal R(\tau))) $ exactly. A commonly adopted alternative is to approximate the expected loss with an empirical loss over a few samples (that is, experiences here). %${\mathcal{L}_{\mathcal N_{b}}^{\rm Tar}}(\pmb \theta_T)$ 
%of say a mini-batch size $N_{b}$ of experiences. 
To this end, we draw a mini-batch of $M_\tau$ experiences uniformly at random from the replay buffer $\mathcal{R}(\tau)$, whose indices are collected in the set $\mathcal{M}_\tau$, i.e., $\{ e(\tau')\}_{\tau' \in \mathcal{M}_\tau}\sim U(\mathcal R(\tau))$.
Upon computing for each of those sampled experiences an output using the target network with parameters $\pmb \theta_\tau^{\rm Tar}$, the empirical loss is 
\begin{align}
\label{eq:minilosstar2}
&{\mathcal{L}^{\rm Tar}}(\pmb \theta_{\tau};\mathcal {M}_\tau) := \frac{1}{ 2M_\tau}\!\sum_{\tau' \in \mathcal{M}_\tau} \Big[{ c(\pmb s(\tau'-1),\pmb a(\tau'))}\notag \\
&+\gamma\! \min_{\pmb a'}  Q^{\rm Tar} (\pmb s(\tau'), \pmb a';\pmb \theta^{\rm Tar}_{\tau})  - Q(\pmb s(\tau'-1), \pmb a(\tau'); \pmb \theta_{\tau})  \Big]^2.
\end{align}

In a nutshell, the weight parameter vector $\pmb \theta_{\tau}$ of the DQN is efficiently updated `on-the-fly' using SGD over the empirical loss ${\mathcal{L}^{\rm Tar}}(\pmb \theta_{\tau};\mathcal {M}_\tau)$, with iterates given by
\begin{equation}
\label{eq:miniSGD}
\pmb \theta_{\tau+1} = \pmb \theta_{\tau} - \beta_\tau \nabla {\mathcal{L}^{\rm Tar}}(\pmb \theta_{\tau};\mathcal {M}_\tau).
\end{equation}

%The \textit{target network} and \textit{experience replay} heuristics indeed help enhance the stability of DQN updates. 
Incorporating \textit{target network} and \textit{experience replay} remedies for stable DRL, our proposed two-timescale voltage regulation scheme is summarized in Alg.~\ref{Alg_a}. 
%	{ 	Algorithm 1 is implemented in an online fashion. 	Per fast-timescale slot, Algorithm 1 solves a linearly constrained quadratic program, whose worst-case computational complexity is $\mathcal{O}(N^3)$ (here $N$ is the number of buses). The complexity increases to $\mathcal{O}(N^4)$ when the exact AC model is used. Per slow-timescale interval, performing a single mini-batch gradient descent update for a feedforward neural network having $d$ unknown weight parameters over a mini-batch $\mathcal{M}$ of experiences, incurs a complexity of $\mathcal{O}(d N|\mathcal{M}|)$. }

\begin{figure}[t]
	\centering
	\includegraphics[width =0.6\textwidth]{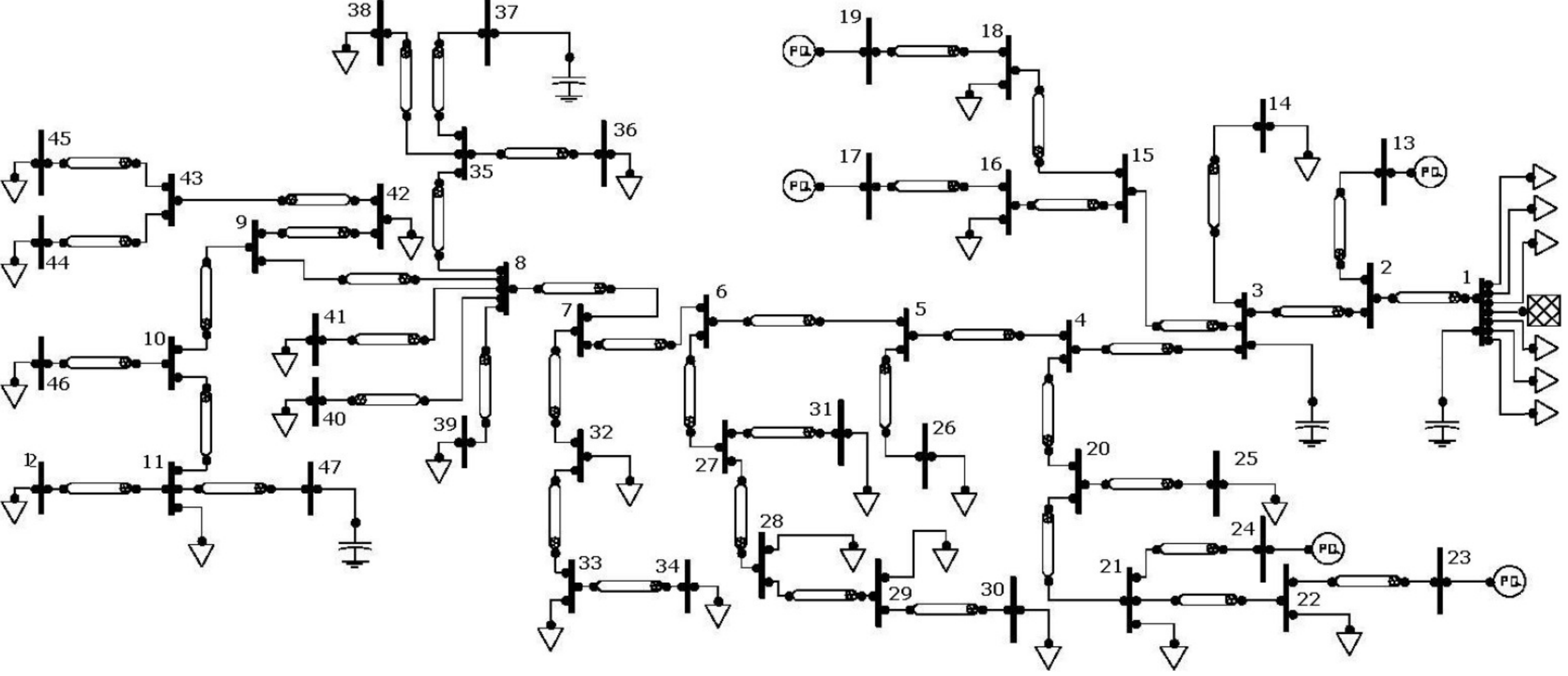}
	\caption{Schematic diagram of the $47$-bus industrial distribution feeder. Bus $1$ is the substation, and the $6$ loads connected to it model other feeders on this substation.
		%		 Buses $1, 3, 37$, and $47$ are equipped with shunt capacitors, while buses $2, 16, 18, 21$, and $22$ are equipped with inverters.
	}
	\label{fig:distributiongrid}
\end{figure}

\begin{figure}
	\centering
	\includegraphics[scale=.5]{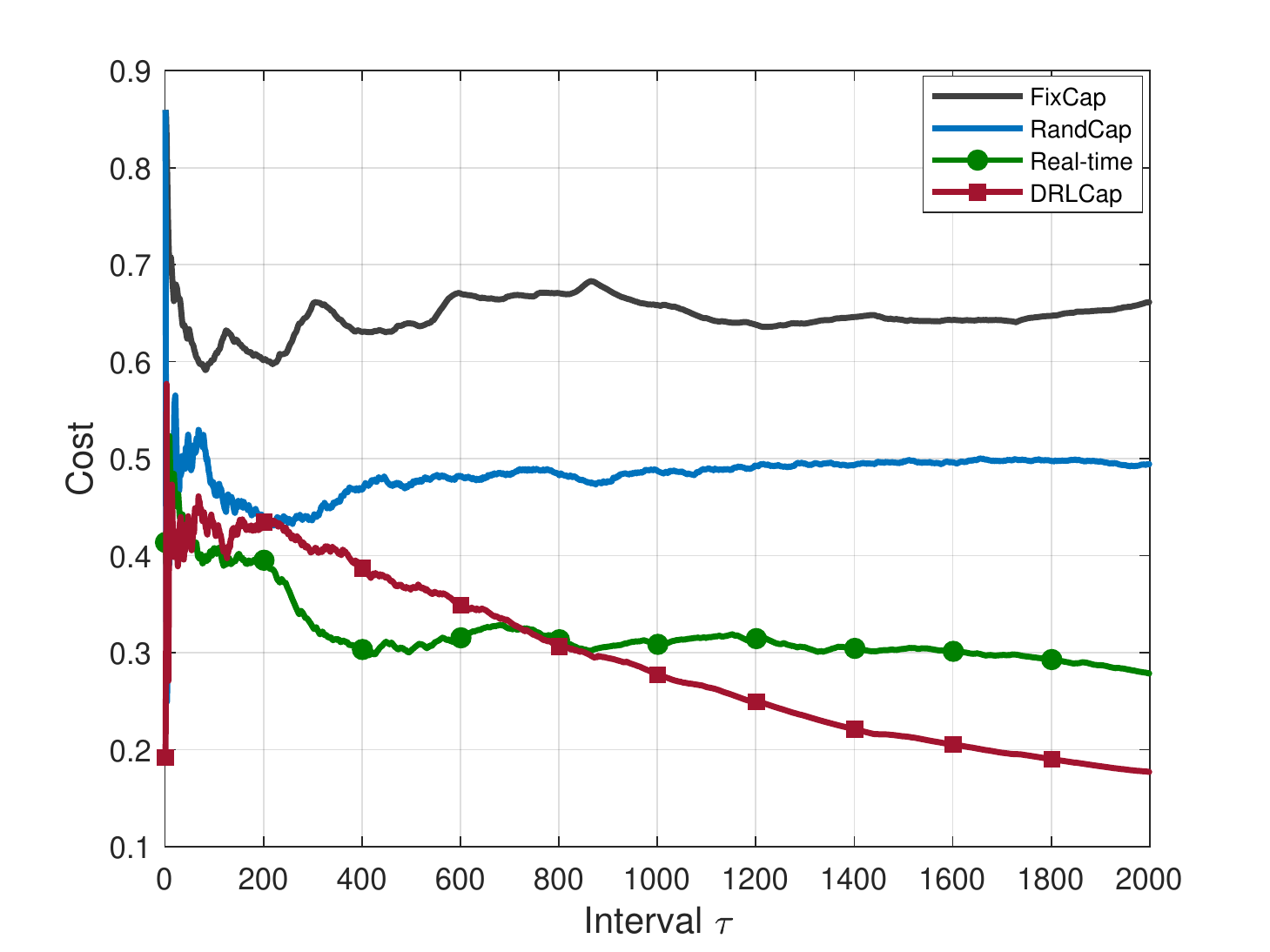}
	\caption{Time-averaged instantaneous costs incurred by the four voltage control schemes.}
	% under the linearized power flow model.
	\label{fig:bus47cap3cost}
\end{figure}

\begin{figure}
	\centering
	\includegraphics[scale=.63]{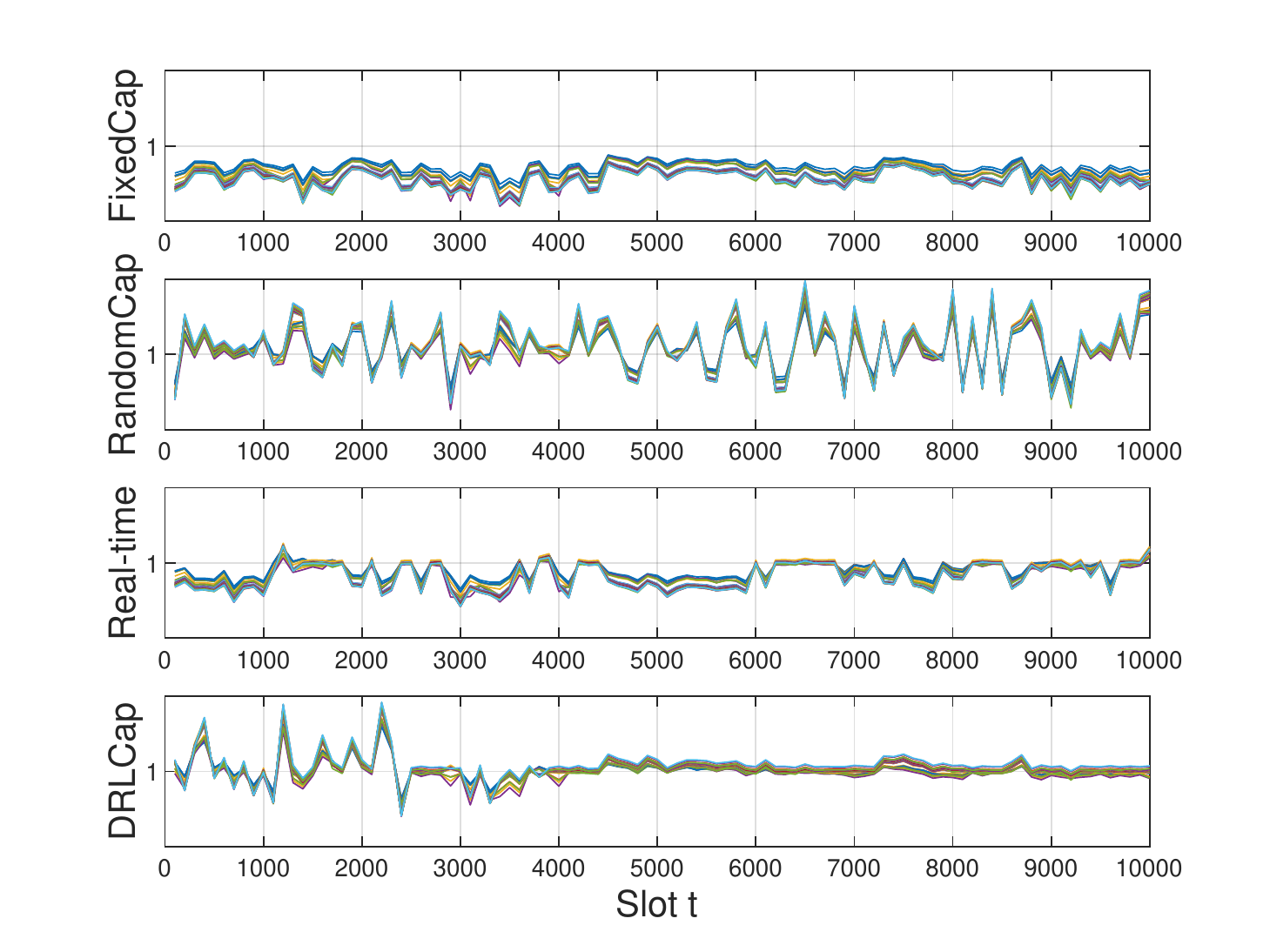}
	\caption{{ Voltage magnitude profiles obtained by the four voltage control schemes over the simulation period of $10,000$ slots.
			%			under the linearized power flow model.
		}}
		\label{fig:bus47cap3volt}
	\end{figure}
	
	\begin{figure}
		\centering
		\includegraphics[scale=.63]{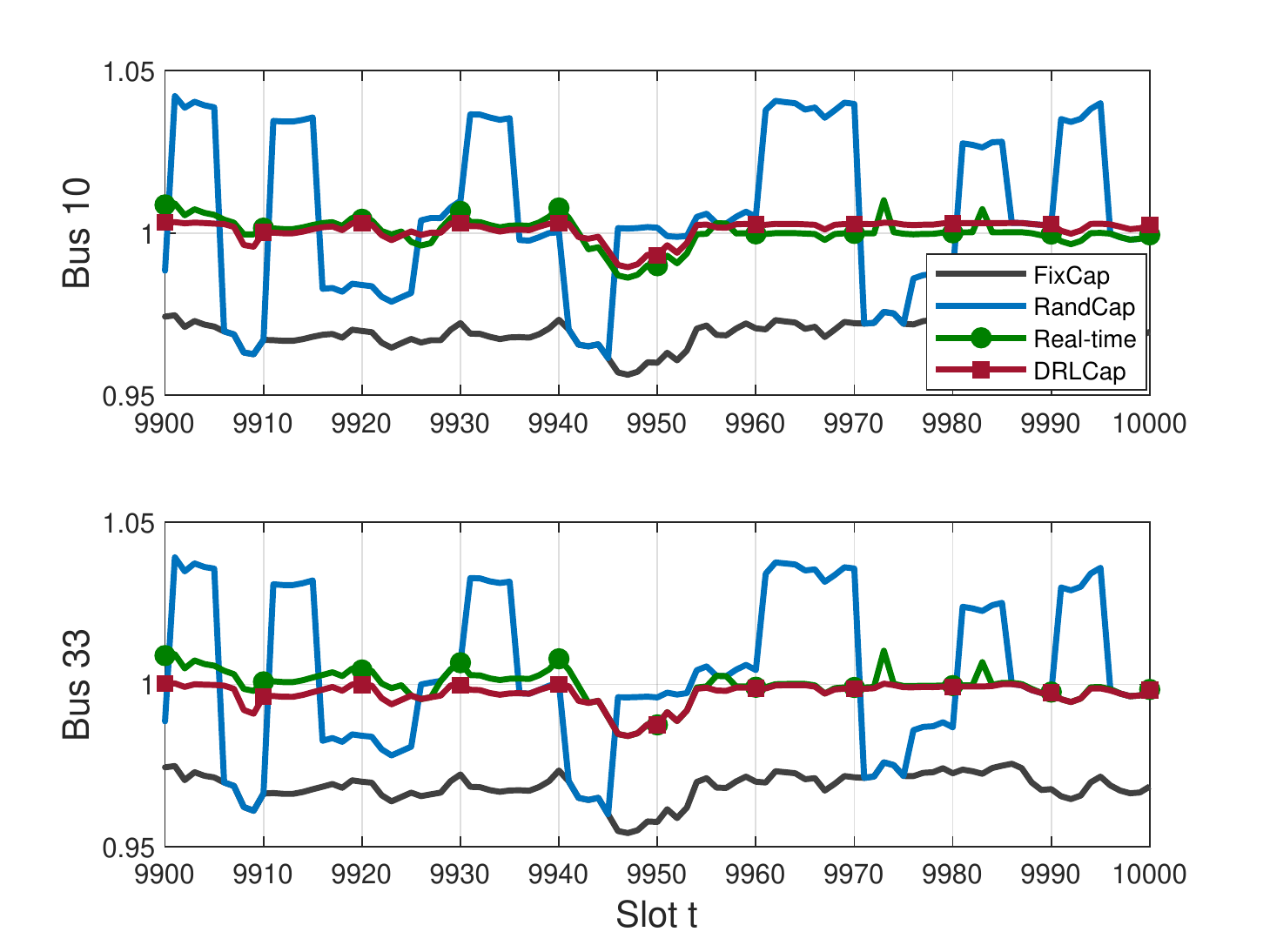}
		\caption{Voltage magnitude profiles obtained by the four voltage control schemes at buses $10$ and $33$ from slot $9,900$ to $10,000$.}
		\label{fig:bus47cap3volttwo}
	\end{figure}
	
	\begin{figure}
		\centering
		\includegraphics[scale=.63]{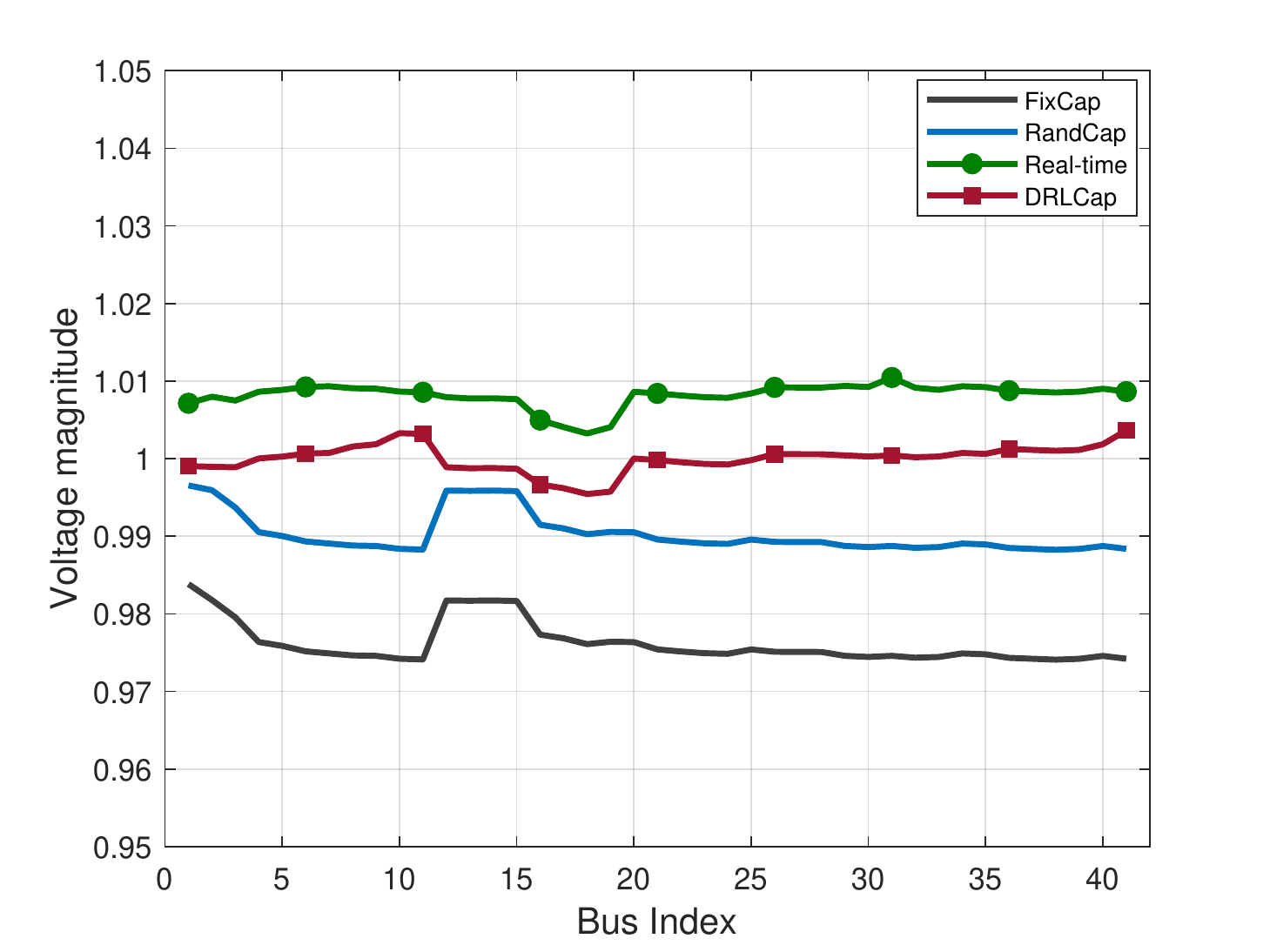}
		\caption{Voltage magnitude profiles at all buses at slot $9,900$ obtained by the four voltage control schemes. 
			%			under the linearized power flow model.
		}
		\label{fig:bus47cap3voltall}
	\end{figure}

	%%%%%%%%%%%%%%%%%%%%%%%%%%%%%%%%%%%%%%%%%%%%%%%%%%%%%%%%%%%%%%%%%%%%%%% 
	\section{Numerical Tests} \label{sec:test}

	In this section, numerical tests on a real-world $47$-bus distribution feeder as well as the IEEE $123$-bus benchmark system are provided to showcase the performance of our proposed DRL-based voltage control scheme (cf. presented in Alg. \ref{Alg_a}). As has already been shown in previous works (e.g., \cite{kekatos2015stochastic,kekatos2016voltage,low2014convex}), the linearized distribution flow model approximates the exact AC model very well; hence, numerical results based on the linearized model were only reported here.  
	
	The first experiment entails the Southern California Edison $47$-bus distribution feeder \cite{farivar2011inverter}, which is depicted in Fig.~\ref{fig:distributiongrid}. This feeder is integrated with four shunt capacitors
	as well as five smart inverters. As the voltage magnitude $v_0$ of the substation bus is regulated to be a constant ($1$ in all our tests) through a voltage transformer, the capacitor at the substation was excluded from our control. Thus, a total of three shunt capacitors along with five smart inverters embedded with large PV plants were engaged in voltage regulation. 
	The rest three capacitors are installed on buses $3$, $37$, and $47$, with capacities $120$, $180$, and $180$ kVar, respectively, while the five large PV plants are located on buses $2$, $16$, $18$, $21$, and $22$, with capacities $300$, $80$, $300$, $400$, and $200$ kW, respectively. 
	To test our scheme in a realistic setting, 
	real consumption as well as solar generation data were obtained from the Smart${^*}$ project collected on August $24, 2011$ \cite{barker2012smart}, which were first preprocessed by following the procedure described in our precursor work \cite{kekatos2015stochastic}.
	%In this section, we apply the proposed two-timescale voltage regulation method to the linear approximation and the full AC model respectively to illustrate its validation. 
	%The novel scheme was tested on a 47-bus feeder with high penetration of renewables~\cite{farivar2011inverter}; see Fig.~\ref{fig:distributiongrid}. The real consumption and solar generation data were obtained from the Smart${^*}$ project collected on August 24, 2011 as delineated next~\cite{barker2012smart}, and preprocessed by following the procedure given in \cite{kekatos2015stochastic}.

	In our tests, to match the availability of real data, each slot $t$ was set to a minute, and each interval $\tau$ was set to five minutes. A power factor of $0.8$ was assumed for all loads.
	The DQN used here consists of three fully connected layers, which has $44$ and $12$ units in the first and second hidden layers, respectively. Although simple, it was found sufficient for the task at hand. ReLU activation functions ($\sigma(x)=\max(x,0)$) were employed in the hidden layers, and logistic sigmoid functions $s(x)=1/(1+e^{-x})$ were used at the output layer. 
	To assess the performance of our proposed scheme, we have simulated three capacitor configuration policies as baselines, that include a fixed capacitor configuration (FixCap), a random capacitor configuration (RandCap), and an (impractical) `real-time' policy. Specifically, the FixCap uses a fixed capacitor configuration throughout, and the RandCap implements random actions to configure the capacitors on every slow time interval; both of which compute the inverter setpoints by solving  \eqref{eq:linear} per slot $t$. The impractical Real-time scheme however, optimizes over inverters and capacitors on a single-timescale, namely at every slot -- hence justifying its `real-time' characterization. To carry out this optimization task, first the binary constraints $y_{k_i}(t)\in \{0,1\}$ are relaxed to box ones $y_{k_i}(t)\in [0,1]$, the resulting convex program is solved using an off-the-shelf routine \cite{cvx}, which is followed by a standard rounding step to recover binary solutions for capacitor configurations \cite{tpd1989wu}.	 
	%		the standard policy is a `static' optimization scheme to joint control both capacitors and inverters simultaneously through  minimizing the instantaneous cost every fast timescale by solving a mixed-integer programming problem. 
	%We also implemented a fixed capacitor configuration policy as well as a randomly switching policy as baselines. As in our proposed approach, 
	%In contrast, both fixed and random capacitor configuration policies first compute the optimal setpoints for inverters by solving \eqref{eq:nonlinear} or \eqref{eq:linear} per fast timescale, then the former employs a fixed capacitor configuration throughout the experiment, while the latter switches its capacitor configuration randomly every slow timescale. }
	
	%\subsection{Tests using the linearized power flow model} 

	%We first illustrated the proposed two-timescale voltage regulation algorithm on a linearized power flow model to get some insights into this algorithm.
	In the first experiment, the DRL-based capacitor configuration (DRLCap) voltage control approach was examined. 
	%	using the linearized power flow model.
	The replay buffer size was set to $R = 10$, the discount factor $\gamma = 0.99$, the mini-batch size $ M_\tau =10$, and the exploration-exploitation parameter $\epsilon_\tau    = {\rm max} \big \{ 1 - 0.1  \times \lfloor \tau / 50 \rfloor, \, 0\big\}.$
	During training, the target network was updated every $B=5$ iterations.
	The time-averaged instantaneous costs $$\frac{1}{\tau}\sum_{i=1}^\tau c(\pmb{s}(i-1),\pmb{a}(i))$$  
	incurred by the four schemes over the first $1\le \tau\le 2,000$ intervals are plotted in Fig.~\ref{fig:bus47cap3cost}. Evidently, the proposed scheme attains a lower cost than FixCap, RandCap, and Real-time after a short period of learning and interacting with the environment. 
	Even though the real-time scheme optimizes both capacitor configurations and inverter setpoints per slot $t$, its suboptimal performance in this case arises from the gap between the convexified problem and the original nonconvex counterpart.
	{  Fig.~\ref{fig:bus47cap3volt} presents the voltage magnitude profiles for all buses regulated by the four schemes  sampled at every $100$ slots}. Again, after a short period ($\sim$\,$4,500$ slots) of training through interacting with the environment, our DRLCap voltage control scheme quickly learns a stable and (near-) optimal policy. 
	In addition, voltage magnitude profiles regulated by FixCap, RandCap, Real-time, and DRLCap at buses $10$ and $33$ from slot $9,900$ to $10,000$ are shown in Fig.~\ref{fig:bus47cap3volttwo}, while the voltage magnitude profiles at
	all buses at slot $9,900$ are presented in Fig.~\ref{fig:bus47cap3voltall}. 
	Curves showcase the effectiveness of our DRLCap scheme in smoothing voltage fluctuations incurred due to large solar generation as well as heavy load demand.

	%our DRL-method effectively regulates voltages despite the despite the unfair comparison between the standard method which changes capacitor configurations per slot and DRL-based scheme which configures capacitors only once throughout an interval, our method more effectively regulates voltage upon converging to a (near-) optimal policy. }

	%To further corroborate the performance of our scheme, the previous experiment was replicated by simulating the exact AC grid model. 
	%%\subsection{Tests using the alternating current grid model }
	%%Fig.~\ref{fig:errorn} represents the convergence of the DQN weight parameter $\pmb \theta_\tau$ to that of the target network one $\pmb \theta^{\rm Tar}_\tau$. 
	%Figure~\ref{fig:bus47cap3costn} depicts the time-averaged instantaneous costs incurred by the \textcolor{blue}{four} simulated schemes, over the first $2,000$ intervals. Curves again show that the \textcolor{blue}{DRLCap results in smaller voltage deviations than its competing alternatives.}
	%%The corresponding actions taken are shown in Fig.~\ref{fig:actionn}. 
	%\textcolor{blue}{The voltage magnitude profiles %found by the four  schemes 
	%at bus $10$ and bus $33$ from slot $9900$ to $10000$ are plotted in Fig.~\ref{fig:bus47cap3volttwon}. Fig.~\ref{fig:bus47cap3voltalln} depicts the voltage magnitude profiles at all buses  at slot $9900$.
	%Both Fig.~\ref{fig:bus47cap3volttwon} and Fig.~\ref{fig:bus47cap3voltalln} corroborate the merits of our two-timescale DRLCap scheme in real-world settings. }
	
	\begin{figure}
		\centering
		\includegraphics[width =0.45 \textwidth]{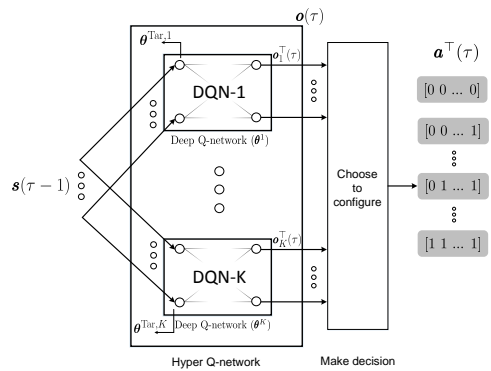}
		\caption{Hyper deep $Q$-network for capacitor configuration.}
		\label{fig:H_DQN}
	\end{figure}
	
	\begin{figure}[t]
		\centering
		\includegraphics[scale = .45 ]{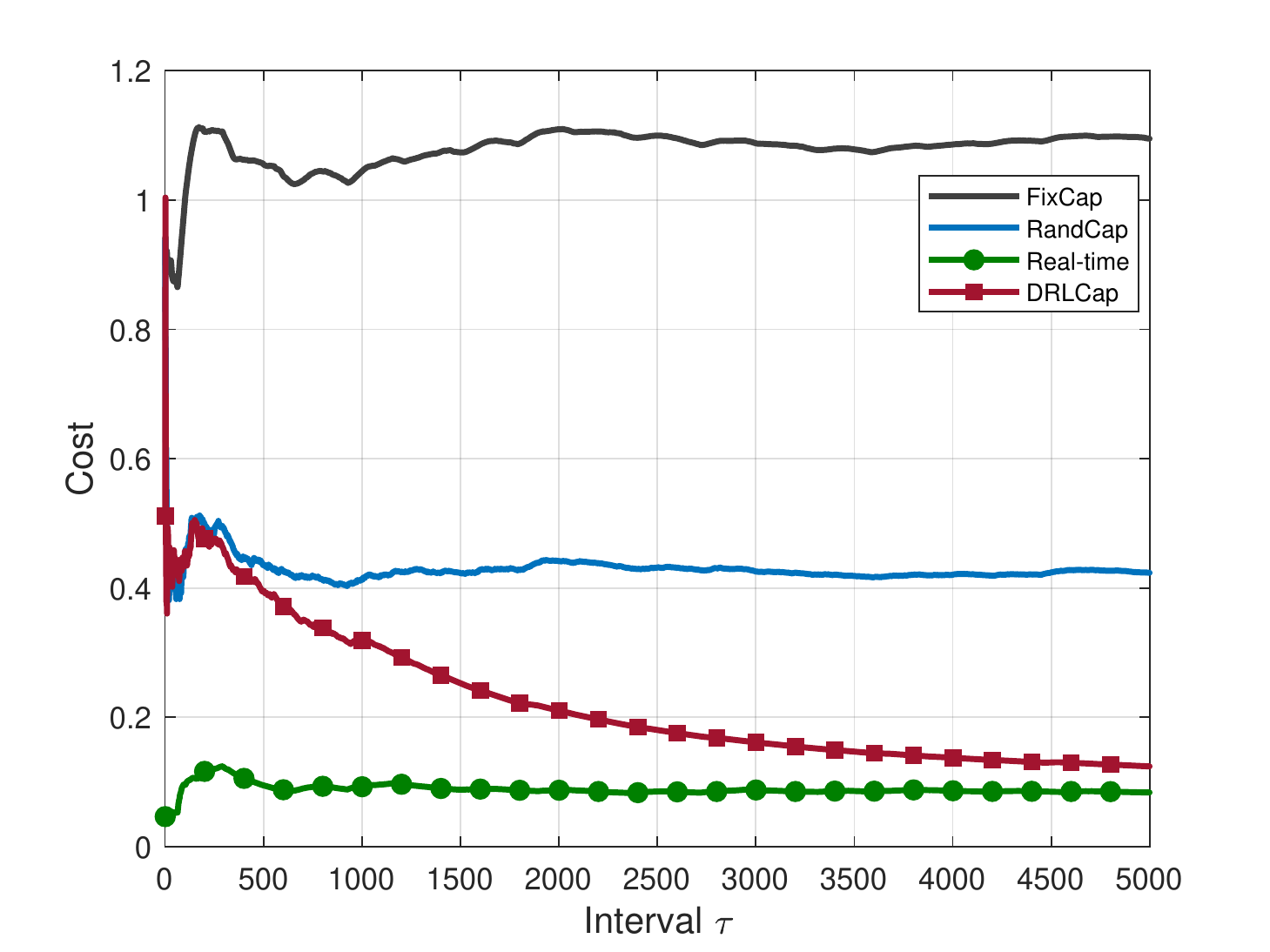}
		\caption{{  Time-averaged instantaneous costs incurred by the four approaches 
				%			with $N_a = 8$ capacitors 
				on the IEEE $123$-bus feeder.}}
		\label{fig:bus123cap8cost}
	\end{figure}
	
	\begin{figure}
		\centering
		\includegraphics[scale=.45]{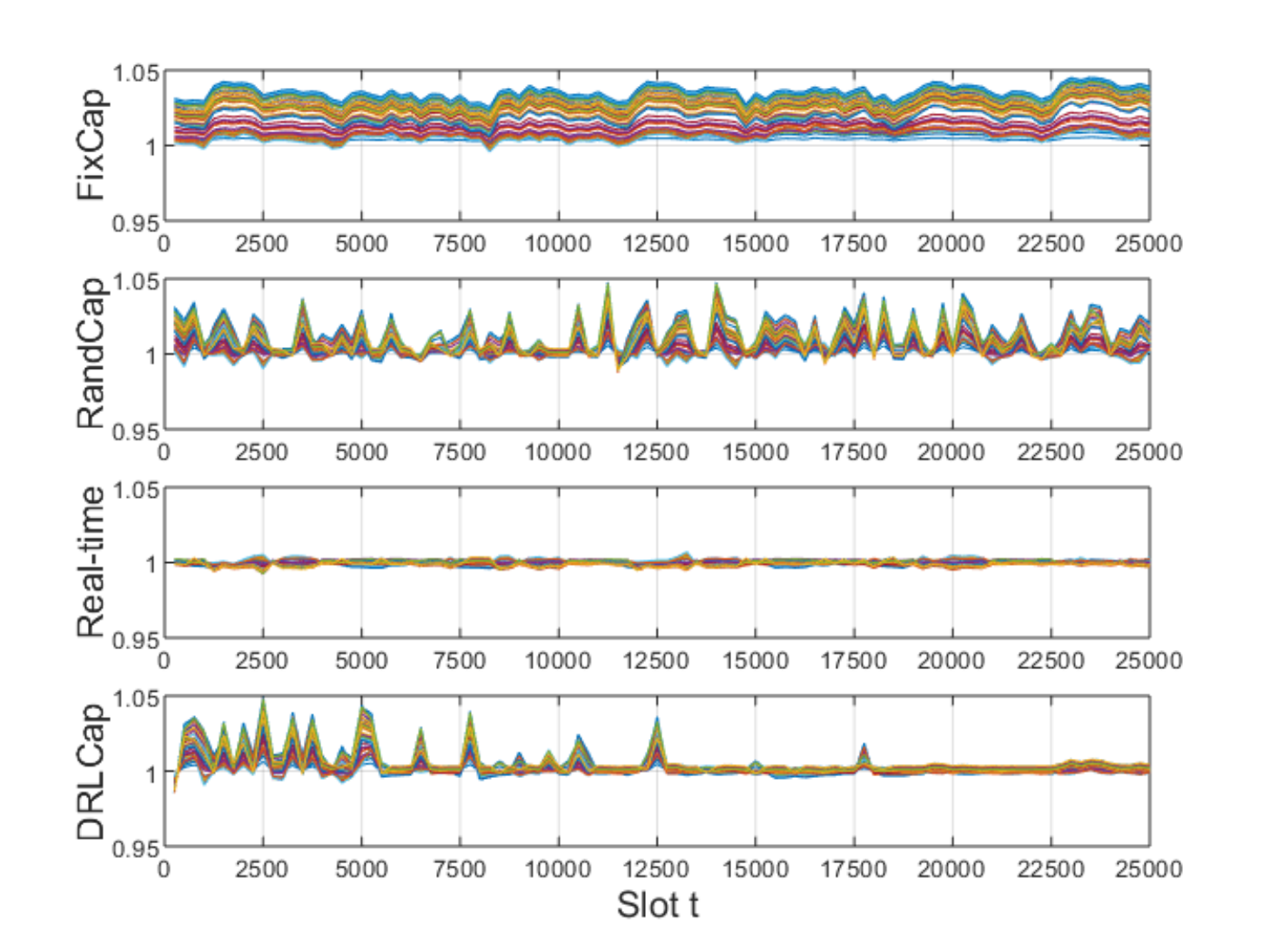}
		\caption{{ Voltage magnitude profiles at all buses over the simulation period of  $25,000$ slots on the IEEE $123$-bus feeder.}
		}
		\label{fig:bus123cap8volt}
	\end{figure}
	
	\begin{figure}[t]
		\centering
		\includegraphics[scale = .45]{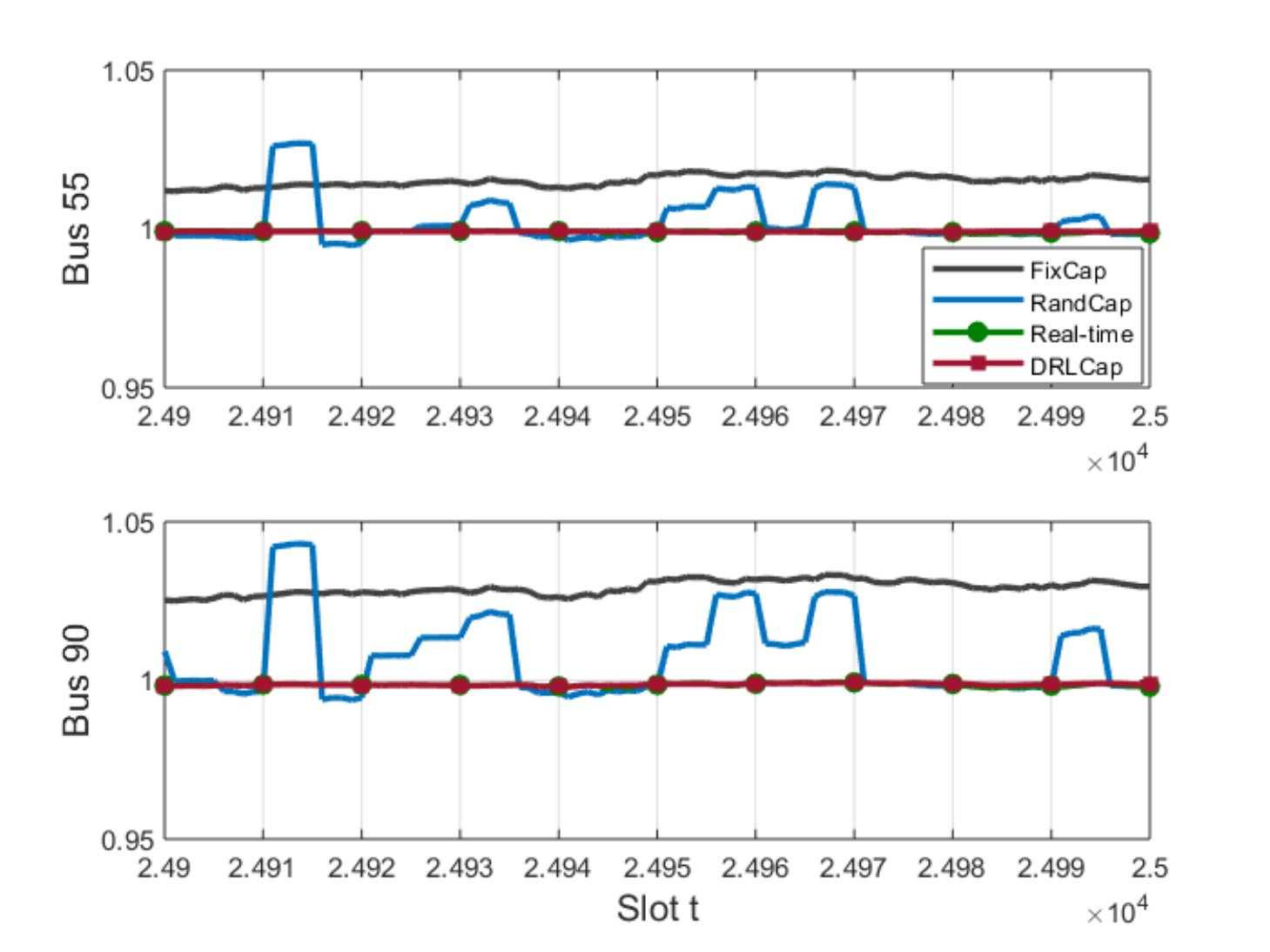}
		\caption{{ Voltage magnitude profiles at buses $55$ and  $90$ from slot $24,900$ to $25,000$ obtained by the four approaches 
				%			with $N_a = 8$ capacitors
				on the IEEE $123$-bus feeder.}}
		\label{fig:bus123cap8volttwo}
	\end{figure}

	\begin{figure}[t]
		\centering
		\includegraphics[scale = .45]{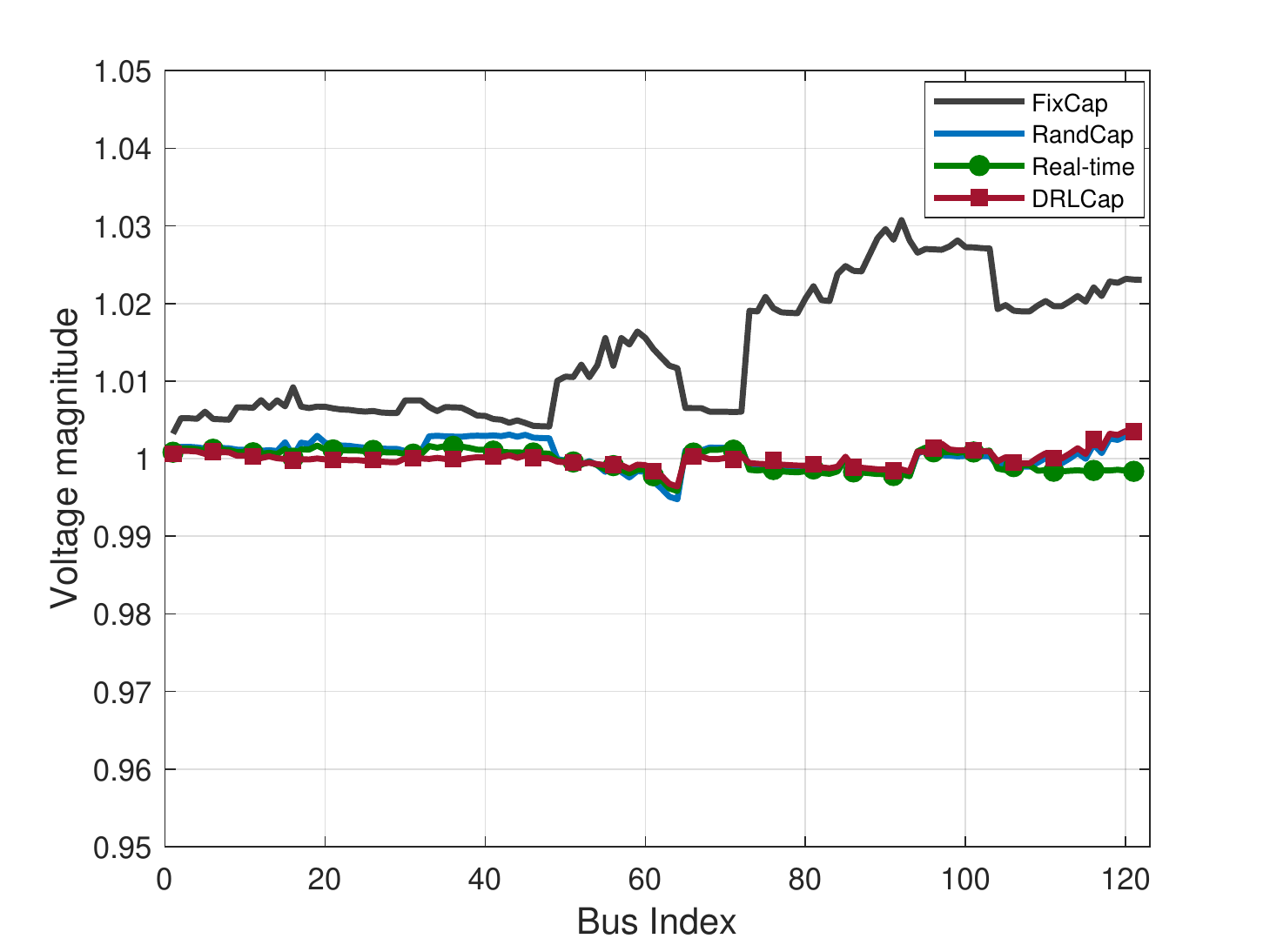}
		\caption{{ Voltage magnitude profiles at all buses on slot $24,900$ obtained by four approaches 
				%			with $N_a = 8$ capacitors 
				on the IEEE $123$-bus feeder.}}
		\label{fig:bus123cap8voltall}
	\end{figure}	
	
	To deal with distribution systems having a moderately large number of capacitors, we further advocate a hyper deep $Q$-network implementation, that endows our DRL-based scheme with scalability. The idea here is to first split the total number $2^{N_a}$ of $Q$-value predictions $\pmb o(\tau)\in\mathbb{R}^{2^{N_a}}$ at the output layer into $K$ smaller groups, each of which is of the same size $2^{N_a}/K$ and is to be predicted by a small-size DQN. This evidently yields the representation $\pmb o(\tau) := [{\pmb o}_1^{\top}(\tau), \ldots, {\pmb o}_K^{\top}(\tau)]^\top \!\!$, where ${\pmb o}_k(\tau) \in {\mathbb R}^{2^{N_a}/K}\!\!$ for $k = 1, \ldots, K$. By running $K$ DQNs in parallel along with their corresponding target networks, each DQN-$k$ generates  predicted $Q$-values $\pmb o_k(\tau)$ for the subset of actions corresponding to $k$th group. Note that all DQNs are fed with the same state vector $\pmb s(\tau-1)$; see also Fig. \ref{fig:H_DQN} for an illustration. 
	% 	Upon gathering all the outputs from the $K$ DQNs together, one obtains $\pmb o(\tau)$ as the predicted Q-values of all actions, and then find action $\pmb a(\tau)$ based on the discussions provided in subsection \ref{subsec:DRL}. 
	
	To examine the scalability and performance of this hyper $Q$-network implementation, additional tests using the IEEE $123$-bus test feeder with $9$ shunt capacitors were performed.
	Again, the capacitor at bus $1$ was excluded from the control, rendering a total number of $2^8=256$ actions (capacitor configurations).
	Renewable (PV) units are located on buses $47$, $49$, $63$, $73$, $104$, $108$, $113$, with capacities $100$, $16$, $70$, $20$, $20$, $30$, and $10$ k, respectively.   
	The $8$ shunt capacitors are installed on buses $3$, $20$, $44$, $93$, $96$, $98$, $100$, and $114$, with capacities $50$, $80$, $100$, $100$, $100$, $100$, $100$, and $60$ kVar.
	In this experiment, we used a total of $K = 64$ equal-sized DQNs to form the hyper $Q$-network, where each DQN implemented a fully connected $3$-layer feed-forward neural network, with ReLU activation functions in the hidden layers, and sigmoid functions at the output. 
	The replay buffer size was set to $R = 50$, the batch size to $M_\tau=8$, and the target network updating period to $B=10$. 
	{ The time-averaged instantaneous costs obtained over a simulation period of $5,000$ intervals 
		%using the linearized power flow model 
		is plotted in Fig.~\ref{fig:bus123cap8cost}. 
		Moreover, voltage magnitude profiles of all buses over the simulation period of $25,000$ slots sampled at every $100$ slots under the four schemes are plotted in Fig.~\ref{fig:bus123cap8volt};
		voltage magnitude profiles at buses $55$ and $90$ from slot $24,900$ to $25,000$ are shown in Fig.~\ref{fig:bus123cap8volttwo}; and, voltage magnitude profiles at all buses on slot $24,900$ are depicted in  \ref{fig:bus123cap8voltall}. } 
	Evidently, the hyper deep $Q$-network based DRL scheme smooths out the voltage fluctuations after a certain period ($\sim 7,000$ slots) of learning, while effectively handling the curse of dimensionality in the control (action) space. Evidently from Figs. \ref{fig:bus123cap8cost} and \ref{fig:bus123cap8voltall}, both the time-averaged immediate cost as well as the voltage profiles of DRLCap converge to those of the impractical `real-time' scheme (which jointly optimizes  inverter setpoints and capacitor configurations per slot). 
	% Comparing these plots with Fig. \ref{fig:bus47cap8volt} also suggest increasing the number of capacitors can further smooth out voltage fluctuations. 

	%Due to space limitations, simulation results on synthetic data are not illustrated here.
	%\qiu{It can be shown that this method can significantly protect the grids from straining by voltage fluctuations due to renewable generation. Due to space limitations, simulation results on synthetic data are not illustrated here.}
	
	%%%%%%%%%%%%%%%%%%%%%%%%%%%%%%%%%%%%%%%%%%%%%%%%%%%%%%%%%%%%%%%%%%%%%%%
	\section{Conclusions}
	\label{sec:conc}
	In this section, joint control of traditional utility-owned equipment and contemporary smart inverters for voltage regulation through reactive power provision was investigated. To account for the different response times of those assets, a two-timescale approach to minimizing bus voltage deviations from their nominal values was put forth, by combining physics- and data-driven stochastic optimization. Load consumption and active power generation dynamics were modeled as MDPs. On a fast timescale, the setpoints of smart inverters were found by minimizing the instantaneous bus voltage deviations, while on a slower timescale, the capacitor banks were configured to minimize the long-term expected voltage deviations using a deep reinforcement learning algorithm. The developed two-timescale voltage regulation scheme was found efficient and easy to implement in practice, through extensive numerical tests on real-world distribution systems using real solar and consumption data. This work also opens up several interesting directions for future research, including deep reinforcement learning for real-time optimal power flow as well as unit commitment.

\section{Gauss-Newton Unrolled Neural Networks and \\Data-driven Priors for Regularized PSSE with Robustness}

\section{Introduction} \label{Sec:Intro}
In today's smart grid, reliability and accuracy of state estimation 
are central for several system control and optimization tasks, including optimal power flow, unit commitment, economic dispatch, and contingency analysis \cite{pssebook2004}.
However, frequent and sizable state variable fluctuations caused by fast variations of renewable generation, increasing deployment of electric vehicles, and human-in-the-loop demand response incentives, are challenging these functions. 

As state variables are difficult to measure directly, the supervisory control and data acquisition (SCADA) system  offers abundant measurements, including voltage magnitudes, power flows, and power injections. Given SCADA measurements, the goal of PSSE is to retrieve the state variables, namely complex voltages at all buses \cite{pssebook2004}.
{PSSE is typically formulated as a  (weighted) least-squares (WLS) or a (weighted) least-absolute-value (WLAV)  problem. The former can be underdetermined, and nonconvex in general \cite{wang2019overview}, while the latter can be formulated as a linear programming problem \cite{gol2014lav, li2020wlav} if network only consists of PMUs. In practice however, power grids must include conventional RTUs as well, which leads to highly complex and non-convex solution.}

To address these challenges, several efforts have been devoted. WLAV-based estimation for instance can be converted into a constrained optimization, for which a sequential linear programming solver was devised in \cite{jabr2003iteratively}, and improved (stochastic) proximal-linear solvers were developed in \cite{gang2019tsg}. On the other hand, focusing on the WLS criterion, the Gauss-Newton solver is widely employed in practice \cite{pssebook2004}. Unfortunately, due to the nonconvexity and quadratic loss function, there are two challenges facing the Gauss-Newton solver: i) sensitivity to initialization; and ii) convergence is generally not guaranteed~\cite{zhu2014power}. Semidefinite programming approaches can mitigate these issues to some extent, at the price of rather heavy computational burden \cite{zhu2014power}. In a nutshell, the grand challenge of these methods, remains to develop fast and robust PSSE solvers attaining or approximating the global optimum.    

To bypass the nonconvex optimization hurdle in power system monitoring and control, recent works have focused on developing data- (and model-) driven  neural network (NN) solutions  \cite{barbeiro2014state,tps2012nnpsse,liang2019tsgpsse,zamzam2019gcn, hu2020physics,tps20langtong,ostrometzky2019physics}.
%A neural network (NN) was trained using historical data to obtain a `smart' initialization for Gauss-Newton iterations in \cite{zamzam2019datadriventps}.
Such NN-based PSSE solvers approximate the mapping from measurements to state variables based on a training set of measurement-state pairs generated using simulators or available from historical data \cite{liang2019tsgpsse}. However, existing NN architectures do not directly account for the power network topology. On the other hand, a common approach to tackling challenging ill-posed problems in image processing has been to regularize the loss function with suitable priors \cite{rudin1992nonlinear}. Popular priors include sparsity, total variation, and low rank \cite{bookregularization}. Recent efforts have also focused on data-driven priors that can be learned from exemplary data \cite{learningprior2013mri,schlemper2017deep, modl2018tmi}.  

Permeating the benefits of \cite{learningprior2013mri,schlemper2017deep} and \cite{modl2018tmi} to power systems, this paper advocates a deep (D) NN-based trainable prior for standard ill-posed PSSE, to promote physically meaningful PSSE solutions. To tackle the resulting regularized PSSE problem, an alternating minimization-based solver is first developed, having Gauss-Newton iterations as a critical algorithmic component. As with Gauss-Newton iterations, our solver requires inverting a matrix per iteration, thus incurring a heavy computational load that may discourage its use for real-time monitoring of large networks. To accommodate real-time operations and building on our previous works \cite{liang2019tsgpsse}, we unroll this alternating minimization solver to construct a new DNN architecture, that we term Gauss-Newton unrolled neural networks  (GNU-NN) with deep priors. As the name suggests, our DNN model consists of a Gauss-Newton iteration as a basic building block, followed by a proximal step to account for the regularization term.  Upon incorporating a graph (G) NN-based prior, our model exploits the structure of the underlying power network. Different from \cite{liang2019tsgpsse}, our GNU-NN  method offers a systematic and flexible framework to incorporate prior information into standard PSSE tasks. 

In practice, measurements collected by the SCADA system may be severely corrupted due to e.g., parameter uncertainty, instrument mis-calibration, and unmonitored topology changes \cite{bad1971tpas, gang2019tsg}. As cyber-physical systems, power networks are also vulnerable to adversarial attacks 
%due to that they are designed without paying enough attention to security 
\cite{fairley2016spectrum,wu2019optimal}, as asserted by the first hacker-caused Ukraine power blackout in $2015$ \cite{ukraincase2016analysis}. Furthermore,  it has recently been demonstrated that adversarial attacks can markedly deteriorate NNs' performance \cite{adv2016kurakin,miller2019adversarial}. Prompted by this, to endow our GNU-NN approach with \textit{robustness} against bad (even adversarial) data, we pursue a principled GNU-NN training method that relies on a distributionally robust optimization formulation. Numerical tests using the IEEE $118$-bus benchmark system corroborate the performance and robustness of the proposed scheme.

%\emph{Outline.} 
%Regarding the remainder of the paper, Section~\ref{sec:prob} introduces the power system model, and formally states the PSSE problem. Section \ref{sec:unro} presents a general framework for incorporating data-driven and topology-aware priors into PSSE, along with an alternating minimization solver for the resultant regularized PSSE. Section \ref{sec:adve} develops an adversarial training method to robustify GNU-GNN against bad data. Numerical tests using the IEEE $118$-bus test feeder are provided in Section \ref{sec:test}, with concluding remarks drawn in Section \ref{sec:conc}. 

\emph{Notation.} Lower- (upper-) case boldface letters denote column vectors (matrices), with the exception of vectors $\bm{V}$, $\bm{P}$ and $\bm{Q}$, and normal letters represent scalars. The $(i,j)$th entry, $i$-th row, and $j$-th column of matrix $\bm X$ are $[\bm X]_{i,j}$, $[\bm X]_{i:}$, and $[\bm X]_{:j}$, respectively. Calligraphic letters are reserved for sets except operators $\mathcal{I}$ and $\mathcal{P}$. Symbol $^{\top}$ stands for transposition; 
$\bf 0$ denotes all-zero vectors of suitable dimensions; and $\| \pmb x \|$ is the $l_2$-norm of vector $\pmb x$.
% Symbol 
%%$\bf 1$ denotes all-one vectors; 
%$\bf 0$ denotes all-zero vectors; $^{\top}$ stands for transposition; and $\| \pmb x \|$ is the $l_2$-norm of~$\pmb x$.

	%%%%%%%%%%%%%%%%%%%%%%%%%%%%%%%%%%%%%%%%%%%%%%%%%%%%%%%%%%%%%%%%%%%%%%%
	\section{Background and Problem Formulation}\label{sec:prob}
	
	%\begin{figure}
	%	\centering
	%	\includegraphics[width =0.26 \textwidth]{powerflow.pdf}
	%	\caption{Bus $n$ and $n'$ are connected with line impedance $y_{nn'}$.}
	%	\label{fig:lineardiagram}
	%\end{figure}	
	
	Consider an electric grid comprising $N$ buses (nodes) with $E$ lines (edges) that can be modeled as a graph $\mathcal{G}:=(\mathcal{N},\mathcal{E}, \bm W)$, where the set $\mathcal{N}:=\{1,\ldots,N\}$ collects all buses, $\mathcal{E}:=\{(n, n')\} \subseteq \mathcal N \times \mathcal N$  all lines, and $\bm W \in \mathbb R^{N \times N}$ is a weight matrix with its $(n, n')$-th entry $[\bm W]_{nn'} = w_{nn'}$  modeling the impedance between buses $n$ and $n'$.
	%impedance between buses $n$ and $n'$. 
	%	 Among possible choices, we use a Gaussian kernel, that is $w_{nn'}=\exp(-k|y_{nn'}|^2)$, where $k$ is a scaling factor, and $y_{nn'}$ is the impedance between buses $n$ and $n'$. 
	In particular, if $(n,n') \in \mathcal E$, then $[\bm W]_{nn'} = w_{nn'}$; and $[\bm W]_{nn'} = 0$ otherwise. 
	%	whose buses are collected into $\mathcal{N}:=\{1,\ldots,N\}$, and lines into $\mathcal{E}:=\{(n, n')\} \subseteq \mathcal N \times \mathcal N$. 
	For each bus $n \in \mathcal N$, let $V_n := v^r_n + jv^i_n$ be its complex voltage with magnitude denoted by $|V_n|$, and $P_n + j Q_n$ its complex power injection. For reference, collect the voltage magnitudes, active and reactive power injections across all buses into the $N$-dimensional column vectors $|\bm  V|$, $\bm P$, and $\bm Q$, respectively.  
	%For each line $(n, n') \in \mathcal L$, let $P^f_{nn'}$ denote the active power flow seen at the `from' end, and $P^t_{nn'}$ the active power flow seen at the `terminal' end, likewise for reactive power flow $Q^f_{nn'}$ (from) and $Q^t_{nn'}$ (terminal); 
	
	\begin{figure*}[t!]
		\centering
		\includegraphics[width =1 \textwidth]{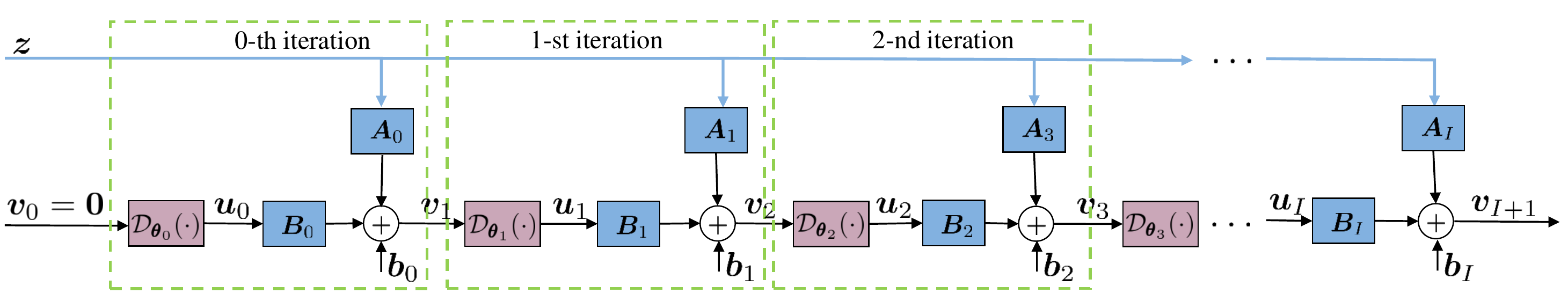}
		\caption{The structure of the proposed GNU-NN.}
		\label{fig:unrollGNN}
	\end{figure*}
	
	System state variables $\bm{v}:=[v^r_{1}~ v^i_{1} ~\dots  v^r_{N} ~ v^i_{N}]^\top \in \mathbb{R}^{2 N}$ can be represented by SCADA measurements, including voltage magnitudes, active and reactive power injections, as well as active and reactive power flows. Let $\mathcal S_V$, $ \mathcal S_P$, $\mathcal S_Q$, $\mathcal{E}_P$, and $\mathcal{E}_Q$ denote the sets of buses or lines where meters of corresponding type are installed. 
	%  us signify the smart meter locations of corresponding types by $\mathcal S_V$, $ \mathcal S_P$ and $\mathcal S_Q$, where %at time slot $t$,  
	%$M$ measurements are observed. 
	% let $\mathcal C_V$, $ \mathcal C_P$ and $\mathcal C_Q$ $\mathcal C^f_P$, $\mathcal C^t_Q$, $\mathcal C^f_P$ and $\mathcal C^t_Q$ denote the smart meter locations of the corresponding type. 
	For a compact representation, 
	let us collect the measurements from all meters into 
	$\bm{z}:=[\{|V_{n}|^2\}_{n \in \mathcal S_V}, \{P_{n}\}_{n \in \mathcal S_P}, \{Q_{n}\}_{n \in \mathcal S_Q}, \{P_{nn'}\}_{(n,n') \in \mathcal E_P},$ $ \{Q_{nn'}\}_{(n,n') \in \mathcal E_Q},]^\top\in \mathbb R^M$. 
	%For brevity, the time index $t$ is dropped.
	Moreover, the $m$-th entry of $\bm{z}:=\{z_{m}\}^M_{m=1}$, can be described by the following model 	
	\begin{equation}\label{eq:measurement}
	z_{m} =  h_{m} (\bm v) + \epsilon_{m}, ~~ \forall m = 1, \ldots, M	 
	\end{equation}
	where  $ h_m(\bm v)=\bm v ^\top \bm H_{m} \bm v $ for some symmetric measurement matrix $\bm H_m \in \mathbb R^{2N \times 2N}$, and $\epsilon_{m}$ captures the modeling error as well as the measurement noise. 
	
	The goal of PSSE is to recover the state vector $\bm v$ from measurements $\bm z$. Specifically, 
	%	for independent and identically distributed noise $\{\epsilon_m\}_{m=1}^M$,
	adopting the least-squares criterion and vectorizing the terms in \eqref{eq:measurement}, PSSE can be formulated as the following nonlinear least-squares (NLS) 
	%	As the  $\ell_{1}$-based losses yield median-based estimators \cite{huber2011robust}, they handle gross errors in the measurements $\bm{z}$ in a relatively benign way. Motivated by this, we consider here  minimizing the $\ell_{1}$ loss of the residuals, which gives the so-called least-absolute-value estimate \cite{lav1982tpas}
	%	\begin{equation}
	%	\min _{\bm v \in \mathbb R^{2N}} \mathbb{E}_{\bm {z} \sim P_{0}}|\bm z - \bm v^{\top} \bm H \bm v|
	%	\end{equation}	
	%	The nominal distribution $P_{0}$ is typically unknown, but instead samples $\left\{z_{n}\right\}_{n=1}^{N} \sim P_{0}$ are given. Therefore, empirical loss minimization is used in practice, where $P_0$ is replaced with empirical distribution $\widehat{P}_{N}$, resulting in following reformulation
	%	
	%	\begin{equation}
	%	\min _{\boldsymbol{\theta} \in \Theta} \mathbb{E}_{\boldsymbol{z} \sim \widehat{P}_{N}}[\ell(\boldsymbol{\theta} ; \boldsymbol{z})]+r(\boldsymbol{\theta})
	%	\end{equation}
	%	
	%	where $\mathbb{E}_{\boldsymbol{z} \sim \widehat{P}_{N}}[\ell(\boldsymbol{\theta} ; \boldsymbol{z})]=N^{-1} \sum_{n=1}^{N} \ell\left(\boldsymbol{\theta} ; \boldsymbol{z}_{n}\right)$
	\begin{equation}\label{eq:exploss}
	{\bm v}^\ast  :  = \arg \underset{\bm v \in \mathbb R^{2N}}{\min} ~ \|\bm{z} -  \bm h(\bm v) \|^2.
	\end{equation}
	A number of algorithms have been developed for solving \eqref{eq:exploss}, including e.g., Gauss-Newton iterations \cite{pssebook2004}, and semidefinite programming-based solvers  \cite{zhu2014power,lan2019fast}. Starting from an initial $\bm v_{0}$, most of these schemes (the former two) iteratively implement a mapping from $\bm v_i$ to $\bm v_{i+1}$, in order to generate a sequence of iterates that hopefully converges to $\bm v^{\ast}$ or some point nearby. In the ensuing subsection, we will focus on the `workhorse' Gauss-Newton PSSE solver. 
	%$\Gamma\left(x^{k}, A, y, \theta\right) \rightarrow x^{k+1}$ is 
	
	\subsection{Gauss-Newton Iterations}
	The Gauss-Newton method is the most commonly used one for minimizing NLS \cite[Sec. 1.5.1]{nolinear1999book}. It relies on Taylor's expansion to linearize the function $\bm h(\bm{v})$. Specifically, at a given point $\bm{v}_{i}$, it linearly approximates  
	\begin{equation}
	\label{eq:lin}
	\tilde{\bm{h}}(\bm{v}, \bm{v}_{i})\approx \bm{h}(\bm{v}_{i})+\bm{J}_{i}(\bm{v}-\bm{v}_{i})
	\end{equation}
	where $\bm{J}_i := \nabla \bm{h}\left(\bm{v}_i\right)$ is the $M \times 2N$ 
	Jacobian of $\bm h$ evaluated at $\bm v_i$, with $[\bm{J}_i]_{m,n} :=\partial \bm{h}_{m} / \partial \bm v_{n}$.
	%; see Wirtinger derivative \cite{kreutz2009complex} for details. %\cite{T. Lipp and S. Boyd, \Variations and extension of the convex{concaveprocedure," Optim. Eng., vol. 17, no. 2, pp. 263{287, Jun. 2016.}
	Subsequently, the Gauss-Newton method approximates the nonlinear term $\bm{h}(\bm{v})$ in \eqref{eq:exploss} via \eqref{eq:lin}, and finds the next iterate as its minimizer; that is,
	\begin{equation}\label{eq:gaussnewton}
	\bm v_{i+1}= \arg \min_{\bm v}~ \left\| \bm{z} - \bm h(\bm v_i)- \bm J_i(\bm v - \bm v_i)\right\|^2.
	\end{equation}
	Clearly, the per-iteration subproblem \eqref{eq:gaussnewton} is convex quadratic. 
	%The optimal $\bm v^{\ast}$ can be found by means of standard convex programming methods. 
	If matrix $\bm{J}_i^{\top} \bm{J}_i$ is invertible, the iterate $\bm v_i$ can be updated in closed-form as
	\begin{equation}\label{eq:closeform}
	\bm{v}_{i+1}=\bm{v}_{i}+\big(\bm{J}_{i}^{\top} \bm{J}_{i}\big)^{-1}\bm{J}_{i}^{\top}(\bm{z}-\bm{h}(\bm{v}_{i}))
	\end{equation}
	until some stopping criterion is satisfied. In practice however, due to the matrix inversion, the Gauss-Newton method becomes computationally expensive; it is also sensitive to initialization, and in certain cases it can even diverge. These limitations discourage its use for real-time monitoring of large-scale networks. To address these limitations, instead of solving every PSSE instance (corresponding to having a new set of measurements in $\bm{z}$) with repeated iterations, an end-to-end approach based on DNNs is pursued next.
	
	%%%%%%%%%%%%%%%%%%%%%%%%%%%%%%%%%%%%
	\section{Unrolled Gauss-Newton with Deep Priors}\label{sec:unro}
	As mentioned earlier, PSSE can be underdetermined and thus ill posed due to e.g., lack of observability. To cope with such a challenge, this section puts forth a flexible topology-aware prior that can be incorporated as a regularizer of the PSSE cost function in~\eqref{eq:exploss}. To solve the resultant regularized PSSE, an alternating minimization-based solver is developed. Subsequently, an end-to-end DNN architecture is constructed by unrolling the alternating minimization solver. Such a novel DNN is built using several layers of unrolled Gauss-Newton iterations followed by proximal steps to account for the regularization term. Interestingly, upon utilizing a GNN-based prior, the power network topology can be exploited in PSSE.

\subsection{Regularized PSSE with Deep Priors}
In practice, recovering $\bm{v}$ from $\bm{z}$ can be ill-posed, for instance when $\bm{J}_i$ is a rectangular matrix. Building on the data-driven deep priors in image denoising \cite{learningprior2013mri,schlemper2017deep,modl2018tmi}, we advocate regularizing any PSSE loss (here, the NLS in \eqref{eq:exploss}) with a trainable prior information, as 
\begin{equation}\label{eq:denoiser}
\min_{\bm{v} \in \mathbb{R}^{2N}}~\|\bm z-\bm{h}(\bm{v})\|^{2}+\lambda \, \| \bm v - \mathcal D(\bm v) \|^2
\end{equation}
where $\lambda \ge 0$ is a tuning hyper-parameter, while the regularizer promotes states $\bm v$ residing close to ${\mathcal D} (\bm v)$. The latter could be a nonlinear $\hat{\bm v}$ estimator (obtained possibly offline) based on training data. 
To encompass a large family of priors, we advocate a DNN-based estimator $\mathcal D_{\pmb \theta}(\bm v)$ with weights $\pmb \theta$ that can be learned from historical (training) data. Taking a Bayesian view, the DNN $\mathcal{D}_{\pmb \theta}(\cdot)$ can ideally output the posterior mean for a given input.  

Although this regularizer can deal with ill conditioning, the PSSE objective in \eqref{eq:denoiser} remains nonconvex. In addition, the nested structure of $\mathcal D_{\bm \theta}(\cdot)$ presents further challenges. Similar to the Gauss-Newton method for NLS in \eqref{eq:exploss}, we will cope with this challenge using an alternating minimization algorithm to iteratively approximate the solution of \eqref{eq:denoiser}. Starting with some initial guess $\bm{v}_0$, each iteration $i$ uses a linearized data consistency term to obtain the next iterate $\bm{v}_{i+1}$; that is,
\begin{align*}
%\label{eq:gaussdenoiser}
\bm v_{i+1}&= \arg \min_{\bm v} \| \bm{z} - \bm h(\bm v_i)\! -  \bm J_i(\bm v - \bm v_i)\|^2  \!+   \lambda  \| \bm v - \mathcal D_{\pmb \theta} ( \bm v_{i}) \|^2\nonumber\\
&=\bm A_i \bm{z}  +\bm B_i \bm u_{i} + \bm b_i  %\label{eq:dcupdt} 
\end{align*}
where we define 
\begin{subequations}\label{eq:paracacul}
	\begin{align}
	\bm A_i :\! &=(\bm J^{\top}_i \bm J_i + \lambda \bm I)^{-1} \bm J^{\top}_i  \notag\\
	\bm B_i :\! &= \lambda (\bm J^{\top}_i \bm J_i + \lambda \bm I)^{-1}  \notag \\
	\bm b_{i}:\! & =(\bm J^{\top}_i \bm J_i + \lambda \bm I)^{-1} \bm J^{\top}_i  ( \bm J_i \bm v_i- \bm h(\bm v_i) )  \notag.
	\end{align}
\end{subequations}
The solution of \eqref{eq:denoiser} can thus be approached by alternating between the ensuing two steps 
\begin{subequations}\label{eq:update}
	\begin{align}
	\bm u_i &= \mathcal D_{\pmb \theta}(\bm v_{i}) \label{eq:denoisupdt} \\
	\bm v_{i+1} &= \bm A_i \bm{z}  +\bm B_i \bm u_{i} + \bm b_i  \label{eq:dcupdt} .
	\end{align}
\end{subequations}

Specifically, with initialization $\bm v_0 = \bm 0$ and input $\bm{z}$, the first iteration yields $\bm v_1 = \bm A_0 \bm z + \bm B_0 \bm u_0 + \bm b_0$.  Upon passing  $\bm{v}_1$ through the DNN $\mathcal{D}_{\bm{\theta}}(\cdot)$, the output $\bm{u}_1$ at the first iteration, which is also the input to the second iteration, is given by $\bm u_1 = \mathcal D_{\pmb \theta}(\bm v_{1})$ [cf. 
\eqref{eq:denoisupdt}]. In principle, state estimates can be obtained by repeating these alternating iterations whenever a new measurement $\bm{z}$ becomes available. However, at every iteration $i$, the Jacobian matrix $\bm{J}_i$ must be evaluated, followed by matrix inversions to form $\bm{A}_i$, $\bm{B}_i$, and $\bm b_i$. The associated computational burden could be thus prohibitive for real-time monitoring tasks of  large-scale power systems.  

For fast implementation, we pursue an end-to-end learning approach that trains a DNN constructed by unrolling iterations of this alternating minimizer to approximate directly the mapping from measurements $\bm{z}$ to states $\bm{v}$; see Fig. \ref{fig:unrollGNN} for an illustration of the resulting GNU-NN architecture. Recall that in order to derive the alternating minimizer, the DNN prior $\mathcal{D}_{\pmb \theta}(\cdot)$ in \eqref{eq:denoisupdt} was assumed pre-trained, with weights $\pmb \theta$ fixed in advance. In our GNU-NN however, we consider all the coefficients $\{\bm {A}_{i}\}_{i=0}^I $, $\{\bm {B}_{i}\}_{i=0}^I $, $\{\bm {b}_{i}\}_{i=0}^I $, as well as the DNN weights $\{{\bm \theta}_i\}_{i=0}^{I}$ to be learnable from data. 

This end-to-end GNU-NN can be trained using backpropagation based on historical or simulated measurements $\{\bm z^t\}_{t=1}^{T}$ and corresponding ground-truth states $\{\bm v^{\ast t}\}_{t=1}^{T}$.
%-state training pairs $\{(\bm z^t, \bm v^{\ast t})\}_{t=1}^{T}$. %\cite{liang2019tsgpsse}.
Entailing only several matrix-vector multiplications, our GNU-NN achieves competitive PSSE performance compared with other iterative solvers such as the Gauss-Newton method. 
Further, relative to the existing data-driven NN approaches, our GNU-NN can avoid  vanishing and exploding gradients. This is possible thanks to direct (a.k.a skipping) connections from the input layer to intermediate and output layers. 

{
	\begin{remark}
		Albeit the problem remains non-convex, and may converge to a local solution, the key advantage of  data-driven-based PSSE  comes from utilizing abundant available historical training data. Specifically, the widely used algorithms such as stochastic gradient descent  algorithm and its variants, have been successful to escape local minima while updating the NN weights. 
		To prevent practical challenges, such as ``overfitting'' and offer better generalization performance, large training data sets are oftentimes used in practice. 
		Another feature of NNs and other machine learning approaches is that
		they alleviate the computational burden at the operation stage by shifting computationally intensive	`hard work' to the off-line training stage. Therefore, the sensitivity, hyper-parameter tuning, and convergence issues are to be tackled mostly during training phase.  After the mapping function between the measurement $\bm z$ and state vector $\bm v$  is learned, estimating the states associated with a fresh set of measurements only requires very simple operations, that is, passing the measurements through the learned NN. This would greatly improve the efficiency of PSSE, bringing real-time state estimation within reach.
		%	Nevertheless, from a theoretical perspective, the convergence of such stochastic algorithms has been studied in the context of adaptive signal processing \cite{benveniste1987}. Unfortunately, most of these proofs seldom extend to non-linear cases, specifically deep NNs.
		%%	 We agree that the convergence analysis of our method theoretically is not well understood, but this indeed is a limitation of all existing DNN methods.
		%	  Many more ideas have recently been developed to effectively train DNNs in practice, such as drop-out \cite{srivastava2014}, batch normalization \cite{ioffe2015}, weight-decay \cite{zhang2017b}, early-stopping \cite{yao2007},
		%	%weight initializing \cite{zhang2017b}, 
		%	etc.. Besides, in the proposed GNU-NN structure, we advocate a trainable regularizer, which well utilizes historical data to improved prediction performance.
	\end{remark}
}

Interestingly, by carefully choosing the specific model for $\mathcal D_{\pmb \theta}(\cdot)$, desirable properties such as scalability and high estimation accuracy can be also effected. For instance, if we use feed forward NNs as $\mathcal D_{\pmb \theta}(\cdot)$, it is possible to obtain a scalable solution for large power networks. However, feed forward NN can only leverage the grid topology indirectly through simulated MATPOWER data. This prompts us to focus on GNNs, which can explicitly capture the topology and the physics of the power network. The resultant Gauss-Newton unrolled with GNN priors (GNU-GNN) is elaborated next.

%%%%%%%%%%%%%%%%%%%%%%%%%%%%%%%%%%%%%%%%%%%%%%%%%%%%%%%%%%%%%%%%%%%%%%%
\subsection{Graph Neural Network Deep Prior}\label{sec:GNNs}
To allow for richly expressive state estimators to serve in our regularization term, we model $\mathcal D_{\pmb \theta}(\cdot)$ through GNNs, that are a prudent choice for networked data. 
GNNs have recently demonstrated remarkable performance in several tasks, including classification, recommendation, and robotics \cite{kipf2016semi}. By operating directly over graphs, GNNs can explicitly leverage the power network topology. Hence, they are attractive options for parameterization in application domains where data adhere to a graph structure~\cite{kipf2016semi}.   

Consider a graph of $N$ nodes with weighted adjacency matrix $\bm W$ capturing node connectivity. Data matrix $\bm X \in \mathbb R^{N \times F}$ with $n$-th row $\bm{x}_n^\top:=[\bm{X}]_{n:}$ representing an $F\times 1$ feature vector of node $n$, is the GNN input. For the PSSE problem at hand, features are real and imaginary parts of the nodal voltage ($F=2$). Upon pre-multiplying the input $\bm X$ by $\bm W$, features are propagated over the network, yielding a diffused version $\check{\bm Y} \in {\mathbb R}^{N\times F}$ that is given by 
\begin{equation}\label{eq:shiftedallbus}
\check{\bm{Y}}=\bm{W X}.
\end{equation} 
%where %the new graph signal $\bm Y \in {\mathbb R}^{N\times F}$ is a diffused measurement, and 
%$\bm W$ encodes the topological information of the underlying graph. 

\begin{remark}
	To model feature propagation, a common option is to rely on the adjacency matrix or any other matrix that preserves the structure of the power network (i.e.
	$\bm W_{nn'}=0$ if $(n,n') \notin \mathcal E$). Examples include the graph Laplacian, the random walk Laplacian, and their normalized versions.
\end{remark}

Basically, the shift operation in \eqref{eq:shiftedallbus} 
linearly combines the $f$-th features of all neighbors to obtain its propagated feature. Specifically for bus $n$, the shifted feature $[\check{\bm Y}]_{nf}$ is  
\begin{equation}\label{eq:onehopshift}
[\check{\bm{Y}}]_{nf} = \sum_{i=1}^{N}[\bm{W}]_{ni}[\bm{X}]_{if}=\sum_{i \in \mathcal{N}_{n}} w_{ni} x_{i}^{f} 
\end{equation}
where ${\mathcal{N}}_{n}=\{i \in \mathcal{N}:(i, n) \in \mathcal{E}\}$ denotes the set of neighboring buses for bus $n$. Clearly, this interpretation generates a diffused copy or shift of $\bm{X}$ over the graph.

The `graph convolution' operation in GNNs exploits topology information to linearly combine features, namely
\begin{equation}
\label{eq:gc}
[\bm{Y}]_{nd}:=[\mathcal{H} \star \bm{X}; \bm{W}]_{nd}:=\sum_{k=0}^{K-1}[\bm{W}^{k} \bm{X}]_{n:} [\bm{H}_k]_{:d} 
\end{equation}
where $\mathcal{H}:=[\bm{H}_0~ \cdots~\bm{H}_{K-1}]$ with ${\bm H}_{k} \in \mathbb{R}^{F\times D}$ concatenating all filter coefficients;~$\bm Y \in \mathbb R^{N\times D}$ is the intermediate (hidden) matrix with $D$ features per bus;~and $\bm{W}^{k} \bm{X}$ linearly combines features of buses within the $k$-hop neighborhood by recursively applying the shift operator $\bm W$.

\begin{figure}
	\centering
	\includegraphics[width =0.5 \textwidth]{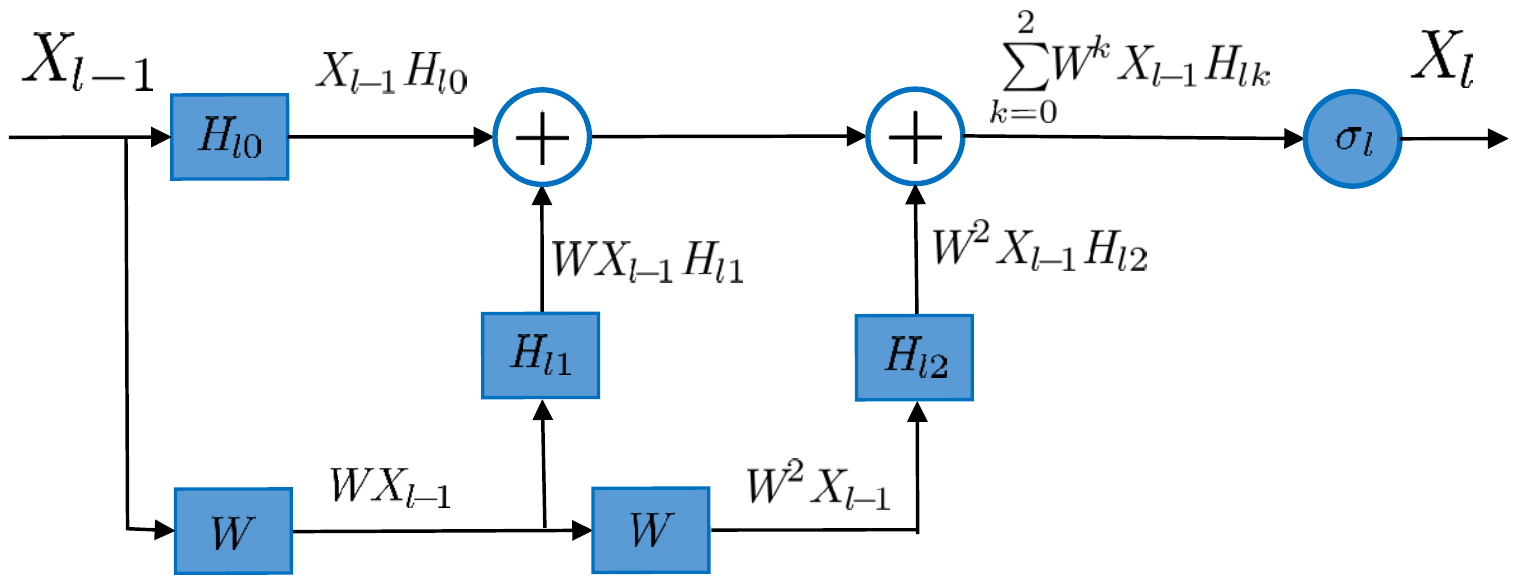}
	\caption{The signal diffuses from layer $l-1$ to $l$ with $K= 3$.}
	\label{fig:GNNflow}
\end{figure}

To obtain a GNN with $L$ hidden layers, let $\bm{X}_{l-1}$ denote the output of the $(l-1)$-st layer, which is also the $l$-th layer input for $l=1, \ldots, L$, and $\bm{X}_0 = \bm{X}$ is the input matrix. The hidden $\bm{Y}_{l} \in \mathbb{R}^{N\times D_{l}}$ with $D_l$ features is obtained by applying the graph convolution operation \eqref{eq:gc} at layer $l$, that is
\begin{equation}
\label{eq:gclayer} 
[\bm{Y}_l]_{nd} =\sum_{k=0}^{K_l-1}[\bm{W}^{k} \bm{X}_{l-1}]_{n:} [\bm{H}_{lk}]_{:g} 
\end{equation}
where $\bm{H}_{lk} \in \mathbb{R}^{F_{l-1}\times F_{l}}$ are the graph convolution coefficients for $k=0, \ldots, K_l-1$. The output $\bm X_l$ at layer $l$ is found by applying a graph convolution followed by a point-wise nonlinear operation $\sigma_{l}(\cdot)$, such as the rectified linear unit (ReLu) $\sigma_{l}(t):=\max\{0,\,t\} $ for $t\in\mathbb{R}$; see Fig. \ref{fig:GNNflow} for a depiction. Rewriting \eqref{eq:gclayer} in a compact form, we arrive at  
\begin{equation}\label{llayershift}
\bm{X}_{l} =\sigma_{l}(\bm Y_{l})=\sigma_{l}\!\left(\sum_{k=0}^{K_l-1} \bm{W}^{k} \bm{X}_{l-1} \bm{H}_{l k}\right).
\end{equation}
The GNN-based PSSE provides a nonlinear functional operator $\bm{X}_L=\bm { \Phi}(\bm{X}_{0} ; \bm{\Theta}, \bm{W})$ that maps the GNN input $\bm{X}_{0}$ to voltage estimates by taking into account the graph structure through $\bm W$, through
\begin{align}
&\bm { {\Phi}}(\bm{X}_{0} ; \bm{\Theta}, \bm{W}) = \label{eq:GNNbasepsse} \\ 
&\sigma_{L}\!\left(\sum_{k=0}^{K_{L }-1} \bm{W}^{k} \!\left(\ldots \!\left(\sigma_{1}\!\left(\sum_{k=0}^{K_1-1} \bm{W}^{k} \bm{X}_{0} \bm{H}_{1 k}\right) \ldots\right)\right)\bm{H}_{L k}\right)\nonumber
\end{align}
where the parameter set $\bm \Theta$ contains all the filter weights; that is, $\bm{\Theta} := \{\bm{H}_{lk}, \forall l, k\}$, and also recall that $\bm X_0 = \bm X$.

\begin{remark}
	With $L$ hidden layers, $F_l$ features and $K_l$ filters per layer, the total number of parameters to be learned is $|\bm{\Theta}| =\sum_{l=1}^{L} K_{l} \times F_{l} \times F_{l-1}$. 
\end{remark}

To accommodate the GNN implementation over the proposed unrolled architecture, at the $i$-th iteration,  we reshape the states $\bm{v}_{i} \in \mathbb{R}^{2N}$ to form the $N \times 2$ GNN input matrix $\bm X_0^{i} \in {\mathbb R}^{N \times 2}$.~Next, we vectorize the GNN output $\bm{X}_L^i \in \mathbb{R}^{N\times 2}$ to obtain the vector $\bm u_i \in \mathbb{R}^{2N}$ (cf. \eqref{eq:denoisupdt}).~For notational brevity, we concatenate all trainable parameters of the GNU-GNN in vector $\bm{\omega}:=[\{\bm{\Theta}_i\}_{i=0}^{I}, \{{\bm A}_i\}_{i=0}^{I}, \{\bm{B}_i\}_{i=0}^{I}, \{\bm{b}^1_i\}_{i=0}^{I}]$, and let $\bm{\pi}(\bm{z};\bm{\omega})$ denote the end-to-end GNU-GNN parametric model, which for given measurements $\bm{z}$ predicts the voltages across all buses, meaning $\hat{\bm{v}}=\bm{\pi}(\bm{z};\bm{\omega})$.~The GNU-GNN weights $\bm \omega$ can be updated using backpropagation, after specifying a certain loss $\ell(\bm{v}^{\ast}, \bm{v}_{I+1})$ measuring how well the estimated voltages $\bm{v}_{I+1}$ by the GNU-GNN matches the ground-truth ones $\bm{v}^{\ast}$. The proposed method is summarized in Alg. \ref{Alg_a}.

\begin{algorithm}[t!]
	\caption{PSSE Solver with GNN Priors.}
	\label{Alg_a}
	\hspace*{0.02in} {\bf Training phase:}
		\newline
		\textbf{Input:} Training samples $\{(\bm z^t, {\bm v}^{\ast t})\}_{t=1}^{T}$
		\newline
		\textbf{Initialize:} 
		\newline $\bm{\omega}^1 :=[\{\bm{\Theta}^1_i\}_{i=0}^{I}, \{{\bm A}^1_i\}_{i=0}^{I}, \{\bm{B}^1_i\}_{i=0}^{I}, \{\bm{b}^1_i\}_{i=0}^{I}]$, $\bm v_0=0$.
		%		$\{\bm A^0_i\}^I_{i=1}$, $\{\bm B^0_i\}^I_{i=1}$, $\{\bm u^t_0\}^T_{t=1}$, and $\{\bm \Theta^0_i\}^I_{i=1}$.
		\newline
		{For \quad $t =1,2,\ldots, T$ } 
		\newline
		Feed $ \bm z^t$  and $\bm v_{0}$ as input into GNU-GNN.
		\newline
		{For \quad $i = 0, 1, \ldots, I$}\footnotemark
		\newline
		Reshape $\bm v_i \in \mathbf R^{2N}$ to get $\bm X_0^{i} \in {\mathbb R}^{N \times 2}$.
		\newline
		Feed $\bm X_0^{i}$ into GNN.
		\newline
		Vectorize the GNN output $\bm{X}_L^i \in \mathbb{R}^{N\times 2}$ to get $\bm u_i$.
		\newline
		Obtain $\bm v_{i+1} \in \mathbf R^{2N}$ using \eqref{eq:dcupdt}.
		\newline
		Obtain $\bm v^t_{I+1}$ using \eqref{eq:dcupdt}.
		\newline
		Minimize the loss $\ell(\bm{v}^{\ast t}, \bm{v}^t_{I+1})$ and update $\bm{\omega}^{t}$.
		\newline
		\textbf{Output:} $\bm{\omega}^{T}$
	\hspace*{0.02in} {\bf Inference phase:} 
%	\begin{algorithmic}[1]
\newline
		 {For $t = T+1, \ldots, T'$}
		\newline
		 Feed real-time $\bm{z}^t$ to the trained GNU-GNN.
		\newline
		Obtain the estimated voltage $\bm v^{t} $.
%	\end{algorithmic}
\end{algorithm}

%%%%%%%%%%%%%%%%%%%%%%%%%%%%%%%%%%%%%%%%%%%%%%%%%%%%%%%%%%%%%%%%%%%%%%%
\section{Robust PSSE Solver}\label{sec:adve}
{ In real-time inference, our proposed GNU-GNN that has been trained using past data, outputs an estimate of the state $\bm v^t$ per time slot $t$ based on the observed measurements $\bm z^t$.~However, due to impulsive communication noise and possibly cyberattacks,  our proposed GNU-GNN in Section \ref{sec:unro} can yield grossly biased estimation results. A natural extension of our approach is to consider these imperfections in the PSSE problem.
	%	 including impulsive communication noises, cyber-attacks, and possibly corrupted data, which leads to biased or grossly wrong estimates. 
	Therefore, after proposing our method to inject prior information and training the DNN for normal input, we robustify our method in the presence of imperfections in this section.}	

{To obtain estimators robust to bad data, classical formulations including H\"uber estimation, H\"uber M-estimation, and Schweppe-H\"uber generalized M-estimation, rely on the premise that measurements obey $\epsilon$-contaminated probability models; see e.g., \cite{wang2019overview}.}~Instead, the present paper postulates that measured and ground-truth voltages are drawn from some nominal yet unknown distribution $P_0$ supported on $\mathcal{S} = \mathcal{Z}\times \mathcal{V}$, that is $(\bm{z}, \bm{v}^{\ast}) \sim P_0$.~Therefore, to obtain the end-to-end GNU-GNN parametric model $\bm
{\pi}(\bm{z};\bm{\omega})$, the trainable parameters $\bm{\omega}$ are optimized by solving $
{\min}_{\,\bm{\omega}}~\mathbb{E}_{P_{0}}\big[\ell(\bm{\pi}(\bm{z};\bm{\omega}),\bm{v}^{\ast })\big]$ \cite{miller2019adversarial}.~In practice, $P_0$ is unknown but i.i.d. training samples $\{ (\bm z^t, \bm v^{\ast t})\}_{t=1}^{T} \sim P_0$ are available.~In this context, our PSSE amounts to solving for the minimizer of the empirical loss as
\begin{equation} \label{eq:emploss}
\!\!\min_{\bm{\omega}}\; \bar{\mathbb{E}}_{\widehat{P}^{(T)}_0} [\ell(\bm{\pi}(\bm{z}^t;\bm{\omega}),\bm{v}^{\ast t})] =  \frac{1}{T} \sum_{t=1}^T \ell(\bm{\pi}(\bm{z}^t;\bm{\omega}),\bm{v}^{\ast t}).
\end{equation} 
\footnotetext{For brevity the superscript $t$ is removed from inner iteration $i$.} 
To cope with uncertain and adversarial environments, the solution of \eqref{eq:emploss} can be robustified by optimizing over a set $\mathcal{P}$ of probability distributions centered around $\widehat{P}^{(T)}_0$, and minimizing the \textit{worst-case} expected loss with respect to the choice of any distribution $P \in \mathcal{P}$.~Concretely, this can be formulated as the following \textit{distributionally robust} optimization
\begin{equation}\label{eq:worstcase}
\min_{\bm{\omega}} \sup _{P \in \mathcal{P}}\; \mathbb{E}_{P}[\ell(\bm{\pi}(\bm{z};\bm{\omega}),\bm{v}^\ast)].
\end{equation}
Compared with \eqref{eq:emploss}, the worst-case formulation in \eqref{eq:worstcase} ensures a reasonable performance across a continuum of distributions in $\mathcal{P}$.~A broad range of ambiguity sets $\mathcal{P}$ could be considered here. Featuring a strong duality enabled by the optimal transport theory \cite{vinali08opttrans}, such distributionally robust optimization approaches have gained popularity in robustifying machine learning models \cite{bandi2014robust}.~Indeed, this tractability is the key impetus for this section.

%including momentum \cite{}, KL divergence \cite{}, statistical test \cite{}, and currently popular Wasserstein distance-based sets \cite{}; see e.g., \cite{} for a thorough discussion. 

%The optimal transport theory was first studied by Monge \cite{monge1781memoire}. In the Monge problem, piles of sand and some holes with the same total volume as sands are geographically distributed over an area. 
%The goal of optimal transport is to find the optimal moves to transfer the entire piles to the holes with the minimum transportation cost. Clearly,  the optimal strategy depends on the cost function, which is a function of distance between piles and holes.

%To formalize, consider probability density functions $P$ and $Q$ defined over support $\mathcal{S}$, and let $\Pi(P,Q)$ be the set of all joint probability distributions with marginals $P$ and $Q$.
{Considering probability density functions $P$ and $Q$ defined over support $\mathcal{S}$, let $\Pi(P,Q)$ be the set of all joint probability distributions with marginals $P$ and $Q$.}
Also let $c: \mathcal{Z} \times \mathcal{Z}  \rightarrow [0, \infty)$ be some cost function representing the cost of transporting a unit of mass from $(\bm{z}, \bm{v}^\ast)$ in $P$ to another element $(\bm{z}', \bm{v}^\ast)$ in $Q$ (here we assume that attacker can compromise the measurements $\bm{z}$ but not the actual system state $\bm{v}^\ast$).~The so-called optimal transport between two distributions $P$ and $Q$ is given by \cite[Page 111]{vinali08opttrans}
\begin{align}
W_c(P,Q) := \; \underset{\pi \in \Pi}{\inf} \, \mathbb{E}_\pi \big[ c(\bm{z},\bm{z}')\big]. 
\label{eq:wassdist}
\end{align}
Intuitively, $W_c(P,Q)$ denotes the minimum cost associated with transporting all the mass from distribution $P$ to $Q$.~Under mild conditions over the cost function and distributions, $W_c$ gives the well-known Wasserstein distance between $P$ and $Q$; see e.g., \cite{sinha2017certify}.
%\begin{remark}
%	If $c(\cdot)$ satisfies the axioms of distance, then $W_c$ defines a distance on the space of probability measures. For instance, if $P$ and $Q$ are defined over a Polish space equipped with metric $d$, then choosing $c(\bm z, \bm z') = d^p(\bm z, \bm z')$ for some $p\in [1, \infty)$ asserts that $W_c^{1/p}(P,Q)$ is the well-known Wasserstein distance of order $p$ between probability measures $P$ and $Q$ \cite[Definition 6.1]{vinali08opttrans}.  
%\end{remark}

Having introduced the distance $W_c$, let us define an uncertainty set for the given empirical distribution $\widehat{P}^{(T)}_0$,~as $\mathcal{P}:= \{P| W_c(P,\widehat{P}^{(T)}_0) \le \rho \}$ that includes all probability distributions having at most $\rho$-distance from $P_0^{(T)}$.~Incorporating $\mathcal{P}$ into \eqref{eq:worstcase} yields the following optimization for distributionally robust GNU-GNN estimation
\begin{subequations}\label{eq:robform}
	\begin{align}\label{eq:robformbobj}
	\min_{\bm{\omega}} \, \sup _{P}\,&~ \mathbb{E}_{P}[\ell(\bm{\pi}(\bm{z};\bm{\omega}),\bm{v}^\ast)]\\
	%\end{align}
	%\begin{equation}
	\label{eq:robformbcons}
	\qquad {\rm s.t.} &~ W_c(P,\widehat{P}^{(T)}_0) \le \rho. 
	\end{align}
\end{subequations}
{Notice that the inner functional optimization in \eqref{eq:robformbobj} runs over all  probability distributions $P$ characterized by \eqref{eq:robformbcons}. It is intractable to optimize directly over the infinite-dimension distribution functions. Fortunately, for continuous loss as well as transportation cost functions, the optimal objective value of the inner maximization is equal to its dual optimal objective value. In addition, the dual problem involves optimization over only a one-dimension variable.} 
%Observe that the inner functional optimization in \eqref{eq:robformbobj} runs over all  probability distributions $P$ characterized by \eqref{eq:robformbcons}.
%Evidently, optimizing directly over the infinite-dimension distribution functions is intractable.
%Fortunately, for continuous loss as well as transportation cost functions, the inner maximization satisfies strong duality condition;~that is, the optimal objective value of the inner maximization is equal to its dual optimal objective value.
%In addition, the dual problem involves optimization over only a one-dimension variable, that can be carried out efficiently. 
These two observations prompt us to solve \eqref{eq:robform} in the dual domain.~To formally obtain this tractable surrogate, we call for a result from \cite{blanchet2019quantifying}.

\begin{proposition}
	\label{prop:strongdual}
	Let the loss $\ell: \bm{\omega} \times \mathcal{Z} \times \mathcal{V} \rightarrow [0,\infty)$, and transportation cost $c:\mathcal{Z} \times \mathcal{Z} \rightarrow [0, \infty)$ be continuous functions. Then, for any given $\widehat{P}^{(T)}_0$, and $\rho > 0$, it holds \begin{align} \label{eq:strongduality}
	\sup_{P\in\mathcal{P}} & \,\mathbb{E}_{P}\!\left[\ell(\bm{\pi}(\bm{z};\bm{\omega}), \bm{v}^\ast)
	\right] =  \\ 
	& \inf_{\gamma \ge 0}  \big\{  \bar{\mathbb{E}}_{{(\bm{z}, \bm{v}^\ast)\sim\widehat{P}^{(T)}_0}} \big[\sup_{\bm \zeta \in {\mathcal Z}}  \ell(\bm{\pi}(\bm{\zeta};\bm{\omega}),\bm{v}^\ast) \!+\! \gamma (\rho - c(\bm z, \bm \zeta)) \big] \big\} 
	\nonumber
	\end{align}
	where $\mathcal{P}:= \left\{P| W_c(P,\widehat{P}^{(T)}_0) \le \rho \right\}$.
\end{proposition}

\begin{remark}
	Thanks to the strong duality, the right-hand side in \eqref{eq:strongduality} simply is a univariate dual reformulation of the primal problem given on the left-hand side. In contrast with the primal formulation, the expectation in the dual domain is taken only over the empirical distribution $\widehat{P}^{(T)}_0$ rather than over any $P \in \mathcal{P}$. Furthermore, since this reformulation circumvents the need for finding the optimal coupling $\pi \in \Pi$ to define $\mathcal{P}$, and  characterizing the primal objective for all $P \in \mathcal{P}$, it is practically appealing and convenient.
\end{remark}

{Capitalizing on Proposition \ref{prop:strongdual}, the inner maximization can be replaced with its dual reformulation. As a consequence, the following distributionally robust PSSE optimization can be arrived at 
	\begin{align}
	\!\!\!\min_{\bm \omega} \!\inf_{\gamma \ge 0}   {\bar{\mathbb{E}}}_{(\bm{z}, \bm{v}^\ast)\sim\widehat{P}^{(T)}_0} \!\Big[\!\sup_{\bm \zeta \in {\mathcal Z}} \ell(\bm{\pi}(\bm{\zeta};\bm{\omega}),\bm{v}^\ast) \!+\! \gamma (\rho \!-\! c(\bm z, \bm \zeta)) \! \Big]\!.
	\label{eq:robustdual}
	\end{align}
	%Before proceeding to solve this learning problem in the next section, two remarks are in order. 
	%\begin{remark}
	%\label{rm:minmax}
	%Although the robust surrogate in \eqref{eq:robustdual} looks similar to minimax (saddle-point) optimization problems, it requires the supremum to be solved separately per observed measurements $\bm{z}$, that cannot readily be handled by existing minimax optimization solvers. 
	%\end{remark} 
	%As remark \ref{rm:minmax} suggests, the problem of interest in \eqref{eq:robustdual} cannot be handle through existing optimization techniques. 
	Finding the optimal solution $(\bm{\omega}^\ast,\gamma^\ast)$ of 
	\eqref{eq:robustdual} is in general challenging, because it requires the supremum to be solved separately per observed measurements $\bm{z}$, that cannot readily be handled by existing minimax optimization solvers.}
A common approach to bypassing this hurdle is to approximate the optimal $\bm{\omega}^\ast$ by solving  \eqref{eq:robustdual} with a preselected and fixed $\gamma>0$  \cite{sinha2017certify}.~Indeed, it has been shown in \cite{sinha2017certify} that for any strongly convex transportation cost function, such as $c(\bm{z}, \bm{z'}) :=\| \bm{z} - \bm{z}'\|^2_p$ for any $p \ge 1$, a sufficiently large $\gamma>0$ ensures that the inner maximization is strongly convex, hence efficiently solvable.  
%This is because the objective function resultant from inner maximization may become non differentiable and nonconvex. The remedy is to fix the penalty $\gamma \ge 0$ to appropriate value so that the inner maximization admits a unique solution. Consequently, iterative solvers for  \eqref{eq:robustdual} can be proposed. These algorithms essentially iterate between minimizing over $\bm{\omega}$ and maximizing over $\bm{\zeta}$, and provably converge to a stationary point \cite{sinha2017certify}. This is an immediate result of Danskin's theorem for minimax optimization problems; see \cite{danskin1966theory} for more details. 
%%\gang{It has been shown that with large enough $\gamma>0$, one can approximate the optimal solution $\bm{\omega}^\ast$ of \eqref{eq:robustdual} iteratively by alternating between minimizing over $\bm{\omega}$ and maximizing over $\bm{\zeta}$ \cite{sinha2017certify}. 
%%Unfortunately, such an approach leads to suboptimal performance as it is difficult to set the optimal $\gamma^\ast$. Certainly, this could be done through cross-validation, which however will incur significant computational burden. 
%Prompted by this result, our approach is to employ a strongly convex function as the transportation cost, such as $c(\bm{z}, \bm{z'}) :=\| \bm{z} - \bm{z}'\|^2_p$ for any $p \ge 1$. Upon appropriately choosing $\gamma$, the inner maximization has a unique solution, therefore the Danskin's theorem can be invoked. As a consequence, we employ an alternating solver for \eqref{eq:robustdual}.   
Note that having a fixed $\gamma$ is tantamount to tuning $\rho$, which in turn \emph{controls} the level of infused \emph{robustness}. Fixing some large enough $\gamma > 0$ in \eqref{eq:robustdual}, our robustified GNU-GNN model can thus be obtained by solving  
\begin{equation}
\min_{\bm{\omega}}~\bar{\mathbb E}_{{(\bm{z}, \bm{v}^\ast)\sim\widehat{P}^{(T)}_0}}\Big[ \sup_{\bm{\zeta}\in\mathcal{Z}} \psi(\bm{\bm{\omega}}, \bm{\zeta}; \bm {z},\bm{v}^\ast) \Big]
\label{eq:objective}
\end{equation}
where 
\begin{equation} 
\label{eq:psi}
\psi(\bm{\bm{\omega}}, \bm{\zeta}; \bm {z},\bm{v}^\ast):=\ell(\bm{\pi}(\bm{\zeta};\bm{\omega}),\bm{v}^\ast) + \gamma (\rho -  c(\bm{z}, \bm \zeta)).
\end{equation}

{Intuitively, \eqref{eq:objective} can be understood as first `adversarially' perturbing the measurements $\bm{z}$ into $\bm{\zeta}^\ast$ 
	by maximizing $\psi(\cdot)$, and then seeking a model that minimizes the empirical loss with respect to even such perturbed inputs. Therefore, the robustness of the sought model is achieved to future data that may be contaminated by adversaries.
	Initialized with some $\bm{\omega}^0$, and given a datum $(\bm{z}^t, \bm{v}^\ast)$, we form $\psi(\cdot)$ (c.f. \eqref{eq:psi}), and implement a single gradient ascent step for the inner maximization as follows 
	\begin{align}
	\bm{\zeta}^t =  \bm{z}^t  + \eta^t \nabla_{\bm{\zeta}} \psi({\bm \omega}^t, \bm  \zeta; \bm{z}^t, \bm{v}^{\ast t})\big|_{\bm{\zeta}={\bm{z}^t}} \label{eq:sgaupdt}
	\end{align}
	where $\eta^t>0$ is the step size.
	Upon evaluating \eqref{eq:sgaupdt}, the perturbed data $\bm{\zeta}^t$ will be taken as input (replacing the `healthy' data $\bm{z}^t$) fed into Algorithm \ref{Alg_a}. Having the loss $\ell(\bm{\pi}(\bm{\zeta}^t;\bm{\omega}),\bm{v}^{\ast t})$ as solely a function of the GNU-GNN weights $\bm{\omega}$, the current iterate $\bm{\omega}^t$ can be updated again by backpropagation. }

%%%%%%%%%%%%%%%%%%%%%%%%%%%%%%%%%%%%%%%%%%%%%%%%%%%%%%%%%%%%%%%%%%%%%%% 
\section{Numerical Tests} \label{sec:test}
{This section tests the estimation performance as well as robustness of our proposed methods.}

\subsection{Simulation Setup}	

The simulations were carried out on an NVIDIA Titan X GPU with a $12$GB RAM. For numerical tests, we used real load consumption data from the $2012$ Global Energy Forecasting Competition (GEFC) \cite{datagefc}.~Using this dataset, training and testing collections were prepared by solving the AC power flow equations using the MATPOWER toolbox.~To match the scale of power demands, we normalized the load data, and fed it into MATPOWER to generate $1,000$ pairs of measurements and ground-truth voltages, $80\%$ of which were used for training while the remaining $20\%$ were employed for testing.~Measurements include all sending-end active power flows, as well as voltage magnitudes, corrupted by additive white Gaussian noise. Standard deviations of the noise added to power flows and voltage magnitudes were set to $0.02$ and $0.01$ \cite{gang2019tsg}, respectively.

\begin{figure}
	\centering
	\includegraphics[width =0.48 \textwidth]{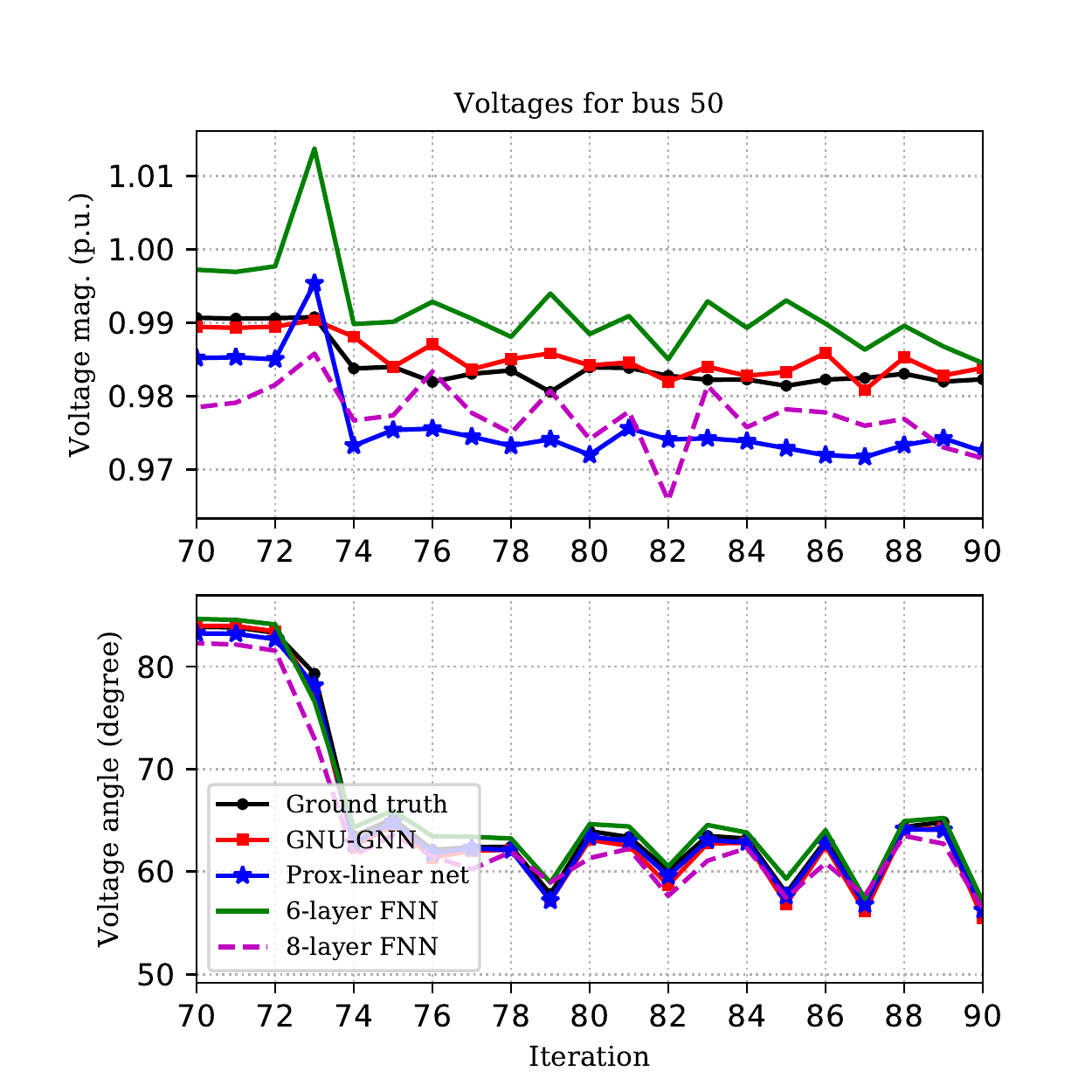}
	\caption{The estimated voltage magnitudes and angles by the four schemes at bus $50$ from slots $70$ to $90$.}
	%	\vspace{-.2 in}
	\label{fig:vol_clean_50}
\end{figure}

A reasonable question to ponder is whether explicitly incorporating the power network topology through a trainable regularizer offers improved performance over competing alternatives.~In addition, it is of interest to study how a distributionally robust training method enhances PSSE performance in the presence of bad data and even adversaries. {To this aim, four baseline PSSE methods were numerically tested, including one optimization-based method Wirtinger-Flow Gauss-Newton algorithm in \cite{gaussnewton}, and three data-driven methods: i) the prox-linear network in~\cite{liang2019tsgpsse};~ii) a $6$-layer vanilla feed-forward (F)NN;~and iii) an $8$-layer FNN.~The weights of these NNs were trained using the Adam optimizer to minimize the H\"uber loss.~The learning rate was fixed to $10^{-3}$ throughout $500$ epochs, and the batch size was set to $32$.  Furthermore, the average estimation accuracy of each algorithm is defined as follows
	\begin{equation}
	\nu=\frac{1}{N} \sum_{n=1}^{N}\left\|\bm v_{n}-\bm v_{n}^{\ast }\right\|_{2}^{2}
	\end{equation}
	where $\bm v_n$ is the estimated voltage profile from the noisy measurements generated using $\bm v_n^{\ast}$.}

\begin{figure}
	\centering
	\includegraphics[width =0.48 \textwidth]{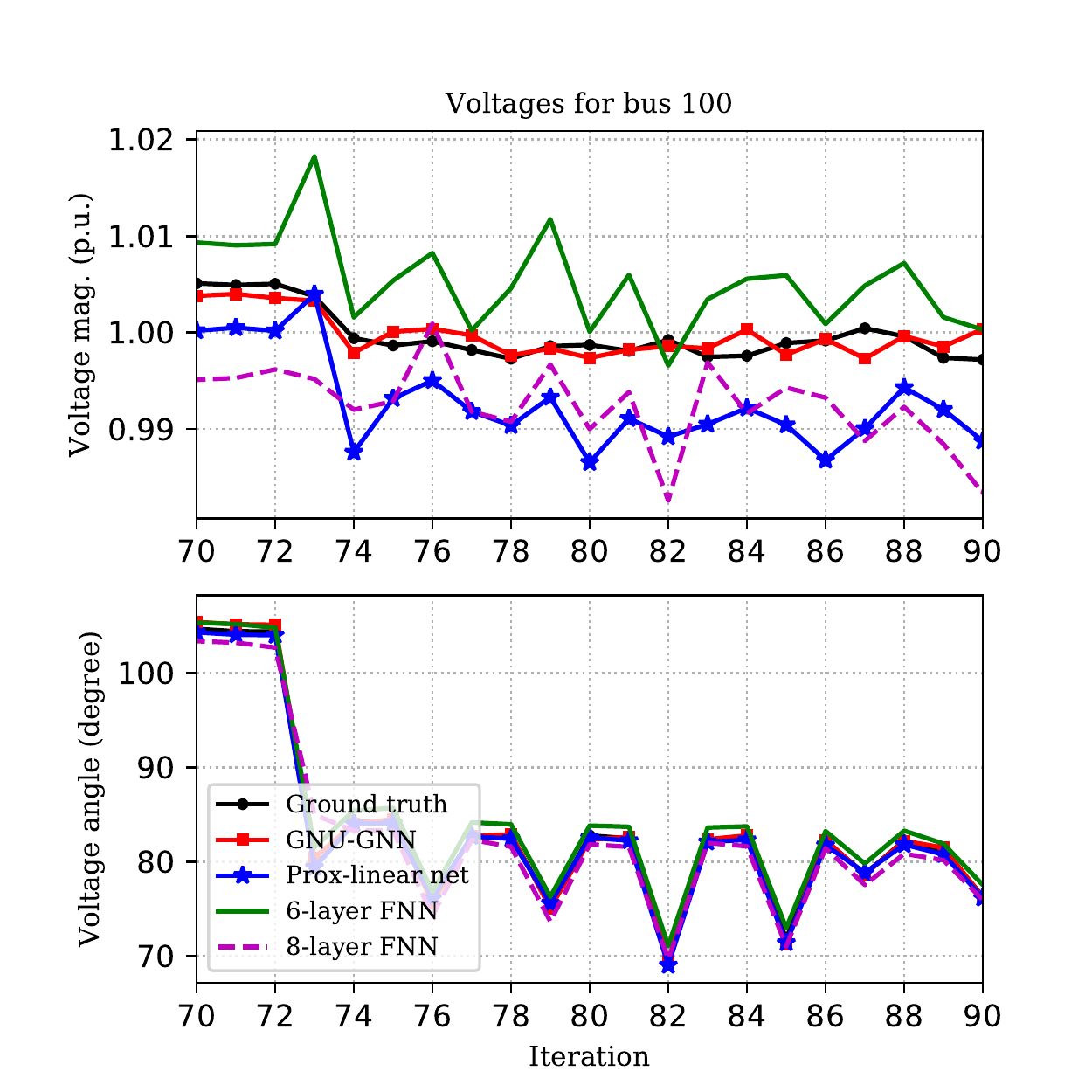}
	\caption{The estimated voltage magnitudes and angles by the four schemes at bus $100$ from slot $70$ to $90$.}
	\label{fig:vol_clean_100}
\end{figure}

\subsection{GNU-GNN for regularized PSSE}
In the first experiment, we implemented GNU-GNN by unrolling $I=6$ iterations of the proposed alternating minimizing solver, respectively. A GNN with $K=2$ hops, and $D=8$ hidden units with ReLU activations per unrolled iteration was used for the deep prior of GNU-GNN. The GNU-GNN architecture was designed to have total number of weight parameters roughly the same as that of the prox-linear network.  The Gauss-Newton algorithm is initialized using the flat voltage profile. {Table I tabulates the average performance of the proposed GNU-GNN approach, the Gauss-Newton method, the  prox-linear network, $6$-layer FNN and $8$-layer FNN over $200$ testing samples. We report the accuracy of estimation on the IEEE $118$-bus feeder, and the running time (s) on IEEE $118$-bus feeder, as well as IEEE $300$-bus feeder. 	
	%	Simple feed-forward neural networks approaches require a lot of training data and computational resources. In addition, they often suffer from exploding or diminishing gradients. This results in bad estimates for the state of the network. For example, the average estimation accuracy of the state using a 4-layer feed-forward NN, with number of neurons similar to the 4- layer GPNN, was for noiseless measurements in Scenario A. Hence, we did not include a feed-forward NN in our comparisons. The Gauss-Newton algorithm is initialized using the flat voltage profile.	
	Clearly, the proposed GNU-GNN approach achieves superior performance where the accuracy of estimation is an order of magnitude better than the state-of-the-art Gauss-Newton approach on $118$-bus feeder. In addition, since the GNU-GNN approach alleviates almost all the computational burden at the estimation time by shifting it to the training time, the running time of the proposed approach is three orders of magnitude less than the optimization-based approach on both IEEE $118$-bus feeder and IEEE $300$-bus feeder.}

%\begin{table}[htp]
%	\vspace{0.1mm}
%	{\color{blue} \caption{\textcolor{blue}{Performance comparison of different state estimation approaches.}}
%		%\vspace{.3cm}
%		\centering
%		\renewcommand{\arraystretch}{1.4} 
%		\label{tab:error}
%		\begin{tabular}{c c c c}
%			%\toprule
%			\hline \textrm{ Method } & $\nu$ ($118$-bus) & \textrm { Time ($118$-bus) } & \textrm { Time ($300$-bus) }\\
%			\hline \textrm { GNU-GNN } & $9.42 \times 10^{-3}$ & $1.54 \times 10^{-2}$ & $ 0.59$ \\
%			\textrm { Prox-linear net} & $1.26 \times 10^{-2}$ &  $1.63 \times 10^{-2}$& $0.75$ \\
%			\textrm { 6-layer FNN} & $8.65 \times 10^{-2}$ &  $2.89 \times 10^{-2}$& $3.88$ \\
%			\textrm { 8-layer FNN} & $1.93 \times 10^{-2}$ & $3.94 \times 10^{-2}$ & $4.29$ \\
%			\textrm { Gauss-Newton} & $5.25 \times 10^{-1}$ & $13.34$ & $152$ \\
%			\hline
%			%\bottomrule
%		\end{tabular}}
%	\end{table}

\begin{figure}[h!]
	\centering
	\includegraphics[width =0.48 \textwidth]{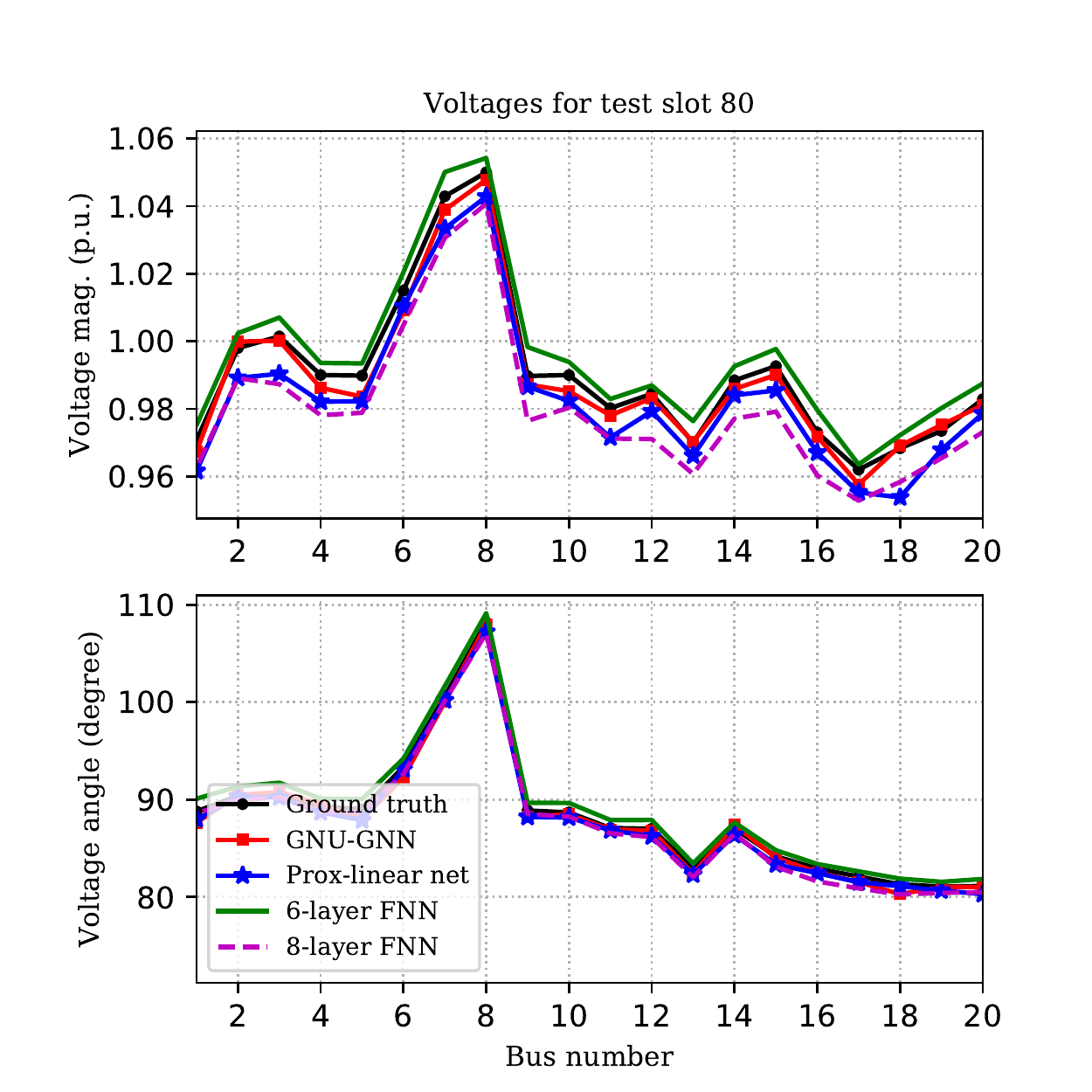}
	\caption{The estimated voltages magnitudes and angles by the four schemes for the first $20$ buses at slot $80$.}
	\label{fig:vol_clean_allbus}
	\vspace{-1em}
\end{figure}

{In order to show the quality of estimates provided by the proposed GNU-GNN method, we present the estimate of the voltage magnitudes and angles on the IEEE $118$-bus feeder. As is shown in Table I, the estimation performance of the Gauss-Newton method is too bad to depict in the same figure with other alternatives. Therefore, we did not include the Gauss-Newton in this set of results. Figs. \ref{fig:vol_clean_50} and \ref{fig:vol_clean_100} show the estimated voltage profiles obtained at buses $50$ and  $100$ from test slots $70$ to $90$, respectively.} The ground-truth and estimated voltages for the first $20$ buses on the test slot $80$ are presented in Fig. \ref{fig:vol_clean_allbus}. 
%In terms of runtime, our GNU-GNN, the prox-linear net, the $6$-layer FNN, and the $8$-layer FNN over $200$ testing samples took $1.5 \times 10^{-2}$s, $1.6 \times 10^{-2}$s, $2.3 \times 10^{-2}$s, and $3.9 \times 10^{-2}$s, respectively. 
These results corroborate the improved performance of our GNU-GNN relative to the simulated PSSE solvers.
%The results show superior estimation performance for the proposed approach.

\begin{figure}
	\centering
	\includegraphics[width =0.48 \textwidth]{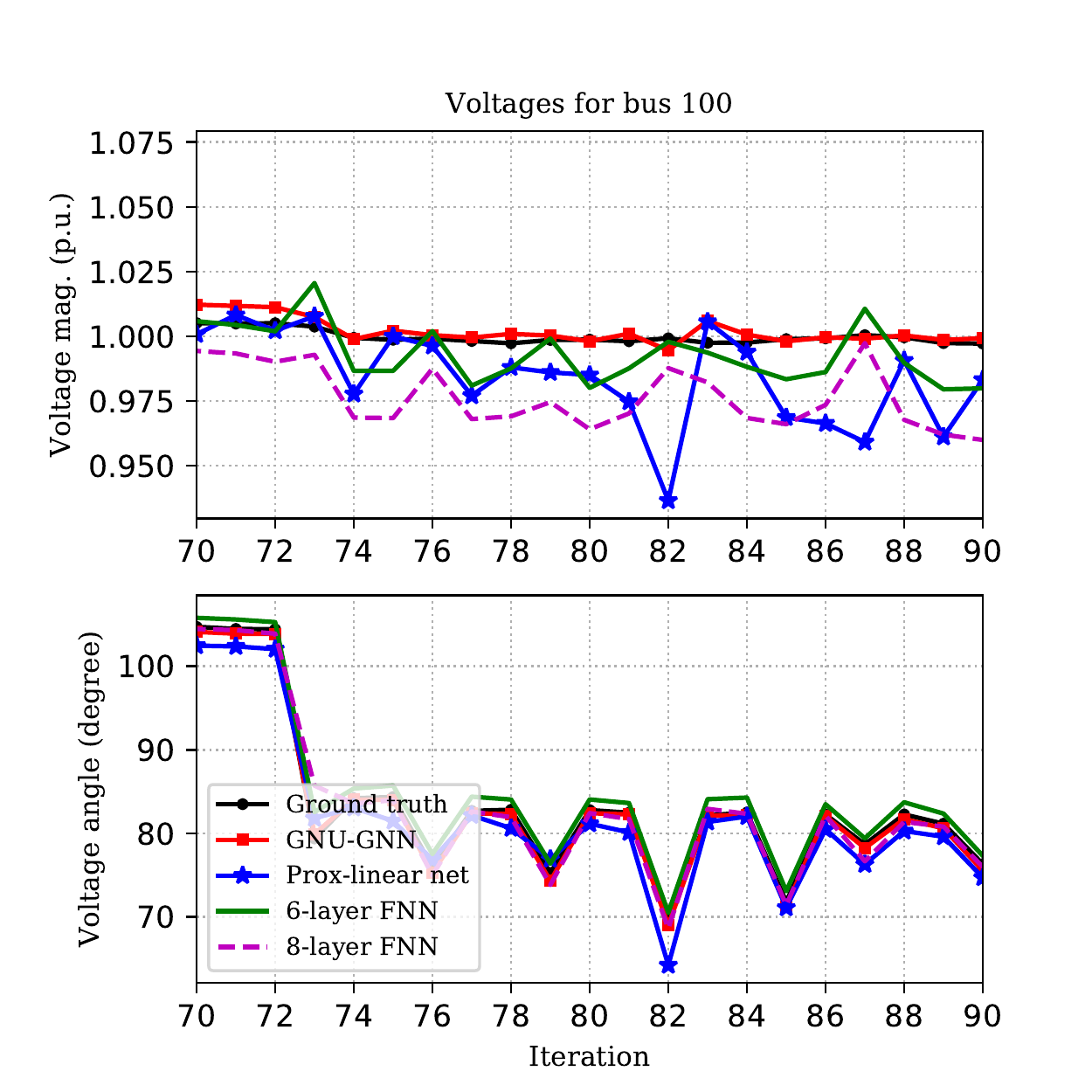}
	\caption{The estimated voltage magnitudes and angles
		by the four schemes under distributional attacks at bus $100$ from slots $70$ to $90$.}
	\label{fig:vol_advdistr_80}
\end{figure}

\subsection{Robust PSSE}
\begin{figure}
	\centering
	\includegraphics[width =0.48 \textwidth]{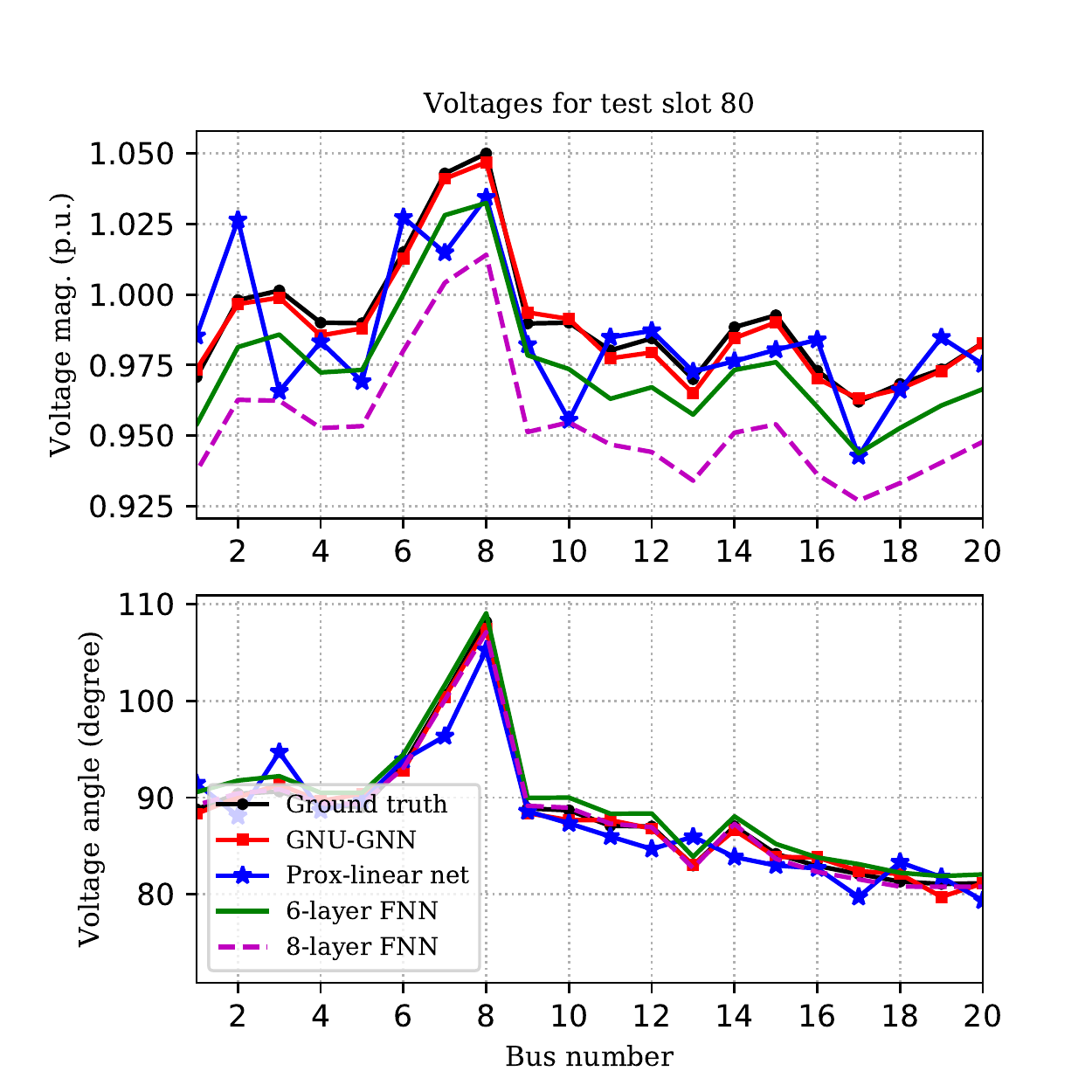}
	\caption{The estimated voltage magnitudes and angles by the four schemes under distributional attacks for the first $20$ buses at slot $80$. }
	\label{fig:vol_advdistr_slot50}
\end{figure}

Despite their remarkable performance in standard PSSE, DNNs may fail to yield reliable and accurate estimates in practice when bad data are present.~Evidently, this challenges their application in safety-critical power networks.~In the experiment of this subsection we examine the robustness of our GNU-GNN trained with the described adversarial learning method  on the IEEE $118$-bus feeder.~To this aim, a distributionally robust learning scheme was implemented to manipulate the input of GNU-GNN, prox-linear net, $6$-layer FNN, and $8$-layer FNN models.~Specifically, under distributional attacks, an ambiguity set $\mathcal{P}$ comprising distributions centered at the nominal data-generating $P_0$ was postulated.~Although the training samples were generated according to $P_0$, testing samples were obtained by drawing samples from a distribution $P \in \mathcal{P}$ that yields the worst empirical loss.~To this end we preprocessed test samples using \eqref{eq:sgaupdt} to generate adversarially perturbed samples.~Figs. \ref{fig:vol_advdistr_80} and \ref{fig:vol_advdistr_slot50} demonstrate the estimated voltage profiles under a distributional attack with a fixed $\gamma = 0.13$ (c.f. \eqref{eq:objective} and \eqref{eq:psi}) and $\ell_2$ transportation cost.~As the plots showcase, our proposed robust training method enjoys guarantees against distributional uncertainties, especially relative to competing alternatives. 
%For relatively small values of $\gamma$ (large adversarial budgets), our method is a heuristic way to  Here we have chosen $\gamma$ 

%%%%%%%%%%%%%%%%%%%%%%%%%%%%%%%%%%%%%%%%%%%%%%%%%%%%%%%%%%%%%%%%%%%%%%%
\section{Conclusions}\label{sec:conc}
This section introduced topology-aware DNN-based regularizers to deal with the ill-posed and nonconvex characteristics of standard PSSE approaches.~An alternating minimization solver was developed to approach the solution of the regularized PSSE objective function,~which is further unrolled to construct a DNN model.~For real-time monitoring of large-scale networks,~the resulting DNN was trained using historical or simulated measured and ground-truth voltages.~A basic building block of our GNU-GNN consists of a Gauss-Newton iteration followed by a proximal step to deal with the regularization term.~Numerical tests showcased the competitive performance of our proposed GNU-GNN relative to several existing ones.~Further,~a distributionally robust training method was presented to endow the GNU-GNN with resilience to bad data that even come from adversarial attacks.

%\include{chapters/event_selection}
%\include{chapters/analysis}

%Conclusion
%%%%%%%%%%%%%%%%%%%%%%%%%%%%%%%%%%%%%%%%%%%%%%%%%%%%%%%%%%%%%%%%%%%%%%%%%%%%%%%
% intro.tex: Introduction to the thesis
%%%%%%%%%%%%%%%%%%%%%%%%%%%%%%%%%%%%%%%%%%%%%%%%%%%%%%%%%%%%%%%%%%%%%%%%%%%%%%%%
\chapter{Summary and Future Directions}\label{chap:concl}
%%%%%%%%%%%%%%%%%%%%%%%%%%%%%%%%%%%%%%%%%%%%%%%%%%%%%%%%%%%%%%%%%%%%%%%%%%%%%%%%

Leveraging recent advances in machine learning, deep learning models in conjunction with  statistical signal processing, this thesis pioneered robust, deep, reinforced learning algorithms with applications in management and control of cyber-physical systems. In this final chapter, we provide a summary of the main results discussed in this thesis, and also point out a few possible directions for future research.

%\begin{itemize}
%
%%\item Chapter 1 introduces the analytic goals pursued in this thesis.
%
%\item Chapter 2 briefly presents the history of, and science behind, the
%subjects presented in this thesis.
%
%\item In Chapter 3 the experiment is outlined.
%
%\item Chapter 4 describes the simulation process used in the analysis.
%
%\item Chapter 5 follows the chain of reconstruction software used to obtain
%meaningful results from data.
%
%\item Chapter 6 hashes out the strategy for analysis and presents the data and
%simulated sets that will be used in the analysis.
%
%\item Chapter 7 demonstrates the implementation of the event selection
%processes.
%
%\item In Chapter 8 those events selected in Chapter 7 are analyzed.
%
%\item Chapter 9 presents a final discussion of the analyses presented in the
%thesis.
%
%\end{itemize}
%%%%%%%%%%%%%%%%%%%%%%%%%%%%%%%%%%%%%%%%%%%%%%%%%%%%%%%%%%%%%%%%%%%%%%%%%%%%%%%%
%\section{Background}

\section{Thesis Summary}
Chapter \ref{chap:af} dealt with distributionally robust learning, where the data distribution was considered unknown. A framework to robustify parametric machine learning models against distributional uncertainties was put forth, where the so-called Wasserstin distance metric was used to quantify the distance between training and testing data generating data distributions. 

Chapter \ref{chap:sslovergraphs} explored robust semi-supervised learning over graphs. To account for uncertainties associated with data distributions, or adversarially manipulated input data, a principled robust learning framework was developed. Using the parametric models of graph neural networks (GNNs), we were able to reconstruct the unobserved nodal values. Experiments corroborated the outstanding performance of the novel method when the input data are corrupted.

Chapter \ref{chap:saf} targeted a network resource allocation problem, namely the caching. The idea of caching was to device some entities in a wireed and wireless network with storage capacity. These devices are to store reusable information during off-the peak instances, and then reuse them during on-peak demand periods. By smartly storing popular contents, these devices efficiently help the network to decrease the operational costs and increase user satisfaction. Especially, we designed a generic setup where a caching unit makes sequential fetch-cache decisions based on dynamic content popularities in local section as well as global network. Critical practical constraints were identified, the aggregated cost across files and time instants was formed, and the optimal adaptive caching was then formulated as: i) classical reinforcement learning algorithm; ii) Deep reinforcement learning problem; and, iii) a stochastic optimization problem.  To address the inherent functional estimation problem that arises in each type of considered problems, while leveraging the underlying problem structure, several computationally efficient algorithms were developed. 

Finally, Chapter \ref{chap:sparse} developed a suite of methods to efficiently monitor, and manage the smart grid. Especially, we started with voltage regulation problem using joint control of traditional utility-owned equipment and contemporary smart inverters to inject reactive power. To account for the different response times of those assets, a two-timescale approach to minimizing bus voltage deviations from their nominal values was put forth, by combining physics- and data-driven stochastic optimization. Load consumption and active power generation dynamics were modeled as MDPs. On a fast timescale, the setpoints of smart inverters were found by minimizing the instantaneous bus voltage deviations, while on a slower timescale, the capacitor banks were configured to minimize the long-term expected voltage deviations using a deep reinforcement learning algorithm. Then we considered the second problem of monitoring the smart grid. In particular, topology-aware DNN-based regularizers were developed to deal with the ill-posed and nonconvex characteristics of standard power system state estimation approaches (PSSE).~An alternating minimization solver was developed to approach the solution of the regularized PSSE objective function,~which was further unrolled to construct a DNN model.

\section{Future Research}

The contributions in this thesis opens up a broad range of interesting directions to explore and new problems to solve. Some of such possible research directions are briefly discussed next. \footnote{Due to space limitations, a few works of this PhD thesis have not been reported here, including \cite{yang2019twotrans, sadeghi2017performance, Yang2021relu, li2021heavy,Yan2022}.}

\subsection{Multi-agent, distributionally robust decentralized RL}
We will investigate consensus-based decentralized optimization for our scalable and distributionally robust RL framework, with the ultimate goal of developing safe, multi-agent RL algorithms operating in complex real-world environments. Autonomous driving for instance, is naturally a multi-agent collaborative setting, where the host vehicle (a.k.a. the planner) must apply sophisticated negotiation skills with other road users (agents), when overtaking, giving way, merging, taking left/right turns, or when pushing ahead in unstructured urban roadways. We will also broaden the scope of our multi-step Lyapunov tool, to obtain non-asymptotic performance guarantees of the proposed multi-agent, distributionally robust RL algorithms. We will also corroborate our analytical results extensive experiments.

\subsection{Robust learning approach to fairness in machine learning.}
Fairness-aware machine learning algorithms seek methods under which the predicted outcome of a classifier is fair or non-discriminatory based on sensitive attributes such as race, sex, religion, etc. Broadly, fairness-aware machine learning algorithms have been categorized as those \textit{pre-processing} techniques designed to modify the input data so that the outcome of any machine learning algorithm applied to that data will be fair \cite{romei2014multidisciplinary}. Preprocessing algorithms considers training data as the cause of the discrimination. This simply is because the training data itself captures historical discrimination or since there are more subtle patterns in the data. Feature modification, data set massaging, and learning unbiased data transformation are examples for this class of methods \cite{feldman2015certifying, calmon2017optimized, kamiran2009classifying}. The \textit{algorithm modification} techniques on the other hand modify an existing algorithm or create a new one that will be fair under any inputs. Algorithm modification target specific learning algorithms, e.g., by imposing additional constraints. These methods have been by far the most common methods to promote fairness. Among popular techniques in this class are the regularization techniques, convex relaxation of fairness promoting constraints, and training separate models for each value of a sensitive attribute \cite{Kamishima2012, zafar2015fairness, calders2010three}. Combined preprocessing and algorithm modification methods are among effective methods at classification \cite{zemel2013learning}. Finally, the \textit{post-processing} techniques enforce the output of any model to be fair. These methods modify the results of a previously trained classifier to achieve the desired results on different groups. For example modifying the labels of leaves in a decision tree to satisfy fairness constraints, or modifying error profiles to name a few \cite{kamiran2010discrimination, hardt2016equality, woodworth2017learning}. Despite their success in dealing with some sensitive features, these proposed methods cannot handle setups where the data is coming from unbalanced mixture of distributions where we should protect at least one of classes. For example, consider data is comming from a mixture of distributions, that is $\{{\bf x}_n,y_n\}_{n=1}^{N} \sim P:=\alpha Q_0 + (1- \alpha) Q_{1}$, where $\alpha \in [0,1)$ is subpopulation portion, and $Q_0$ and $Q_1$ are \textit{unknown} subpopulations. The classical learning approaches do not guarantee to ensure \textit{equitable} performance for data from both $Q_0$ and $Q_1$, especially for small $\alpha$. To offer a fair model, we instead focus on the worst-case risk that finds model parameters $\bm{\theta}$ by minimizing the loss over the latent subpopulation $Q_0$
\begin{equation}
\underset{{\bm \theta} \in \Theta}{\operatorname{minimize}}\; \mathbb{E}_{{\bf x} \sim Q_{0}}\mathbb{E}[\ell({\bm \theta};({\bf x}, y))]
\end{equation}
Since $\alpha$ and $Q_0$ are unknown, it is impossible to compute the loss here from observed data. Therefore we postulate a lower bound $\alpha_0 \in (0, 1/2)$ on the subpopulation proportion $\alpha$ and consider a set of potential minority subpopulations $\mathcal{P}_{\alpha_{0}}:=\left\{Q_{0}: P=\alpha Q_{0}+(1-\alpha) Q_{1} \text { for some } \alpha \geq \alpha_{0} \right\}$, and target to solve the worst case loss, formulated as follows
\begin{equation}
\underset{\bm \theta \in \Theta}{\operatorname{minimize}} \sup _{Q_{0} \in \mathcal{P}_{\alpha_{0}}} \mathbb{E}[\ell(\bm \theta ;({\bf x}, y))]. 
\end{equation}

Relying on the proposed techniques described in this T2, we will provide tractable approaches to solve this optimization problem.

\subsection{Communication- and computation-efficient robust federated learning}

The recently growing need to learn from massive datasets that are distributed across multiple sites, has propelled research to replace a single learner with multiple learners (a.k.a. workers) exchanging information with a server to learn the sought learning model while abiding with the privacy of local data. Albeit appealing for its scalability, to endow this so-termed federated learning with robustness too, we must address the server-workers communication overhead that is known to constitute the bottleneck in this setup. This becomes aggravated in deep learning, where one may have to deal with millions of unknown parameters. 

To outline our research outlook in this setting, consider $M$ workers with each worker $m \in \mathcal{M}$ collecting samples $\{\mathbf{z}_t(m)\}_{t=1}^{T}$, and a globally shared model $\bm{\theta}$ that is to be updated at the server by aggregating gradients computed locally per worker. Bandwidth and privacy concerns discourage uploading these distributed data to the server, which necessitates training to be performed by having workers communicating \textit{iteratively} with the server.	To endow such a distributed learning approach with robustness, we will consider solving the following optimization 
problem in a distributed fashion 
\begin{equation}
\underset{\bm{\theta}\in\Theta
}{{\rm min}}\;\,  \sup_{P\in\mathcal{P}}\;\mathbb{E}_{\mathbf{z}\sim P}\!\left[\ell(\mathbf{z};\bm{\theta})
\right],\quad {\textrm{s. to.}}\quad \mathcal{P}:= \Big\{P \Big| \sum_{m=1}^{M} W_c(P,\widehat{P}^{(T)}(m) \le \rho \Big\}
\label{eq:minsupdistr}
\end{equation}  
where $W_c(P,\widehat{P}^{(T)}(m))$ is the distance between $P$ and the locally available distribution $\widehat{P}^{(T)}(m)$ per learner $m$. A critical task here is to efficiently handle the communication overhead while guaranteeing the desired robustness. To this aim, we will develop a general framework building on the considered distributionally robust approach. Tapping into our expertise in communication-efficient decentralized learning and wireless sensor networks \cite{quant2008spincomMschu,quant2008luo, berb2016censor,berb2018censor}, we will develop methods to integrate distributional robustness in a large-scale parallel architecture with communication-friendly learning schemes through \textit{quantization}, and \textit{censoring}. The quantized gradients computed locally at the learners will be transmitted to the server at a controllably low cost; while censoring will save communication costs in learner-server rounds by simply skipping less informative gradients. To maximize the communication efficiency, we will further investigate two-way communication compression, meaning we will compress both the upload and download information to a limited number of bits. It will be interesting to delineate the tradeoffs emerging between robustness, overhead reduction, and convergence rate of the learning iterates. To this end, we will investigate performance both analytically, as well as with thorough numerical tests.

%%%%%%%%%%%%%%%%%%%%%%%%%%%%%%%%%%%%%%%%%%%%%%%%%%%%%%%%%%%%%%%%%%%%%%%%%%%%%%%%

%%%%%%%%%%%%%%%%%%%%%%%%%%%%%%%%%%%%%%%%%%%%%%%%%%%%%%%%%%%%%%%%%%%%%%%%%%%%%%%%
% Bibliography
%%%%%%%%%%%%%%%%%%%%%%%%%%%%%%%%%%%%%%%%%%%%%%%%%%%%%%%%%%%%%%%%%%%%%%%%%%%%%%%%
%\bibliographystyle{IEEEtranS}

%\addcontentsline{toc}{chapter}{Bibliography}
\bibliography{myabrv,apower}

\appendix
%\chapter*{Appendix}
%\chapter*{Appendix}

%%%%%%%%%%%%%%%%%%%%%%%%%%%%%%%%%%%%%%%%%%%%%%%%%%%%%%%%%%%%%%%%%%%%%%%%%%%%%%%%%%%%%%%%%%%%%%%%%%%%%%%%%%%%%%%%%%%%%
%\newpage

%\appendix
%\renewcommand{\thesection}{\Arabic{section}}

\chapter{Proofs for Chapter \ref{chap:af}}

\subsection{Proof of Lemma \ref{lem:smooth}}
\label{app:lemm1}
Since function $\bm \zeta \mapsto \psi(\bar{\bm  \theta}, \bm \zeta; \bm{z})$ is $\lambda$-strongly concave, then $\bm \zeta_{\ast}(\bar{\bm{\theta}}) = \sup_{\bm \zeta \in \mathcal{Z}} \psi(\bar{\bm  \theta}, \bm  \zeta; \bm{z})$ is unique. In addition, the first-order optimality condition gives $\langle \nabla_{\bm \zeta} \psi(\bar{\bm{\theta}}, \bm \zeta_\ast(\bar{\bm{\theta}}); \bm z), \bm{\zeta} - \bm{\zeta}_{\ast}(\bar{\bm{\theta}})\rangle \le 0$. Let us define $\bm \zeta^1_{\ast} = \bm{\zeta}_{\ast}(\bar{\bm{\theta}}_1)$, $\bm \zeta^2_{\ast} = \bm{\zeta}_{\ast}(\bar{\bm{\theta}}_2)$, and use the strong concavity for any $\bar{\bm{\theta}}_1$ and $\bar{\bm{\theta}}_2$, to write
\begin{align}
\nonumber
\psi(\bar{\bm \theta}_2, \bm \zeta^2_{\ast}; \bm{z}) 
& 
\le 
\psi(\bar{\bm \theta}_2, \bm \zeta^1_{\ast}; \bm{z}) + \langle \nabla_{\bm \zeta} \psi(\bar{\bm \theta}_2, \bm \zeta^1_{\ast}; \bm{z}), \bm \zeta^2_{\ast} - \bm \zeta^1_{\ast} \rangle   
\\
& 
\quad -\frac{\lambda}{2} \|\bm \zeta^2_{\ast} -  \bm \zeta^1_{\ast} \|^2 \label{app:lema1}
\end{align}
and
\begin{align}
\nonumber
\psi(\bar{\bm \theta}_2, \bm \zeta^1_{\ast}; \bm{z}) 
& \le \psi(\bar{\bm \theta}_2, \bm \zeta^2_{\ast}; \bm{z}) +  \langle \nabla_{\bm \zeta} \psi(\bar{\bm \theta}_2, \bm \zeta^2_{\ast}; \bm{z}), \bm \zeta^1_{\ast} - \bm \zeta^2_{\ast} \rangle 
\\
& \quad 
-\frac{\lambda}{2} \|\bm \zeta^2_{\ast} - \bm \zeta^1_{\ast} \|^2 \nonumber \\
& \le  \psi(\bar{\bm \theta}_2, \bm \zeta^2_{\ast}; \bm{z}) -\frac{\lambda}{2} \|\bm \zeta^2_{\ast} - \bm \zeta^1_{\ast} \|^2 \label{app:lema2}
\end{align}

where the last inequality is a consequence of the first-order optimality condition. Summing \eqref{app:lema1} and \eqref{app:lema2}, we find that
\begin{align}
\lambda \|\bm \zeta^2_{\ast} - \bm \zeta^1_{\ast} \|^2  & \le \langle \nabla_{\bm \zeta} \psi(\bar{\bm \theta}_2, \bm \zeta^1_{\ast}; \bm{z}), \bm \zeta^2_{\ast} - \bm \zeta^1_{\ast} \rangle \nonumber \\ 
& \le \langle \nabla_{\bm \zeta} \psi(\bar{\bm \theta}_2, \bm \zeta^1_{\ast}; \bm{z}), \bm \zeta^2_{\ast} - \bm \zeta^1_{\ast} \rangle 
\\
&
\quad 
-   \langle \nabla_{\bm \zeta}\psi(\bar{\bm \theta}_1, \bm \zeta^1_{\ast}; \bm{z}), \bm \zeta^2_{\ast} - \bm \zeta^1_{\ast} \rangle \nonumber \\ 
& = \langle \nabla_{\bm \zeta} \psi(\bar{\bm \theta}_2, \bm \zeta^1_{\ast}; \bm{z}) -\nabla_{\bm \zeta}\psi(\bar{\bm \theta}_1, \bm \zeta^1_{\ast}; \bm{z}) , \bm \zeta^2_{\ast} - \bm \zeta^1_{\ast} \rangle.  
\end{align}
And using H\"older's inequality, we obtain that
\begin{align}
\nonumber
\lambda \|\bm \zeta^2_{\ast}  - \bm \zeta^1_{\ast} \|^2  
& 
\le  \| \nabla_{\bm \zeta} \psi(\bar{\bm \theta}_2, \bm \zeta^1_{\ast}; \bm{z}) -\nabla_{\bm \zeta}\psi(\bar{\bm \theta}_1, \bm \zeta^1_{\ast}; \bm{z}) \|_{\star} \\
& 
\quad \times \|\bm \zeta^2_{\ast} - \bm \zeta^1_{\ast} \|
\end{align}
from which we deduce 
\begin{align}
\|\bm \zeta^2_{\ast} - \bm \zeta^1_{\ast} \| \le \frac{1}{\lambda} \| \nabla_{\bm \zeta} \psi(\bar{\bm \theta}_2, \bm \zeta^1_{\ast}; \bm{z}) -\nabla_{\bm \zeta}\psi(\bar{\bm \theta}_1, \bm \zeta^1_{\ast}; \bm{z}) \|_{\star}. \label{applem:xlimits}
\end{align}
Using $\psi(\bar{\bm{\theta}}, \bm{\zeta}; \bm{z}):= \ell(\bm{\theta}; \bm{\zeta}) + \gamma(\rho - c(\bm{z}, \bm{\zeta}))$, we have that
\begin{align}
& 
\| \nabla_{\bm \zeta} \psi(\bar{\bm \theta}_2,  \bm \zeta^1_{\ast}; \bm{z})  -\nabla_{\bm \zeta}\psi(\bar{\bm \theta}_1,   \bm \zeta^1_{\ast}; \bm{z}) \|_{\star} \nonumber \\ 
&
=\, \| \nabla_{\bm \zeta} \ell(\bm \theta_2; \bm \zeta^1_\ast) - \nabla_{\bm{\zeta}} \ell(\bm \theta_1; \bm \zeta^1_\ast)  
\\
& \quad
+ \gamma_1 \nabla_{\bm{\zeta}}  c(\bm z, \bm \zeta^1_\ast) - \gamma_2 \nabla_{\bm{\zeta}}  c(\bm z, \bm \zeta^1_\ast) \|_{\star} \nonumber \\
&
\le\,  \| \nabla_{\bm \zeta} \ell(\bm \theta_2; \bm \zeta^1_\ast) - \nabla_{\bm{\zeta}} \ell(\bm \theta_1; \bm \zeta^1_\ast) \|_\star  
\\
& \quad
+ \| \gamma_1 \nabla_{\bm{\zeta}}  c(\bm z, \bm \zeta^1_{\ast}) - \gamma_2 \nabla_{\bm{\zeta}}  c(\bm z, \bm \zeta^1_{\ast}) \|_{\star} \nonumber \\ 
&
\le\,  L_{\bm{z}\bm{\theta}} \|\bm{\theta}_2 -\bm{\bm{\theta}}_1 \| + \| \nabla_{\bm \zeta}  c(\bm z, \bm \zeta^1_{\ast}) \|_\star \, \| \gamma_2 -\gamma_1 \| \label{applem:crosslipsch}.
\end{align}
Substituting \eqref{applem:crosslipsch} into \eqref{applem:xlimits}, yields
\begin{align}
\|\bm \zeta^2_{\ast} - \bm \zeta^1_{\ast} \| 
& \le \frac{L_{\bm{z}\bm{\theta}}}{\lambda} \|\bm{\theta}_2 -\bm{\bm{\theta}}_1 \| + \frac{1}{\lambda} \| \nabla_{\bm \zeta}  c(\bm z, \bm \zeta^1_{\ast})\|_\star \, \| \gamma_2 -\gamma_1 \|  \nonumber  \\ 
& \le \frac{L_{\bm{z}\bm{\theta}}}{\lambda}  \|\bm{\theta}_2 -\bm{\bm{\theta}}_1 \| + \frac{L_c}{\lambda}\, \| \gamma_2 -\gamma_1 \| \label{app:lemafirtineql}
\end{align}
where the last inequality holds because $\bm{\zeta} \mapsto c(\bm{z},\bm{\zeta})$ is $L_c$-Lipschitz as per Assumption \ref{as:cstrongcvx}.  

To obtain \eqref{eq:lemmapsi}, we first suppose without loss of generality that only a single datum $\bm{z}$ is given, and in order to prove existence of the gradient of $\bar{\psi}(\bar{\bm{\theta}}, \bm{z})$ with respect to $\bar{\bm{\theta}}$, we resort to the Danskin's theorem as follows.  

\emph{Danskin's Theorem \cite{danskin1966theory}.} Consider the following minimax optimization problem
\begin{equation}
\min_{\bm{\theta}\in\Theta} \max_{\bm{\zeta}\in\mathcal{X}} f(\bm{\theta},\bm{\zeta}) 
\end{equation}
where $\mathcal{X}$ is a nonempty compact set, and $f: \Theta \times {\mathcal{X}} \to [0,\infty)$ is such that $f(\cdot, \bm{\zeta})$ is differentiable for any $\bm{\zeta} \in \mathcal{X}$, and $\nabla_{\bm{\theta}} f(\bm{\theta},\bm{\zeta})$ is continuous on $\Theta \times \mathcal{X}$. With $\mathcal{S(\bm{\theta})}:=\{\bm{\zeta}_\ast| \bm{\zeta}_\ast=\arg \max_{\bm{\zeta}} f(\bm{\theta},\bm{\zeta})\}$, the function 
\begin{equation*}
\bar{f}(\bm{\theta}):=\max_{\bm{\zeta}\in\mathcal{Z}} f(\bm{\theta},\bm{\zeta})
\end{equation*}
is locally Lipschitz and directionally differentiable, where the directional derivatives satisfy
\begin{equation}
\bar{f}(\bm{\theta}, \bm{d}) = \sup_{\bm{\zeta}\in \mathcal{S(\bm{\theta})}} \langle \bm{d}, \nabla_{\bm{\theta}}f(\bm{\theta},\bm{\zeta}) \rangle.
\end{equation}
For a given $\bm{\theta}$, if the set $\mathcal{S}(\bm{\theta})$ is a singleton, then the function $\bar{f}(\bm{\theta})$ is differentiable at $\bm{\theta}$ with gradient 
\begin{equation}
\nabla_{\bm{\theta}} \bar{f}(\bm{\theta})= \nabla_{\bm{\theta}} f(\bm{\theta},\bm{\zeta}_\ast(\bm{\theta})).
\end{equation}
Given $\bm{\theta}$, and the $\mu$-strongly convex $c(\bm{z},\cdot)$, function $\psi(\bar{\bm{\theta}}, \cdot; \bm{z})$ is concave if $L_{\bm{z z}} - \gamma {\mu} <0$, which holds true for $\gamma_0> {L_{\bm{zz}}}/{\mu}$.
%This holds true for a cost $c(\bm{z},\cdot)$ with large enough $\mu \gg 1$.  
Replacing $\bar{f}(\bm{\theta},\bm{\zeta})$ with $\psi(\bar{\bm{\theta}}, \bm \zeta; \bm{z})$, and given the concavity of $\bm \zeta \mapsto \psi(\bar{\bm{\theta}}, \bm \zeta; \bm{z})$, we have that $\bar{\psi}(\bar{\bm{\theta}}; \bm{z})$ is a continuous function with gradient
\begin{equation}
\nabla_{\bar{\bm{\theta}}} \bar{\psi}(\bar{\bm{\theta}};\bm{z})= \nabla_{\bar{\bm{\theta}}} \bar{\psi}(\bar{\bm{\theta}}, \bm{\zeta}_\ast(\bar{\bm{\theta}};\bm{z});\bm{z}).
\end{equation}
We can then obtain the second inequality, as 
\begin{align}
&
\| \nabla_{\bar{\bm{\theta}}} \psi(\bar{\bm{\theta}}_1, \bm{\zeta}^1_\ast;\bm{z}) - \nabla_{\bar{\bm{\theta}}} \psi(\bar{\bm{\theta}}_2, \bm{\zeta}^2_\ast;\bm{z}) \| \nonumber \\ 
\le\,
& 
\| \nabla_{\bar{\bm{\theta}}} \psi(\bar{\bm{\theta}}_1, \bm{\zeta}^1_\ast;\bm{z}) - \nabla_{\bar{\bm{\theta}}} \psi(\bar{\bm{\theta}}_1, \bm{\zeta}^2_\ast;\bm{z}) \|   
\nonumber \\
& \quad  
+ \| \nabla_{\bar{\bm{\theta}}} \psi(\bar{\bm{\theta}}_1, \bm{\zeta}^2_\ast;\bm{z}) - \nabla_{\bar{\bm{\theta}}} \psi(\bar{\bm{\theta}}_2, \bm{\zeta}^2_\ast;\bm{z}) \|  \\ 
\le\,
&
\left\| \begin{bmatrix}
\nabla_{\bm{\theta}} \ell(\bm{\theta}_1,\bm{\zeta}^1_\ast) - \nabla_{\bm{\theta}} \ell(\bm{\theta}_1,\bm{\zeta}^2_\ast) \\ c(\bm{z},\bm{\zeta}^2_\ast) - c(\bm{z},\bm{\zeta}^1_\ast)
\end{bmatrix} \right\|    
\nonumber \\
& \quad +
\left\| \begin{bmatrix}
\nabla_{\bm{\theta}} \ell(\bm{\theta}_1,\bm{\zeta}^2_\ast) - \nabla_{\bm{\theta}} \ell(\bm{\theta}_2,\bm{\zeta}^2_\ast) \\ 
0
\end{bmatrix} \right\| \\ 
\le\,
& L_{\bm{\theta z}} \|\bm{\zeta}^1_\ast - \bm{\zeta}^2_\ast \| + L_c \|\bm{\zeta}^1_\ast - \bm{\zeta}^2_\ast \| + L_{\bm{\theta \theta}} \|\bm{\theta_1} -\bm{\theta_2}\|  \nonumber \\ 
\le\,
& 
\Big(L_{\bm{\theta \theta}}+\frac{L_{\bm \theta z }L_{\bm{z}\bm{\theta}} + L_c L_{\bm{z}\bm{\theta}}}{\lambda}\Big)  \|\bm{\theta}_2 -\bm{\bm{\theta}}_1 \| 
\nonumber \\
& \quad 
+ \frac{ L_{\bm \theta z} L_c + L^2_c}{\lambda}\, \| \gamma_2 -\gamma_1 \|
\end{align}
where we again used inequality \eqref{app:lemafirtineql}. As a technical note, if the considered model is a neural network with a non-smooth activation function, the loss will not be continuously differentiable. However, we will not encounter this challenge often in practice. 

\subsection{Proof of Theorem \ref{thm:convspgd}}
\label{app:thm1}
With slight abuse of notation, define for convenience $F(\bm \theta, \gamma) := f(\bm \theta, \gamma) + r(\bm \theta) + h(\gamma) $, where $h(\gamma)$ is the indicator function  
\begin{equation}
h(\gamma) =\begin{cases}
0, & \text{  if } \gamma \in  \Gamma \\ 
\infty,& \text{ if } \gamma \notin  \Gamma 
\end{cases}
\label{eq:gamind}
\end{equation} with
$\Gamma := \{\gamma| \gamma \ge \gamma_0\}$, and for ease of representation we use $\bar{r}(\bar{\bm \theta}) := r(\bm \theta) + h(\gamma)$. Having an $L_f$--smooth function $f$, yields
\begin{align} 
f(\bar{\bm \theta}^{t+1}) &\le f(\bar{\bm \theta}^{t}) + \big \langle \nabla f(\bar{\bm \theta}^t), \bar{\bm \theta}^{t+1} - \bar{\bm \theta}^{t} \big \rangle + \frac{L_f}{2} \| \bar{\bm \theta}^{t+1} - \bar{\bm \theta}^{t} \|^2. \label{eq:smoothf1} 
\end{align}
For a given ${\bm z}^t$, the gradients are 

\begin{align*}
\bm g^\ast(\bar{\bm \theta}^t) := & \begin{bmatrix}
\nabla _{\bm  \theta}\psi(\bm  {\theta}^t, \gamma, {\bm \zeta}_\ast(\bar{\bm \theta}^t;\bm{z}^t);\bm{z}^t)
\\ 
\partial_\gamma \psi(\bm  {\theta}^t, \gamma, {\bm \zeta}_\ast(\bar{\bm \theta}^t;\bm{z}^t);\bm{z}^t)
\end{bmatrix} 
\\
= & \begin{bmatrix}
\nabla _{\bm  \theta}\psi(\bm  {\theta}^t, \gamma, {\bm \zeta}_\ast(\bar{\bm \theta}^t;\bm{z}^t);\bm{z}^t)
\\ 
\rho- c(\bm  z^t, \bm  {\bm \zeta}_\ast(\bar{\bm \theta}^t;\bm{z}^t)
\end{bmatrix}.
%\label{eq:gepsdef}
\end{align*}
and 
\begin{align*}
\bm  g^\epsilon(\bar  {\bm \theta}^t) := & \begin{bmatrix}
\nabla _{\bm  \theta}\psi(\bm  {\theta}^t, \gamma, {\bm \zeta}_\epsilon(\bar{\bm \theta}^t;\bm{z}^t);\bm{z}^t)
\\ 
\partial_\gamma \psi(\bm  {\theta}^t, \gamma, {\bm \zeta}_\epsilon(\bar{\bm \theta}^t;\bm{z}^t);\bm{z}^t)
\end{bmatrix}  
\\
= & 
\begin{bmatrix}
\nabla _{\bm  \theta}\psi(\bm  {\theta}^t, \gamma, {\bm \zeta}_\epsilon(\bar{\bm \theta}^t;\bm{z}^t);\bm{z}^t)
\\ 
\rho- c(\bm  z^t, \bm  {\bm \zeta}_\epsilon(\bar{\bm \theta}^t;\bm{z}^t)
\end{bmatrix}
%\label{eq:gepsdef}
\end{align*}
obtained by an oracle at the optimal $\bm \zeta_\ast$ and the $\epsilon$-optimal $\bm \zeta_\epsilon$ solvers, respectively.  
Now, we define the error vector ${\bm \delta}(\bar{\bm \theta}^t) := \nabla f(\bar{\bm \theta}^t) -\bm  g^{\epsilon}(\bar{\bm \theta}^t)$, and replace this into \eqref{eq:smoothf1}, to obtain
\begin{align}
\nonumber 
f(\bar{\bm \theta}^{t+1}) & \le f(\bar{\bm \theta}^{t}) + \big\langle \bm  g^\epsilon(\bar{\bm \theta}^t)+ \bm  \delta(\bar{\bm \theta}^t), \bar{\bm \theta}^{t+1} - \bar{\bm \theta}^{t} \big\rangle  
\\
& \quad + \frac{L_f}{2} \big\| \bar{\bm \theta}^{t+1} - \bar{\bm \theta}^{t} \big\|^2. \label{eq:f_uprbnd1}
\end{align}

The following properties hold equivalently for the proximal operator, and for any $\bm x, \bm y$ 
\begin{align}
\label{eq:prox_prop}
\bm  u = \textrm{prox}_{\alpha r} (\bm  x)  \iff \big\langle \bm  x-\bm  u,  \bm  y - \bm  u \big \rangle \le \alpha r(\bm  y) - \alpha r(\bm  u). 
\end{align}
With $\bm  u=\bar{\bm \theta}^{t+1}$ and $\bm  x =\bar{\bm \theta}^t - \alpha_t \bm  g^\epsilon (\bar {\bm \theta}^t)$ in \eqref{eq:prox_prop}, it holds that 
\begin{equation*}
\big\langle \bar{\bm \theta}^t - \alpha_t \bm  g^\epsilon (\bar{\bm \theta}^t) - \bar{\bm \theta}^{t+1}, \bar{\bm \theta}^t - \bar{\bm \theta}^{t+1} \big\rangle \le \alpha_t \bar{r}(\bar{\bm \theta}^t) - \alpha_t \bar{r}(\bar{\bm \theta}^{t+1})
\end{equation*}
and upon rearranging, we obtain 
\begin{equation}
\big \langle \bm  g^\epsilon (\bar{\bm \theta}^t), \bar{\bm \theta}^{t+1} - \bar{\bm \theta}^t \big\rangle \le \bar{r}(\bar{\bm \theta}^t) - \bar{r}(\bar{\bm \theta}^{t+1}) - \frac{1}{\alpha_t} \big\| \bar{\bm \theta}^{t+1} - \bar{\bm \theta}^{t}\big\|^2.
\label{eq:gepsupperbound_1}
\end{equation}
Adding inequalities in \eqref{eq:gepsupperbound_1} and \eqref{eq:f_uprbnd1} gives
\begin{align*}  
f(\bar{\bm \theta}^{t+1}) & \le f(\bar{\bm \theta}^{t}) + \big \langle \bm  \delta(\bar{\bm \theta}^t), \bar{\bm \theta}^{t+1} - \bar{\bm \theta}^{t}\big \rangle + \frac{L_f}{2} \big\| \bar{\bm \theta}^{t+1} - \bar{\bm \theta}^{t}\big\|^2
\\
& \quad   + \bar{r}(\bar{\bm \theta}^t) - \bar{r}(\bar{\bm \theta}^{t+1}) - \frac{1}{\alpha_t} \big\|\bar{\bm \theta}^{t+1} - \bar{\bm \theta}^{t}\big\|^2 
\end{align*}
and with  $F(\bar{\bm{\theta}}):= f(\bar{\bm{\theta}})+\bar{r}(\bar{\bm{\theta}})$, we can write
\begin{align} \nonumber 
F(\bar{\bm \theta}^{t+1}) -  & F(\bar{\bm \theta}^{t}) \le \big \langle \bm  \delta(\bar{\bm \theta}^t), \bar{\bm \theta}^{t+1} - \bar{\bm \theta}^{t} \big\rangle 
\\
& \quad + \Big(\frac{L_f}{2}- \frac{1}{\alpha_t}\Big) \big\| \bar{\bm \theta}^{t+1} - \bar{\bm \theta}^{t}\big\|^2.
\end{align}
Using Young's inequality for any $\eta>0$ gives $\big \langle \bm  \delta(\bar{\bm \theta}^t), \bar{\bm \theta}^{t+1} - \bar{\bm \theta}^{t} \big\rangle \le \frac{\eta}{2} \| \bar{\bm \theta}^{t+1} - \bar{\bm \theta}^{t}\|^2 +   \frac{1}{2\eta} \| \delta(\bar{\bm \theta}^t)\|^2$, and hence     

\begin{align} 
F(\bar{\bm  \theta}^{t+1}) -  & F(\bar{\bm  \theta}^{t}) \le  \Big(\frac{L_f + \eta}{2}\!-\! \frac{1}{\alpha_t}\Big) \big\|\bar{\bm  \theta}^{t+1} \!-\! \bar{\bm  \theta}^{t}\big\|^2 + \frac{\big\|\bm  \delta(\bar{\bm  \theta}^t)\big\|^2}{2 \eta}
\label{eq:bnd_itrsF1}.
\end{align}

Next, we will bound $\bm\delta(\bar{\bm  \theta}^t) := \nabla f (\bar{\bm  \theta}^t) - \bm g^\epsilon(\bar{\bm  \theta}^t)$. By adding and subtracting $\bm g^\ast(\bar{\bm  \theta}^t)$ to the right hand side, we find 
\begin{equation}
\label{eq:bounddelta1}
\|\bm\delta(\bar{\bm  \theta}^t)\|^2 \le 2 \|\nabla f (\bar{\bm  \theta}^t) - \bm g^\ast(\bar{\bm  \theta}^t)\|^2 + 2 \|\bm g^\ast(\bar{\bm  \theta}^t) - \bm g^\epsilon(\bar{\bm  \theta}^t)\|^2. 
\end{equation}
The Lipschitz smoothness of the gradient, implies that
\begin{align}
& \big\|\bm g^\ast(\bar{\bm  \theta}^t) - \bm g^\epsilon(\bar{\bm  \theta}^t)\big\|^2  
\\
=\, &\Bigg\| \begin{bmatrix}
\nabla _{\bm \theta}\psi(\bm {\theta}^t, \gamma, {\bm \zeta}_\ast(\bar{\bm  \theta}^t;\bm{z}^t);\bm{z}^t)
\\ 
\rho- c(\bm z^t, \bm {\bm \zeta}_\ast(\bar{\bm  \theta}^t;\bm{z}^t)
\end{bmatrix} - \begin{bmatrix}
\nabla _{\bm \theta}\psi(\bm {\theta}^t, \gamma, {\bm \zeta}_\epsilon(\bar{\bm  \theta}^t;\bm{z}^t);\bm{z}^t)
\\ 
\rho- c(\bm z^t, \bm {\bm \zeta}_\epsilon(\bar{\bm  \theta}^t;\bm{z}^t)
\end{bmatrix} \Bigg\|^2 \nonumber \\ 
=\,& \big \|\nabla _{\bm \theta}\psi(\bm {\theta}^t, \gamma, {\bm \zeta}_\ast(\bar{\bm  \theta}^t;\bm{z}^t);\bm{z}^t)- \nabla _{\bm \theta}\psi(\bm {\theta}^t, \gamma, {\bm \zeta}_\epsilon(\bar{\bm  \theta}^t;\bm{z}^t);\bm{z}^t) \big \|^2  \nonumber \\ 
&   + \big \| c(\bm z^t, \bm \zeta^t_\ast) - c(\bm z^t,\bm \zeta^t_\epsilon) \big \|^2 \nonumber \\ 
\overset{(\rm a)}{\le} \,& \Big(\frac{ L_{\bm \theta \bm  z}^2  }{\lambda^t}+L_c\Big) \|\bm \zeta^t_\ast-\bm \zeta^t_\epsilon \|^2  \nonumber \\ 
\overset{(\rm b)}{\le} \,& \Big(\frac{ L_{\bm \theta \bm  z}^2  }{\lambda^t}+L_c \Big) \epsilon \nonumber \\
\le\,& \Big(\frac{ L_{\bm \theta \bm  z}^2  }{\lambda_{0}}+L_c \Big)\epsilon \label{eq:boundg1}
\end{align}
where $(\rm a)$ uses the $\lambda^t = \mu \gamma^t -L_{\bm z \bm z}$ strong-concavity of ${\bm \zeta}\mapsto \psi(\bar{\bm \theta}, \gamma, \bm{\zeta}; \bm z)$, and the second term is bounded by $L_c \|\bm \zeta_{\ast}^t - \bm \zeta_{\epsilon}^t\|^2$ according to Assumption \ref{as:cstrongcvx}. The last inequality holds for $\lambda_{0}:= \mu \gamma_0 - L_{\bm{zz}}$, where we used \eqref{eq:gamind} to bound $\gamma^t \ge \gamma_{0} > L_{\bm z\bm z}$. So far, we have established that
\begin{equation}
\|\bm g^\ast(\bar{\bm  \theta}^t) - \bm g^\epsilon(\bar{\bm  \theta}^t)\|^2 \le \frac{ L_{\bar{\bm  \theta} \bm  z}^2  \epsilon}{\lambda_0}
\label{eq:ggbound}
\end{equation}

where for notational convenience we let  $ L_{\bar{\bm \theta} \bm   z}^2  := L_{\bm  \theta \bm  z}^2 + \lambda_0 L_c$. Substituting \eqref{eq:ggbound} into \eqref{eq:bounddelta1}, the error can be bounded as 
\begin{equation}
\label{eq:bound_delta2}
\|\bm\delta(\bar{\bm \theta}^t)\|^2 \le 2 \| \nabla f (\bar{\bm \theta}^t) - \bm g^\ast(\bar{\bm \theta}^t)\|^2 + \frac{ 2 L_{\bar{\bm \theta} \bm  z}^2  \epsilon}{\lambda_0}.
\end{equation}
Combining \eqref{eq:bnd_itrsF1} and \eqref{eq:boundg1} yields
\begin{align}  
F(\bar{\bm \theta}^{t+1}) -  F(\bar{\bm \theta}^{t}) 
& 
\le \Big( \frac{L_f+\eta}{2}- \frac{1}{\alpha_t}\Big) \big\|\bar{\bm \theta}^{t+1} - \bar{\bm \theta}^{t}\big\|^2 
\\ 	\nonumber 
& \quad 
+  \frac{1}{\eta} \big\|\nabla f (\bar{\bm \theta}^t) - \bm g^\ast(\bar{\bm \theta}^t)\big\|^2 + \frac{ L_{\bar{\bm \theta} \bm  z}^2  \epsilon}{\eta \lambda_0}.
\end{align}
Considering a constant step size $\alpha$ and summing these inequalities over $t= 1, \ldots, T$ yields
\begin{align} 
& 
\Big( \frac{1}{\alpha} - \frac{L_f + \eta}{2}\Big) \sum_{t = 0}^{T} \big\|\bar{\bm \theta}^{t+1} - \bar{\bm \theta}^{t}\big\|^2  \le F(\bar{\bm \theta}^{0}) -  F(\bar{\bm \theta}^{T})  
\nonumber \\ 
& \quad +  \frac{1}{\eta} \sum_{t=0}^{T} \big\|\nabla f (\bar{\bm \theta}^t) - \bm g^\ast(\bar{\bm \theta}^t)\big\|^2 + \frac{ (T+1) L_{\bar{\bm \theta} \bm  z}^2  \epsilon}{\lambda_0}.
\label{eq:boundtheta1}
\end{align}
From the proximal gradient update
\begin{equation}
\bar{\bm \theta}^{t+1} = \arg \min_{\bm  \theta} \, \alpha \bar{r}(\bm  \theta) + \alpha \big\langle \bm  \theta - \bar{\bm \theta}^t, \bm g^\epsilon (\bar{\bm \theta}^t) \big \rangle + \frac{1}{2} \big\|\bm  \theta - \bar{\bm \theta}^{t}\big\|^2 \label{eqapp:proxupdate}
\end{equation}
the optimality of $\bar{\bm \theta}^{t+1}$ in \eqref{eqapp:proxupdate}, implies that
\begin{equation*}
\bar{r}(\bar{\bm \theta}^{t+1}) + \big\langle \bar{\bm \theta}^{t+1} - \bar{\bm \theta}^t, \bm g^\epsilon (\bar{\bm \theta}^t) \big\rangle + \frac{1}{2 \alpha} \big\|\bar{\bm \theta}^{t+1} - \bar{\bm \theta}^{t}\big\|^2 \le \bar{r}(\bar{\bm \theta}^{t})
\end{equation*}
which combined with the smoothness of $f$ (c.f. \eqref{eq:smoothf1}) yields
%\begin{equation}
%f(\bar{\bm \theta}^{t+1}) \le f(\bar{\bm \theta}^{t}) + \big\langle \nabla f(\bar{\bm \theta}^t), \bar{\bm \theta}^{t+1} \! - \bar{\bm \theta}^{t} \big \rangle + \frac{L_f}{2} \|\bar{\bm \theta}^{t+1} - \bar{\bm \theta}^{t}\|^2 \nonumber
%\end{equation}
\begin{align}
& \big\langle \bar{\bm \theta}^{t+1} - \bar{\bm \theta}^t , \bm g^\epsilon (\bar{\bm \theta}^t) - \nabla f(\bar{\bm \theta}^t) \big\rangle +  \Big(\frac{1}{2 \alpha} - \frac{L_f}{2}\Big) \big\|\bar{\bm \theta}^{t+1} - \bar{\bm \theta}^{t}\big\|^2  
\nonumber \\ 
& \qquad \le F(\bar{\bm \theta}^{t}) - F(\bar{\bm \theta}^{t+1}) 
\nonumber
\end{align}
Subtracting $\langle \bar{\bm \theta}^{t+1} - \bar{\bm \theta}^t, \nabla f(\bar{\bm \theta}^{t+1}) \rangle$ from both sides gives
\begin{align*}
& 
\big\langle \bar{\bm \theta}^{t+1}\! -   
\,	\bar{\bm \theta}^t,  \bm g^\epsilon (\bar{\bm \theta}^t)  - \nabla f(\bar{\bm \theta}^{t+1}) \big \rangle 
+  \Big(\frac{1}{2 \alpha} \! -\frac{L_f}{2}\Big) \big\|\bar{\bm \theta}^{t+1} - \bar{\bm \theta}^{t}\big\|^2 \nonumber  \\ 
& \qquad
\le\,
F(\bar{\bm \theta}^{t}) - F(\bar{\bm \theta}^{t+1}) -  \big\langle  \bar{\bm \theta}^{t+1} \!- \bar{\bm \theta}^t, \nabla f (\bar{\bm \theta}^{t+1}) - \nabla f(\bar{\bm \theta}^t) \big\rangle.
\end{align*}
Considering $\big\|\bm g^\epsilon (\bar{\bm \theta}^t) - \nabla f(\bar{\bm \theta}^{t+1}) + \frac{1}{\alpha} (\bar{\bm \theta}^{t+1} - \bar{\bm \theta}^t)\big\|^2$ on the left hand side, and adding relevant terms to the right hand side, we arrive at
\begin{align}
& \Big\|\bm g^\epsilon (\bar{\bm \theta}^t) - \nabla f(\bar{\bm \theta}^{t+1}) + \frac{1}{\alpha} (\bar{\bm \theta}^{t+1} - \bar{\bm \theta}^t)\Big\|^2 \nonumber \\
\le\, 
&   \big\|\bm g^\epsilon (\bar{\bm \theta}^t) - \nabla f(\bar{\bm \theta}^{t+1})\big\|^2 
+ \frac{1}{\alpha^2} \big\|\bar{\bm \theta}^{t+1} - \bar{\bm \theta}^t\big\|^2 
\\
& \quad   + \Big(\frac{L_f}{\alpha} - \frac{1}{\alpha^2}\Big)\big\|\bar{\bm \theta}^{t+1} - \bar{\bm \theta}^{t}\big\|^2 + \frac{2}{\alpha} \big(F(\bar{\bm \theta}^{t}) - F(\bar{\bm \theta}^{t+1})\big)
\nonumber \\ 
&  \quad  - \frac{2}{\alpha}\big\langle  \bar{\bm \theta}^{t+1} - \bar{\bm \theta}^t, \nabla f (\bar{\bm \theta}^{t+1}) - \nabla f(\bar{\bm \theta}^t) \big\rangle  \nonumber \\ 
\le \,
&  \big\|\bm g^\epsilon (\bar{\bm \theta}^t) - \nabla f(\bar{\bm \theta}^{t})\|^2 + \frac{1}{\alpha^2} \big\|\bar{\bm \theta}^{t+1} - \bar{\bm \theta}^t\big\|^2  
\\
& \quad 
+ \Big(\frac{L_f}{\alpha} - \frac{1}{\alpha^2}\Big) \big\|\bar{\bm \theta}^{t+1} - \bar{\bm \theta}^{t}\big\|^2 + \frac{2}{\alpha}\big(F(\bar{\bm \theta}^{t}) - F(\bar{\bm \theta}^{t+1})\big) \nonumber \\ 
& \quad - \frac{2}{\alpha} \big\langle \bar{\bm \theta}^{t+1} - \bar{\bm \theta}^t), \nabla f (\bar{\bm \theta}^{t+1}) - \nabla f(\bar{\bm \theta}^t) \big \rangle \nonumber \\
\le \,
&  \big\|\bm g^\epsilon (\bar{\bm \theta}^t) - \nabla f(\bar{\bm \theta}^{t})\|^2  + \frac{1}{\alpha^2} \big\|\bar{\bm \theta}^{t+1} - \bar{\bm \theta}^{t}\big\|^2 
\\
& \quad 
+ \Big(\frac{L_f}{\alpha} - \frac{1}{\alpha^2}\Big) \big\|\bar{\bm \theta}^{t+1} - \bar{\bm \theta}^{t}\big\|^2 + \frac{2}{\alpha}\big(F(\bar{\bm \theta}^{t}) - F(\bar{\bm \theta}^{t+1})\big) \nonumber \\ 
& \quad + \frac{\eta}{\alpha} \big\|\bar{\bm \theta}^{t+1} - \bar{\bm \theta}^{t}\big\|^2 +\frac{L_f^2}{\eta} \big\|\bar{\bm \theta}^{t+1} - \bar{\bm \theta}^{t}\big\|^2
\nonumber
\end{align}
where the last inequality is obtained by applying Young's inequality, and then using the $L_f$-Lipschitz continuity of~$f(\cdot)$. By simplifying the last inequality, we obtain
\begin{align}
&  
\Big\|\bm g^\epsilon (\bar{\bm \theta}^t) - \nabla f(\bar{\bm \theta}^{t+1}) + \frac{1}{\alpha} (\bar{\bm \theta}^{t+1} \!\!\!\! - \bar{\bm \theta}^t)\Big\|^2  
\!\!\!
\le \big\|\bm g^\epsilon (\bar{\bm \theta}^t) - \nabla f(\bar{\bm \theta}^{t})\big\|^2 
\nonumber 
\\
& 
+ \frac{2}{\alpha}\big(F(\bar{\bm \theta}^{t}) - F(\bar{\bm \theta}^{t+1})\big) 
+ \Big(\frac{L_f^2}{\eta} + \frac{L_f+\eta}{\alpha} \Big) \big\|\bar{\bm \theta}^{t+1} - \bar{\bm \theta}^{t}\big\|^2.
\nonumber 
\end{align}
The first term in the right hand side can be bounded by adding and subtracting $\bm{g}^\ast(\bar{\bm{\theta}}^t)$ and using \eqref{eq:ggbound}, to arrive at 
\begin{align}
& \nonumber
\Big\|\bm g^\epsilon  (\bar{\bm \theta}^t) - \nabla f(\bar{\bm \theta}^{t+1}) + \frac{1}{\alpha} (\bar{\bm \theta}^{t+1}   - \bar{\bm \theta}^t)\Big\|^2 
\\
& \le 2 \big\|\nabla f (\bar{\bm \theta}^t) - \bm g^\ast(\bar{\bm \theta}^t)\big\|^2     \frac{2 L_{\bar{\bm \theta}}^2 \epsilon }{\lambda_0}  +  \frac{2}{\alpha}\big(F(\bar{\bm \theta}^{t}) - F(\bar{\bm \theta}^{t+1})\big) \nonumber \\ 
& \quad + \Big(\frac{L_f^2}{\eta} + \frac{L_f+\eta}{\alpha} \Big) \big\|\bar{\bm \theta}^{t+1} - \bar{\bm \theta}^{t}\big\|^2. \label{eq:subgradientbound}
\end{align}
Summing these inequalities over $t=1, \ldots, T$, we find 
\begin{align}
\,&\sum_{t=0}^{T} \Big\|\bm g^\epsilon (\bar{\bm \theta}^t) - \nabla f(\bar{\bm \theta}^{t+1})  + \frac{1}{\alpha} (\bar{\bm \theta}^{t+1} - \bar{\bm \theta}^t)\Big\|^2 \nonumber \\ 
\le\,&  \sum_{t=0}^{T} \big\|\nabla f (\bar{\bm \theta}^t) - \bm g^\ast(\bar{\bm \theta}^t)\big\|^2 +  \frac{2 (T+1) L_{\bar{\bm \theta}}^2 \epsilon }{\lambda_0}  \\ \nonumber
& + \frac{2}{\alpha}\big[F(\bar{\bm \theta}^{0}) - F(\bar{\bm \theta}^{T})\big] + \Big(\frac{L_f^2}{\eta} + \frac{L_f+\eta}{\alpha} \Big) \sum_{t=0}^{T} \big\|\bar{\bm \theta}^{t+1} - \bar{\bm \theta}^{t}\big\|^2. 
\end{align}
Using \eqref{eq:boundtheta1} to bound the last term yields  
\begin{align}
\,&\sum_{t=0}^{T}   \Big\|\bm g^\epsilon (\bar{\bm \theta}^t) - \nabla f(\bar{\bm \theta}^{t+1})  + \frac{1}{\alpha} (\bar{\bm \theta}^{t+1}    - \bar{\bm \theta}^t)\Big\|^2 \nonumber \\ 
\le\,& \,  2 \sum_{t=0}^{T} \left\|\nabla f (\bar{\bm \theta}^t) - \bm g^\ast(\bar{\bm \theta}^t)\right\|^2 + \frac{2 (T+1) L_{\bar{\bm \theta}}^2 \epsilon }{\lambda_0}  + \frac{2}{\alpha} \Delta_F + \beta \Delta_F \nonumber \\
&   + \frac{\beta}{\eta} \sum_{t=0}^{T} \big\|\nabla f (\bar{\bm \theta}^t) - \bm g^\ast(\bar{\bm \theta}^t)\big\|^2 + \frac{\beta (T+1)L^2_{\bar{\bm \theta}\bm  z}\epsilon}{\lambda_0} 
\end{align}
where $\beta = (\frac{L_f^2}{\eta} + \frac{L_f+\eta}{\alpha}) \frac{2 \alpha}{2-(L_f+\eta)\alpha}$. By taking expectation of both sides of this inequality, we obtain
\begin{align}
\label{eq:bnd_sbdif}
&\frac{1}{T+1}\mathbb E \Big[ \sum_{t=0}^{T} \Big\|\bm g^\epsilon (\bar{\bm \theta}^t)  - \nabla f(\bar{\bm \theta}^{t+1})   +   \frac{1}{\alpha} (\bar{\bm \theta}^{t+1}    - \bar{\bm \theta}^t)\Big\|^2 \Big]  \nonumber \\
\le\,&  
\Big(\frac{2}{\alpha} +\beta \Big) \frac{\Delta_F}{T+1} + \Big( \frac{\beta}{\eta}+2\Big) \sigma^2 \! + \frac{(\beta+2)L_{\bar{\bm \theta}}^2 \epsilon}{\lambda_0} 
\end{align} 
where we have used $\mathbb{E} [\|\nabla f (\bar{\bm \theta}^t) - \bm g^\ast(\bar{\bm \theta}^t)\|_2^2] \le \sigma^2$, which holds according to Assumption \ref{as:grdestimate}. By \cite [Theorem 10]{rockafellar2009variational} and \cite{xu2019non}, we know that 
\begin{equation}
-\bm{g}^\epsilon(\bar{\bm{\theta}}^t) - \frac{1}{\alpha}(\bar{\bm{\theta}}^{t+1}-\bar{\bm{\theta}}^t) \in \partial \bar r(\bar{\bm{\theta}}^{t+1}) 
\end{equation}
which gives
\begin{align*}
\nabla f(\bar{\bm{\theta}}^{t+1}) - \bm{g}^\epsilon  (\bar{\bm{\theta}}^t) - \frac{1}{\alpha}(\bar{\bm{\theta}}^{t+1}-\bar{\bm{\theta}}^t) 
&\in    \nabla f(\bar{\bm{\theta}}^{t+1}) + \partial \bar r(\bar{\bm{\theta}}^{t+1}) 
\\
& := \partial F(\bar{\bm{\theta}}^{t+1}).
\end{align*}
Upon replacing the latter in the left hand side of \eqref{eq:bnd_sbdif}, 
and recalling the definition of distance, we deduce that  
\begin{align*}
\mathbb{E}\big[\textrm{dist}(\bm{0},  \partial \hat{F}(\bar{\bm  \theta}^{t'}))\big]&\le\!   \big(\frac{2}{\alpha}\! +\!\beta \big) \frac{\Delta_F}{T}\! +\! \big( \frac{\beta}{\eta}\!+\!2\big) \sigma^2 \!+\! \frac{(\beta+2)L_{\bar{\bm \theta z}}^2 \epsilon}{\lambda_0}
\end{align*}
where $t'$ is randomly drawn from $ t'\in\{1,2,\ldots, T+1\}$, which concludes the proof.

\subsection{Proof of Theorem \ref{alg:spgda}}
\label{thm:spgda}
Instead of resorting to an oracle to obtain an $\epsilon$-optimal solver for the surrogate loss, here we utilize a single step stochastic gradient ascent with mini-batch size $M$ to solve the maximization step. Consequently, the updates become 
\begin{equation}
\bar{\bm{\theta}}^{t+1} = \textrm{prox}_{\alpha_t r}
\big(\bar{\bm{\theta}}^{t} -  \alpha_t \bm g^{t}(\bar{\bm \theta}^t) \big)  
\end{equation}
where $\bm g^{t}(\bar{\bm \theta}^t)  := \frac{1}{M} \sum_{m=1}^{M} \bm g (\bar{\bm \theta}^{t}, \bm \zeta_m^{t}; \bm z_m)$. Letting $\bm \delta(\bar{\bm{\theta}}^t) := \nabla f(\bar{\bm{\theta}}^t) - \bm g^t(\bar{\bm{\theta}}^t)$, and using the $L_f$-smoothness of $f(\bar{\bm{\theta}})$, we obtain  
\begin{align} 
& f(\bar{\bm \theta}^{t+1})  \le f(\bar{\bm{\theta}}^{t}) + \big \langle \nabla f(\bar{\bm{\theta}}^t), \bar{\bm{\theta}}^{t+1} - \bar{\bm{\theta}}^{t} \big \rangle  + \frac{L_f}{2} \| \bar{\bm{\theta}}^{t+1} - \bar{\bm{\theta}}^{t} \|^2 \nonumber \\
& \le f(\bar{\bm{\theta}}^{t}) + \big\langle \bm  g^t (\bar{\bm{\theta}}^t)+ \bm  \delta(\bar{\bm{\theta}}^t) , \bar{\bm{\theta}}^{t+1} - \bar{\bm{\theta}}^{t}\big \rangle  + \frac{L_f}{2} \| \bar{\bm{\theta}}^{t+1} - \bar{\bm{\theta}}^{t} \|^2. 
\label{eq:fsmooth}
\end{align}
Next, we substitute $ \bar{\bm{\theta}}^{t+1} \rightarrow \bm  u $, $  \bm{\theta}^t \rightarrow \bm{y}$, and $\bar{\bm{\theta}}^t - \alpha_t \bm  g^t (\bar{\bm{\theta}}^t) \rightarrow \bm  x$ in \eqref{eq:prox_prop}, to arrive at 
\begin{equation*}
\big \langle \bar{\bm{\theta}}^t - \alpha_t \bm  g^t (\bar{\bm{\theta}}^t) - \bar{\bm{\theta}}^{t+1} , \bar{\bm{\theta}}^t - \bar{\bm{\theta}}^{t+1}\big \rangle \le \alpha_t \bar{r}(\bar{\bm{\theta}}^t) - \alpha_t \bar{r}(\bar{\bm{\theta}}^{t+1}) 
\end{equation*}
which leads to 
\begin{equation*}
\big \langle \bm g^t (\bar{\bm{\theta}}^t), \bar{\bm{\theta}}^{t+1} - \bar{\bm{\theta}}^t \big \rangle \le \bar{r}(\bar{\bm{\theta}}^t) - \bar{r}(\bar{\bm{\theta}}^{t+1}) - \frac{1}{\alpha_t} \| \bar{\bm{\theta}}^{t+1} - \bar{\bm{\theta}}^{t}\|^2.
\end{equation*}
Substituting the latter into \eqref{eq:fsmooth}, gives
\begin{align*}
% \nonumber 
f(\bar{\bm{\theta}}^{t+1}) & \le f(\bar{\bm{\theta}}^{t}) + \big \langle \bm  \delta(\bar{\bm{\theta}}^t), \bar{\bm{\theta}}^{t+1} - \bar{\bm{\theta}}^{t}\big \rangle + \frac{L_f}{2} \| \bar{\bm{\theta}}^{t+1} - \bar{\bm{\theta}}^{t}\|^2 \nonumber \\ 
& \quad + \bar{r}(\bar{\bm{\theta}}^t) - \bar{r}(\bar{\bm{\theta}}^{t+1}) - \frac{1}{\alpha_t} \|\bar{\bm{\theta}}^{t+1} - \bar{\bm{\theta}}^{t}\|^2
\end{align*}
and with $F(\bm \theta) := f(\bm \theta) + \bar{r}(\bm \theta)$, we have 
\begin{align} \nonumber 
F(\bar{\bm{\theta}}^{t+1}) -  F(\bar{\bm{\theta}}^{t}) 
& \le \big\langle \bm  \delta(\bar{\bm{\theta}}^t), \bar{\bm{\theta}}^{t+1} - \bar{\bm{\theta}}^{t}\big \rangle
\\
& \quad + \Big( \frac{L_f}{2}- \frac{1}{\alpha_t} \Big) \| \bar{\bm{\theta}}^{t+1} - \bar{\bm{\theta}}^{t}\|^2.
\end{align}
Using Young's inequality $ \big\langle \bm  \delta(\bar{\bm{\theta}}^t), \bar{\bm{\theta}}^{t+1} - \bar{\bm{\theta}}^{t}\big \rangle \le \frac{1}{2} \|\bm  \delta(\bar{\bm{\theta}}^t) \|^2+\frac{1}{2}\|\bar{\bm{\theta}}^{t+1} - \bar{\bm{\theta}}^{t} \|^2$ implies that 
\begin{align} 
F(\bar{\bm{\theta}}^{t+1}) -  & F(\bar{\bm{\theta}}^{t}) \le \Big(\frac{L_f \!+ \!1}{2}- \frac{1}{\alpha_t} \Big) \|\bar{\bm{\theta}}^{t+1} - \bar{\bm{\theta}}^{t}\|^2 + \frac{\|\bm  \delta(\bar{\bm{\theta}}^t)\|^2}{2}
\label{eq:bnd_itrsF}
\end{align}
and after adding the term $\big \langle \bar{\bm{\theta}}^{t+1} - \bar{\bm{\theta}}^t, \nabla f(\bar{\bm{\theta}}^{t+1}) \big \rangle$ to both sides in \eqref{eq:bnd_itrsF}, and simplifying terms, yields   
\begin{align}
\nonumber
& \big \langle \bar{\bm{\theta}}^{t+1} - \bar{\bm{\theta}}^t,  \bm g^t (\bar{\bm{\theta}}^t) - \nabla f(\bar{\bm{\theta}}^{t+1})\big \rangle 
\\
& \le  -\Big (\frac{1}{2 \alpha_t} - \frac{L_f}{2} \Big) \|\bar{\bm{\theta}}^{t+1} - \bar{\bm{\theta}}^{t}\|^2 + F(\bar{\bm{\theta}}^{t}) - F(\bar{\bm{\theta}}^{t+1})  \nonumber 
\\  & \quad - \big \langle\bar{\bm{\theta}}^{t+1} - \bar{\bm{\theta}}^t, \bm \nabla f (\bar{\bm{\theta}}^{t+1}) - \nabla f(\bar{\bm{\theta}}^t)\big \rangle.
\end{align}
Completing the square yields
\begin{align}
\nonumber
& 
\big\|\bm g^t (\bar{\bm{\theta}}^t) - \nabla f(\bar{\bm{\theta}}^{t+1}) + \frac{1}{\alpha_t} (\bar{\bm{\theta}}^{t+1} - \bar{\bm{\theta}}^t)\big\|^2 \nonumber \\ 
\le \,
&  \|\bm g^t (\bar{\bm{\theta}}^t) - \nabla f(\bar{\bm{\theta}}^{t+1})\|^2 
+ \frac{1}{\alpha_t^2} \|\bar{\bm{\theta}}^{t+1} - \bar{\bm{\theta}}^t\|^2
\nonumber \\
&
+ \Big(\frac{L_f}{\alpha_t} - \frac{1}{\alpha_t^2}\Big) \|\bar{\bm{\theta}}^{t+1} - \bar{\bm{\theta}}^{t}\|^2 +\frac{2 (F(\bar{\bm{\theta}}^{t}) - F(\bar{\bm{\theta}}^{t+1}))}{\alpha_t} \nonumber \\ 
&
- \frac{2}{\alpha_t} \big \langle \bar{\bm{\theta}}^{t+1} - \bar{\bm{\theta}}^t, \nabla f (\bar{\bm{\theta}}^{t+1}) - \nabla f(\bar{\bm{\theta}}^t)\big \rangle \nonumber \\ 
\le\,
&
2 \|\bm g^t (\bar{\bm{\theta}}^t) - \nabla f(\bar{\bm{\theta}}^{t})\|^2 + 2 \|\nabla f (\bar{\bm{\theta}}^t) - \nabla f(\bar{\bm{\theta}}^{t+1})\|^2  
\nonumber  \\
& 
+ \frac{1}{\alpha_t^2} \|\bar{\bm{\theta}}^{t+1} - \bar{\bm{\theta}}^t\|^2 + \Big(\frac{L_f}{\alpha_t} - \frac{1}{\alpha_t^2}\Big) \|\bar{\bm{\theta}}^{t+1} - \bar{\bm{\theta}}^{t}\|^2   
\nonumber \\ 
& 
+ \frac{2 (F(\bar{\bm{\theta}}^{t}) - F(\bar{\bm{\theta}}^{t+1}))}{\alpha_t} - \frac{2}{\alpha_t} \langle  \bar{\bm{\theta}}^{t+1} - \bar{\bm{\theta}}^t , \nabla f (\bar{\bm{\theta}}^{t+1}) - \nabla f(\bar{\bm{\theta}}^t) \rangle \nonumber \\ 
\le\,
& 2 \|\bm g^t (\bar{\bm{\theta}}^t) - \nabla f(\bar{\bm{\theta}}^{t})\|^2 + 2 L_f^2 \|\bar{\bm{\theta}}^{t+1} - \bar{\bm{\theta}}^{t}\|^2
\nonumber  \\
& 
+ \frac{1}{\alpha_t^2} \|\bar{\bm{\theta}}^{t+1} - \bar{\bm{\theta}}^{t}\|^2 + \Big(\frac{L_f}{\alpha_t} - \frac{1}{\alpha_t^2}\Big) \|\bar{\bm{\theta}}^{t+1} - \bar{\bm{\theta}}^{t}\|^2 \nonumber \\ 
& 
+ \frac{2 (F(\bar{\bm{\theta}}^{t}) - F(\bar{\bm{\theta}}^{t+1}))}{\alpha_t} + \frac{2 L_f}{\alpha_t} \|\bar{\bm{\theta}}^{t+1} - \bar{\bm{\theta}}^{t}\|^2\nonumber \\ 
\le\,
&  2 \|\bm g^t (\bar{\bm{\theta}}^t) - \nabla f(\bar{\bm{\theta}}^{t})\|^2 + \frac{2(F(\bar{\bm{\theta}}^{t}) - F(\bar{\bm{\theta}}^{t+1}))}{\alpha_t}  + \nonumber \\
& 
\frac{3L_f + 2L_f^2 \alpha_t }{\alpha_t} \|\bar{\bm{\theta}}^{t+1} - \bar{\bm{\theta}}^{t}\|^2.
\label{eq:subdifbound}
\end{align}

Recalling that $\bm \delta(\bar{\bm{\theta}}^t) := \nabla f(\bar{\bm{\theta}}^t) - \bm g^t(\bar{\bm{\theta}}^t)$, we can bound the first term as
\begin{align}\label{eq:graderrorbound}
\,&\mathbb E \Big[\|\bm g^{t} (\bar{\bm{\theta}}^{t})  - \nabla f(\bar{\bm{\theta}}^{t})\|^2  \,\big|\bm\theta^t\Big] \nonumber \\ 
=\,& \mathbb E \Big[\left \| \bm g^{\ast}(\bar{\bm{\theta}}^{t})  - \nabla f(\bar{\bm{\theta}}^{t}) + \bm \delta^{t} \right \|^2\big|\bm\theta^t\Big]  \nonumber \\ 
=\, &  \| \bm g^{\ast}(\bar{\bm{\theta}}^{t})  - \nabla f(\bar{\bm{\theta}}^{t}) \|^2 + \| \bm \delta^{t}  \|^2 + 2 \mathbb E \Big[\left \langle \bm g^{\ast}(\bar{\bm{\theta}}^{t})  - \nabla f(\bar{\bm{\theta}}^{t}) ,\bm \delta^{t} \right \rangle\big|\bm{\theta}^t\Big]
\end{align}
where the third equality is obtained by expanding the square term, and using $\mathbb E \big[\langle \bm g^{\ast}(\bar{\bm{\theta}}^{t})  - \nabla f(\bar{\bm{\theta}}^{t}) ,\bm \delta^{t} \rangle  \big| \bar{\bm{\theta}}^t \big] = \bm 0$. We will further bound the right hand side here as follows. Recalling that $\bm \delta^{t} =  \frac{1}{M} \sum_{m=1}^{M} \bm g (\bar {\bm \theta}^{t}, \bm \zeta_m^{t}; \bm z_m)  - \bm g^{\ast}(\bar{\bm \theta}^{t})$, where  $\bm g^\ast (\bm \theta^t) := \frac{1}{M} \sum_{m=1}^M \nabla_{\bar{\bm{\theta}}} \psi(\bar{\bm{\theta}}^{t}, \bm  \zeta_m^{\ast t}; \bm  z_m)$, it holds that  
\begin{align}
& \mathbb{E}\Big[\Big\| \frac{1}{M} \sum_{m=1}^{M}\Big[ \bm g (\bar{\bm \theta}^{t}, \bm \zeta_m^{t}; \bm z_m)  -\bm{g}^\ast(\bar{\bm{\theta}}^t) \big]\Big\|^2\Big|\bar{\bm{\theta}}^t, \bm \zeta_m^t \Big]\nonumber\\
=\,& \frac{1}{M^2}\!\sum_{m=1}^M\! \mathbb{E} \Big[\big\| \nabla_{\bar{\bm{\theta}}} \psi(\bar{\bm{\theta}}^{t},\! \bm  \zeta_m^{t};\! \bm  z_m) \! -\!\nabla_{\bar{\bm{\theta}}} \psi(\bar{\bm{\theta}}^{t}, \!\bm  \zeta_m^{\ast t} ; \!\bm  z_m)\big\|^2\big|\bar{\bm{\theta}}^t, \bm \zeta_m^t \Big]
\nonumber\\
\le\,&  
\frac{L^2_{\bm \theta \bm z}}{M^2}   \sum_{m=1}^{M} \!\left\| \bm \zeta_m^{t}  - {\bm \zeta^{\ast t}_m} \right \|^2
\label{eq:gtgstar}
\end{align} 
where the second equality is because the samples $\{\bm{z}_m\}_{m=1}^M$ are i.i.d., and last inequality holds due to the Lipschitz smoothness of $\psi(\cdot)$. Since $\bm{\zeta}_m^t$ is obtained by a single gradient ascent update over a $\mu$-strongly concave function, we have that 
\begin{equation}
\frac{L^2_{\bm \theta \bm z}}{M^2}   \sum_{m=1}^{M} \!\left\| \bm \zeta_m^{t}  - {\bm \zeta^{\ast t}_m} \right \|^2 \le \frac{L^2_{\bm \theta \bm z}}{M}  \Big[\! \left(1-\alpha_t \mu \right) D^2 + \alpha_t^2 B^2 \Big]
\end{equation}
where $D$ is the diameter of the feasible set, and $\alpha_t>0$ is the step size. The following holds for the expected error term 
\begin{equation}
\mathbb{E}\Big[\|\bm \delta^{t}\|^2\big|\bar{\bm{\theta}}^t, \bm \zeta_m^t\Big] \le \frac{L^2_{\bm \theta \bm z}}{M}  \Big[\! \left(1-\alpha_t \mu \right) D^2 + \alpha_t^2 B^2 \Big] 
\end{equation}
and using it in \eqref{eq:graderrorbound}, we arrive at 
\begin{align}
& \mathbb E \Big[\|\bm g^{t} (\bar{\bm{\theta}}^{t}) -  \nabla f(\bar{\bm{\theta}}^{t})\|^2\big|\bm\theta^t\Big] 
\le 2 \| \bm g^{\ast}(\bar{\bm{\theta}}^{t})  - \nabla f(\bar{\bm{\theta}}^{t}) \|^2 
\\
& \quad 
+ \frac{L^2_{\bar{\bm{\theta}} \bm z}}{M}  \Big[\! \left(1-\alpha_t \mu \right) D^2 + \alpha_t^2 B^2 \Big]. 
\end{align}
Substituting the last inequality  into \eqref{eq:subdifbound} boils down to 
\begin{align}
& \mathbb E \Big[ \big\|\bm g^t (\bar{\bm{\theta}}^t) -   \nabla  f(\bar{\bm{\theta}}^{t+1}) + \frac{1}{\alpha_t} (\bar{\bm{\theta}}^{t+1} - \bar{\bm{\theta}}^t)\big\|^2 \big| \bar{\bm{\theta}}^t \Big]\nonumber\\
\le \,& 4  \| \bm g^{\ast}(\bar{\bm{\theta}}^{t})  - \nabla f(\bar{\bm{\theta}}^{t}) \|^2  +\frac{3L_f + 2L_f^2 \alpha_t }{\alpha_t} \mathbb E \Big[ \|\bar{\bm{\theta}}^{t+1} - \bar{\bm{\theta}}^{t}\|^2 \big| \bar{\bm{\theta}}^t\Big]  \nonumber \\ 
& +\frac{2 F(\bar{\bm{\theta}}^{t}) - 2 \mathbb E \big[ F(\bar{\bm{\theta}}^{t+1})\big| \bar{\bm{\theta}}^t  \big]}{\alpha_t} +\frac{L^2_{\bar{\bm{\theta}} \bm z}}{M}  \Big[\! \left(1-\alpha_t \mu \right) D^2 + \alpha_t^2 B^2 \Big]  .
\end{align}

Taking again expectation over $\bar{\bm{\theta}}^t$ on both sides, yields  
\begin{align}
\label{eq:subdiffsgda}
\, & \mathbb E  \Big\|\bm g^t (\bar{\bm{\theta}}^t)   - \nabla f(\bar{\bm{\theta}}^{t+1}) + \frac{1}{\alpha_t} (\bar{\bm{\theta}}^{t+1} - \bar{\bm{\theta}}^t)\Big\|^2 
\\  
\le \,&  4  \mathbb E \Big[ \| \bm g^{\ast}(\bar{\bm{\theta}}^{t})  - \nabla f(\bar{\bm{\theta}}^{t}) \|^2 \Big] + \frac{L^2_{\bar{\bm{\theta}} \bm z}}{M}  \Big[\! \left(1-\alpha_t \mu \right) D^2 + \alpha_t^2 B^2 \Big]  \nonumber \\ 
& \! + \mathbb E \Big[ \frac{2 F(\bar{\bm{\theta}}^{t}) - 2 F(\bar{\bm{\theta}}^{t+1})}{\alpha_t} + \frac{3L_f + 2L_f^2 \alpha_t }{\alpha_t}  \|\bar{\bm{\theta}}^{t+1} - \bar{\bm{\theta}}^{t}\|^2 \Big].  \nonumber 
\end{align}
Recalling that $\mathbb E [\|\bm \psi^\ast (\bar{\bm{\theta}}^{t}, \bm \zeta_m^{t}; \bm z_m)-\nabla f(\bar{\bm{\theta}}^t)\|^2] \le \sigma^2$, and that $\bm g^\ast(\bar{\bm{\theta}}^t) = \frac{1}{M} \sum_{m=1}^{M} \bm \psi (\bar{\bm{\theta}}^{t}, \bm \zeta_m^{\ast t}; \bm z_m)$, the first term on the right hand side can be bounded by $\frac{4\sigma^2}{M}$. For a fixed learning rate $\alpha>0$, summing inequalities \eqref{eq:subdiffsgda} from $t=0,\ldots,T$, yields
\begin{align}
&\frac{1}{T+1}\mathbb{E}   \Big[\sum_{t=0}^T\big\|\bm g^t (\bar{\bm{\theta}}^t)  - \nabla f(\bar{\bm{\theta}}^{t+1}) + \frac{1}{\alpha_t} (\bar{\bm{\theta}}^{t+1} - \bar{\bm{\theta}}^t)\big\|^2
\Big] \nonumber \\ 
\le\, 
&  \frac{2}{\alpha(T+1)}\big(F(\bm{\theta^0})-\mathbb E [F(\bm\theta^T)]\big) \! + \! \frac{2L^2_{\bar{\bm{\theta}} \bm z}}{M}  [(1-\alpha \mu ) D^2 + \alpha^2 B^2 ]  
\nonumber 
\\ 
& 
+ \frac{3L_f + 2L_f^2 \alpha }{\alpha}  \frac{1}{T+1} \mathbb E \Big[ \sum_{t=0}^T    \|\bar{\bm{\theta}}^{t+1} - \bar{\bm{\theta}}^{t}\|^2 \Big] +
\frac{4\sigma^2}{M}
\nonumber \\ 
\le \,& \frac{1}{T+1} \bigg\{\frac{2}{\alpha} + \frac{6L_f + 4L_f^2 \alpha }{[2- \alpha(L_f+\beta)]}\bigg\} ( F(\bar{\bm{\theta}}^{0}) -  \mathbb E [F(\bar{\bm{\theta}}^{T})] ) + \frac{4 \sigma^2}{M} \nonumber \\ 
&  +\frac{2  L^2_{\bar{\bm{\theta}} \bm z}}{M}\bigg\{ 1+ \!\frac{ 3L_f + 2L_f^2 \alpha }{2(2- \alpha(L_f+\beta))}  \bigg\} \Big[\! \left(1-\alpha \mu \right) D^2 + \alpha^2 B^2 \Big]. 
\end{align}
Consider now replacing $F(\bar{\bm{\theta}}^0) -  F(\bar{\bm{\theta}}^T)$ with $\Delta_F = F(\bar{\bm{\theta}}^0) -\inf_{\bar{\bm{\theta}}} \bm F(\bar{\bm{\theta}})$, and note that $\bm g^t (\bar{\bm{\theta}}^t) - \nabla f(\bar{\bm{\theta}}^{t+1}) + \frac{1}{\alpha_t} (\bar{\bm{\theta}}^{t+1} - \bar{\bm{\theta}}^t) \in \partial {F}(\bar{\bm{\theta}}^{t+1})$, where $\partial {F}$ denotes the set of subgradients of $F$. It then becomes clear that
\begin{align*}
& 
\mathbb E \big[ \textrm{dist}(0, \partial F)^2 \big] 
\\
\le\,&  \frac{1}{T+1}\mathbb{E}\Big[\sum_{t=0}^T\Big\|\bm g^t (\bar{\bm{\theta}}^t) - \nabla f(\bar{\bm{\theta}}^{t+1}) + \frac{1}{\alpha_t} (\bar{\bm{\theta}}^{t+1} - \bar{\bm{\theta}}^t)\Big\|^2
\Big] \nonumber \\
\le\,&  \frac{\zeta}{T+1} \Delta_F +\frac{2  L^2_{\bar{\bm{\theta}} \bm z} \nu}{N} \Big[\! \left(1-\alpha \mu \right) D^2 + \alpha^2 B^2 \Big] + \frac{4 \sigma^2}{M}
\end{align*}
where $\zeta = \frac{2}{\alpha} + \frac{6L_f + 4L_f^2 \alpha }{(2- \alpha(L_f+\beta))}$ and $\nu = 1+ \frac{3L_f + 2L_f^2 \alpha }{2(2- \alpha(L_f+\beta))}$, which concludes the proof.

%\section{Proof of Lemma \ref{lem:up}}\label{seca:proofup}

%%%%%%%%%%%%%%%%%%%%%%%%%%%%%%%%%%%%%%%%%%%%%%%%%%%%%%%%%%%%%%%%%%%%%%%%%%%%%%%%

%%%%%%%%%%%%%%%%%%%%%%%%%%%%%%%%%%%%%%%%%%%%%%%%%%%%%%%%%%%%%%%%%%%%%%%%%%%%%%%%
% Appendices
%%%%%%%%%%%%%%%%%%%%%%%%%%%%%%%%%%%%%%%%%%%%%%%%%%%%%%%%%%%%%%%%%%%%%%%%%%%%%%%%
%\appendix
%\include{chapters/app_glossary}
%%%%%%%%%%%%%%%%%%%%%%%%%%%%%%%%%%%%%%%%%%%%%%%%%%%%%%%%%%%%%%%%%%%%%%%%%%%%%%%%

%%%%%%%%%%%%%%%%%%%%%%%%%%%%%%%%%%%%%%%%%%%%%%%%%%%%%%%%%%%%%%%%%%%%%%%%%%%%%%%%
% End the Document
%%%%%%%%%%%%%%%%%%%%%%%%%%%%%%%%%%%%%%%%%%%%%%%%%%%%%%%%%%%%%%%%%%%%%%%%%%%%%%%%
\end{document}